\documentclass[12pt,lot, lof]{puthesis}
\newcommand{\proquestmode}{}

\title{Neural Field Representations of Mobile Computational Photography}

\submitted{March 2025}  
\copyrightyear{2025}  
\author{Ilya Chugunov}
\adviser{Professor Felix Heide}  
\department{Computer Science}

\usepackage{xcolor}
\usepackage{graphicx}
\usepackage{bm}
\definecolor{URLBlue}{RGB}{105, 105, 250}
\definecolor{URLBlack}{RGB}{0, 0, 0}
\usepackage{cancel}
\usepackage{amsmath}
\usepackage{amssymb}
\usepackage{booktabs}

\let\originalleft\left
\let\originalright\right
\def\left#1{\mathopen{}\originalleft#1}
\def\right#1{\originalright#1\mathclose{}}

\newcommand{\ucomma}{%
  \leavevmode 
  \kern-0.08em\text{,}\kern0.08em 
}

\usepackage[utf8]{inputenc} 
\usepackage[russian,english]{babel}

\usepackage{graphicx}

\usepackage{verbatim}
\usepackage{makecell}

\usepackage{multirow}
\usepackage{longtable}
\usepackage{wrapfig}
\usepackage{caption}
\usepackage{hyperref}
\hypersetup{
    colorlinks=true, 
    linkcolor=URLBlue, 
    citecolor=URLBlue, 
    filecolor=URLBlue, 
    urlcolor=URLBlue
    }
\usepackage{booktabs}

\ifdefined\printmode

\usepackage{url}

\else

\ifdefined\proquestmode

\usepackage{hyperref}
\hypersetup{bookmarksnumbered}

\makeatletter
\hypersetup{pdftitle=\@title,pdfauthor=\@author}
\makeatother

\else

\usepackage{hyperref}
\hypersetup{colorlinks,bookmarksnumbered}

\makeatletter
\hypersetup{pdftitle=\@title,pdfauthor=\@author}
\makeatother

\fi 
\fi 

\ifodd 0
\renewcommand{\maketitlepage}{}
\renewcommand*{\makecopyrightpage}{}
\renewcommand*{\makeabstract}{}

\else

\abstract{
Over the past two decades, mobile imaging has experienced a profound transformation, with cell phones rapidly eclipsing all other forms of digital photography in popularity. Today’s cell phones are equipped with a diverse range of imaging technologies -- laser depth ranging, multi-focal camera arrays, and split-pixel sensors -- alongside non-visual sensors such as gyroscopes, accelerometers, and magnetometers. This, combined with on-board integrated chips for image and signal processing, makes the cell phone a versatile pocket-sized computational imaging platform. 

Parallel to this, we have seen in recent years how neural fields -- small neural networks trained to map continuous spatial input coordinates to output signals -- enable the reconstruction of complex scenes without explicit data representations such as pixel arrays or point clouds. 
In this thesis, I demonstrate how carefully designed neural field models can compactly represent complex geometry and lighting effects. Enabling applications such as depth estimation, layer separation, and image stitching directly from collected in-the-wild mobile photography data. These methods outperform state-of-the-art approaches without relying on complex pre-processing steps, labeled ground truth data, or machine learning priors. Instead, they leverage well-constructed, self-regularized models that tackle challenging inverse problems through stochastic gradient descent, fitting directly to raw measurements from a smartphone.
}

\acknowledgements{
I'm incredibly grateful for the opportunities I've received during my years at Princeton, and for all the folks I've been able to work with along the way. I thank my advisor, Felix Heide, for giving me the tools to become a competent researcher, introducing me to a broad slice of the computational imaging world from photons to silicon. I thank Jiawen Chen and my industry collaborators at Adobe, Meta, and Google research for helping me find the direction of work that would shape my PhD. I thank my committee members -- Adam Finkelstein, Szymon Rusinkiewicz, Jia Deng, and Ellen Zhong -- both for their time on and for their boundless advice beyond the scope of this thesis. I thank Olga Russakovsky and Sanjeev Arora for the experiences they gave to me to grow as a STEM educator. 

I'm grateful for the many, many amazing colleagues, collaborators, and mentors I've had along the way. For Seung-Hwan, who taught me that the ``F'' in ``4F System'' does not in fact stand for ``Fish''; for Alex, who taught me that Blender is just Python in a trench coat; and for all the undergrads, the grads, the postdocs, and the post-postdocs I've had the blessing to meet.

Above all, I could not have done this without the support of my family, friends, dogs, and cat.{\selectlanguage{russian} Миша и Марина, спасибо что всегда напоминали мне что после дождя выходит солнце, но даже в солнечный день стоит взять с собой зонт.} Grisha and Regine, thank you for giving me a shoulder to lean on, and a floor to sleep on. Zheng, thank you for being the candlelight in the office, making everything a little brighter and a little warmer. James, Greg, Dinesh, even if we're not in the same house, state, or continent, you'll always be my roommates. Umka, I wish you could have seen Princeton, it had everything you wanted: free food and free squirrels. \\ 

\noindent \textit{This research was supported by NSF GRFP (Award Number 2039656).}
}

\dedication{To Mikhail and Marina.}

\fi  

\begin{document}
{\hypersetup{linkcolor=black}
\makefrontmatter
}

\doublespacing
\chapter{Introduction\label{ch:intro}}
In the following sections, we will explore the fundamentals of mobile computational photography (from photon collection to image processing), neural field representations, and how we can apply the latter to solve problems in the former. But before that, let's start with a discussion of -- from a research, educational, and industry perspective -- what exactly makes this a compelling problem space to work in.  

\section{Motivation}

\noindent \textbf{Dataset Generation.} Most images on the internet, 94\% according to ~\cite{broz2024photostats}, are generated from cell phone photography. This means most internet image datasets ~\cite{deng2009imagenet, lin2014microsoft} are in large part really cell phone data collections. And just as high-quality image datasets drive advancements in computer vision tasks, the generation of high-quality cell phone data drives the creation of downstream image datasets. As an example, it would be extremely difficult to train generative image models~\cite{rombach2022high} to produce visually appealing night-time scenes without sensors and low-level processing algorithms capable of reconstructing images (training data) in low-light settings~\cite{liba2019handheld}. And while physically based rendering can produce realistic simulated data, we rely on digital photography to generate the environmental and object textures to build these scenes -- even for fully procedurally generated approaches~\cite{raistrick2023infinite}, photographs provide invaluable reference geometries to model the assets off of.

\begin{figure}[t!]
  \centering
  \includegraphics[width=\linewidth]{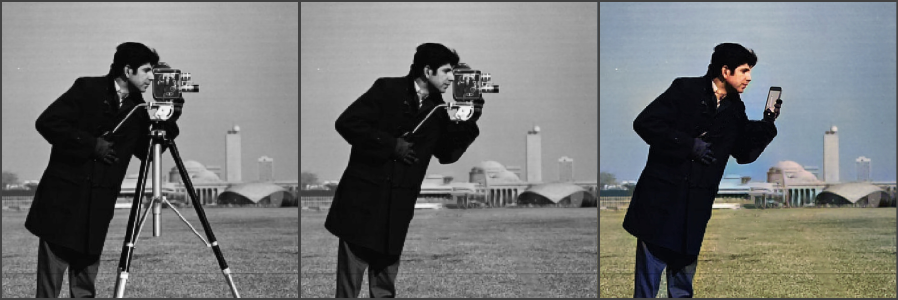}
  \caption{From fixed, tripod-stabilized systems to portable personal cameras and modern, multi-sensor, internet-connected handheld computational imaging platforms, photography has undergone significant evolution over the past century. (Original image courtesy of ~\cite{schreiber1978image})}
  \label{fig:0_cameraman}
\end{figure}

Consequently, as mobile imaging technologies evolve -- continuing the process illustrated in Figure \ref{fig:0_cameraman}) -- so does the data they produce. Given the sheer number of mobile devices in use globally, this trend has a profoundly democratizing effect. The introduction of LiDAR-based depth sensors in Apple iPhones~\cite{chugunov2022implicit} means that \textit{tens of millions} of ordinary people can now participate in the creation of depth datasets. Split-pixel fast-autofocus sensors in both mobile and professional cameras~\cite{shi2024split} means hundreds of millions of devices can now effectively record micro-baseline stereo measurements~\cite{joshi2014micro}. And the proliferation of multi-camera configurations in cell phones means a higher diversity of lens geometry data than ever before --- from short focal length fisheye captures to 5x magnification telephoto images. As new sensor, optical, and image processing developments are made, such as miniaturized hyperspectral or polarization mobile imagers, the stream of data from these will invariably lead to developments in computer vision applications. Both in creating novel problem spaces (e.g., handheld material classification) and in simplifying existing problems (e.g., using LiDAR-driven metric depth for object segmentation and tracking). Thus, research in mobile computational photography has the exciting potential to
\begin{wrapfigure}{r}{0.4\textwidth}
    \hspace*{0.5em} 
    \begin{minipage}{0.4\textwidth} 
        \includegraphics[width=\linewidth]{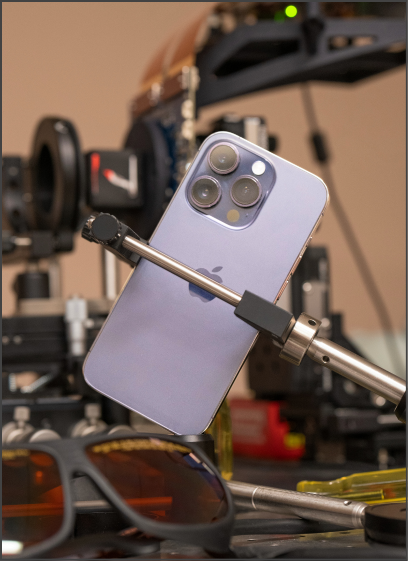} 
        \caption{The modern cell phone is a portable computational imaging platform that fits as well into a research laboratory as well as it does into your pocket.}
        \vspace{-2em}
        \label{fig:rightfigure}
    \end{minipage}
\end{wrapfigure}
 shape the future of computer vision by enabling researchers to access richer, more varied datasets and richer, more varied problem spaces.

\noindent \textbf{Pocket Computational Imaging.} For most people, the term ``computational imaging" evokes the image of an optics table cluttered with lenses and lasers, if it evokes anything at all. And while it is true that a large part of the research in the field is in scientific or biomedical imaging domains -- on \textit{specialized} topics such as ``fast dynamic 3D magnetic resonance spectroscopic imaging~\cite{larson2011fast}" or ``multiplexed coded illumination for Fourier ptychography microscopy~\cite{tian2014multiplexed}" -- computational imaging extends far beyond the laboratory setting. Every-day applications of it are all around us, seamlessly integrated into the devices we use daily -- for example, the night mode in your cell phone~\cite{liba2019handheld} is very much a form of computational imaging. Yet, particularly in graphics and machine learning communities, there can exist a divide between ``imaging people," who often focus on images as \textbf{outputs} of a process, and ``vision people," who tend to view images as \textbf{inputs} to a method. In my experience, mobile computational photography can serve as a strong and accessible link between these disciplines.


The images used in computer vision applications, captured by cell phones such as the one illustrated in Figure ~\ref{fig:rightfigure}, are themselves the product of complex imaging processes. The final result of photons reflected off object surfaces, focused through wafer-thin lenses, filtered by color, and converted first to electrons and then to digital readouts through nano-fabricated silicon semiconductor circuits. Over the course of my thesis work, I wrote a number of data capture applications for both iOS and Android devices designed to extract as much raw camera data as possible to simplify downstream reconstruction tasks. While the ultimate research goals were inverse graphics applications, such as removing reflections from glass or stitching images together, the engineering and data capture work provided a valuable view into the fundamentals of the imaging process. Encountering and circumventing challenges from real-world non-idealities such as sensor read noise, lens distortion, and motion blur across varying exposure settings and lens geometries. Many of the lessons learned from this process were directly transferable to laboratory settings -- e.g., accurate calibration is as valuable for getting accurate depth maps~\cite{chugunov2023shakes} as it is for optically encoded inverse reconstruction~\cite{shi2024split}. Most importantly, in contrast to an MRI machine or a scanning electron microscope, most researchers already own a cell phone and are familiar with its operation. This makes mobile computational photography an \textit{extremely cost-effective} space for students and young researchers to explore and understand key principles in optics and signal processing. 

\noindent \textbf{High Throughput ``Mobile Devices".} The methods I present in this thesis operate on data captured from commodity cell phones, but I frame this work within a much broader definition of ``mobile devices". Cell phones have undergone a meteoric rise in popularity over the last decades as the primary communication, information, and photography tools for billions of people. However, there is no guarantee that this trend will continue indefinitely. The commercialization of VR devices -- primarily for entertainment -- has already happened, and in the coming decades products such as lightweight AR glasses may take over the role currently held by cell phones~\cite{minaee2022modern}. I categorize all of these as mobile devices, as they are devices which -- whether in your hand or on your head -- are inherently mobile, moveable. Furthermore, many of the questions addressed in mobile computational photography -- e.g., how to reconstruct scenes with large rotational camera motion -- can prove just as relevant for data collected from drones, robots, autonomous vehicles, and other self-navigating platforms. 

	As the number of devices, number of sensors per device, and number of signal outputs per sensor all continue to rise, so does the need for methods to compress and extract salient information out of this flood of data. As we'll see in the next section, it’s quite telling when flagship cell phone cameras natively capture images at a resolution of 3000 $\times$ 4000px by default, only to remove 3/4 of these pixels before saving the image. This highlights a fundamental tradeoff in modern imaging systems: balancing the ability to capture high spatial and temporal resolution details with the limitations of storage, transmission, and computational resources. As novel sensors emerge and mature -- such as single photon detectors which can produce hundreds of gigabytes of information per second~\cite{zhang2024streaming} -- this challenge will continue to be exacerbated. The test-time optimization approaches in this thesis offer a scalable solution to this challenge by decomposing large, otherwise intractable datasets into millions of efficient and manageable ray-based operations. These methods extract and condense information directly from the captured data, embedding it into the implicit representations~\cite{sitzmann2020implicit} encoded within the weights of neural field models. The design and deployment of these models has the potential to redefine how we process and reconstruct data from emerging sensor technologies, paving the way for more effective solutions to large inverse imaging problems in both everyday and scientific applications.

\section{Background} The following sections will cover the fundamentals of mobile photography\footnote{I highly recommend \textit{Mobile Computational Photography: A Tour}~\cite{delbracio2021mobile} for further reading.} in the context of a potentially dynamic 3D environment represented by neural field models. In Sec. \ref{ch:signalgeneration} \textit{Signal Generation} we will examine how photons from the scene are captured on a camera sensor and converted to digital signals. Next, in Sec. \ref{ch:imageprocessing} \textit{Image Processing} we will see how these digital signals can be collected and converted into ``eye-pleasing" images through processes such as \textit{demosaicing} and \textit{burst fusion}. In Sec. \ref{ch:3dprojection} \textit{3D Projection} we will discuss how these images correspond to views of a 3D environment and how their pixels can be mapped back to points in the scene. Then in Sec. \ref{ch:neuralfields}, we conclude with an overview of neural field representations and their application in modeling 2D, 3D, and higher-dimensional scene data.

\subsection{Signal Generation\label{ch:signalgeneration}}
\begin{figure}[t!]
  \centering
  \includegraphics[width=0.9\linewidth]{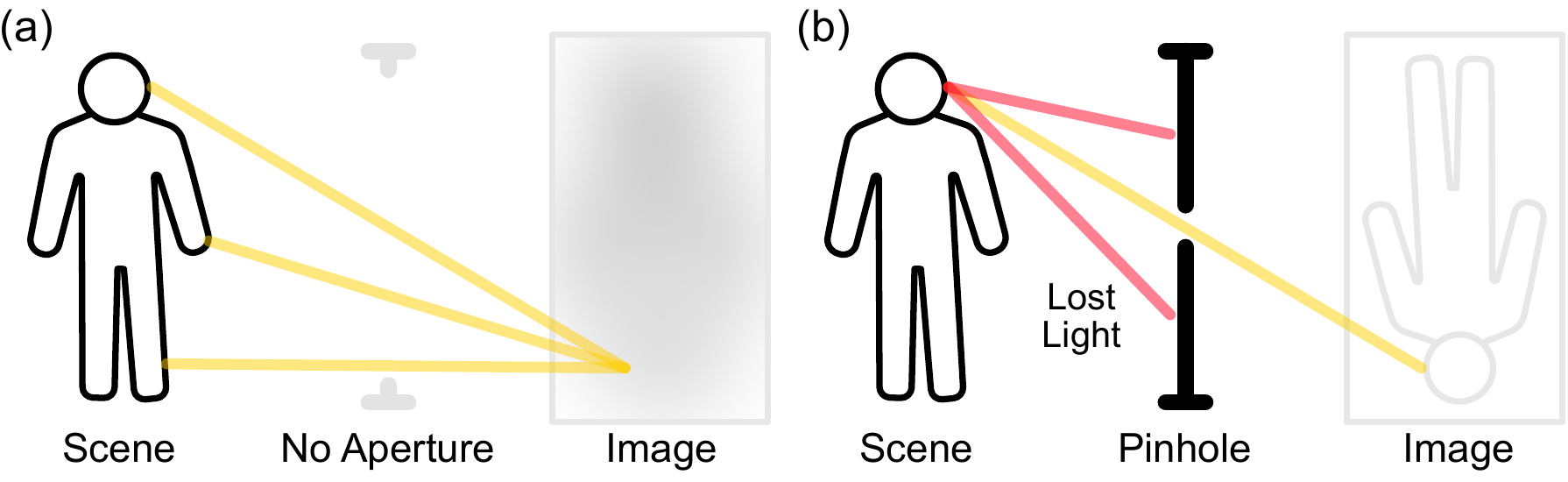}
  \caption{(a) Without an aperture to block light, all rays from the scene reach the imaging plane, resulting in a blurred projection of overlapping rays (blending the head, arm, and foot of the subject). (b) A pinhole camera restricts light to a narrow beam from each point in the scene, allowing distinct rays to pass through and form a sharp inverted image, at the cost of throwing away a majority of the light from the scene.}
  \label{fig:pinhole-lens}
\end{figure}

A natural place to start a discussion of imaging is \textit{light collection}, getting photons from the scene  to an image. Humans have been performing \textit{light collection} to form images long before the advent of digital image sensors, or analog film, or any other ``light recording devices". In fact, for a few centuries with the help of the \textit{camera obscura}~\cite{richmond1982camera}, humans \textit{were} the recording device. Using a simple pinhole camera, illustrated in Fig.~\ref{fig:pinhole-lens} (b), an artist could project the image of a scene onto a flat surface to use as reference for a painting or drawing. The reason this process forms an image is because the pinhole allows only a narrow beam of light from each point in the scene to pass through and project onto a specific location on the imaging plane, blocking the other rays of light that would otherwise overlap with it and turn the image into a blurry mix of colors; see Fig.~\ref{fig:pinhole-lens} (a). This approach, however, throws away the 
\begin{wrapfigure}{r}{0.5\textwidth}
    \hspace*{0.5em} 
    \begin{minipage}{0.5\textwidth} 
        \includegraphics[width=\linewidth]{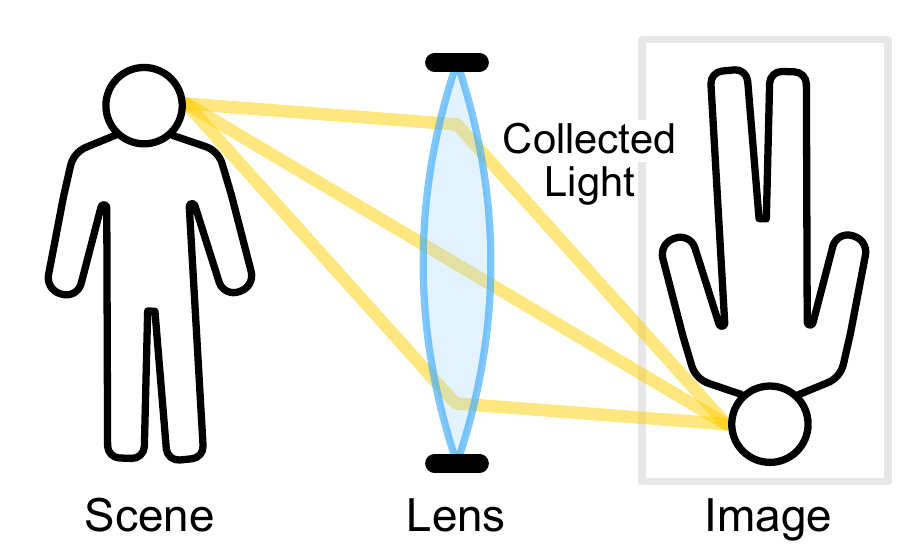} 
        \caption{A lens collects and focuses light from a point in the scene onto a point in the imaging plane, allowing a larger amount of light to form both a bright and sharp image.}
        \vspace{-2em}
        \label{fig:lens-imaging}
    \end{minipage}
\end{wrapfigure}
majority of the light from the scene, resulting in a very dark image -- hence why the image plane of a camera obscura was typically enclosed in a very dark box, so the user could actually see it.

A camera lens, shown in Fig.~\ref{fig:lens-imaging}, can in many ways be thought of as a physical analogue to a ``larger pinhole," gathering light from a wider area while still focusing it sharply onto the imaging plane to form a brighter image. In practice, however, making a lens actually act like a pinhole is an \textit{extremely} non-trivial task that has generated entire multi-billion dollar industries in optical design, materials, and fabrication research. This is because of the complexity of light propagation: as light interacts with material interfaces, it refracts into, reflects off of, and disperses through the medium, requiring careful design to control and align these effects for precise image formation. Modern mobile imaging camera systems often contain around a \textit{dozen} individual high-precision micro-fabricated lens elements~\cite{bliedtner2011wafer} all working in tandem to correct for these aberrations and produce a sharp final image. However, not all these effects can be perfectly corrected for, and real world lens systems do not perfectly simplify to a pinhole model. This is where \textit{calibration data} is essential, quantitatively describing how the actual constructed camera system deviates from a perfect model, and we will extensively use device calibration data to simplify and improve the accuracy of the models presented in the following chapters. One particularly useful piece of data is a set of \textit{lens distortion values}. As light rays pass through the lens elements in a camera system, they are bent and no longer follow a straight path between a point in the scene and its corresponding point in the image, as they would in a pinhole camera model. These lens distortion values describe how light rays must be adjusted to account for this bending, a correction crucial for accurately projecting scene points to their corresponding locations in space, as we will see in Sec.~\ref{ch:3dprojection}.

\begin{figure}[t!]
  \centering
  \includegraphics[width=\linewidth]{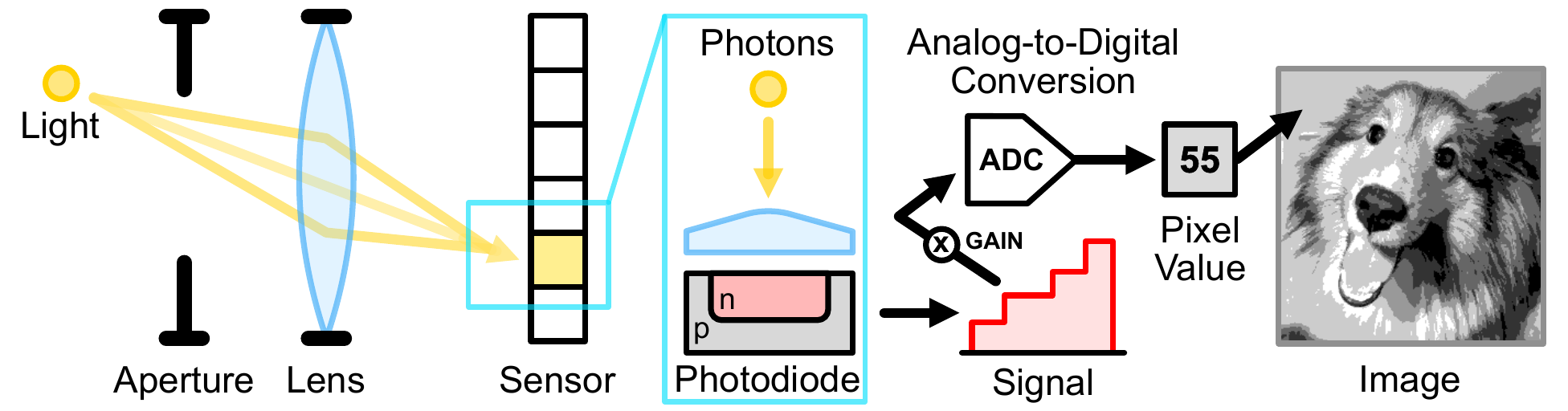}
  \caption{Simplified overview of image formation in a digital camera. Light enters through the aperture and is focused by the lens onto the sensor, where it is converted into an electrical signal by a photodiode. The analog signal is then digitized using an Analog-to-Digital Converter (ADC) to produce pixel values, ultimately forming a digital image.}
  \label{fig:signal-generation}
\end{figure}

Illustrated in Fig.~\ref{fig:signal-generation}, after focusing light onto an image plane, we need a few more steps to convert it into a digital signal and computer-interpretable image. Key to this is process is the \textit{photodiode}, a device that converts light into electrical signal via the photoelectric effect. While the specifics of this process involve complex quantum behavior\footnote{The work~\cite{einstein1905heuristic} for which Einstein won the 1921 Nobel prize in physics.} -- photons absorbed by semiconductor material exciting electrons from the valence band to the conduction band -- for our purposes, we can treat the photodiode as a \textit{device that converts photons into electrons}. This electrical signal is very weak, however, and needs to be aggregated, multiplied, and converted into a computer-readable digital signal before we can write it to an image. The first practical implementation of this concept was in CCDs (charge-coupled devices)~\cite{boyle1970charge}, which transfer electrical charges pixel by pixel across the sensor to a shared ADC (analog-to-digital converter). This ADC iteratively compares an input continuous voltage to a set of known reference voltages values to convert it into a binary value -- e.g., input voltage 2.51V $\approx$ \textbf{1} * (2V) + \textbf{0} * (1V) + \textbf{1} * (0.5V) = \textbf{101}. While most modern sensors now use CMOS (complementary metal-oxide-semiconductor) technology~\cite{fossum1997cmos}, which can be mass-produced using processes similar to how we manufacture microprocessors, they behave functionally very similar to CCDs\footnote{The key difference is that CMOS sensors have one ADC per \textit{row} of pixels, allowing rows to independently and asynchronously collect photons and read out signals}.

\begin{figure}[t!]
  \centering
  \includegraphics[width=\linewidth]{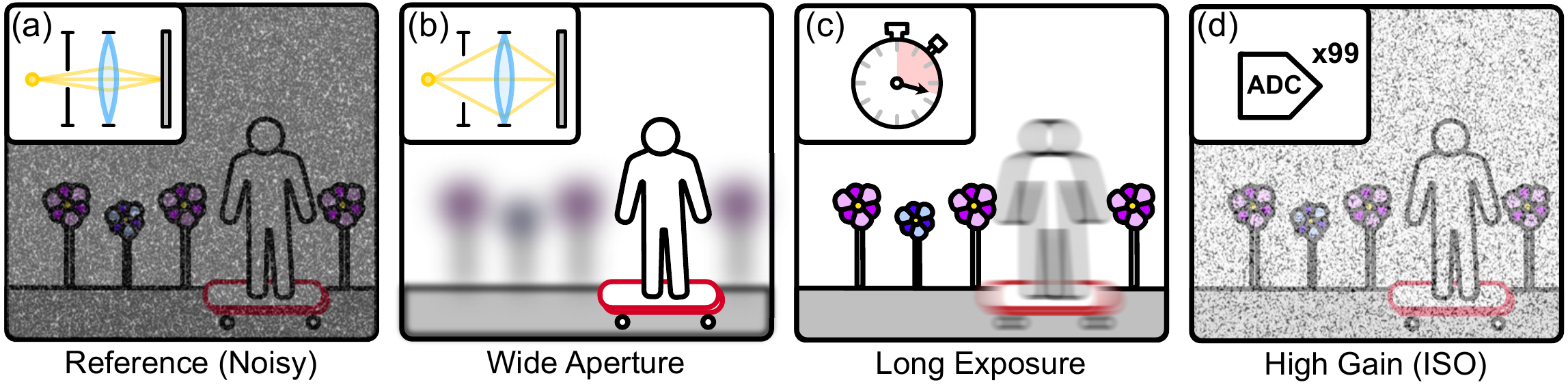}
  \caption{Comparison of ways to increase light capture in low-light imaging and their trade-offs. (a) Reference image with significant noise due to insufficient light, (b) wide aperture capture with blurred background due to reduced depth of field, (c) long exposure capture with motion blur, (d) high gain image with amplified noise.}
  \label{fig:light-levels}
\end{figure}

Unfortunately, these sensors are not perfect measurement devices, and the light they're measuring is itself not a perfect continuous function but rather a collection of randomly distributed photon packets. 
These two main sources of noise are often referred to as \textit{read} noise, random errors from sensor electronics modeled as Gaussian distribution, and \textit{shot}\footnote{The term ``shot" here is in reference to particles scattering like ``shots" aimed at a target.} noise, arising from the random arrival of photon packets, modeled as Poisson distribution:
\begin{equation}
\vspace{-0.5em}
\text{Total Noise} = \mathcal{N}_{\text{read}}(\mu_{\text{read}}, \sigma_{\text{read}}^2) + \mathcal{P}_{\text{shot}}(\mu_{\text{shot}}) + \textit{\small Other Noise Sources}
\vspace{-0.5em}
\end{equation}
Where \textit{Other Noise Sources} include factors like thermal noise (heat-generated electrons in the sensor) or quantization noise (imprecise analog-to-digital conversion). While sensor manufacturers do their best to minimize these noise sources, in low-light settings with limited signal  -- low signal-to-noise ratio (SNR) -- we will inevitably end up with a low quality measurement, Fig.~\ref{fig:light-levels} (a). We can boost the signal in three ways, each has its own tradeoffs. In Fig.~\ref{fig:light-levels} (b) we expand the aperture of the camera, while this lets more light through the lens and onto the sensor, it also makes our system behave less like a pinhole  -- which keeps light rays aligned in a tight beam -- and instead lets this light diverge and blurs the background out-of-focus portions of the scene. In Fig.~\ref{fig:light-levels} (c), we lengthen the exposure time -- the duration for which the sensor collects photons -- to collect more light and brighten the image. The tradeoff here is that if there is motion in the scene, for example the figure on the skateboard moving from left to right, they'll be smudged in the image as their reflected photons arrive at different pixels over time. Lastly, in Fig.~\ref{fig:light-levels} (d), we raise the sensor's gain -- commonly referred to as the camera's \textit{ISO} -- essentially multiplying the signal by some factor $N$. There's actually no \textit{imaging} tradeoff here\footnote{Except for potentially clipping bright values, as we'll see in the following section.}, the picture will appear just as sharp but brighter. The catch is that this factor $N$ will \textit{also} multiply the noise, as it's applied \textit{after} the signal is read off the sensor, meaning this does not change the overall SNR of the system. This highlights one of the major challenges in computational photography, particularly for downstream vision and graphics tasks which seek to maximize reconstruction quality. Achieving sharp, all-in-focus images with minimal motion blur and minimal noise is a balancing act -- one often requiring computational solutions and image post-processing -- as these goals are fundamentally at odds with one another.

\begin{figure}[b!]
  \centering
  \includegraphics[width=\linewidth]{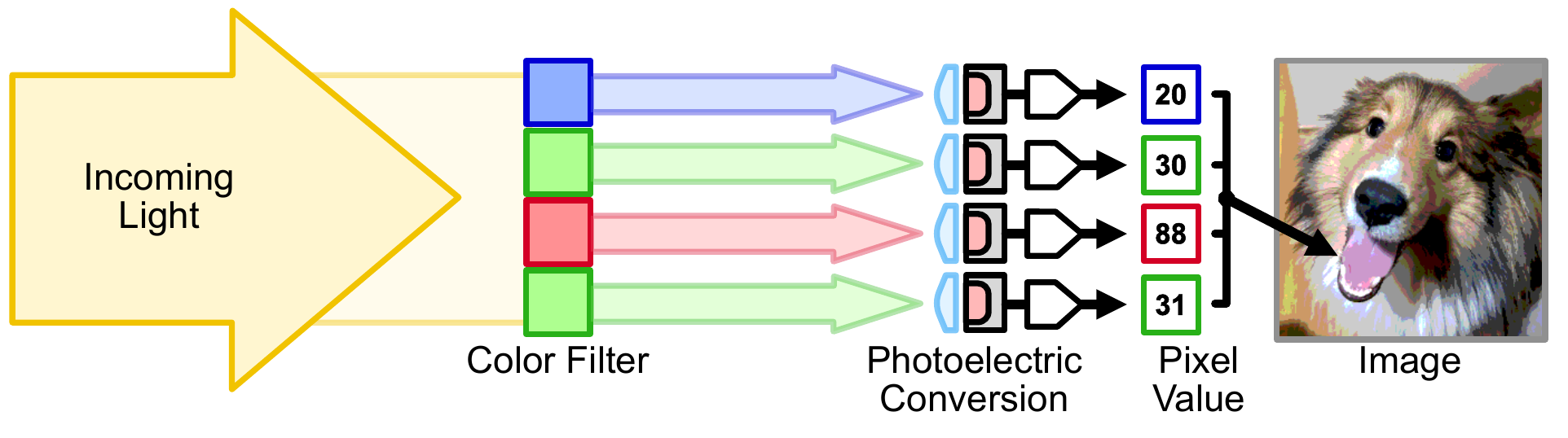}
  \caption{Illustration of color image formation in a digital camera. Incoming light passes through color filters, separating it into red, green, and blue components. Each color channel undergoes photoelectric conversion in the sensor, translating light intensity into digital signals which are combined to form the RGB pixel value in the final image.}
  \label{fig:color-filter}
\end{figure}	

\begin{figure}[t!]
    \centering
    \begin{minipage}{0.5\textwidth}
        \includegraphics[width=\linewidth]{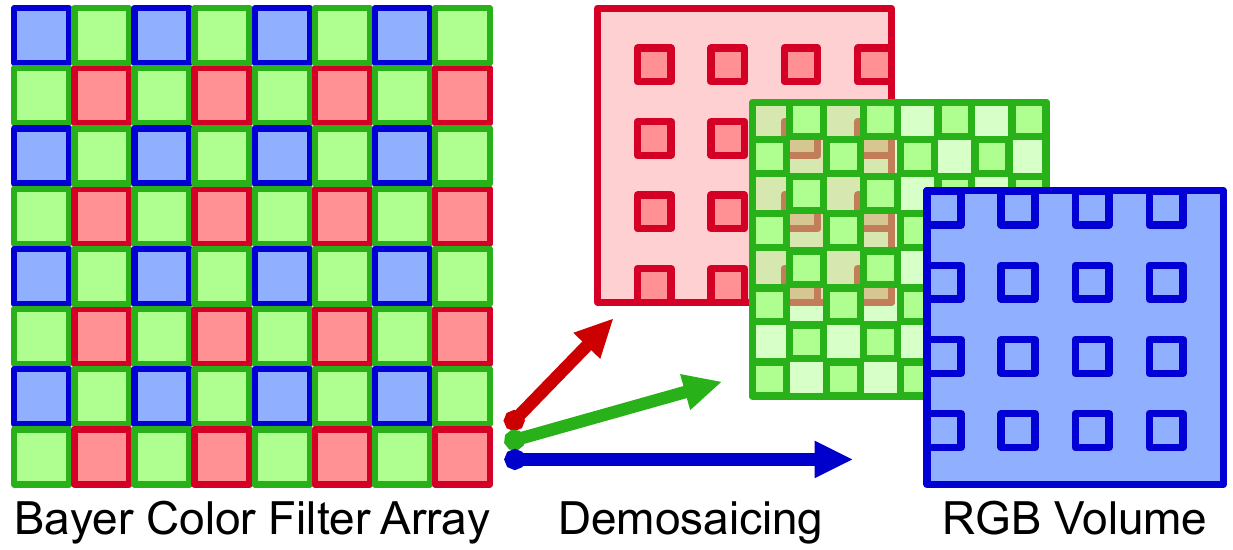}
    \end{minipage}%
    \hfill
    \begin{minipage}{0.45\textwidth}
        \vspace{1em} 
        \caption{As each pixel in the Bayer CFA captures only one color (red, green, or blue), the data must be \textit{demosaiced} to reconstruct a full RGB volume, with missing color values interpolated at each pixel location.}
        \label{fig:demosaicing}
    \end{minipage}
    \vspace*{-1em}
\end{figure}

Lastly, there is the topic of color imaging. Unfortunately, fabricating color-sensitive photodiodes -- ones capable of differentiating wavelengths of light through the photoelectric process -- proves extremely challenging~\cite{yoon2022color}. Instead camera sensors implement color imaging by placing a color filters on top of the photodiodes, filtering out all light except a specific color before measuring whatever photons remain with a regular photodiode. These color filters are typically arranged in a 
Bayer~\cite{bayer1976color} Color Filter Array (CFA), as shown in Fig.~\ref{fig:demosaicing}, with each 2-by-2 square of the pixel grid containing one blue, one red, and two green filters\footnote{This design reflects the observation that human retinal cells are most sensitive to green light, making it the primary contributor to perceived image brightness~\cite{stockman1999spectral}.}. These individual values can then be combined and rendered on the screen as color images, after some \textit{image processing}.

\subsection{Image Processing\label{ch:imageprocessing}}
In a modern mobile imaging device, there can be more than a dozens steps that need to be done -- typically by a dedicated piece of hardware called an Image Signal Processor (ISP) -- before an image can go from the sensor to the screen. Illustrated in Fig.~\ref{fig:demosaicing}, one of these steps is \textit{demosaicing}, separating the interleaved color data in the Bayer array into three separate RGB arrays. As this leaves gaps between the measurements -- e.g., we don't know how much blue light would have been detected at a location with a red filter -- numerous demosaicing algorithms have been developed to recover the missing color information~\cite{hirakawa2005adaptive, malvar2004vng}, with a common simple approach being direct interpolation~\cite{duchon1979lanczos}. \\

 \begin{wrapfigure}{r}{0.3\textwidth}
    \hspace*{0.5em} 
    \begin{minipage}{0.3\textwidth} 
        \includegraphics[width=\linewidth]{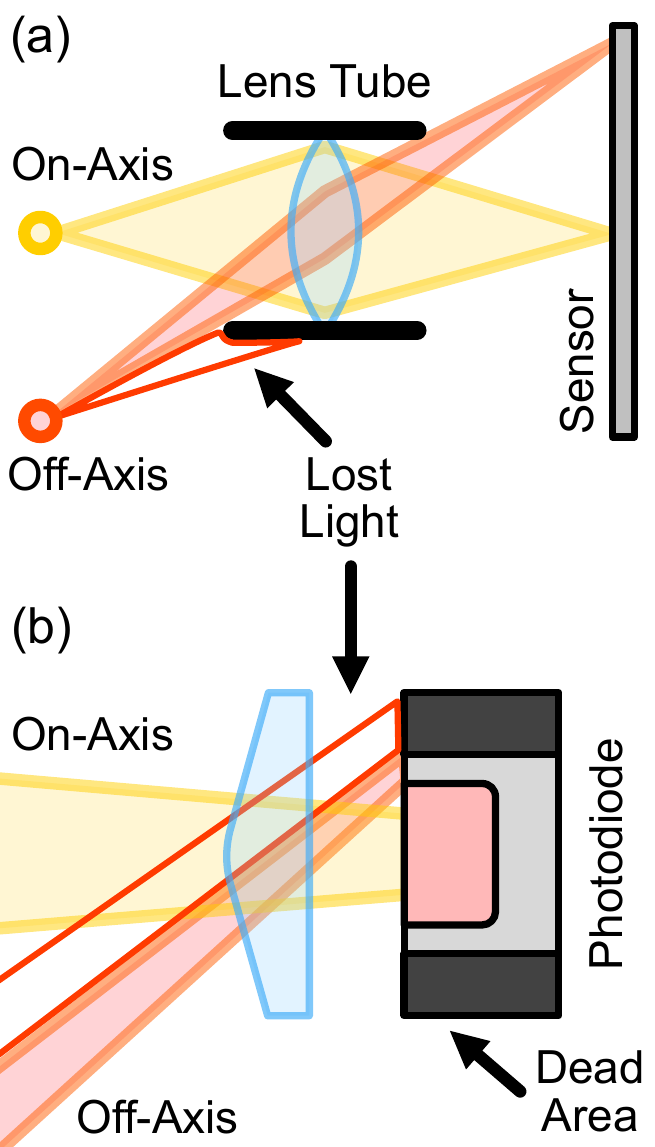} 
        \caption{Lens shading can occur as (a) off-axis light rays are partially blocked by the lens tube, resulting in lost light before reaching the sensor or (b) rays fall into dead areas of the sensor, not recorded by the photodiode.}
        \label{fig:lens-shading}
    \end{minipage}
\end{wrapfigure}

\textit{Lens shading} is another issue requiring post-capture correction, where the edges of the sensor array receive less light than the center, resulting in visible vignetting in the image. Fig.~\ref{fig:lens-shading} depicts two potential causes of this. In Fig.~\ref{fig:lens-shading} (a) part of the off-axis light is blocked by the physical structure of the camera, hitting the side of the device instead of entering the lens. In Fig.~\ref{fig:lens-shading} (b) the light loss occurs because a part of the light beam -- which does not focused to an infinitely small point -- falls off the active area of the photodiode onto the circuitry around it which does not detect light. There exist other optical and physical causes of lens shading, such as the dispersion of light passing through thicker sections of the lens material, but these phenomena are notably under-explored in academic literature\footnote{Much of my understanding here stems from collaboration with mobile electronics companies.}. Yet, particularly for high-fidelity 3D reconstruction, this is an important effect to calibrate for, as otherwise pixels representing the same color in real life may appear to have varying and incompatible colors or shading in the captured images.

\begin{figure}[t!]
  \centering
  \includegraphics[width=0.8\linewidth]{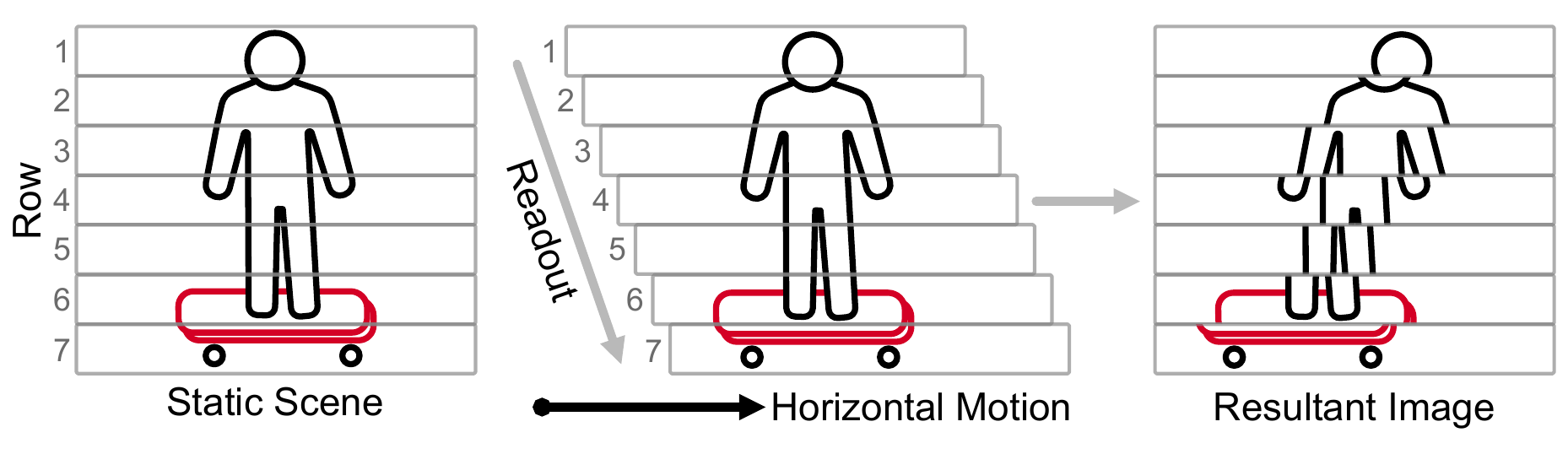}
  \caption{Rolling shutter effect caused by row-wise image readout. In a static scene, rows are read sequentially without distortion. However, when there is horizontal motion during readout, the object’s position shifts between rows, resulting in a skewed image.}
  \label{fig:rolling-shutter}
\end{figure}

\textit{Rolling shutter} is a phenomenon caused by the way CMOS sensors read out data. Instead of capturing the entire scene at once, the sensor records data row by row, resulting in the final image being equivalent to thousands of strips of individually captured images stitched together. When the scene and the camera remain relatively static, this is not an issue, as the time between rows being recorded is typically on the order of microseconds and very little pixel misalignment occurs. However, if there is significant motion, the final image can appear skewed -- as illustrated in  Fig.~\ref{fig:rolling-shutter} -- when rows captured at different, incompatible times are naively stacked together. Unfortunately, this effect cannot be easily calibrated out, as we would have to jointly estimate the camera and world's motion to map pixels back to where they should have been observed at the beginning of the capture. Instead, the inevitable effects of rolling shutter are often embedded directly into 3D reconstruction models as offsets to where the pixels are projected in space~\cite{im2015high, hedborg2012rolling}, or even taken advantage of to distill additional information about the scene~\cite{antipa2019video, gu2010coded}.

\begin{figure}[t!]
  \centering
  \includegraphics[width=1\linewidth]{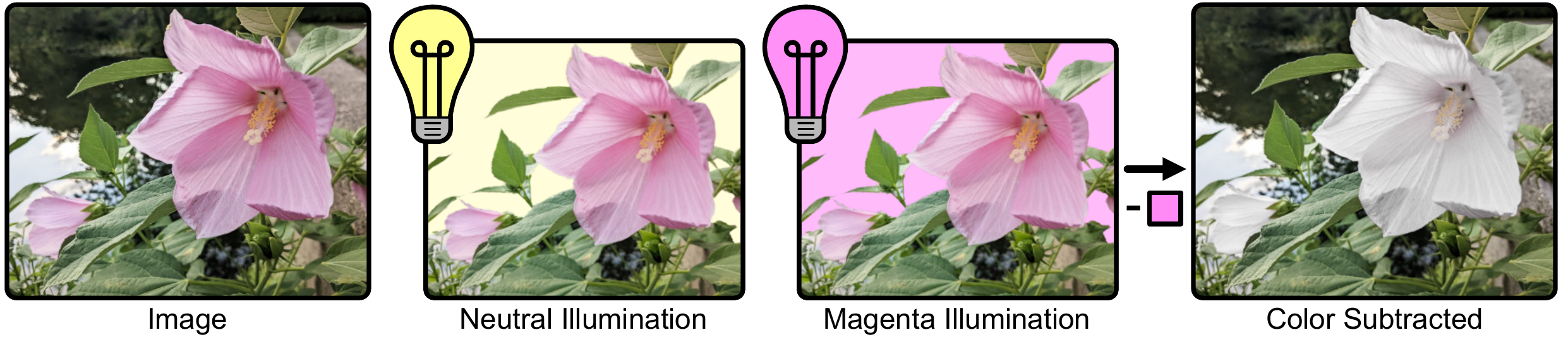}
  \caption{The image of an pink-colored flower could plausibly correspond to either reality of a pink flower under neutral illumination of white flower under magenta illumination. }
  \label{fig:color-correction}
\end{figure}

\textit{Color correction} is another seemingly simple problem that proves extremely complex to calibrate. The photons our sensor captures are the product of the physical properties of scene --  surface colors, absorption, reflection, geometry, and texture (e.g., shiny or rough) -- and the properties of the light entering the scene, its color and distribution (e.g., diffuse or directional). Abstracting away most of these properties, we can fundamentally think of our image as a visualization of $I(\lambda)$, the intensity of light at wavelength $\lambda$, where 
\begin{equation}
    I(\lambda) = E(\lambda) R(\lambda) = (E(\lambda) - \delta_1)(R(\lambda) + \delta_2), \quad \delta_2 = \frac{\delta_1 R(\lambda)}{E(\lambda)- \delta_1}).
\end{equation}
Here $E(\lambda)$ is the power of the illumination at wavelength $\lambda$ and $R(\lambda)$ is the amount of light the surface reflects at that wavelength. It follows that we can come up with any number of offsets $\delta_1$ and $\delta_2$ to the illumination and reflectance to keep this equation balanced. Meaning that, without some external way to measure the illumination or material properties of the scene, there is no way to know the \textit{true} color of contents on an image. An example\footnote{Another example that went viral on the internet in 2015 was ``$\char"0023$TheDress"~\cite{BRAINARD2015R551}, where viewers could not agree whether a dress in an image was blue or gold in real life. } of this problem is depicted in Fig.~\ref{fig:color-correction}, where we cannot distinguish if the image is of a pink flower under neutral illumination or a white flower under magenta illumination. Given this ambiguity, methods rely on assumptions about illumination and reflectance~\cite{buchsbaum1980spatial}, statistical priors~\cite{van2007edge}, or machine learning models trained on paired image datasets~\cite{bianco2015color} to correct observed scene colors.

\begin{figure}[t!]
  \centering
  \includegraphics[width=1\linewidth]{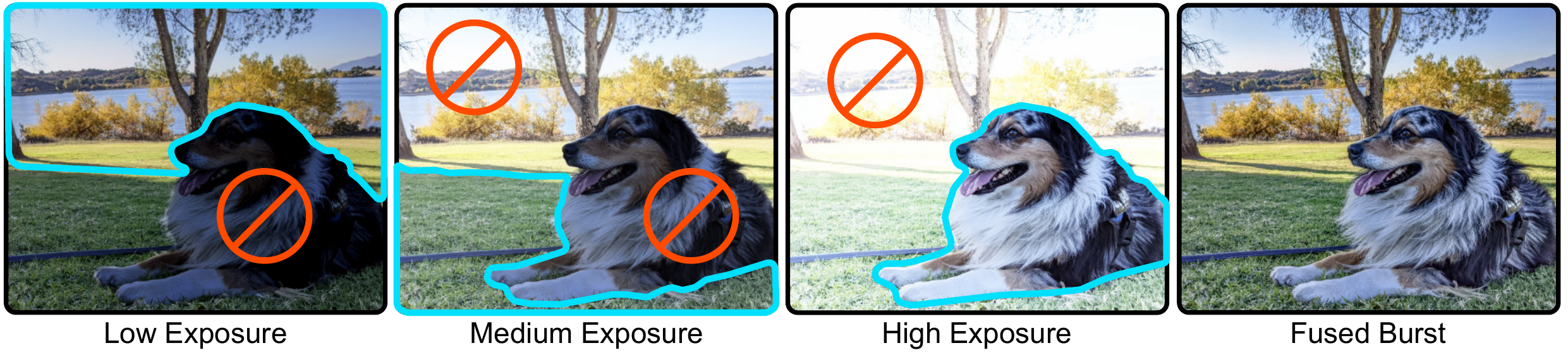}
  \caption{ Illustration of burst imaging for high dynamic range (HDR) reconstruction. A low exposure image can capture highlights but loses detail in the shadows, medium exposure similarly captures midtones, and high exposure reveals shadows. By fusing regions from all three exposures we can preserve details across the full dynamic range of the image. }
  \label{fig:burst-imaging}
\end{figure}

Often, a single image does not fully capture everything we want from a scene, and here we turn to \textit{burst imaging} to recover additional information. We capture a burst -- a rapid succession --  of frames, potentially with different exposure and camera settings, and computationally fuse them into a higher-fidelity reconstruction. 
In Fig.~\ref{fig:burst-imaging} we see a common example of burst imaging for high-dynamic range (HDR) image reconstruction. Digital image sensors have a strict physical limits on the minimum and maximum amount of light they can effectively measure. So for scenes with both bright and dark content -- e.g., direct sunlight and shade --  only a portion of the image might be properly exposed for any choice of camera settings. By capturing a sequence of different exposure settings -- ``bracketing" -- we can carve out the well exposed regions of each image and computationally combine them with a burst fusion pipeline~\cite{mertens2009exposure} to create an image that recovers all the observed details in the scene. In fact, it's very likely that with the wide deployment of mobile HDR photography applications, low-light and ``night-mode" settings~\cite{hasinoff2016burst}, and burst telephoto and super-resolution~\cite{lecouat2022high} methods, the majority of all digital photography is now created by burst fusion pipelines. The research I present in the following chapters focuses on an extension of this burst imaging setting to longer sequences of 40+ recorded frames. As more time passes between frames, scene content can move and illumination can change. Since these sequences also produce a significant amount of data -- when each frame has 12 million pixels, a two-second recording is nearly a gigabyte of raw data -- my work focuses on developing compressive representations capable of interpreting and reconstructing this large amount of potentially time-varying image data.

\subsection{3D Projection\label{ch:3dprojection}}
\begin{figure}[t!]
  \centering
  \includegraphics[width=0.9\linewidth]{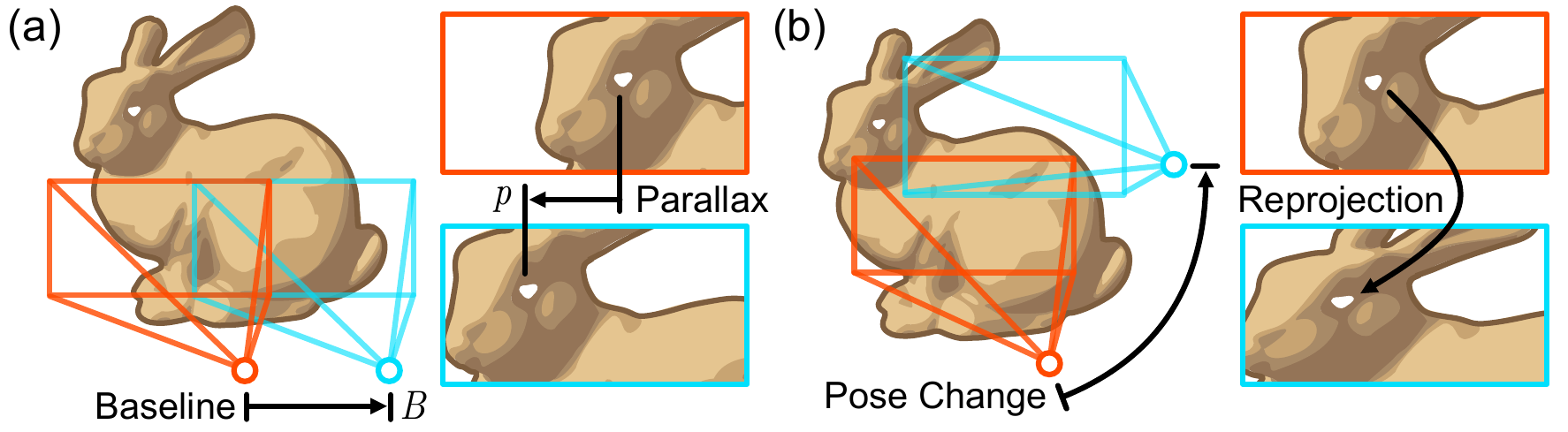}
  \caption{(a) In the simplest case, where the camera undergoes only horizontal motion, the scene appears to exhibit parallax as points shift left or right based on their depth. (b) For a more general change in camera pose, points may experience stretching, skewing, or rotation as their relationships are governed by a more complex set of 3D transformations.}
  \label{fig:stereo-bunny}
\end{figure}

To generate a consistent reconstruction of a scene from a sequence of images with changing viewpoints, we have to solve a pixel correspondence problem -- matching which pixels in the image correspond to the same point in 3D space. In the simplest case, Fig.~\ref{fig:stereo-bunny} (a), where nothing in the scene is moving on its own and the camera undergoes only horizontal motion, the relationship between a points location in the image and the cameras translation in space is
\begin{equation}
	d = \frac{fB}{z}, \quad\quad z = \frac{fB}{d}
\end{equation} 
Here $d$ is the \textit{disparity}, the number of pixels the point appears to move left when the camera moves a $B$ baseline distance right, $z$ is the depth of the point from the camera, and $f$ is the focal length of the camera. From here we can immediately gather that the amount a point moves between images is proportional to how far we move the camera, and inversely proportional to how far away that point is from the camera. Flipping the equation, this means that if we can match two points in the image and calculate their disparity, and know the camera's focal length and baseline, we can estimate where exactly that point is in 3D space. This is how stereo depth estimation methods operate: identifying corresponding feature points -- either through engineered~\cite{sinha2007multi} or learned~\cite{tankovich2021hitnet} approaches -- between pairs of images to compute their disparity.

If the camera undergoes more than just horizontal translation between images, as shown in Fig.~\ref{fig:stereo-bunny} (b), the point locations are no longer represented by simple parallax but instead are described by a set of 3D reprojections.
\begin{figure}[b!]
    \centering
    \begin{minipage}{0.6\textwidth}
        \includegraphics[width=\linewidth]{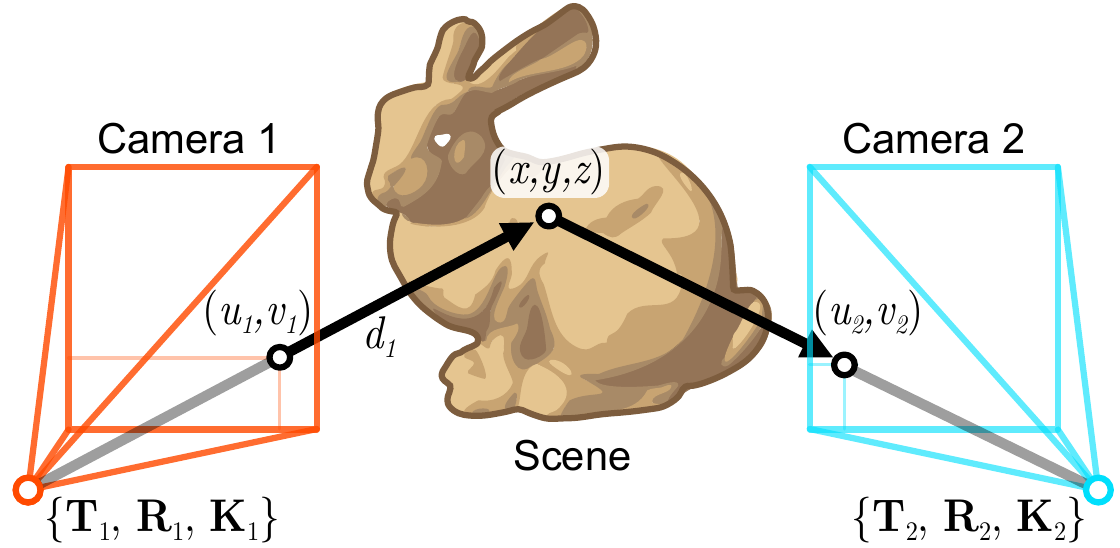}
    \end{minipage}%
    \hfill
    \begin{minipage}{0.35\textwidth}
        \vspace{2em} 
        \caption{Illustration of 3D point reprojection between two camera views. A point $(u_1, v_1)$ from camera corresponds to a point $(x,y,z)$ in 3D space, that projects onto point $(u_2,v_2)$ in the second camera view.}
    \label{fig:3d-reprojection}        
    \label{fig:demosaicing}
    \end{minipage}
\end{figure}
 Given a camera with translation $\mathbf{T}_1$, rotation $\mathbf{R}_1$, and intrinsics $\mathbf{K}_1$ where
\begin{equation}
\mathbf{T} = 
\begin{bmatrix}
    t_x \\
    t_y \\
    t_z
\end{bmatrix}, \quad
    \mathbf{R} = 
    \begin{bmatrix}
        r_{11} & r_{12} & r_{13} \\
        r_{21} & r_{22} & r_{23} \\
        r_{31} & r_{32} & r_{33}
    \end{bmatrix}, \quad
    \mathbf{K} = 
    \begin{bmatrix}
        f_x & 0 & c_x \\
        0 & f_y & c_y \\
        0 & 0 & 1
    \end{bmatrix}.
\end{equation}
$\mathbf{T}$ and $\mathbf{R}$ are transformation matrices that describe the cameras position in 3D space, and $\mathbf{K}$ contains the cameras focal length\footnote{If a lens is not perfectly spherically symmetric, it may have different focal lengths along the vertical and horizontal axes.} parameters $(f_x, f_y)$ and center point $(c_x, c_y)$. Starting with a point with image coordinates $(u_1,v_1)$ that is a depth $d_1$ from the camera, we can project this to a second camera with parameters $\left\{\mathbf{T}_2, \mathbf{R}_2, \mathbf{K}_2\right\}$ via
\begin{equation}
\begin{bmatrix} 
    \,\,u_2\,\, \\ 
    v_2 \\ 
    1 
\end{bmatrix} 
= \mathbf{K}_2 
\mathbf{R}_2 
\begin{bmatrix} 
    \,\,x\,\,\, \\ 
    y \\ 
    z 
\end{bmatrix} 
+ \mathbf{T}_2, \quad
\begin{bmatrix} 
    \,\,x \,\,\,\\ 
    y \\ 
    z 
\end{bmatrix} = 
\mathbf{R}_2^{-1} 
\left( 
    \frac{1}{d_1} \mathbf{K}_1^{-1} 
    \begin{bmatrix} 
        \,\,u_1\,\, \\ 
        v_1 \\ 
        1 
    \end{bmatrix} 
    - \mathbf{T}_1 
\right).
\end{equation}
\label{eq:reprojection}
Here the operation to go from image coordinate $(u_1,v_1)$ to a 3D location $(x,y,z)$ is referred to as \textit{backward projection}, and to from $(x,y,z)$ to $(u_2,v_2)$ is \textit{forward projection}. One thing to note in the process are that we \textit{must} know the depth $d_1$ of an image point in order to project it to a point 3D space, as otherwise $(x,y,z)$ could be anywhere along the line connecting the camera center to $(u_1,v_1)$ -- its \textit{epipolar} line\footnote{I highly recommend \textit{Multiple View Geometry in Computer Vision}~\cite{10.5555/861369} for an overview of the principles of epipolar geometry.}. A second observation is that we \textit{don't} need the depth of the point the view of the second camera to project it to $(u_2,v_2)$; as that information is inherently contained in the relationship between $(x,y,z)$ and the camera's pose. These observations are critical for understanding competing inverse reconstruction approaches -- e.g., Neural Radiance Fields~\cite{mildenhall2021nerf} and Gaussian Splatting~\cite{kerbl20233d}. Fundamentally, backward projection requires finding the intersection of the epipolar line with the scene’s underlying geometry, a process that is can be both computationally and analytically complex. In contrast, forward projection is relatively straightforward when a point's 3D position is known. For this reason, the work in Chapter \ref{ch:depth}, which focuses on fine geometry estimation, leverages forward projection from an optimized 3D surface. In contrast, the methods in Chapters \ref{ch:layer} and \ref{ch:pan} which rely on backward projection, avoid complex geometrical representations to accelerate reconstruction.

\begin{figure}[t!]
  \centering
  \includegraphics[width=0.9\linewidth]{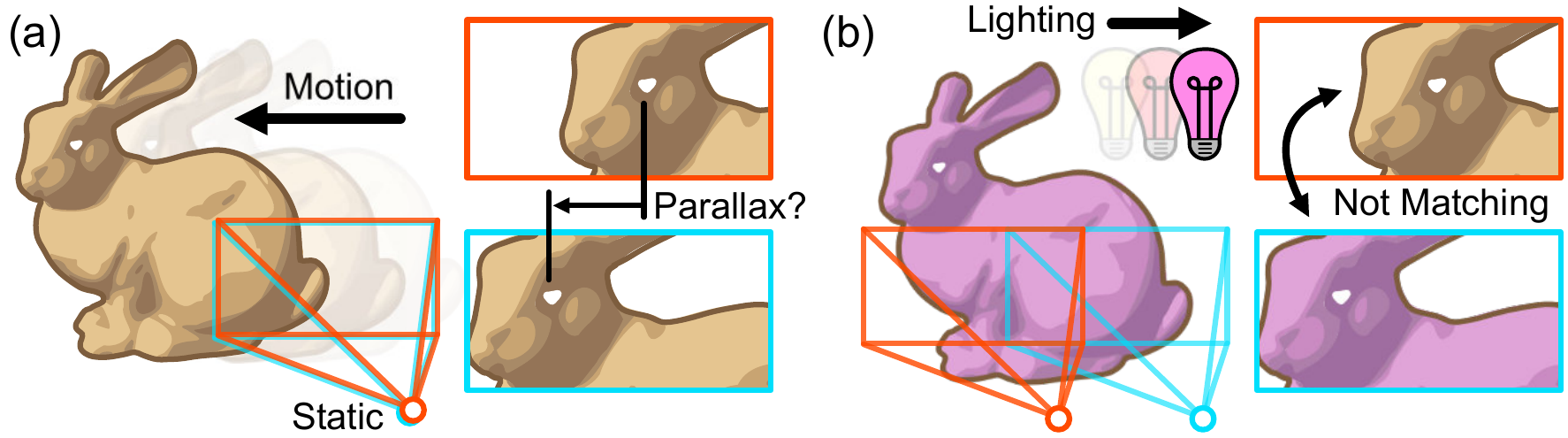}
  \caption{Scene reconstruction challenges. (a) Even if the camera remains static, scene motion -- the rabbit moving from right to left -- can create the illusion of depth parallax. (b) If the illumination in the scene changes between images, it may no longer be possible to match feature points based on color alone.}
  \label{fig:stereo-failure}
\end{figure}

Depicted in Fig.~\ref{fig:stereo-failure} are two challenges for scene reconstruction posed by real-world conditions, which we explore solutions for in Chapter \ref{ch:pan}. In Fig.~\ref{fig:stereo-failure} (a) we see that if the scene itself exhibits motion, this can be difficult to disambiguate from depth parallax. In fact, just as we saw in Sec.~\ref{ch:imageprocessing} that it is impossible to disambiguate lighting from reflectance without additional information, it can similarly be impossible to differentiate camera from world motion from images alone -- e.g., if the view is fully taken up by a the side of truck we can not differentiate the truck moving left from our camera moving right. In Fig.~\ref{fig:stereo-failure} (b) we observe that if the illumination in the scene changes between images -- e.g., the lighting shifts from neutral to magenta -- features in the images may no longer visually match. This makes it challenging to determine whether points in the two images correspond to the same location or structure in the 3D scene. Similar to the case for color correction, robust reconstruction approaches must rely on assumptions about the scene's structure and motion~\cite{akhter2008nonrigid} or learned priors~\cite{sarlin2020superglue} to differentiate scene changes from true underlying geometry.	

\subsection{Neural Fields\label{ch:neuralfields}}
\begin{figure}[t!]
  \centering
  \includegraphics[width=\linewidth]{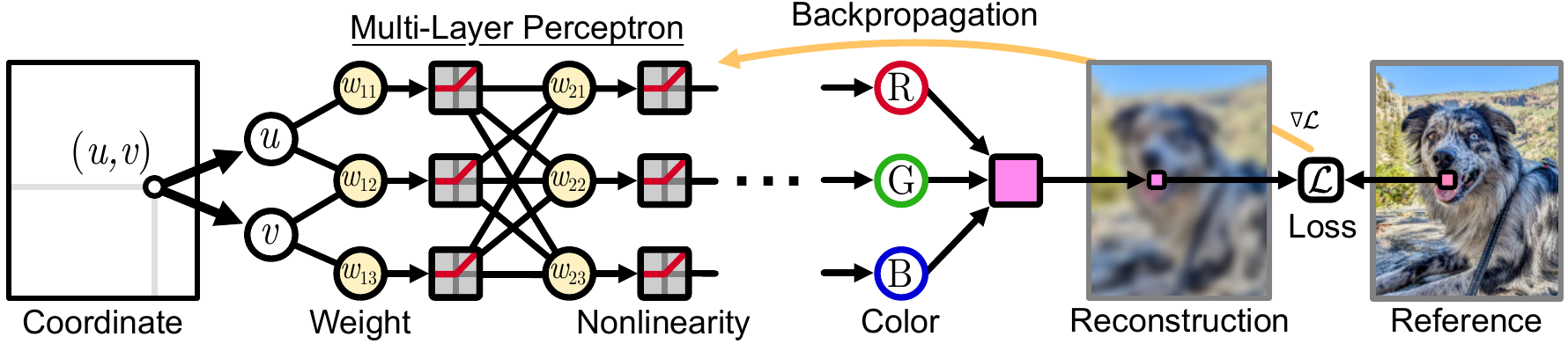}
  \caption{Overview of a neural field model of a color image. An image coordinate $(u,v)$ is input into a multi-layer perceptron (MLP. The output, computed through weight layers and nonlinear activation functions, is interpreted as an RGB color and compared with the reference image. The resulting loss is backpropagated to update the network weights.   }
  \label{fig:neural-field}
\end{figure}

A \textit{neural field} -- or \textit{coordinate network} -- is a continuous function parametrized in the weights of a neural network model which is optimized to map spatial coordinates to some desired output. The network is typically a small \textit{multi-layer perceptron} (MLP), with on the order of tens of thousands of parameters in a fully connected architecture composed of several layers, with nonlinear activation functions applied between them. Alternatively stated, an MLP applies a sequence of matrix multiplications and non-linear functions to an input $N$-dimensional coordinate vector until arriving at an $M$-dimensional output. This output is then compared to a reference, allowing a loss to be calculated and backpropagated through the MLP to update its weights, nudging the network to better approximate the reference data. This process is illustrated for an image fitting example in Fig.~\ref{fig:neural-field}. It is important to clarify that, although neural field models can be incorporated into generative frameworks~\cite{schwarz2020graf}, they are inherently \textbf{not generative models}. Unlike models trained to learn a probabilistic distribution over data -- e.g., a UNet trained to map images to object labels -- neural fields are designed to approximate a single dataset or signal. They act as functional approximators~\cite{hornik1989multilayer}, similar to traditional low-rank decompositions like singular value decomposition (SVD)~\cite{hansen2006deblurring} or compressed data formats, where the network's weights serve as a compact encoding of the data like bits stored in a .zip file. Thus neural field representations of data are often referred to as \textit{implicit} representations because they represent the data indirectly through the interaction of internal weights, rather than explicitly storing the data itself. 

One key property of neural field representations is that, as they take spatial coordinates as input, we do not need to ``expand" the representation to evaluate it. Unlike a traditional low-rank representation, where evaluating a specific point typically requires reconstructing the full space by combining basis vectors and coefficients, a neural field can directly compute the output at any given input coordinate without needing to reconstruct the entire function. The downside is that, unlike an explicit representation like an image array, a neural field is also not ``pre-evaluated". To render an image on screen we must first evaluate it at a grid of coordinates  -- a potentially computational intensive task -- and assemble the outputs into an RGB array.
\begin{figure}[t!]
  \centering
  \includegraphics[width=\linewidth]{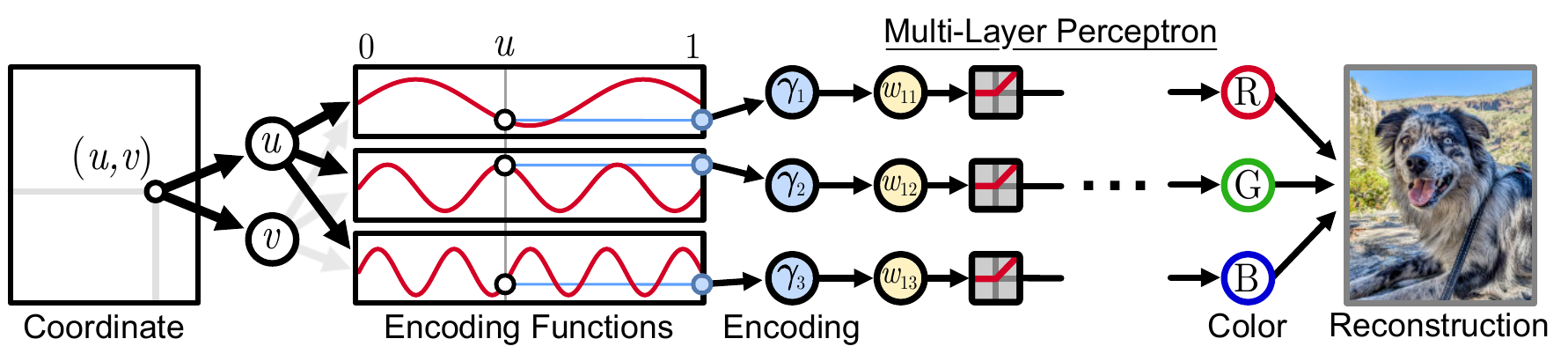}
  \caption{Overview of coordinate encoding. The input spatial coordinate is used to sample a set of encoding functions, and the resulting values are concatenated into a higher-dimensional representation, which is then passed as input to the MLP for optimization.}
  \label{fig:frequency-encoding}
\end{figure}
While neural fields can be trained directly on spatial coordinates -- e.g., a 3-dimensional vector of $(x,y,z)$ points -- they exhibit a problem during training known as \textit{spectral bias}~\cite{tancik2020fourier}. This refers to the network's difficulty to learn high-frequency components of the target function when the inputs -- smoothly varying coordinates -- do not have have high-frequency components. To combat this, most approaches implement \textit{coordinate encoding}, illustrated in Fig.~\ref{fig:frequency-encoding}. Input coordinates are first mapped through some set of functions to a higher dimensional encoding vector -- with higher frequency components --  before passing it into the MLP for optimization. 
\begin{equation}
	\bm{y} = f(\theta, \gamma(\bm{x})), \quad \gamma(\bm{x}) = \left[\bm{x}, \sin(2^0 \pi \bm{x}), \cos(2^0 \pi \bm{x}), \sin(2^1 \pi \bm{x}),  \dots,  \cos(2^{L-1} \pi \bm{x})\right]
\end{equation}\label{eq:encoding}
Eq.~\ref{eq:encoding} demonstrated \textit{Fourier feature encoding}~\cite{tancik2020fourier}, a widely-used form of coordinate encoding where the input coordinate $\bm{x}$ is transformed into a higher-dimensional encoding $\gamma(\bm{x})$ using sine and cosine functions at multiple frequencies. The neural field model $f()$ then applies its internal weights $\theta$ to the encoded coordinate $\gamma(\bm{x})$ to generate the output signal $\bm{y}$. Of note here is that this Fourier feature encoding does have any additional learnable parameters, we are not increasing the size of the model, only transforming its inputs to improve optimization. 
\begin{figure}[t!]
  \centering
  \includegraphics[width=\linewidth]{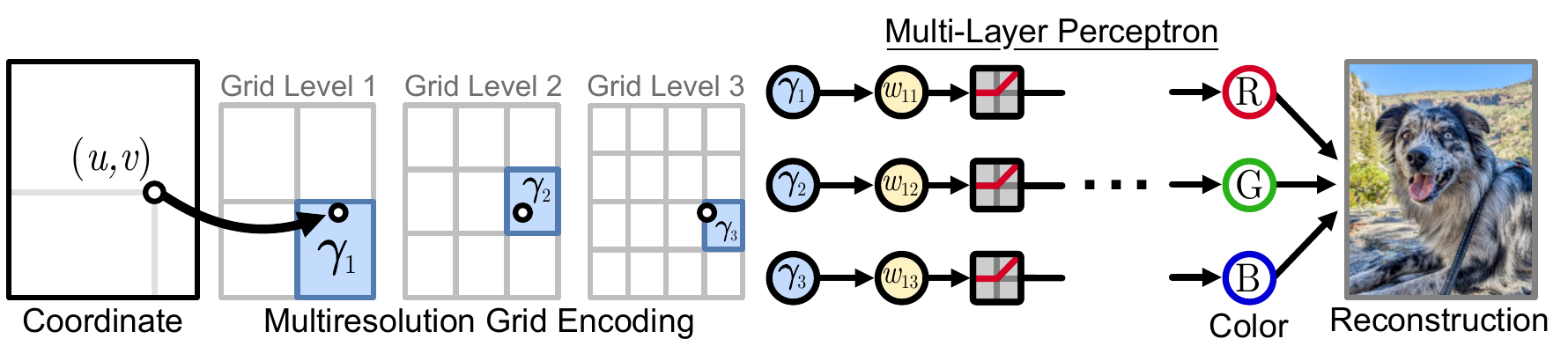}
  \caption{Overview of multiresolution grid encoding. The input spatial coordinate is used to index into multiple grid arrays at different scales, retrieving a feature vector from each level. These feature vectors -- also learned -- are concatenated and input into the MLP.}
  \label{fig:hash-encoding}
\end{figure}

In contrast, the multiresolution grid encoding shown in Fig.~\ref{fig:hash-encoding}, \textit{does} increase the model size. Here, the input spatial coordinate is used to retrieve learnable feature vectors stored in arrays at multiple resolutions -- similar to features in an image pyramid~\cite{adelson1984pyramid} -- which are concatenated and passed into the network. This means in addition to optimizing network weights $\theta$, our neural field model now jointly optimizes learnabole feature weights $\theta_\gamma$ in the encoding. This is referred to as a \textit{hybrid} representation,
\begin{figure}[t!]
    \centering
    \begin{minipage}{0.42\textwidth}
        \includegraphics[width=\linewidth]{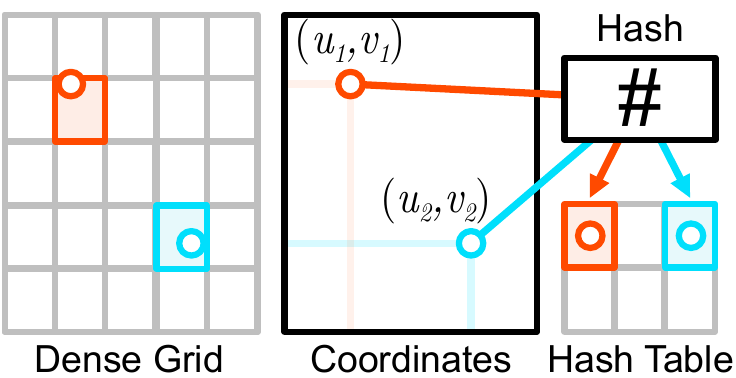}
    \end{minipage}%
    \hfill
    \begin{minipage}{0.55\textwidth}
        \vspace{1em} 
        \caption{Rather than storing values in a dense grid -- where a value is stored for each location in a grid-sized table -- a hash grid uses a hash function to map spatial coordinates to a smaller hash table. In this representation, the position of an entry in the table is no longer directly correlated to the location of the input coordinate.}
    \label{fig:hash-table}
    \end{minipage}
\end{figure}
one where the output function is a combination of both information implicitly encoded in the network weights and explicitly stored in the encoding. This can greatly increase both the speed and final reconstruction quality of neural field models, at the cost of increased memory requirements for storing high-resolution feature grids. To address this, Müller et al.~\cite{muller2022instant} introduced the use of an efficient hash-table backing for these encoding grids for highly accelerated neural field training. Depicted in Fig.~\ref{fig:hash-table}, rather than storing a feature for every location in the grid, this approach first passed the coordinate through a a hash function to map to a random index in a much smaller data table. While this inevitably leads to hash collisions -- where multiple coordinate pairs map to the same index in the table -- these collisions can effectively be resolved by the trained network~\cite{takikawa2023compact}. And as only the smaller hash table needs to be stored, rather than the entire grid, we can often reduce memory usage by orders of magnitude. 

We will see in the next chapters how these neural field representations with multiresolution hash grids encodings can be used to represent and disentangle complex 3D lighting and geometry in mobile computational photography.

\chapter{Micro-Baseline Depth\label{ch:depth}}
This chapter explores the application of neural field models for depth reconstruction from ``micro-baseline" data: image sequences with minimal view variation -- on the order of only several pixels -- between frames. Unlike traditional stereo imaging data which requires either multiple cameras or large user motion to generate image pairs, this micro-baseline data is created naturally every time someone takes a photo with a mobile device. During two seconds of view-finding, a photographer's unstabilized hand is likely to drift on the order of a centimeter in space. Thus we investigate how much geometric information about the world can be recovered from this otherwise discarded data. Given the minimal parallax between images, we must go beyond sparse feature matching and instead track the dense motion of millions of pixels across frames to accurately estimate depth. To achieve this, we perform test-time optimization of a dense neural field, distilling the information from these tiny updates into a compact and interpretable 3D surface representation.

\vspace{1em}
\hrule
\vspace{1em}
\noindent  \textit{This chapter is based on the work ``Shakes on a Plane: Unsupervised Depth Estimation from Unstabilized Photography"~\cite{chugunov2023shakes} by Ilya Chugunov, Yuxuan Zhang, and Felix Heide presented at CVPR 2023.}

\section{Prior Work}
This work directly builds on \textit{``The Implicit Values of A Good Hand Shake: Handheld Multi-Frame Neural Depth Refinement"~\cite{chugunov2022implicit} by Ilya Chugunov, Yuxuan Zhang, Zhihao Xia, Xuaner Zhang, Jiawen Chen, and Felix Heide presented at CVPR 2022.}

In \cite{chugunov2022implicit}, we explore the same problem space: micro-baseline depth reconstruction from images collected during photo view-finding. However, we rely on initial depth measurements and estimated camera position provided by the iPhone 12 Pro, which features an on-board LiDAR-based depth camera not available to other cell phone manufacturers. While this work established the foundations of the problem space -- e.g., the distribution of motion experienced during view-finding, which we include for reference in Sec.~\ref{sechndr:hand_vis} -- it relies on strong assumptions about the data provided by the mobile device for initialization and requires a complex forward-backward projection model to utilize it effectively. In this follow-up work, we drop both many of these requirements -- the proposed method can operate on images alone with no other sensor information -- and develop a much more compact forward projection model with improved 3D reconstruction quality.

\graphicspath{{chapters/1/SoaP/}}
 \vspace{-1em}
\section{Introduction}
Over the last century we saw not only the rise and fall in popularity of film and DSLR photography, but of standalone cameras themselves. We've moved into an era of ubiquitous multi-sensor, multi-core, multi-use, mobile-imaging platforms~\cite{delbracio2021mobile}. Modern cellphones offer double-digit megapixel image streams at high framerates; optical image stabilization; on-board motion measurement devices such as accelerometers, gyroscopes, and magnetometers; and, most recently, integrated active depth sensors~\cite{luetzenburg2021evaluation}. This latest addition speaks to a parallel boom in the field of depth imaging and 3D reconstruction~\cite{han2019image,zollhofer2018state}. As users often photograph people, plants, food items, and other complex 3D shapes, depth can play a key role in object understanding tasks such as detection, segmentation, and tracking~\cite{ji2021calibrated,schwarz2018rgb,yang2022RGBD}. 3D information can also help compensate for non-ideal camera hardware and imaging settings through scene relighting~\cite{pandey2021total,guo2019relightables,yang2021s3net}, simulated depth-of-field effects~\cite{wadhwa2018synthetic,abuolaim2020defocus, wang2018deeplens}, and frame interpolation~\cite{bao2019depth}. Beyond helping improve or understand RGB content, depth itself is a valuable output for simulating objects in augmented reality~\cite{serrano2019motion,bertel2020omniphotos,luo2020consistent,du2020depthlab} and interactive experiences~\cite{hedman2018instant,kopf2020one}.

\begin{figure}[t]
    \centering
    \includegraphics[width=0.75\linewidth]{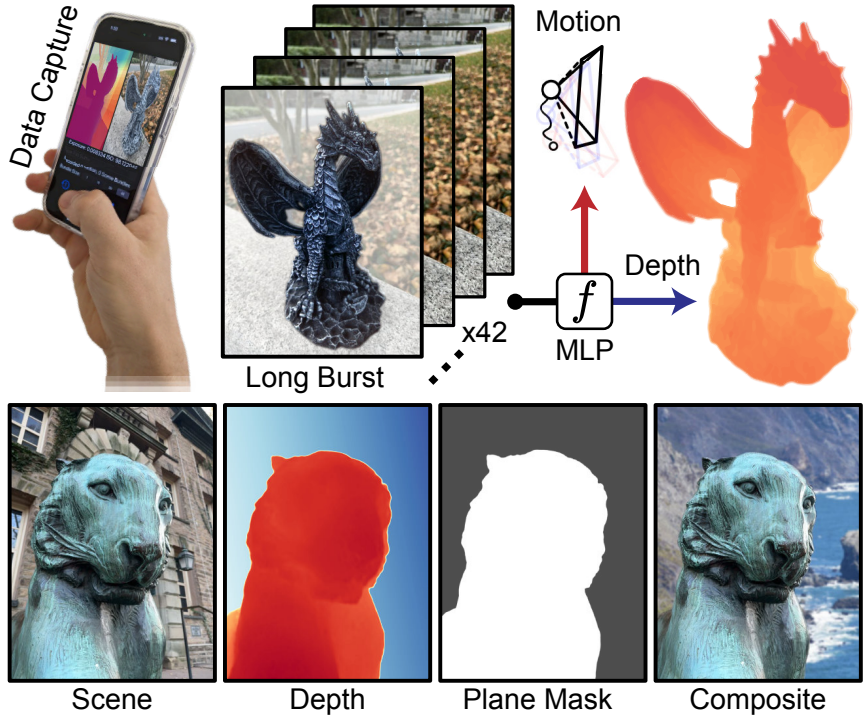}
    \caption{Our neural RGB-D model fits to a single \emph{long-burst} image stack to distill high quality depth and camera motion. The model's \emph{depth-on-a-plane} decomposition can facilitate easy background masking, segmentation, and image compositing.}
    \label{figsoap:my_label}
\end{figure}

Depth reconstruction can be broadly divided into \emph{passive} and \emph{active} approaches. \emph{Passive} monocular depth estimation methods leverage training data to learn shape priors~\cite{ranftl2021vision, hu2019revisiting, bhat2021adabins} -- e.g., what image features imply curved versus flat objects or occluding versus occluded structures -- but have a hard time generalizing to out-of-distribution scenes~\cite{ranftl2020towards,miangoleh2021boosting}. Multi-view depth estimation methods lower this dependence on learned priors by leveraging parallax information from camera motion~\cite{fonder2021m4depth, ummenhofer2017demon} or multiple cameras~\cite{tankovich2021hitnet, marr1976cooperative} to recover geometrically-guided depth. The recent explosion in neural radiance field approaches~\cite{mildenhall2020nerf, yu2021pixelnerf, tancik2022block, mildenhall2022nerf} can be seen a branch of multi-view stereo where a system of explicit geometric constraints is swapped for a more general learned scene model. Rather than classic feature extraction and matching, these models are fit directly to image data to distill dense \textit{implicit} 3D information.

 \emph{Active} depth methods such as pulsed time-of-flight~\cite{mccarthy2009long} (e.g., LiDAR), correlation time-of-flight~\cite{lange20003d}, and structured light~\cite{zhang2018high, scharstein2003high} use \emph{controlled illumination} to help with depth reconstruction. While these methods are less reliant on image content than \emph{passive} ones, they also come with complex circuitry and increased power demands~\cite{horaud2016overview}. Thus, miniaturization for mobile applications results in very low-resolution \emph{sub-kilopixel sensors}~\cite{hegblom2022column, warren2018low,callenberg2021low}. The Apple iPhone 12-14 Pro devices, which feature one of these miniaturized sensors, use depth derived from RGB, available at 12 \emph{mega-pixel} resolution, to recover scene details lost in the sparse LiDAR measurements. While how exactly they use the RGB stream is unknown, occluding camera sensors reveals that the estimated geometry is the result of \emph{monocular} RGB-guided depth reconstruction.

Returning to the context of mobile imaging, even several seconds of continuous mode photography, which we refer to as a ``long-burst", contain only millimeter-scale view variation from natural hand tremor~\cite{chugunov2022implicit}. While these \emph{micro-baseline}~\cite{joshi2014micro} shifts are effectively used in burst superresolution and denoising methods~\cite{wronski2019handheld, qian2019rethinking} as indirect observations of content between sensor pixels, 3D parallax effects on pixel motion are commonly ignored in these models as the depth recovered from this data is too coarse for sub-pixel refinement~\cite{yu20143d, joshi2014micro,im2015high}. A recent work~\cite{chugunov2022implicit} demonstrates high-quality object reconstructions refined with long-burst RGB data, but relies on the iPhone 12 Pro LiDAR sensor for initial depth estimates and device poses, not available on many other cellphones. They treat these poses as ground truth and explicitly solve for depth through minimization of photometric reprojection loss.

In this work, we devise an unsupervised end-to-end approach to jointly estimate high-quality object depth and camera motion from more easily attainable unstabilized two-second captures of 12-megapixel RAW frames and gyroscope data. Our method requires no depth initialization or pose inputs, only a long-burst. We formulate the problem as an image synthesis task, similar to neural radiance methods~\cite{mildenhall2020nerf}, decomposed into explicit geometric projection through continuous depth and pose models. In contrast to recent neural radiance methods, which typically estimate poses in a pre-processing step, we jointly distill relative depth and pose estimates as a product of simply fitting our model to long-burst data and minimizing photometric loss. In summary, we make the following contributions:
\begin{itemize}
    \item An end-to-end neural RGB-D scene fitting approach that distills high-fidelity affine depth and camera pose estimates from unstabilized long-burst photography.\vspace{-0.25em}
    \item A smartphone data collection application to capture RAW images, camera intrinsics, and gyroscope data for our method, as well as processed RGB frames, low-resolution depth maps, and other camera metadata.    
    \item Evaluations which demonstrate that our approach outperforms existing single and multi-frame image-only depth estimation approaches, with comparisons to high-precision structured light scans to validate the accuracy of our reconstructed object geometries.
\end{itemize}
Code, data, videos, and additional materials are available on our project website: {\color{URLBlue}\href{https://light.princeton.edu/soap/}{https://light.princeton.edu/soap}}

\section{Related Work} 
There exist a wide array of both \emph{active} and \emph{passive} depth estimation methods: ones that recover depth with the help of a controlled illumination source, and ones that use only naturally collected light. We review related work in both categories before discussing neural scene representations.

\vspace{1em}
\noindent\textbf{Active Depth Reconstruction.}\hspace{0.1em} Structured light and active stereo methods rely on patterned illumination to directly infer object shape~\cite{zhang2018high, fanello2016hyperdepth} and/or improve stereo feature matching~\cite{scharstein2003high}. In contrast, time-of-flight (ToF) depth sensors use the round trip time of photons themselves -- how long it takes light to reach and return from an object -- to infer depth. \emph{Indirect} ToF does this by calculating phase changes in continuously modulated light~\cite{lange20003d,hansard2012time,kolb2010time}, while \emph{direct} ToF times how long a pulse of light is in flight to estimate depth~\cite{morimoto2020megapixel,mccarthy2009long}. The LiDAR system found in the iPhone 12-14 Pro devices is a type of direct ToF sensor built on low-cost single-photon detectors~\cite{callenberg2021low} and solid-state vertical-cavity surface-emitting laser technology~\cite{warren2018low}. While active LiDAR depth measurements can help produce \emph{metric} depth estimates -- without scale ambiguity -- existing mobile depth sensors have very limited sub-kilopixel spatial resolution, are sensitive to surface reflectance, and are not commonly found on other mobile devices.

\vspace{1em}
\noindent\textbf{Passive Depth Reconstruction.}\hspace{0.1em} Single-image passive methods leverage the correlation between visual and geometric features to estimate 3D structure. Examples include depth from shading~\cite{barron2014shape, xiong2014shading}, focus cues~\cite{xiong1993depth}, or generic learned priors~\cite{ranftl2021vision, hu2019revisiting, bhat2021adabins}. Learned methods have shown great success in producing visually coherent results, but rely heavily on labeled training data and produce unpredictable outputs for out-of-distribution samples. Multi-view and structure from motion works leverage epipolar geometry~\cite{hartley2003multiple}, the relationship between camera and image motion, to extract 3D information from multiple images. Methods typically either directly match RGB features~\cite{sinha2007multi,galliani2015massively}, or higher-level learned features~\cite{tankovich2021hitnet,lipson2021raft}, in search of depth and/or camera parameters which maximize \emph{photometric consistency} between frames. COLMAP~\cite{schonberger2016structure} is a widely adopted multi-view method which many neural radiance works~\cite{mildenhall2020nerf, mildenhall2022nerf} rely on for camera pose estimates. In the case of long-burst photography, this problem becomes significantly more challenging as many different depth solutions produce identical images under small view variations. Work in this space either relies on interpolation between sparse feature matches~\cite{yu20143d, joshi2014micro,im2015high} or additional hardware~\cite{chugunov2022implicit}  to produce complete depth estimates. Our work builds on these methods to produce both dense depth and high-accuracy camera motion estimates from long-burst image data alone, with a single model trained end-to-end rather than a sequence of disjoint data processing steps.

\vspace{1em}
\noindent\textbf{Neural Scene Representations.}\hspace{0.1em}
Recent work in the area of novel view synthesis has demonstrated that explicit models -- e.g. voxel grids, point clouds, or depth maps -- are not a necessary backbone to generate high-fidelity representations of 3D space. Rather, the neural radiance family of works, including NeRF~\cite{mildenhall2020nerf} and its extensions \cite{chen2021mvsnerf, barron2021mip, muller2022instant}, learn an implicit representation of a 3D scene by fitting a multi-layer perceptron (an MLP)~\cite{hornik1989multilayer} to a set of input images through gradient descent. Similar to multi-view stereo, these methods optimize for photometric loss, ensuring output colors match the underlying RGB data, but they typically don't produce depth maps or camera poses as outputs. On the contrary, most neural radiance methods require camera poses as inputs, obtained from COLMAP~\cite{schonberger2016structure} in a separate pre-processing step. Our setting of long-burst unstabilized photography not only lacks ground truth camera poses, but also provides very little view variation from which to estimate them. While neural scene representation works exist which learn camera poses~\cite{lin2021barf, wang2021nerf}, or operate in the burst photography setting~\cite{pearl2022nan}, to our best knowledge this is the first work to jointly do both. The most similar recent work by Chugunov et al.~\cite{chugunov2022implicit} uses poses derived from the iPhone 12 Pro ARKit library to learn an implicit representation of depth, but \emph{does not have an image generation model}, and is functionally closer to a direct multi-view stereo approach. In contrast, our work uses a neural representation of RGB as an optimization vehicle to distill high quality continuous representations of both depth and camera poses, with loss backpropogated through an explicit 3D projection model.

\section{Long-Burst Photography}
\begin{figure}[t]
    \centering
    \includegraphics[width=0.8\linewidth]{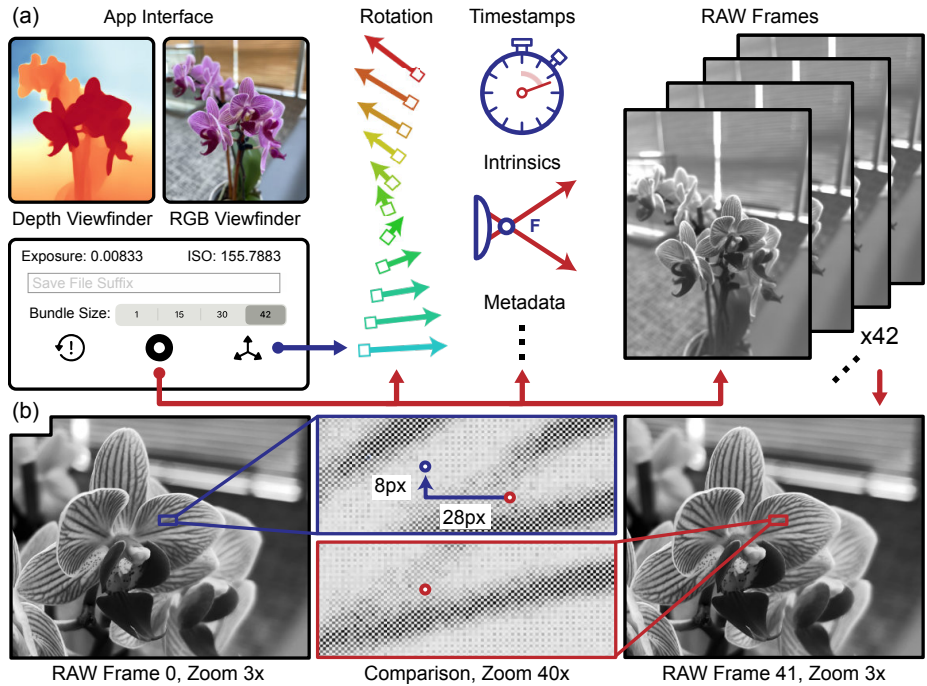}
    \caption{ (a) The interface of our app for recording long-burst data. (b) Aligned RAW frames, which illustrate the scale of parallax motion created by natural hand tremor in a two-second capture: a few dozen pixels for an object 30cm from the camera. }
    \label{figsoap:app}
\end{figure}

\noindent\textbf{Problem Setting.}\hspace{0.1em} Burst photography refers to the imaging setting where for each button press from the user the camera records multiple frames in rapid succession, sometimes varying parameters such as ISO and exposure time during capture to create a \emph{bracketed sequence}~\cite{mertens2009exposure}. Burst imaging pipelines investigate how these frames can be merged back into a single higher-fidelity image~\cite{delbracio2021mobile}. These pipelines \emph{typically operate with 2-8 frame captures} and have proven key to high-quality mobile imaging in low-light~\cite{liba2019handheld, hasinoff2016burst}, high dynamic range imaging with low dynamic range sensors~\cite{hasinoff2016burst,gallo2015locally}, and image superresolution, demosaicing, and denoising~\cite{wronski2019handheld, weissman2005universal}. On the other end of the imaging spectrum we have video processing literature, which operates on sequences hundreds or thousands of frames in length~\cite{monakhova2022dancing} and/or with large camera motion~\cite{kopf2021robust}. Between these two settings we have what we refer to as ``long-burst" photography, several seconds of continuous capture with small view variation. Features built into default mobile camera applications such as Android Motion Photos and Apple Live Photos, which both record three seconds of frames around a button press, demonstrate the ubiquity of \emph{long-burst} data. They are captured spontaneously, without user interaction, during natural handheld photography. In this work we capture two-second long-bursts, which result in 42 recorded frames with an average 6mm maximum effective stereo baseline. As seen in Fig.~\ref{figsoap:app} (b), this produces on the order of several dozen pixels of disparity for close-range objects ($<$0.5m). For an in-depth discussion of motion from natural hand tremor we refer the reader to Chugunov et al.~\cite{chugunov2022implicit}.

\noindent\textbf{Data Collection.}\hspace{0.1em} As there were no commodity mobile applications that allowed for continuous streaming of Bayer RAW frames and metadata, we designed our own data collection tool for long-burst recording. Shown in Fig.~\ref{figsoap:app} (a), it features a live viewfinder with previews of RGB, device depth, and auto-adjusted ISO and exposure values. On a button press, we lock ISO, exposure, and focus, and record a two-second, 42 frame long-burst to the device. Our method uses recorded timestamps, camera intrinsics, gyroscope-driven device rotation estimates, and 12-megapixel RAW frames. However, our app also records processed RGB frames, low-resolution depth maps, and other metadata which we use for validation and visualization. 

\noindent\textbf{RAW Images.\footnote{\emph{RAW} here refers to sensor data after basic corrections such as compensating for broken and non-uniform pixels, not ``raw-raw'' data~\cite{pulli_2022}.}}\hspace{0.1em} A modern mobile image signal processing pipeline can have more than a dozen steps between light hitting the CMOS sensor and a photo appearing on screen: denoising, demosaicing, and gamma correction to name a few~\cite{delbracio2021mobile}. While these steps, when finely-tuned, can produce eye-pleasing results, they also pose a problem to down-stream computer vision tasks: they break linear noise assumptions, correlate pixel neighborhoods, and lower the overall dynamic range of the content (quantizing the 10- to 14-bit sensor measurements down to 8-bit color depth image files)~\cite{brooks2019unprocessing}.
In our work we are concerned with the tracking and reconstruction of small image features undergoing small continuous motion from natural hand tremor, and so apply minimal processing to our image data, using linear interpolation to only fill the gaps between Bayer measurements. We preserve the full 14-bit color depth, and fit our depth plus image model directly to this 4032px$\times$3024px$\times$3 channel$\times$42 frame volume.


\section{Unsupervised Depth Estimation}
In this section, we propose a method for depth estimation from long-burst data. We first lay out the projection model our method relies on, before introducing the scene model, loss functions, and training procedure used to optimize it.

\vspace{0.5em}
\noindent\textbf{Projection Model.}\hspace{0.1em} Given an image stack $I(u,v,\textsc{n})$, where $u,v\in[0,1]$ are continuous image coordinates and $\textsc{n}\in[0,1,...41]$ is the frame number, we aim to {condense the information in $I(u,v,\textsc{n})$ to a single compact projection model}. Given that the motion between frames is small, and image content is largely overlapping, we opt for an RGB-D representation which models each frame of $I(u,v,\textsc{n})$ as the deformation of some \emph{reference} image $I(u,v)$ projected through depth $D(u,v)$ with a change in camera pose $P(\textsc{n})$. We expand this process for a single point at coordinates $u,v$ in the reference frame. Let
\begin{equation}\label{eqsoap:sample_img}
C = [R,G,B]^\top = I(u,v),\quad d = D(u, v)
\end{equation}
be a sampled colored point $C$ at depth $d$. Before we can project this point to new frame, we must first convert it from camera $(u,v)$ to world $(x,y,z)$ coordinates. We assume a pinhole camera model to \emph{un-project} this point via
 \vspace{-0.2em}
\begin{equation}\label{eqsoap:unprojection}
\left[\arraycolsep=2.0pt
\begin{array}{c}
x \\
y \\
z \\
1
\end{array}\right]
= \bm{\pi}^{-1}\left(
\left[\arraycolsep=2.0pt
\begin{array}{c}
u \\
v \\
d \\
\end{array}\right]
;K\right)
=
\left[\arraycolsep=2.0pt
\begin{array}{c}
d(u-c_x) / f_x \\
d(v-c_y) / f_y \\
d \\
1
\end{array}\right],
\end{equation}
 \vspace{-0.2em}
where $K$ are the corresponding camera intrinsics with focal point $(f_x, f_y)$ and principal point $(c_x, c_y)$. We transform this point from the reference frame to target frame $\textsc{n}$, with camera pose $P(\textsc{n})$, via
 \vspace{-0.2em}
\begin{equation}\label{eqsoap:pose}
\left[\arraycolsep=2.0pt
\begin{array}{c}
x^{\textsc{n}} \\
y^{\textsc{n}} \\
z^{\textsc{n}} \\
\end{array}\right]
=
\left[R(\textsc{n})\,|\,T(\textsc{n}) \right]
\left[\arraycolsep=2.0pt
\begin{array}{c}
x \\
y \\
z \\
1
\end{array}\right]
=
\left[P(\textsc{n})\right]
\left[\arraycolsep=2.0pt
\begin{array}{c}
x \\
y \\
z \\
1
\end{array}\right].
\end{equation}
\begin{figure}[t]
    \centering
    \includegraphics[width=0.9\linewidth]{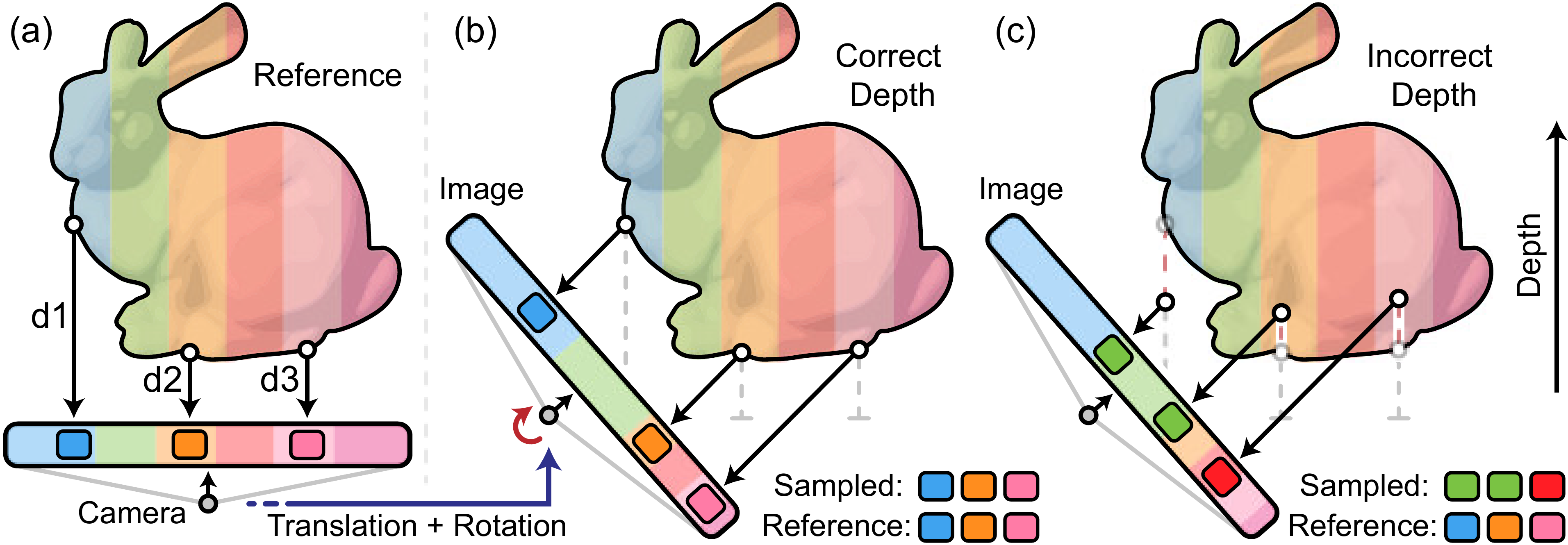}
    \caption{A 2D example of reprojection and sampling. When a reference view (a) is projected to new view with known camera rotation and translation, if the points' depths are accurately estimated they project to sample matching colors in the new image (b). If depths are inaccurate, as in (c), they do not sample corresponding colors, and instead incur \emph{photometric loss}.}
    \label{figsoap:raw_projection}
\end{figure}
\noindent Here, $P(\textsc{n})$ is decomposed into a $3{\times}3$ rotation matrix $R(\textsc{n})$ and $3{\times}1$ translation vector $T(\textsc{n})=\left[t_x, t_y, t_z\right]^\top$. Reverse of the process in Eq.~\eqref{eqsoap:unprojection}, we now \emph{project} this point from the world coordinates $(x^\textsc{n},y^\textsc{n},z^\textsc{n})$ in  frame $\textsc{n}$ to camera coordinates $(u^\textsc{n},v^\textsc{n})$ in the same frame as
 \vspace{-0.2em}
\begin{equation}\label{eqsoap:projection}
\left[\arraycolsep=2.0pt
\begin{array}{c}
u^\textsc{n} \\
v^\textsc{n} \\
\end{array}\right]
= \bm{\pi}\left(
\left[\arraycolsep=2.0pt
\begin{array}{c}
x^{\textsc{n}} \\
y^{\textsc{n}} \\
z^{\textsc{n}} \\
\end{array}\right]
;K(\textsc{n})\right)
=
\left[\arraycolsep=2.0pt\def\arraystretch{1.2}
\begin{array}{c}
(f^\textsc{n}_{x} x^\textsc{n}) / z^\textsc{n} + c^\textsc{n}_{x}\\
(f^\textsc{n}_{y} y^\textsc{n}) / z^\textsc{n} + c^\textsc{n}_{y}
\end{array}\right],
\end{equation}
\begin{figure*}[t]
 \vspace{-0.9em}
    \centering
    \includegraphics[width=\linewidth]{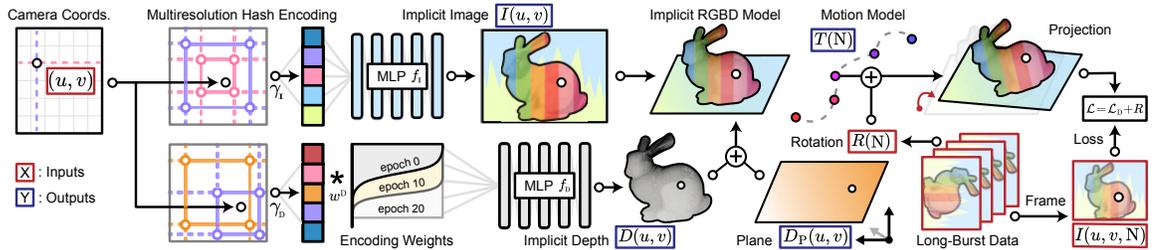}
    \caption{We model a long-burst capture as a single, fully-differentiable forward model comprised of an implicit image $I(u,v)$ projected through implicit depth $D(u,v)$ with motion model $[R(\textsc{n})|T(\textsc{n})]$. Calculating reprojection loss with respect to the the captured image stack $I(u,v,\textsc{n})$, we train this model end-to-end and distill high quality depth and camera motion estimates directly from the burst data.}
    \label{figsoap:method}
\end{figure*}
\noindent where $K(\textsc{n})$ are the frame intrinsics. We can now use these coordinates to sample a point from the full image stack
\begin{equation}\label{eqsoap:sample_volume}
C^\textsc{n} = I(u^\textsc{n},v^\textsc{n},\textsc{n}), \quad \mathcal{L}_{photo} = |C - C^\textsc{n}|.
\end{equation}
Here $\mathcal{L}_{photo}$ is \emph{photometric loss}, the difference in color between the point we started with in the reference frame and what we sampled from frame $\textsc{n}$. Given ideal multiview imaging conditions -- no occlusions, imaging noise, or changes in scene lighting -- if depth $d$ and pose change $P(\textsc{n})$ are correct, we will incur no photometric loss $\mathcal{L}_{photo}\,{=}\,0$ as we sample matching points in both frames. This is visualized in Fig.~\ref{figsoap:raw_projection}. \emph{Inverting this observation}, we can solve for unknown $D(u,v)$ and $P(\textsc{n})$ by finding ones that \emph{minimize photometric loss}~\cite{schonberger2016structure}.

\vspace{0.5em}
\noindent\textbf{Implicit Image Model.}\hspace{0.1em} In our problem setting, we are given a long-burst image stack $I(u,v,\textsc{n})$ and device rotation values $R(\textsc{n})$, supplied by an on-board gyroscope, and are tasked with recovering depth $D(u,v)$ and translation $T(\textsc{n})$ which make these observations consistent. Given the sheer number of pixels in $I(u,v,\textsc{n})$, in our case over \emph{500 million}, exhaustively matching and minimizing pixel-to-pixel loss is both computationally intractable and ill-posed. Under small camera motion, many depth solutions for a pixel can map it to identical-colored pixels in the image, especially in textureless parts of the scene. Traditional multi-view stereo (MVS) and bundle adjustment methods tackle this problem with feature extraction and matching~\cite{triggs1999bundle}, optimizing over a \emph{cost-volume} orders of magnitude smaller than the full image space. Here we strongly diverge from previous small motion works~\cite{im2015high,chugunov2022implicit,yu20143d}. Rather than divide the problem into feature extraction and matching, or extract features at all, we propose a single fully differentiable forward model trained \emph{end-to-end}. Depth is distilled as a product of fitting this \emph{neural scene model} to long-burst data. We start by redefining $I(u,v)$ from a static reference image to a learned implicit representation
\begin{align}\label{eqsoap:implicit_img}
    I(u,v) &= f_\textsc{i}(\gamma_\textsc{i}(u,v;\; \mathrm{params_{\gamma\textsc{i}}});\;\theta_\textsc{i}) \nonumber\\ 
    \mathrm{params_{\gamma\textsc{i}}} &= \left\{N_{min}^{\gamma\textsc{i}},\, N_{max}^{\gamma\textsc{i}},\,L^{\gamma\textsc{i}},\, F^{\gamma\textsc{i}},\,T^{\gamma\textsc{i}}\right\} 
\end{align}
where $f_\textsc{i}$ is a multi-layer perceptron (MLP)~\cite{hornik1989multilayer} with learned weights $\theta$. This MLP learns a mapping from $\gamma_\textsc{i}(u,v)$, a positional encoding of sampled camera coordinates, to image color. Specifically, we borrow the multi-resolution hash encoding from Müller et al.\cite{muller2022instant} for its spatial aggregation properties. The parameters in $params_{\gamma\textsc{i}}$ determine the minimum $N_{min}^{\gamma\textsc{i}}$ and maximum $N_{max}^{\gamma\textsc{i}}$ grid resolutions, number of grid levels $L^{\gamma\textsc{i}}$, number of feature dimensions per level $F^{\gamma\textsc{i}}$, and overall hash table size $T^{\gamma\textsc{i}}$. 

\vspace{0.5em}
\noindent\textbf{Implicit Depth on a Plane Model.}\hspace{0.1em} Our depth model is a similar implicit representation with a \emph{learned planar offset}
\begin{align}\label{eqsoap:depth}
    d &= D(u,v) = D_\textsc{p}(u,v) + f_\textsc{d}(\gamma_\textsc{d}(u,v; \mathrm{params}_{\gamma\textsc{d}}); \theta_\textsc{d})^+ \nonumber\\
    d_\textsc{p} &= D_\textsc{p}(u,v) = \mathrm{a}u + \mathrm{b}v + \mathrm{c},
\end{align}
where $\{\mathrm{a}, \mathrm{b}, \mathrm{c}\}$ are the learned plane coefficients, and $^+$ is the ReLU operation $max(0,x)$. Here $D_\textsc{p}(u,v)$ acts as the depth of the scene background -- the surface on or in front of which objects are placed -- which is often devoid of parallax cues. Then $f_\textsc{d}$ reconstructs the depth of the scene foreground content recovered from parallax in $I(u,v,\textsc{n})$. While it may seem that we are \emph{increasing} the complexity of the problem, as we now have to learn $I(u,v)$ in addition to $D(u,v)$, this model actually simplifies the learning task when compared to a static $I(u,v)$. Rather than solving for a perfect image from the get-go, $f_\textsc{i}$ can move between intermediate representations of the scene with blurry, noisy, and misaligned content, and is gradually refined during training.

\vspace{0.5em}
\noindent\textbf{Camera Motion Model.}\hspace{0.1em} Given the continuous, smooth, low-velocity motion observed in natural hand tremor~\cite{chugunov2022implicit}, we opt for a low-parameter B\'ezier curve motion model
\begin{align}\label{eqsoap:spline_model}
    &T(\textsc{n})  = \mathrm{B}(\textsc{n}; \mathbf{P}^\textsc{t}, N^\textsc{t}_c), \; R(\textsc{n}) = R_{d}(\textsc{n}) + \eta_\textsc{r}\mathrm{B}(\textsc{n}; \mathbf{P}^\textsc{r}, N^\textsc{r}_c)\nonumber\\
    &\mathrm{B}(t;\mathbf{P}, N_c) = \sum_{i=0}^{N_c}
    \left(\arraycolsep=2.0pt
    \begin{array}{l}
    N_c \\
    \;i
    \end{array}
    \right)(1-t)^{N_c-i} t^i \mathbf{P}_i,
\end{align}
with $N_c$ number of control points $\mathbf{P}_i$. Translation estimates $T(\textsc{n})$ are learned from scratch, whereas rotations $R(\textsc{n})$ are initialized as device values $R_d(\textsc{n})$ with learned offsets weighted by $\eta_\textsc{r}$. Under the small angle approximation~\cite{im2015high}, we parameterize the rotational offsets $\mathbf{P^\textsc{r}}$ as 
\begin{equation}
    \mathbf{P^\textsc{r}_i} = \left[\begin{array}{ccc}
0 & -r^z & r^y \\
r^z & 0 & -r^x \\
-r^y & r^x & 0
\end{array}\right].
\end{equation}
The choice of $N_c$ controls the dimensionality of the curve on which motion lies -- e.g. $N_c\,{=}\,1$ restricts motion to be linear, $N_c\,{=}\,2$ is quadratic, and $N_c\,{=}\,42$ trivially overfits the data with a control point for each frame.

\vspace{0.5em}
\noindent\textbf{Loss and Regularization.}\hspace{0.1em} Putting all of the above together we arrive at the full forward model, illustrated in Fig.~\ref{figsoap:method}. Given that all of our operations -- from re-projection to B\'ezier interpolation -- are fully differentiable, \emph{we train all these components simultaneously, end-to-end, through stochastic gradient descent}. But to do this, we need an objective to minimize. We employ a weighted composite loss
\begin{align}
    \mathcal{L} &= \mathcal{L}_\textsc{d} + \alpha_\textsc{p}(\mathcal{L}_\textsc{p}/\mathcal{L}_\textsc{d})\mathcal{R}, \quad \alpha_\textsc{p}>0,\;\beta_\textsc{p}\geq1 \label{eqsoap:loss1}\\
    \mathcal{L}_\textsc{d} &= |(C - C^\textsc{n}_\textsc{d})/(\mathrm{sg}(C) + \epsilon_\textsc{c})|^2 \label{eqsoap:loss2}\\
    \mathcal{L}_\textsc{p} &= |(C - C^\textsc{p}_\textsc{d})/(\mathrm{sg}(C) + \epsilon_\textsc{c})|^2 \label{eqsoap:loss3}\\
    \mathcal{R} &= |1 - d/d_\textsc{p}|^2 \label{eqsoap:loss4}\\
    C &= I(u,v), \quad C^\textsc{n} = I(u,v,\textsc{n}), \quad C^\textsc{n}_\textsc{p} = I(u,v,\textsc{n})_\textsc{p} \nonumber
\end{align}
Here $d$ is the depth output by our combined depth model, and $d_\textsc{p}$ is the depth of only the planar component as in Eq.~\eqref{eqsoap:depth}. $C$ is a colored point sampled from our implicit image model, $C^\textsc{n}$ is the point sampled from the image stack $I(u,v,\textsc{n})$ following Eqs.~\eqref{eqsoap:sample_img}\,--\,\eqref{eqsoap:sample_volume} for depth $d=d$, and $C^\textsc{n}_\textsc{p}$ is the point sampled following Eqs.~\eqref{eqsoap:sample_img}\,--\,\eqref{eqsoap:sample_volume} for the plane depth $d=d_\textsc{p}$. The regularization term \eqref{eqsoap:loss4} penalizes the magnitude of $f_\textsc{d}$, pulling the depth output towards the plane model. Losses \eqref{eqsoap:loss2} and \eqref{eqsoap:loss3} are relative square photometric errors between sampled colored points, where $\mathrm{sg}$ is the stop-gradient operator preventing the denominator $C$'s gradient from being back-propagated. This normalization by the approximate luminance of sampled points is effective in aiding the unbiased reconstruction of HDR images~\cite{lehtinen2018noise2noise}, and we refer the reader to derivations in Mildenhall et. al~\cite{mildenhall2022nerf} on its relation to tone-mapping. In \eqref{eqsoap:loss1}, we combine the photometric loss term $\mathcal{L}_\textsc{d}$, which seeks to maximize overall image reconstruction quality, with a weighted regularization $R$ which penalizes divergence from the planar model. When  $\mathcal{L}_\textsc{p}\approx\mathcal{L}_\textsc{d}$ -- i.e. the depth offset from $f_d$ is not improving reconstruction quality over a simple plane -- the model is strongly penalized. As $\mathcal{L}_\textsc{d}$ decreases relative to $\mathcal{L}_\textsc{p}$ -- the implicit depth model $f_d$ improves reconstruction quality -- this penalty falls off. In this way, regions that are blurred, textureless, or otherwise have no meaningful parallax information are pulled towards a spatially consistent plane solution rather than spurious depth predictions from $f_d$. As otherwise, in these regions, there is no photometric penalty for incorrect and noisy depth estimates. The parameter $\alpha_\textsc{p}$ controls the strength of this regularization.

\begin{figure}[t]
    \centering
    \includegraphics[width=0.85\linewidth]{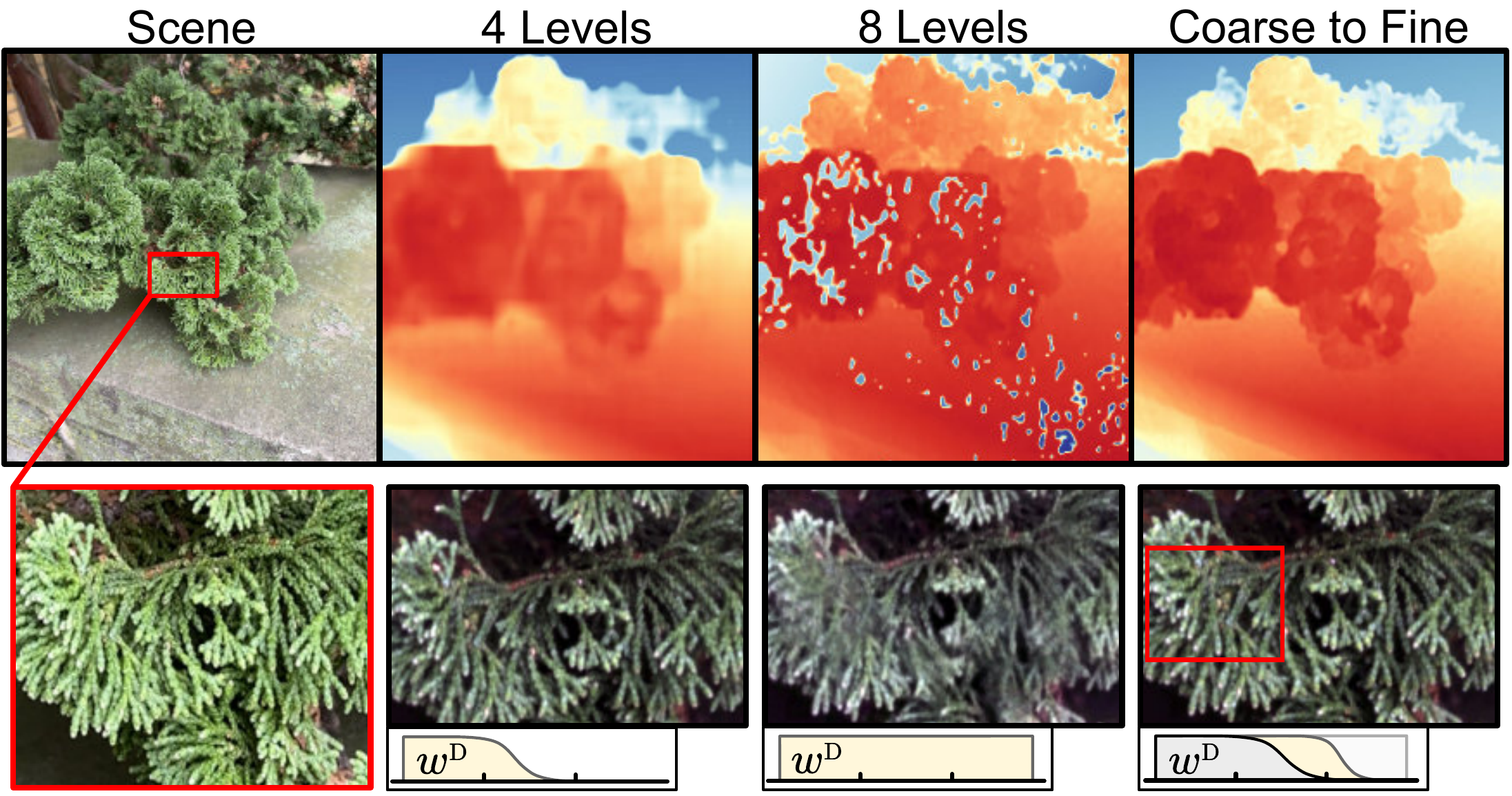}
    \caption{Ablation experiment on the effect of masked encoding levels. Using 4 encoding levels for depth leads to expected lower-resolution depth reconstructions. However, in \emph{8 Levels} where weights are not swept from coarse-to-fine, the reconstruction acquires sharp depth artifacts due to a positive feedback loop during training: high-frequency image gradients from $I(u,v)$ produce discontinuities in estimated depth $D(u,v)$, which in turn produce high-frequency images gradients in $I(u,v)$.}
    \label{figsoap:feature_masking}
\end{figure}

\vspace{0.5em}\noindent\textbf{Coarse-to-Fine Reconstruction.}\hspace{0.1em} First estimating low-resolution depths for whole objects before refining features such as edges and internal structures is a tried-and-true technique for improving depth reconstruction quality and consistency~\cite{chang2018pyramid,eigen2014depth}. However, one typical caveat of implicit scene representations is the difficulty of performing spatial aggregation -- an image pyramid is not well-defined for a continuous representation with no concept of pixel neighborhoods. Rather than try to aggregate outputs, we recognize that the multi-resolution hash encoding $\gamma_\textsc{d}(u,v)$ gives us control over the scale of reconstructed features. By masking the encoding $w^\textsc{d}\gamma_\textsc{d}(u,v)$ with weights $w^\textsc{d}_i\in[0,1]$ we can restrict the effective spatial resolution of the implicit depth network $f_d$, as two coordinates that map to the same masked encoding are treated as identical points by $f_d$. During training, we evolve this weight vector as
\begin{align}
    w^\textsc{d}_i &= 1/(1+exp(-ki))\nonumber\\
    k &= -k_{min} + (\mathrm{epoch} \cdot k_{max})/\mathrm{max\_epochs} 
\end{align}
which smoothly sweeps from passing only low-resolution grid encodings to all grid encodings during training, with $k_{min}$ and $k_{max}$ controlling the rate of this sweep. The effects of this masking are visualized in Fig.~\ref{figsoap:feature_masking}.

\vspace{0.5em}\noindent\textbf{Training and Implementation Details.}\hspace{0.1em} For simplicity of notation we have thusfar only worked with a single projected point. In practice, during a single forward pass of the model we perform \emph{one-to-all} projection of a batch of 1024 points at a time from the reference $I(u,v)$ to \emph{all 42 frames} in $I(u,v,\textsc{n}$). We perform stochastic gradient descent on $\mathcal{L}$ with the Adam optimizer~\cite{kingma2014adam}. Our implementation is built on tiny-cuda-dnn~\cite{muller2021real}, and on a single Nvidia A100 GPU has a training time of approximately $15$ minutes per scene. Our encoding parameters are $N_{min}^{\gamma\textsc{ i}}=8,N_{max}^{\gamma\textsc{i}}=2048,L^{\gamma\textsc{i}}=16,F^{\gamma\textsc{i}}=4,T^{\gamma\textsc{i}}=2^{22}$ and $N_{min}^{\gamma\textsc{ d}}=8,N_{max}^{\gamma\textsc{d}}=128,L^{\gamma\textsc{d}}=8, F^{\gamma\textsc{d}}=4,T^{\gamma\textsc{d}}=2^{14}$, as depth has significantly less high-frequency features than $I(u,v)$. The networks $f_\textsc{i}$ and $f_\textsc{d}$ are both 5-layer 128 neuron MLPs with ReLU activations. For the rotation offset weight we choose $\eta_\textsc{r}=10^{-4}$; regularization weight $\alpha_\textsc{p}=10^{-4}$ and $\epsilon_\textsc{c}=10^{-3}$; encoding weight parameters $k_{min}=-100, k_{max}=200$; and number of control points $N^\textsc{t}_c=N^\textsc{r}_c=21$, one for every two frames. We provide additional training details, and an extensive set of ablation experiments in the Supplemental Document to illustrate the effects of these parameters and how the above values were chosen. Our data capture app is built on the AVFoundation library in iOS 16 and tested with iPhone 12-, 13-, and 14-Pro devices. For consistency, a single 14 Pro device was used for all data captured in this work. RAW capture is hardware/API limited to ${\sim}21$~FPS, hence a two-second long-burst contains 42 frames.
\vspace{-0.5em}

\begin{figure*}[t]
 \vspace{-1.9em}
    \centering
    \includegraphics[width=\linewidth]{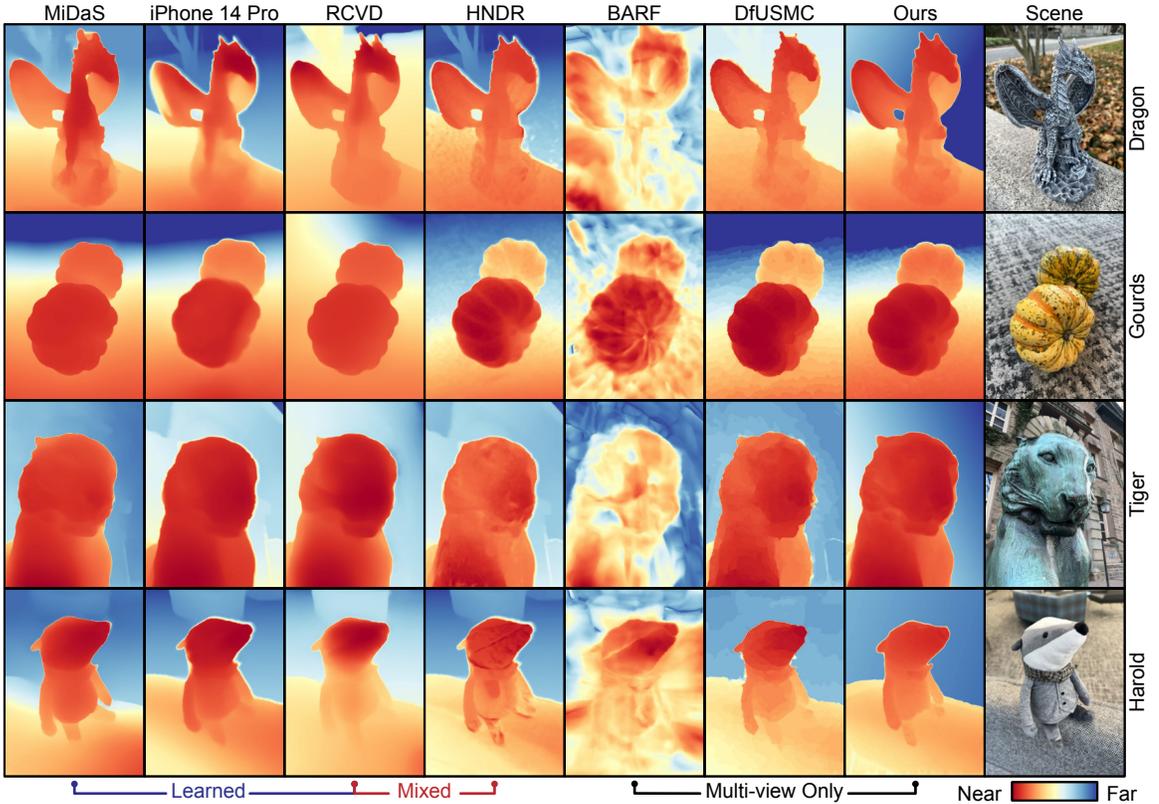}
    \caption{Qualitative comparison of reconstruction results of indoor and outdoor scenes for a range of learned, purely multi-view, and mixed depth estimation methods. Given the mix of depth scales, results are re-scaled by minimizing relative mean square error.}
    \label{figsoap:main_results}
\end{figure*}
\section{Assessment}\label{secsoap:results}
\begin{figure*}[t]
 \vspace{-1em}
    \centering
    \includegraphics[width=\linewidth]{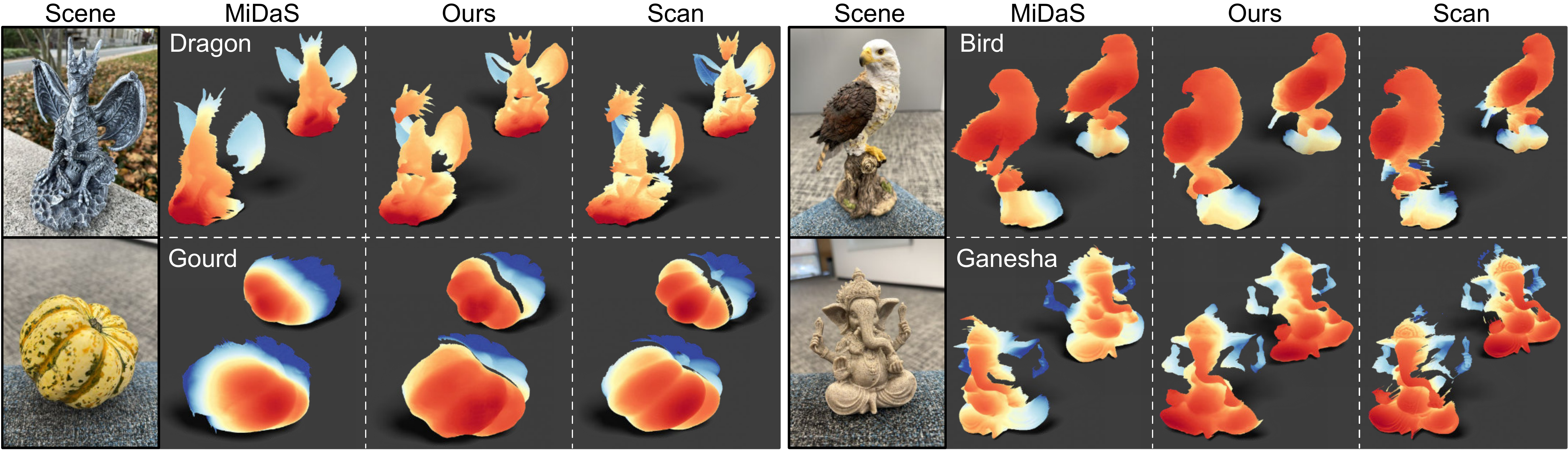}
    \resizebox{\linewidth}{!}{
     \begin{tabular}[b]{lccccccclccccccc}
     \midrule
          & MiDaS & iPhone & RCVD & HNDR & BARF & SfSM & Ours & & MiDaS & iPhone & RCVD & HNDR & BARF & DfUSMC & Ours\\
	\midrule
	\midrule
    \textit{Dragon} & .447$/$.224  & .482$/$.355 & .485$/$.237 & .470$/$.345 & .285$/$.339 & .166$/$.118 & \textbf{.129$/$.073} & \textit{Bird} & .283$/$.199  & .321$/$.282 & .322$/$.255 & .284$/$.274 & .172$/$.159 & .125$/$.161 & \textbf{.082$/$.058}    \\
    \textit{Gourd} & .233$/$.195  & .266$/$.229 & .235$/$.181 & .264$/$.225 & .821$/$.369 & .167$/$.141 & \textbf{.086$/$.078} & \textit{Ganesha} & .232$/$.194  & .283$/$.224 & .306$/$.239 & .275$/$.230 & .376$/$.328 & .132$/$.176 & \textbf{.094$/$.104}    \\
    \bottomrule
  \end{tabular}
  }
    \caption{Object reconstructions visualized as rendered meshes, with associated depth metrics formatted as \emph{relative absolute error / scale invariant error}. Edges over $10{\times}$ the length of their neighbors were culled to avoid connecting mesh features in occluded regions. }
    \label{figsoap:mesh_results}
    \vspace{-1em}
\end{figure*}

\noindent\textbf{Evaluation.} We compare our approach to the most similar purely multi-view methods BARF~\cite{lin2021barf} and Depth From Uncalibrated Small Motion Clip (DfUSMC)~\cite{ha2016high}, both of which estimate depth and camera motion directly from an input image stack. We note that BARF also has a similar implicit image generation model. We also compare to learned monocular methods: iPhone's 14 Pro's native depth output and MiDaS~\cite{ranftl2020towards}, a robust single-image approach. Lastly, we compare to Robust Consistent Video Depth Interpolation (RCVD)~\cite{kopf2021robust} and Handheld Multi-frame Neural Depth Refinement (HNDR)~\cite{chugunov2022implicit}, which both use multi-view information to refine initial depth estimates initialized from a learned monocular approach. The latter of which is most directly related to our approach as it targets close-range objects imaged with multi-view information from natural hand tremor, but relies on iPhone LiDAR hardware for depth initialization and pose estimation. All baselines were run on processed RGB data synchronously acquired by our data capture app, except for HNDR which required its own data capture software that we ran in tandem to ours. We note that other neural scene volume methods such as~\cite{mildenhall2022nerf} require COLMAP as a pre-processing step, which \emph{fails to find pose solutions for our long-burst data}. To assess absolute performance and geometric consistency, we scan a select set of complex 3D objects, illustrated in Fig.~\ref{figsoap:mesh_results}, with a commercial high-precision turntable structured light scanner (Einscan SP). We use this data to generate ground truth object meshes, which we register and render to depth with matching camera parameters to the real captures. For quantitative depth assessment, we use relative absolute error and scale invariant error, commonly used in monocular depth literature~\cite{Ummenhofer2017}; see the Supplemental Document for details.

\begin{figure}[t]
    \centering
    \includegraphics[width=0.9\linewidth]{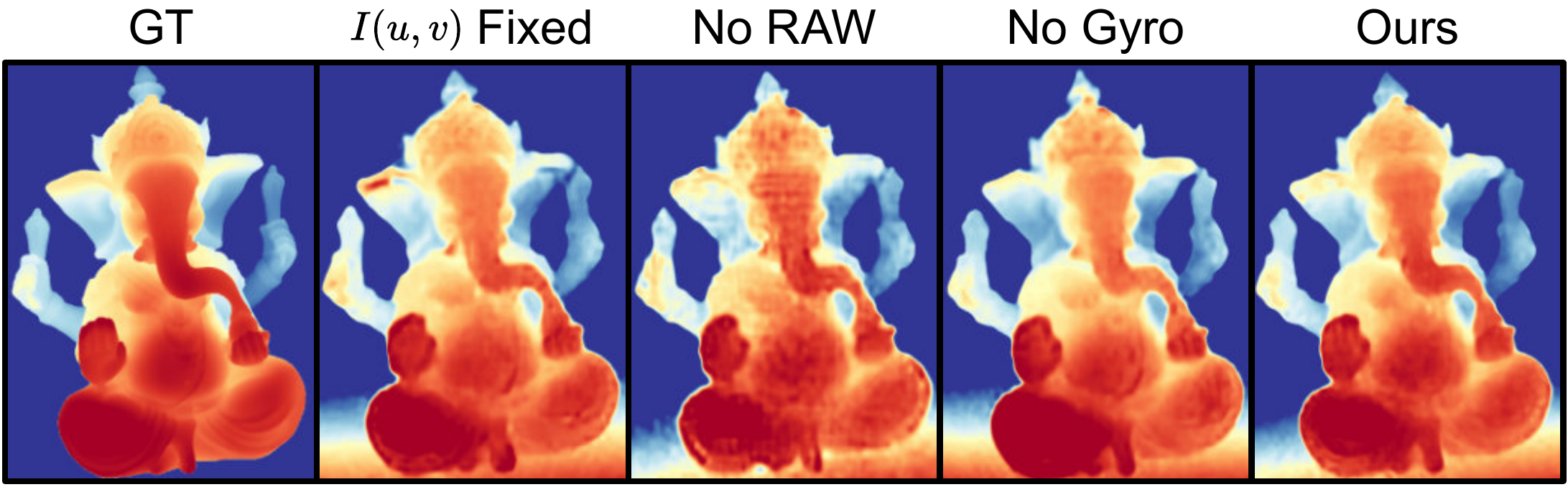}
        \resizebox{\linewidth}{!}{
     \begin{tabular}[b]{lcccclcccc}
     \midrule
          & Fixed & No RAW & No Gyro & Ours & &Fixed & No RAW & No Gyro & Ours \\
	\midrule
	\midrule
    Dragon & .150  & .142 & \textbf{.125} & .129 & Bird & .113 & .148 & .089 & \textbf{.082 } \\
     & $\overline{.081}$  & $\overline{.097}$ & $\overline{.080}$ & $\overline{\textbf{.073}}$ &  & $\overline{.099}$  & $\overline{.122}$ & $\overline{.078}$ & $\overline{\textbf{.058}}$  \\
     \hline
     Gourd & .091  & .093 & .088 & \textbf{.086} & Ganesha & .141  & .111 & .121 & .\textbf{094} \\
     & $\overline{\textbf{.078}}$  & $\overline{.081}$ & $\overline{.080}$ & $\overline{\textbf{.078}}$ &  & $\overline{.117}$  & $\overline{.104}$ & .$\overline{\textbf{094}}$ & $\overline{.104}$  \\
    \bottomrule
  \end{tabular}
  }
    \caption{Ablation study on the effects of fixing the image representation $I(u,v)$ to be the first long-burst frame $I(u,v,0)$, replacing the RAW data in $I(u,v,\textsc{n})$ with processed 8-bit RGB, or removing device initial rotation estimates from our model. Metrics formatted as \emph{relative absolute error / scale invariant error}.}
    \label{figsoap:rgb_comparisons}
\end{figure}

\noindent\textbf{Reconstruction Quality.} Tested on a variety of scenes, illustrated in Fig.~\ref{figsoap:main_results}, we demonstrate high-quality object depth reconstruction outperforming existing learned, mixed, and multi-view only methods. Of particular note is how we are able to reconstruct small features such as \emph{Dragon}'s tail, \emph{Harold}'s scarf, and the ear of the \emph{Tiger} statue consistent to the underlying scene geometry. This is in contrast to methods such as RCVD or HNDR which either neglect to reconstruct the \emph{Tiger}'s ear or reconstruct it behind its head. Our coarse-to-fine approach also allows us to reconstruct scenes with larger low-texture regions, such as \emph{Harold}'s head, which produces striped depth artifacts for HNDR as it can only refine depth within a patch-size of high-contrast edges. Our depth on a plane decomposition avoids spurious depth solutions in low-parallax regions around objects, cleanly segmenting them from their background. This plane segmentation, and it's applications to image and depth matting, are further discussed in the Supplemental Document. In contrast to DfUSMC, which relies on sparse feature matches and RGB-guided filtering to in-paint contiguous depth regions, our unified end-to-end model \emph{directly} produces complete and continuous depth maps.

 In Fig.~\ref{figsoap:mesh_results}, we highlight our method's ability reconstruct complex objects. While from a \emph{single image} the learned monocular method MiDaS produces visually consistent depth results, from a \emph{single long-burst} our approach directly solves for geometrically accurate affine depth. This difference is most clearly seen in the \emph{Dragon} object, whose wings are reconstructed at completely incorrect depths by MiDAS, disjoint from the rest of the object. This improved object reconstruction is also reflected in the quantitative depth metrics, in which we outperform all comparison methods. Another note is that the most structurally similar method to ours, BARF -- which also learns an implicit image model and distills camera poses in a coarse-to-fine encoding approach -- fails to produce reasonable reconstructions. We suspect this is related to the findings of Gao et al.~\cite{gao2022monocular}, that NeRFs do not necessarily obey projective geometry during view synthesis for highly overlapping image content.

\noindent\textbf{Implicit Values of a Learned RAW Model.} In Fig.~\ref{figsoap:rgb_comparisons}, we observe the quantitative and qualitative effects of removing various key method components. For the \emph{No Gyro} tests we replace device rotations $R_d(\textsc{n})$ with identity rotation for all frames $\textsc{n}$ and learn offsets as usual. We find that while the use of a fixed reference image, 8-bit RGB, or no gyro measurements can reduce our model's average reconstruction quality, all these experiments still converge to acceptable depth solutions. This is especially true of the \emph{No Gyro} experiments, which for many scenes result in \emph{near identical} reconstructions. This further validates our model's ability to independently learn high quality camera pose estimates, and demonstrates its modularity with respect to input data and optimization settings -- applicable even to settings where RAW images and device motion data are not available.


\vspace{-0.5em}\section{Discussion}
In this work, we demonstrate that from only a stack of images acquired during long-burst photography, with parallax information from natural hand-tremor, it is possible to recover high-quality, geometrically-accurate object depth.\\
\noindent\textbf{Forward Models.} Our static single-plane RGB-D representation could potentially be modified to include differentiable models of object motion, deformation, or occlusion. \\
\noindent\textbf{Image Refinement.} We use the learned image $I(u,v)$ as a vehicle for depth optimization, but it could be possible to flip this and use the learned depth $D(u,v)$ as a vehicle for aggregating RGB content (e.g., denoising or deblurring).
\noindent\textbf{From Pixels to Features.} Low-texture or distant image regions have insufficient parallax cues for ray-based depth estimation. Learned local feature embeddings could help aggregate spatial information for improved reconstruction.
%
\section{Implementation Details}
\label{secsoap:implementation}
\begin{figure}[t!]
    \centering
    \includegraphics[width=0.85\linewidth]{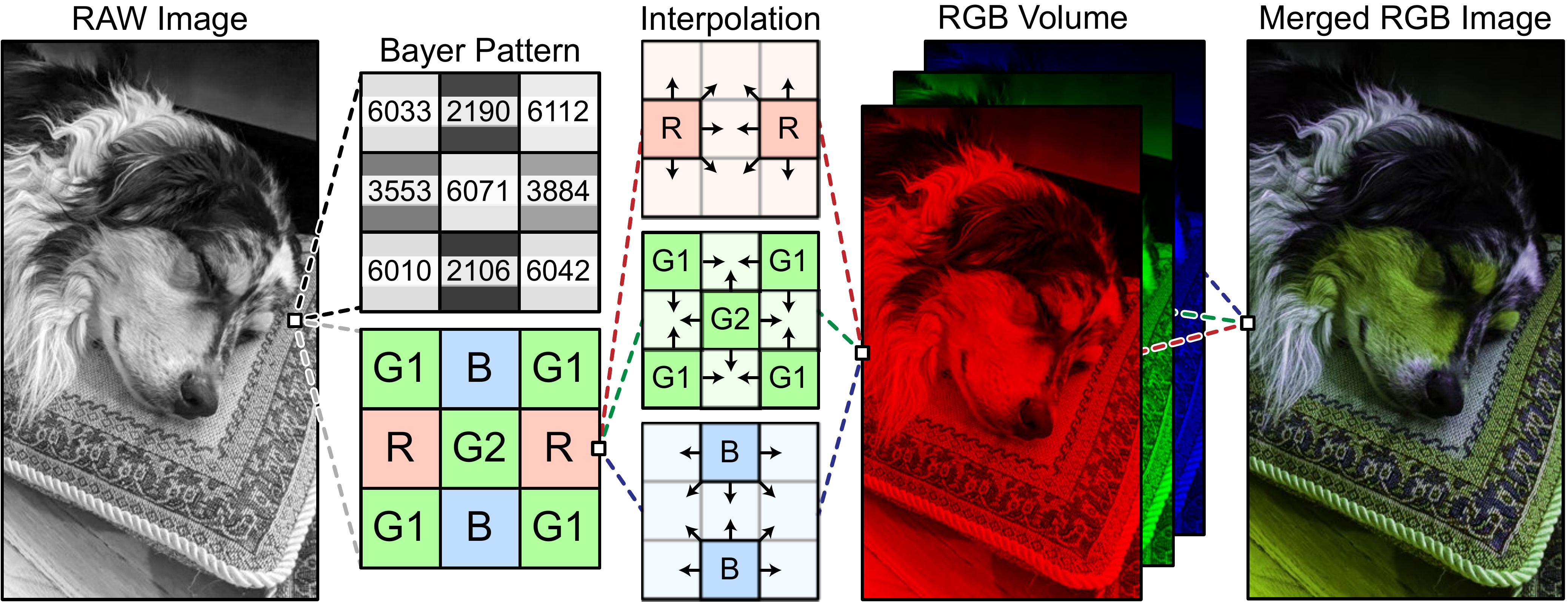}
    \caption{The bayer color filter array on a camera sensor produces a spatially "mosaicked" RAW image, where each 2$\times$2 block contains a blue, red, and two green pixels. Rather than mix channel content to "demosaick" the image, we separate these channels into three planes and only linearly interpolate gaps between measured pixels, preserving the original RAW values.}
    \label{figsoap:bayer_array}
\end{figure}

\begin{figure}[t!]
    \centering
    \includegraphics[width=0.85\linewidth]{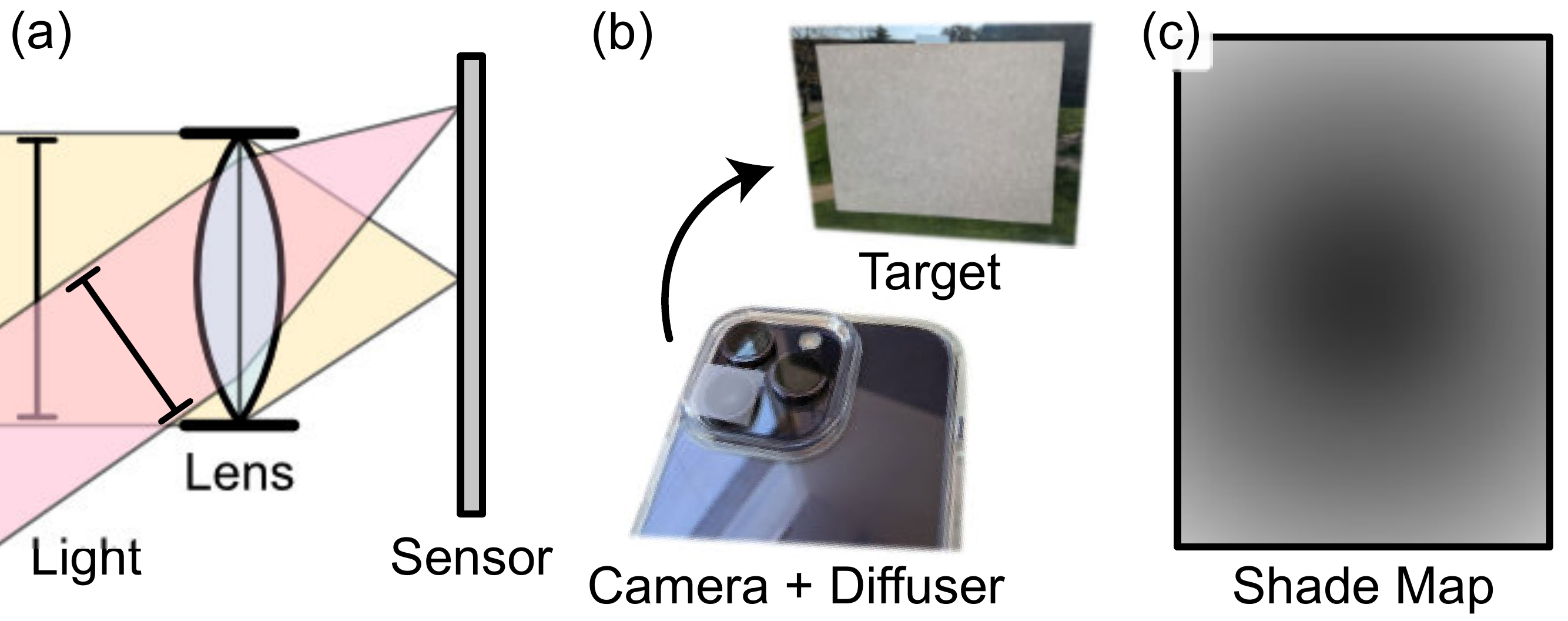}
    \caption{(a) Lens shading is an effect caused by the geometry of the camera lens assembly, where regions close to the edge of the image sensor receive less total light than the center. (b) Camera manufacturers calibrate for this by capturing what should be a uniformly bright scene and (c) generating a \emph{shade map} to compensate for the observed fall-off in brightness in the image.}
    \label{figsoap:lens_shading}
\end{figure}

\noindent\textbf{Long-Burst Data.} We acquire long-burst data through a custom app built on the AVFoundation camera framework for iOS 16. While the vanilla AVFoundation framework offered a default method to capture burst or bracketed sequences, it was limited to only four frame sequences with significant overhead between captures, necessitating custom streaming code to save a longer continuous sequence of RAW data. A restriction we could not lift, however, is the inability to stream RAW captures from multiple cameras simultaneously. If this were possible, one could potentially use parallax and focus cues between two synchronized camera streams -- for example the wide and ultra-wide cameras -- to further improve reconstruction in the overlap of their fields of view. During capture, we record the following: Bayer CFA RAWs (42 frames $4032{\times}3024$px), processed RGB images (42 frames $1920{\times}1440$px), depth maps (42 frames $320{\times}240$px), frame timestamps, ISO, exposure time, brightness estimates, black level, white level, camera intrinsics, lens distortion tables, device acceleration estimates (${\sim}200$ measurements at 100Hz), device rotation estimates (${\sim}200$), and motion data timestamps (${\sim}200$). To preserve measured RAW values, we convert the single channel Bayer CFA images to three channel RGB volumes as shown in Fig.~\ref{figsoap:bayer_array}, linearly interpolating to fill missing values. To account for lens shading effects in bright scenes we also estimate a shade map with the help of a simple diffuser and uniform light source, illustrated in Fig.~\ref{figsoap:lens_shading}. We note that neglecting to compensate for lens shading can disrupt depth estimation in the corners of the image as matching pixels no longer have uniform brightness between frames.

\vspace{0.6em}
\noindent\textbf{Training.} We sample 1024 points $(u,v)$ per iteration of training, projecting these to $42{\times}1024$ points in the image stack $I(u,v,\textsc{n})$, corresponding to 1024 points per frame. We perform 256 iterations per epoch, for 100 epochs of training with the Adam optimizer~\cite{kingma2014adam} with betas (0.9, 0.99) and epsilon $10^{-15}$. We exponentially decay learning rate during training with a factor of 0.98 per epoch. Training on a single Nvidia A100 takes approximately 15 minutes.

\begin{figure}[t]
    \centering
    \includegraphics[width=0.7\linewidth]{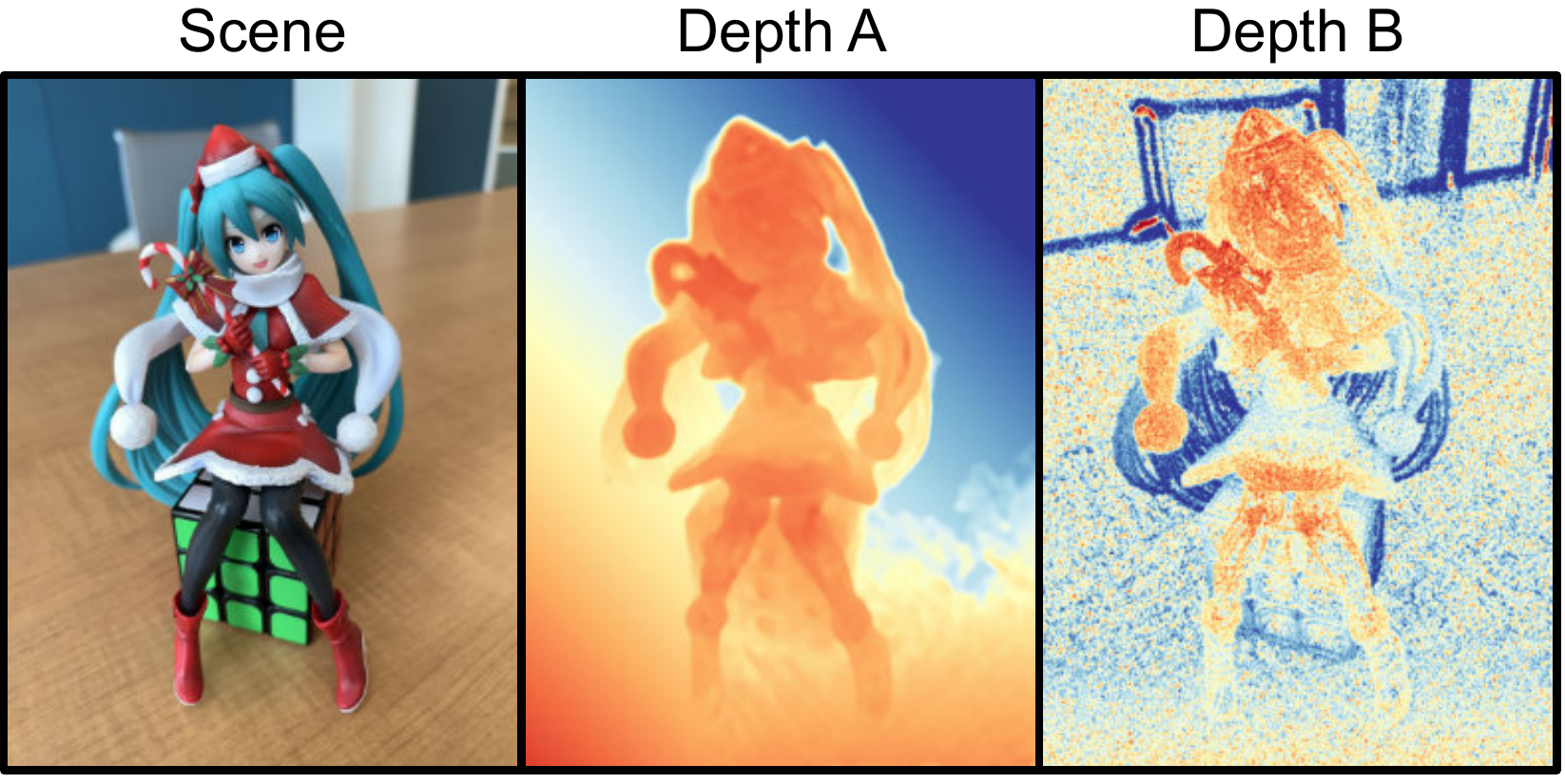}
    \caption{In this example both Depth A and B produce \emph{identical} reprojection error, but where Depth A models smooth geometry which warps the image between frames to model parallax, Depth B performs a brute-force mapping of individual pixels in the reference frame to points with similar values in the image stack.}
    \label{figsoap:reprojection_error}
\end{figure}

\vspace{0.6em}
\noindent\textbf{Evaluation.} To generate depth maps we sample $D(u,v)$ at a grid of $(\textsc{h},\textsc{w})\,{=}\,(1920,1440)$ points $(u,v)\,{\in}\,[0,1]$. To reduce noise introduced by the stochastic training process we median filter this result with kernel size 13 before visualization. For depth evaluation, we use relative absolute error \emph{L1-rel} and scale invariant error \emph{sc-inv} metrics, that is
\begin{equation}
\text{L1-rel}(d, \hat d) = \frac{1}{\textsc{h}\textsc{w}}\sum_{u.v} \frac{|d(u,v) - \hat d(u,v)|}{\hat d(u,v)}\nonumber ,
\end{equation}
\noindent and
\begin{align}
\text{sc-inv}(d, \hat d) &= \sqrt{\frac{1}{\textsc{h}\textsc{w}} \sum_{u,v} \delta(u,v)^2-\frac{1}{(\textsc{h}\textsc{w})^2}(\sum_{u,v} \hat \delta(u,v))^2}\nonumber\\
\delta(u,v) &= \log(d(u,v)) - \log(\hat d(u,v)),\nonumber
\end{align}
which are often used in the monocular depth estimation literature~\cite{Ummenhofer2017} to compare approaches with varying scales and representations of depth. For methods such as MiDaS and RCVD we first convert inverse depth to depth before applying these metrics. We purposely avoid using photometric loss or reprojection error as comparison metrics~\cite{chugunov2022implicit} for similar arguments as discussed in Gao et al.~\cite{gao2022monocular}.
\begin{equation}
    \text{reprojection\_error}=\frac{1}{\textsc{h}\textsc{w}}\sum_{u,v,\textsc{n}} |I(u,v) - I(u^\textsc{n}, v^\textsc{n},\textsc{n})| \nonumber.
\end{equation}
Frames in a long-burst contain ${>}$90\% overlapping scene content, and so many non-physical solutions for depth will produce identical reprojection error as compared to more geometrically plausible depth maps. This is illustrated in Fig.~\ref{figsoap:reprojection_error}, where by ``tearing" the image -- compressing patches of similar colored pixels from the reference frame -- the non-physical depth incurs no additional photometric penalty, and so results in an identical reprojection error to a far more qualitatively plausible depth reconstruction.

\section{Additional Ablation Experiments}
\label{secsoap:ablation}
\begin{figure*}[t!]
    \centering
    \includegraphics[width=\linewidth]{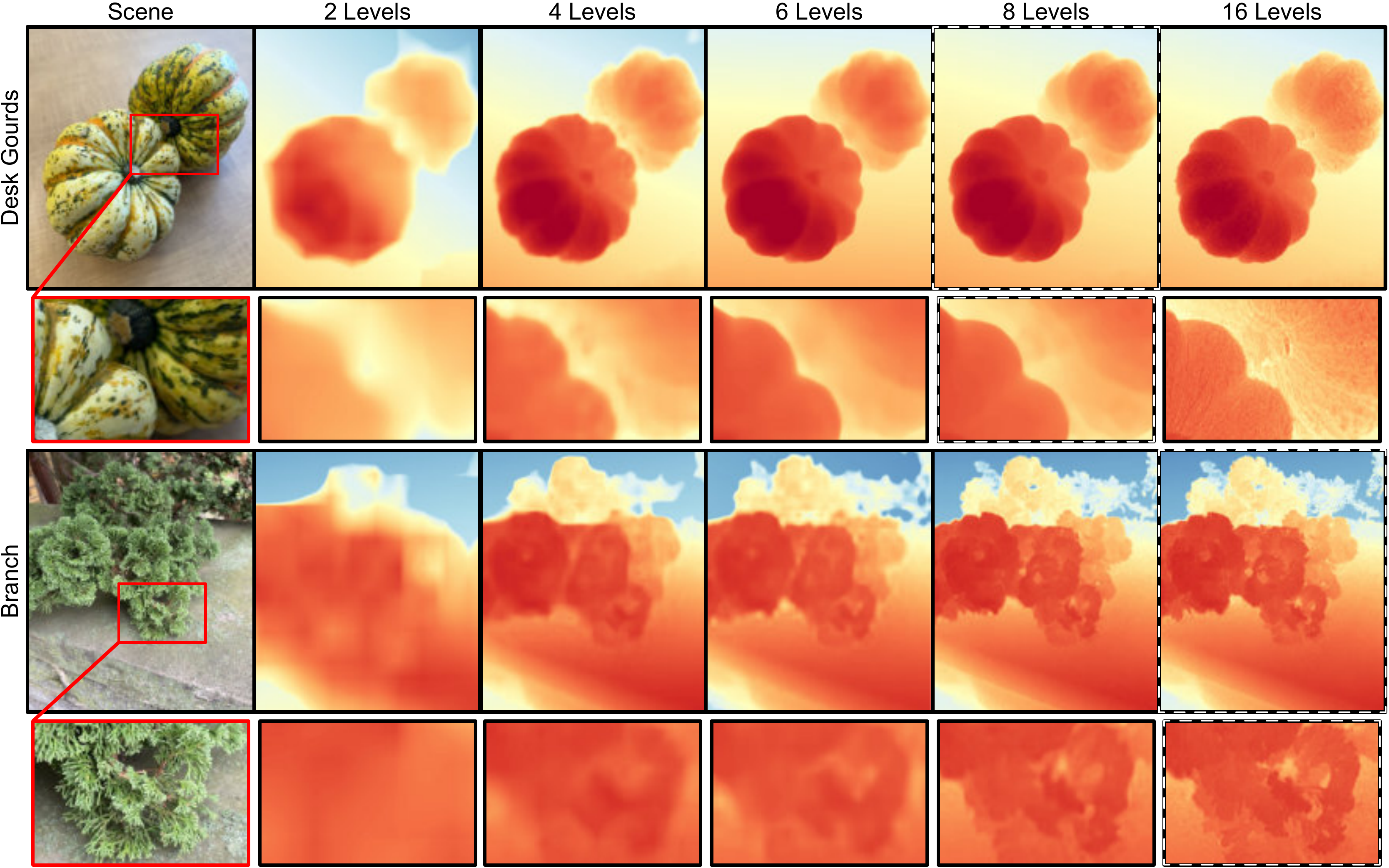}
    \caption{Ablation study on the effect of the number of levels $L^{\gamma\textsc{d}}$, and effective max resolution $N_{max}^{\gamma\textsc{d}}$, in the multiresolution hash encoding $\gamma_\textsc{d}$ on reconstruction. Here, given a scale factor of $\sqrt{2}$ between levels, $L^{\gamma\textsc{d}}=2,4,6,8,16$ correspond to $N_{max}^{\gamma\textsc{d}}=16,32,128,2048$. The qualitatively best reconstructions are highlighted with a dashed border.}
    \label{figsoap:encoding_layers}
\end{figure*}

\begin{figure*}[t!]
    \centering
    \includegraphics[width=\linewidth]{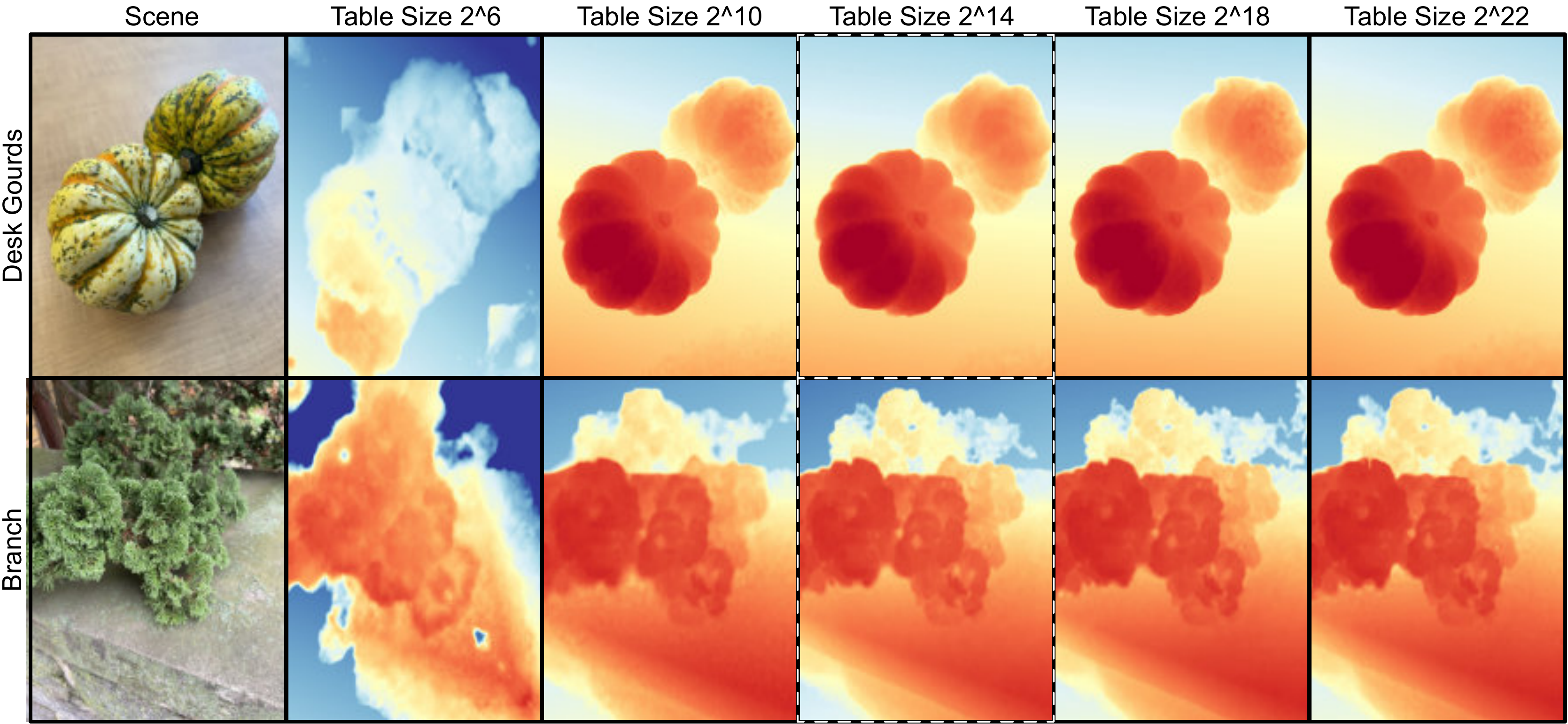}
    \caption{Ablation study on the effect of hash table size $T^{\gamma\textsc{d}}$ on reconstruction quality. Selected $T^{\gamma\textsc{d}}$ is highlighted with a dashed border.}
    \label{figsoap:encoding_hash}
\end{figure*}

\noindent\textbf{Encoding.} In this work we use the multiresolution hash encoding $\gamma_\textsc{d}$ to directly control what spatial information our implicit depth representation $f_\textsc{d}$ has access to during training. This in turn controls the scale of depth features we reconstruct, and presents a similar problem to choosing the scale factors in an image pyramid~\cite{adelson1984pyramid}. As we see in Fig.~\ref{figsoap:encoding_layers}, increasing the number of levels $L^{\gamma\textsc{d}}$ and effective max resolution $N_{max}^{\gamma\textsc{d}}$ increases the spatial frequency of reconstructed depth features. Scenes such as \emph{Branch} contain both high-frequency image and depth content, thin textured needles, and are best reconstructed by a fine resolution grid with $L^{\gamma\textsc{d}}=16$. The \emph{Desk Gourds}, however, have small image features in the patterns on the gourds, but relatively low-frequency depth features. Setting $L^{\gamma\textsc{d}}=16$ allows the network to overfit to these features and bleed image texture into the depth reconstruction. We select $L^{\gamma\textsc{d}}=8$ as a compromise between these imaging settings, but in practice, different scenes have different optimal encoding parameters for maximum reconstruction quality. We find hash table size $T^{\gamma\textsc{d}}$ significantly easier to tune, as choosing an overly large table size primarily affects model storage size, rather than reconstruction quality. We thus choose $T^{\gamma\textsc{d}}=2^{14}$, the smallest table size which does not lower the detail of depth reconstruction, as shown in Fig.~\ref{figsoap:encoding_hash}.

\begin{figure*}[htp!]
\vspace*{-1.2em}
    \centering
    \includegraphics[width=\linewidth]{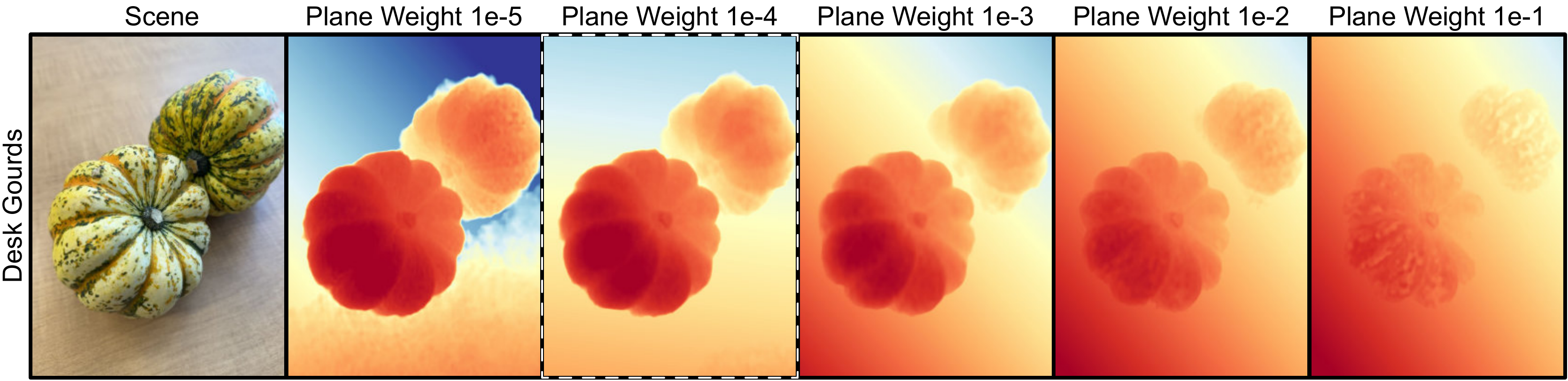}
    \caption{Ablation study on the effects of regularization weight $\alpha_\textsc{p}$ on reconstruction quality. Selected $\alpha_\textsc{p}$ highlighted with a dashed border.}
    \label{figsoap:plane_weight}
\bigskip
    \includegraphics[width=\linewidth]{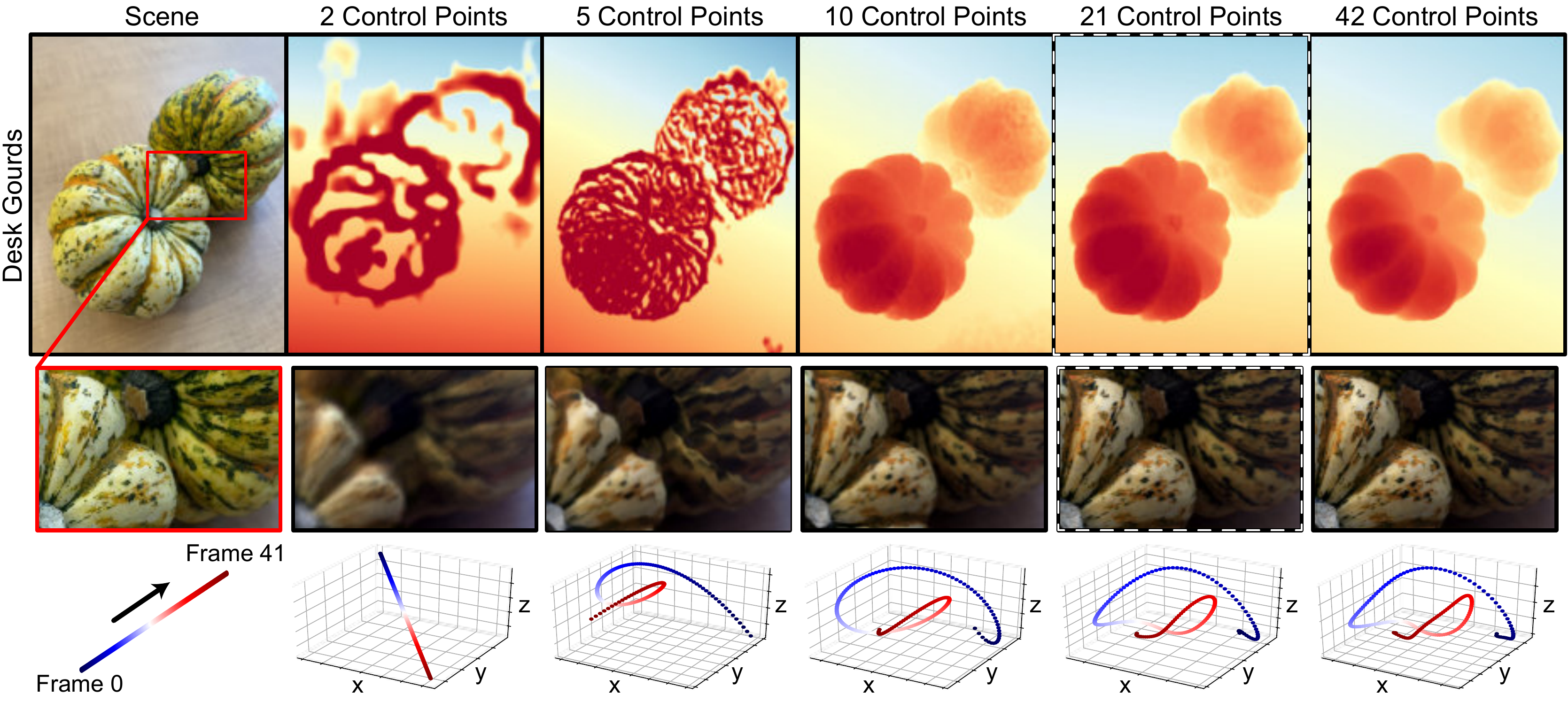}
    \caption{Ablation study on the effect of the number of chosen control points $N^\textsc{t}_c\,{=}\,N^\textsc{r}_c$ on reconstruction quality, with image reconstructions $I(u,v)$ and estimated motion paths plotted below. The selected number of control points is highlighted with a dashed border.}
    \label{figsoap:control_points}
\bigskip

    \includegraphics[width=\linewidth]{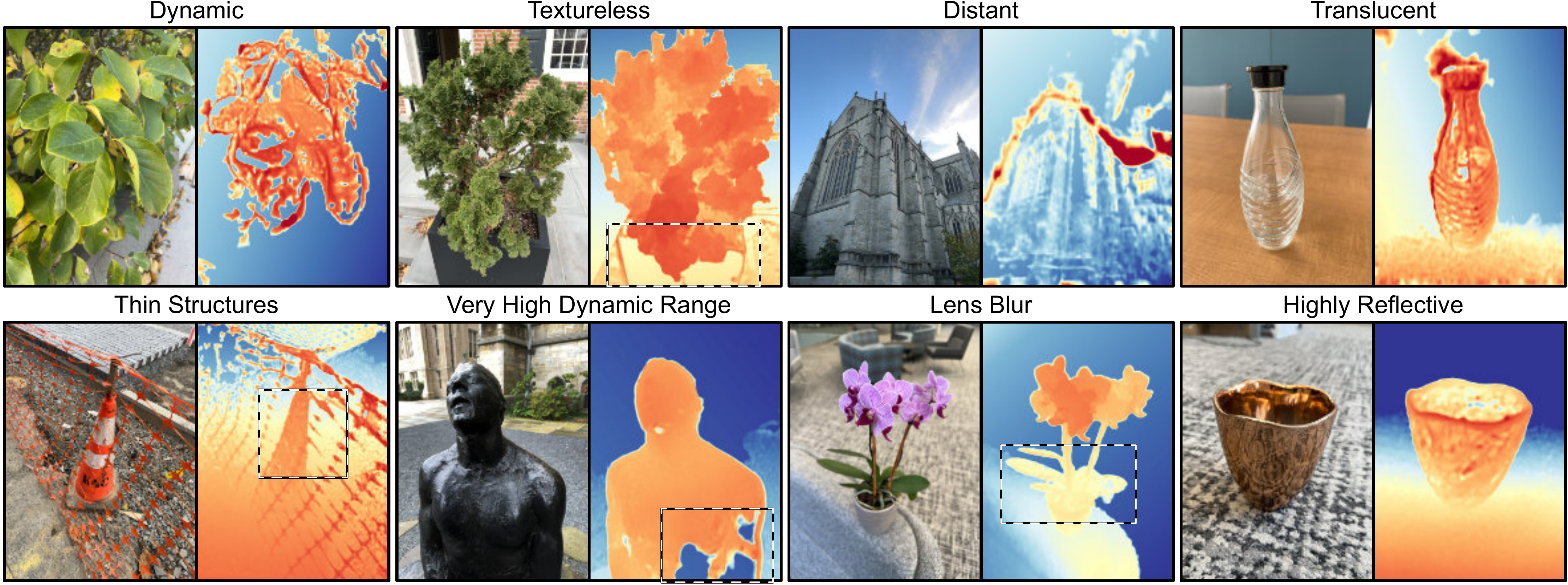}
    \caption{Depth reconstruction results for a set of challenging imaging scenarios. Not visible is the large motion of leaves in the \emph{Dynamic} scene, captured during high wind. Areas of interest are highlighted with a dashed border.}
    \vspace*{-0.5em}
    \label{figsoap:difficult_scenes}
\end{figure*}

\noindent\textbf{Depth Model.} The main adjustable parameter in our forward model is the plane regularization weight $\alpha_\textsc{p}$. This plane regularization affects depth reconstruction bidirectionally, regions with little parallax information are pulled towards the plane to remove spurious depth estimates, but in order to minimize depth offset, the plane is also pulled towards the reconstructed foreground objects. The effect of this can be seen in Fig.~\ref{figsoap:plane_weight}, where for very low $\alpha_\textsc{p}\,{\leq}\,10^{-5}$ this plane does not align with the foreground depth, and instead drifts into the background, causing a discontinuity in the reconstruction. Conversely, for large $\alpha_\textsc{p}\,{\geq}\,10^{-3}$, this regularization is so strong that the plane begins to cut into the foreground objects, flattening regions with low parallax information. We find $\alpha_\textsc{p}\,{=}\,10^{-4}$ to work well for a wide range of scenes, ``gluing'' the depth plane to the limit of reconstructed objects. We note that in scenes such as \emph{Desk Gourds}, and as we will see later with synthetic data, this plane accurately reconstructs the real geometry of the background. However, for many settings it is more akin to a \emph{segmentation mask} than depth, designating the area which we cannot reconstruct using parallax information.

\begin{figure*}[htp!]
    \centering
    \includegraphics[width=\linewidth]{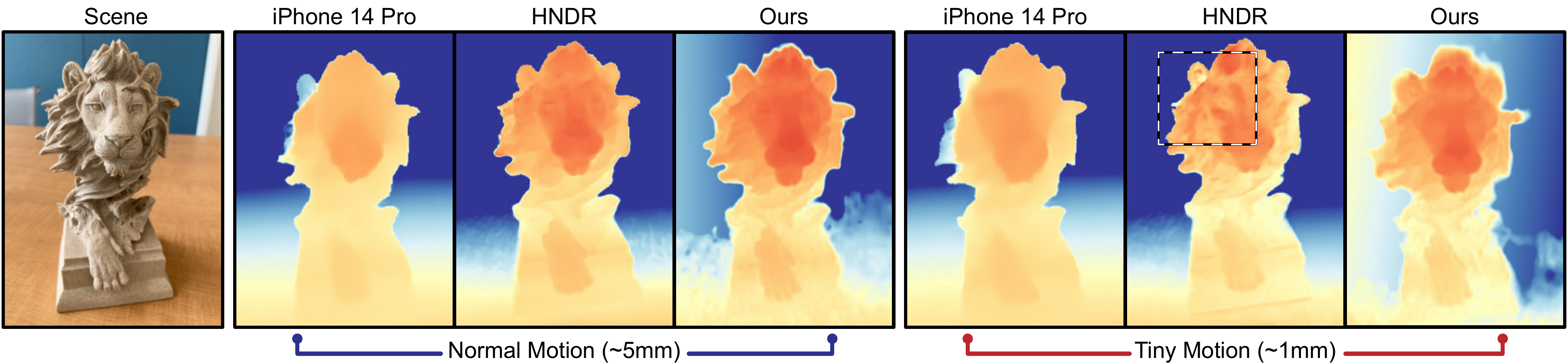}
    \caption{Depth reconstruction results for long-bursts captured with normal (approximately 5 millimeter maximum effective stereo baseline) and minimal (on the scale of a millimeter) hand shake motion. Major depth artifacts are highlighted with a dashed border.}
    \label{figsoap:small_motion}
\end{figure*}

\begin{figure}[t]
    \centering
    \includegraphics[width=0.9\linewidth]{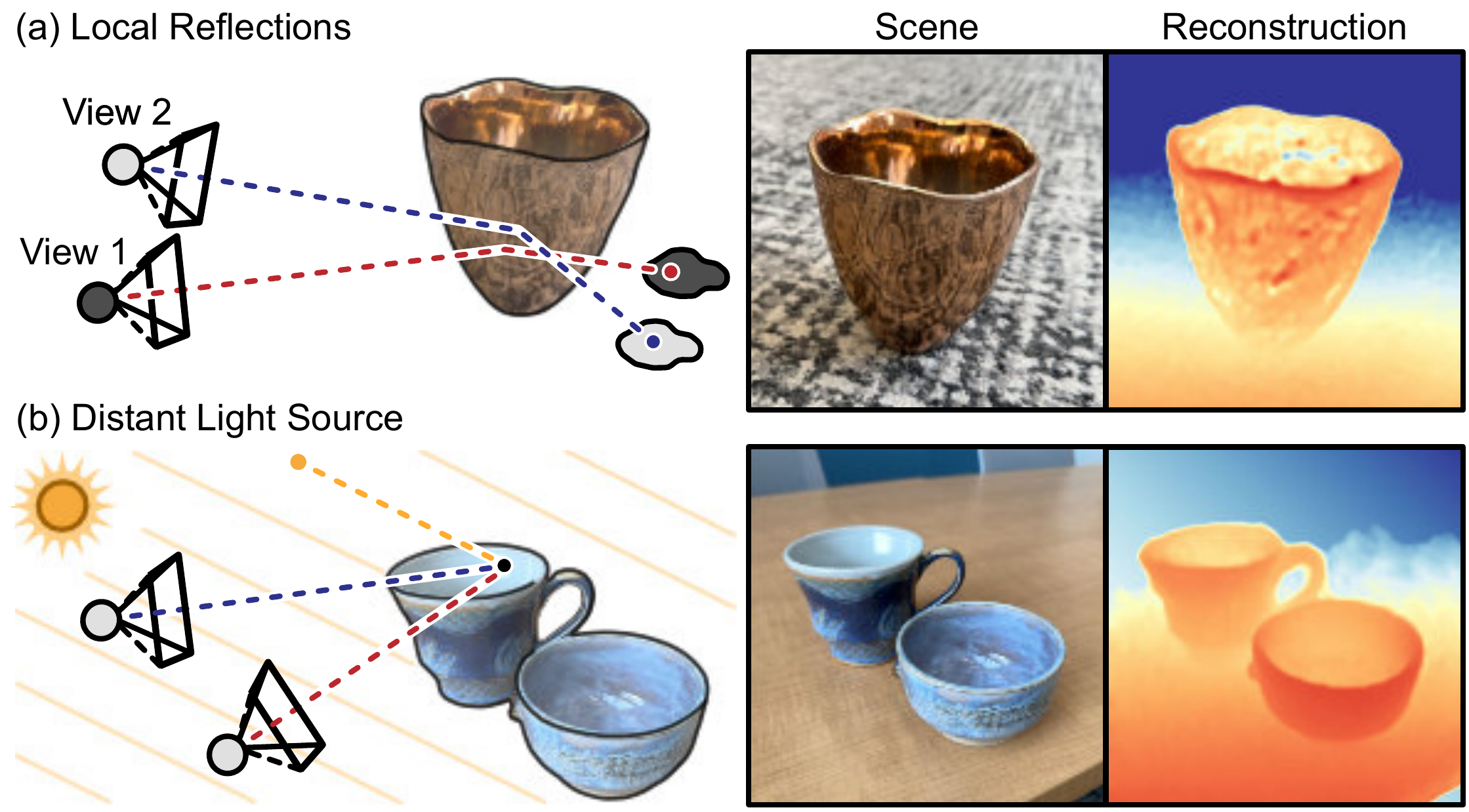}
    \caption{(a) Objects which reflect local scene content, in this example a mirror-finish copper pot reflecting the carpet around it, can completely break view consistency assumptions used for depth reconstruction. Even small view angle changes result in light paths which sample completely inconsistent colors in the surrounding environment. (b) In contrast, objects with specularities caused by a light source at effectively infinity, in this example polished ceramic reflecting sunlight, do not fully break view consistency.}
    \label{figsoap:non-lambertian}
\end{figure}

\noindent\textbf{Motion Model.} We use a B\'ezier curve model to represent translation between frames, as natural hand-tremor draws a continuous low-velocity path during capture. By limiting the number of control points $N_c$ in this model we can enforce smoothness constraints on this motion, the effects of which are illustrated in Fig.~\ref{figsoap:control_points}. Not surprisingly, using too few control points does not allow us to faithfully model camera motion and results in blurry image reconstruction and inconsistent depth estimates. We thus choose the smallest number of control points which leads to successful image and depth reconstruction. We note that while for \emph{Desk Gourds} reconstruction succeeded with $N_c=42$, for many scenes setting $N_c\,{\geq}\,42$ leads to \emph{very unstable} training as the over-defined motion model can generate erratic high-velocity motion between frames.

\begin{figure*}[htp!]
    \centering
    \includegraphics[width=\linewidth]{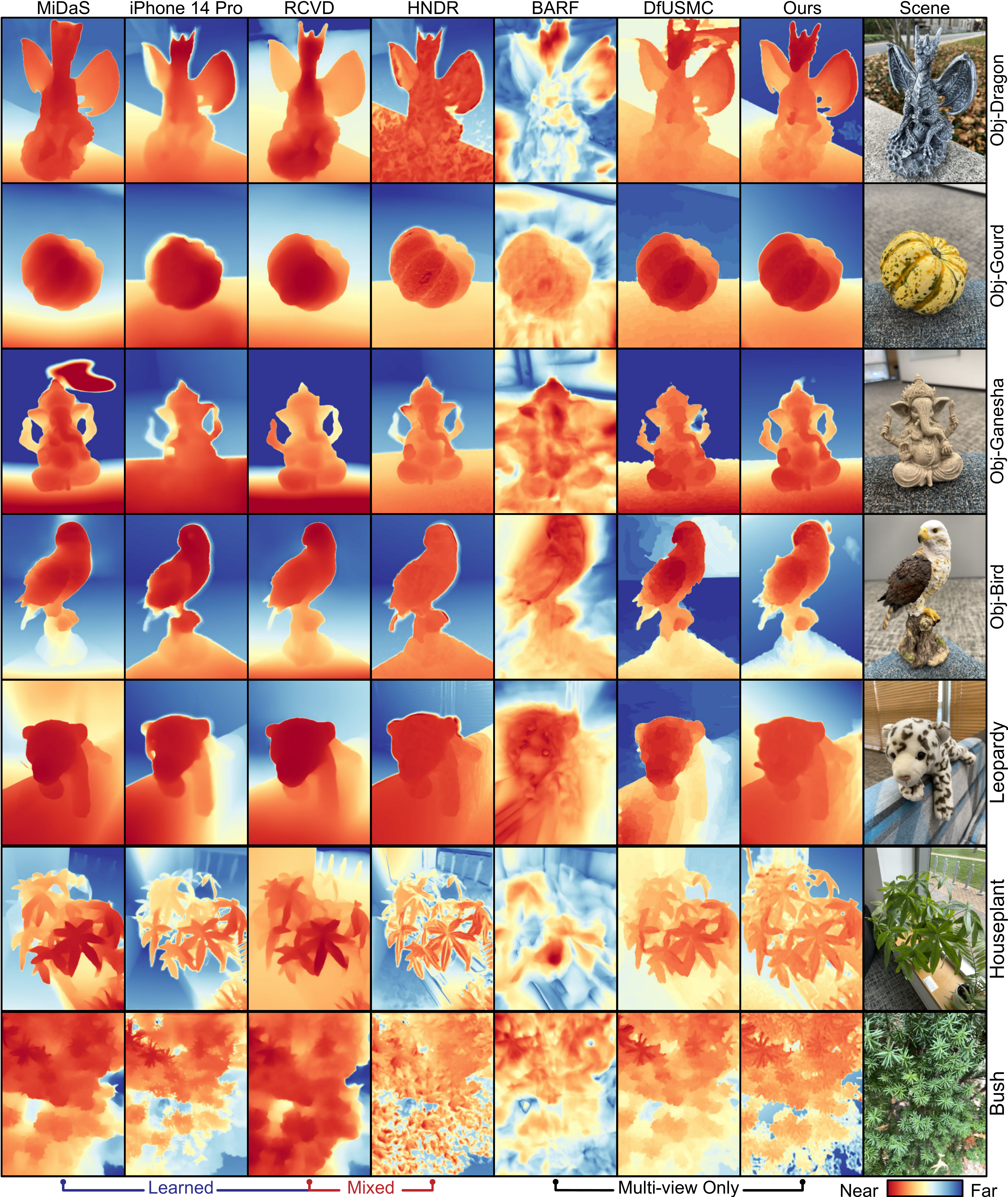}
    \caption{Reconstruction on 7 additional scenes for our method and a mix of learned, purely multi-view, and mixed depth estimation methods. Given the mix of depth representations, results are re-scaled by minimizing relative mean square error.}
    \label{figsoap:results_supp}
\end{figure*}

\section{Additional Reconstruction Results}
\noindent\textbf{Challenging Imaging Scenarios.} Given the fundamental building blocks of our approach, namely that it performs multi-view depth estimation through ray reprojection, some scenes will naturally be more difficult to reconstruct than others. As shown in Fig.~\ref{figsoap:difficult_scenes}, each of these scenarios presents its own set of challenges and direction of study. In the \emph{Dynamic} scene, we fail to reconstruct accurate depth for the majority of the plant leaves as they undergo deformation far larger than the parallax effects we observe in the long-burst. Our forward model has no way to model this deformation, and it is notoriously difficult to separate the effects of object motion from camera motion. The \emph{Textureless} and \emph{Distant} scenes present two sides of a similar problem, insufficient parallax information. While we are able to reconstruct the plant in \emph{Textureless}, the textureless planter provides no multi-view information from which to estimate depth except for along its edges, which we can track relative to the motion of the background. The church in \emph{Distant} is so far from the camera that it exhibits only fractions of pixel in disparity over the entire long-burst. In both these scenarios we need a mechanism to aggregate information in image space to make up for the lack of parallax. In \emph{Textureless} this would be in-painting the planters depth based on its edges, and in \emph{Distant} we would need to look at the deformation of larger image patches to estimate sub-pixel motion. The \emph{Thin Structures} reconstruction is partially successful, as in the foreground region we are able to track and reconstruct the depth of the thin orange mesh, but breaks down when it begins to overlap with the traffic cone. We suspect this is because our forward model is a single-layer RGB-D representation, with no explicit way to model for occlusions. In the region of the traffic cone is has to decide between reconstructing the cone or the mesh in front of it in the long-burst $I(u,v,\textsc{n})$ data, not both. Here, a layered depth representation could potentially solve this, but greatly increases the complexity of the problem as we would now need to learn an alpha map for each frame $\textsc{n}$ to sample these layers. For the \emph{Very High Dynamic Range} scene, we have specular reflections three orders of magnitude brighter than the shadowed portions of the statue. While using the fixed auto exposure and ISO settings we are able to reconstruct a large portion of the statue body with our RAW data. To reconstruct all the regions of the scene, including the dimly-lit body, our model could potentially be augmented to incorporate bracketed image data with varying exposure, similar to Mildenhall et al.~\cite{mildenhall2022nerf}, and perform joint HDR image volume and depth reconstruction. The \emph{Lens Blur} scene shows a loss in reconstruction quality due to portions of the scene being blurred by a shallow depth of field from the camera. Depth-from-defocus cues~\cite{xiong1993depth} could potentially help regularize reconstruction in these areas which are otherwise devoid of fine image features. Lastly, the \emph{Translucent} and \emph{Highly Reflective} settings both violate view consistency. Namely, changes in pixel colors can no longer be attributed solely to parallax or camera motion, and can be caused by seeing through or around the objects. We further discuss the reconstruction of non-lambertian objects in the next section.

\noindent\textbf{Non-Lambertian Reconstruction.}  While we focus on the reconstruction of primarily lambertian scenes -- matte, diffusely-reflective objects -- non-lambertian scenes provide an interesting set of both imaging challenges and opportunities. We first divide this setting into two categories: \emph{local reflections} and \emph{distant light sources}, illustrated in Fig.~\ref{figsoap:non-lambertian}. In the first setting, sampled light paths and colors can drastically change for even small view variations. As photometric matching tries to match reflected content, which does not follow the parallax motion of the reflective object itself, this produces erroneous depth estimates for objects such as the copper pot in Fig.~\ref{figsoap:non-lambertian} (a). With a distant light source, however, small changes in view angle result in the same apparent specularities as the path from the camera center to the illuminator remains connected. These specularities thus act as image texture, and exhibit the same parallax effects as the surface of the object. As seen in Fig.~\ref{figsoap:non-lambertian} (b) and the \emph{Tiger} scene, this does not disrupt depth reconstruction. This finding, that specularities from distant light sources act as object texture and local reflections do not, points towards an avenue of work in lighting separation and reflection removal. Regions which do not fit a static RGB-D model and incur large photometric penalties regardless of their depth, could be separated into view-dependent texture plus reflection components for later manipulation.

\noindent\textbf{Small Camera Motion.}  As hand shake is a naturally random process, long-burst captures have varying effective stereo baseline. While on average we can expect 5-6 millimeters of baseline~\cite{chugunov2022implicit}, if we are unlucky -- e.g. the user is not taking a breath and is rigidly holding the phone with two hands close to their body -- this motion can be as small as a millimeter. Illustrated in Fig.~\ref{figsoap:small_motion}, we see how our end--to-end camera pose estimation still converges in the minimal baseline setting, and how we are able to produce useable -- albeit blurrier -- depth estimates. This is in contrast to HNDR~\cite{chugunov2022implicit}, which uses the imperfect ARKit pose estimates for reprojection and produces major artifacts because of it, mapping incorrect pixel matches to spurious depths solutions. This demonstrates the value of continuous pose refinement, as mobile SLAM algorithms and COLMAP~\cite{schonberger2016structure} \emph{do not produce ground truth poses}.

\noindent\textbf{Additional Results.} \label{secsoap:additional_results}
Fig.~\ref{figsoap:results_supp} provides additional qualitative comparisons of our proposed approach to a wide set of baseline methods. This includes the four target objects used to demonstrate object reconstruction in the main text, prefixed with \emph{Obj-}. The visualizations also reflect the challenges in evaluating methods purely from depth maps, as geometric inconsistencies that are apparent in the mesh projections -- such as the distorted arms of \emph{Obj-Ganesha} -- are much harder to identify in these 2D visualizations. In addition to these objects, we include 3 scenes \emph{Leopardy, Bush, and Houseplant}, which demonstrate successful reconstruction with deceptive image features, small depth features, and large field of view respectively. Of particular note is how we are able to reconstruct the needles of the \emph{Bush} scene and individual leaves of \emph{Houseplant}, where other methods blend features at different depth levels together. 

\section{Synthetic Evaluation}
\label{secsoap:synthetic}
\begin{figure*}[ht!]
    \centering
    \includegraphics[width=\linewidth]{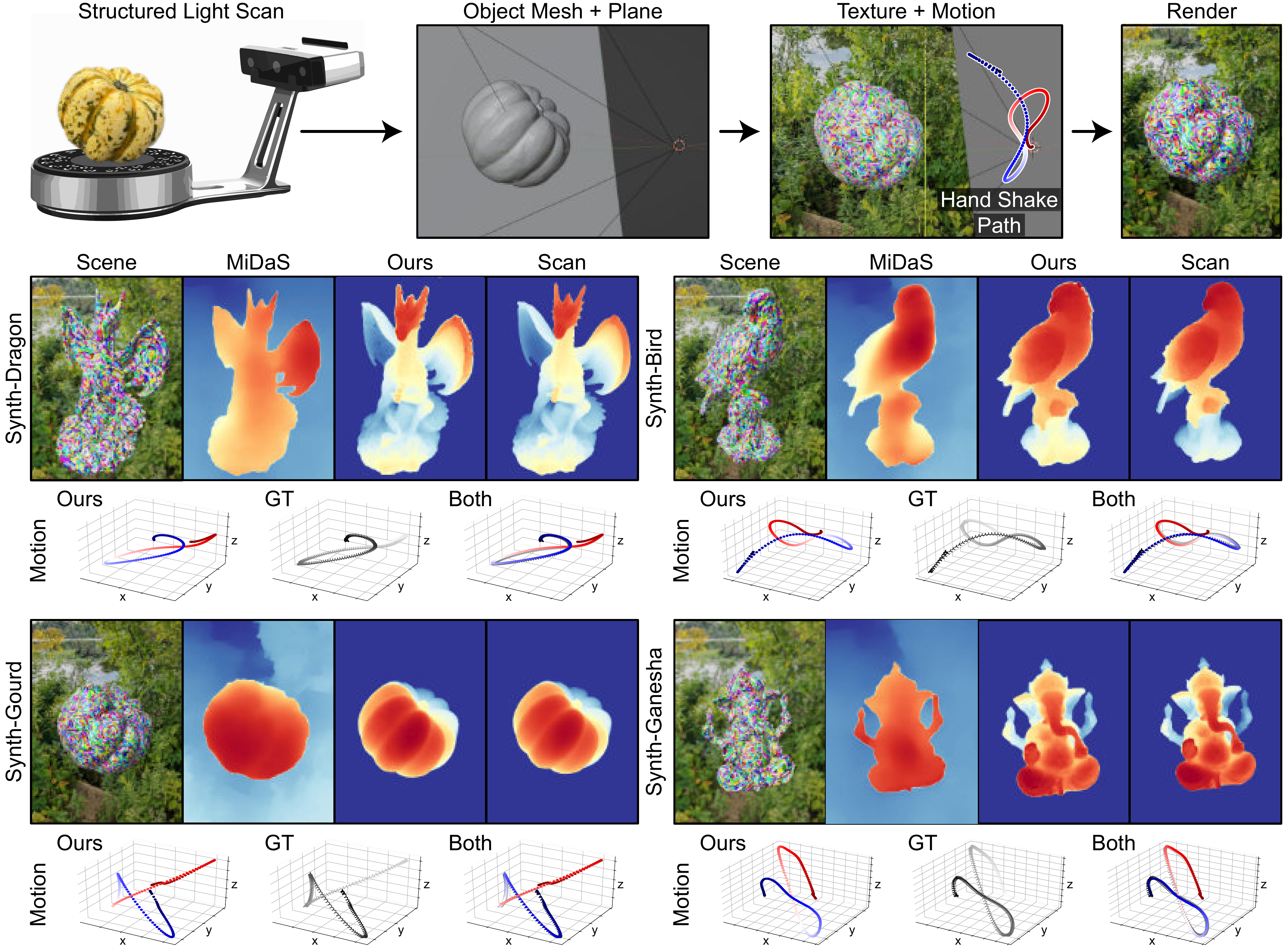}
    \caption{Depth reconstruction and motion estimation results for a set of simulated textured objects with realistic hand-tremor motion. Motion estimates are re-normalized and overlaid to demonstrate the accuracy in estimated camera trajectory to ground truth data.}
    \label{figsoap:synthetic_results}
\end{figure*}

\vspace{0.6em} \noindent\textbf{Setup.} To further validate our approach we use the high-fidelity structured light object scans we acquired for quantitative evaluation to generate simulated long-burst captures. Illustrated in Fig~\ref{figsoap:synthetic_results}, we apply a Voronoi color texture to the surface of these meshes, and place them in front of a tilted background plane with an outdoor image texture. We add depth-of-field effects and match camera intrinsics to our real captures -- using the ARKit poses captured by the software from Chugunov et al.~\cite{chugunov2022implicit} to generate realistic hand tremor motion paths -- and render frames at 16-bit color depth with Blender's Eevee engine. This synthetic data allows us to not only validate the fidelity of our object reconstructions, but also our estimated camera motion paths, for which we cannot otherwise get ground truth during ordinary captures.
\\

\noindent\textbf{Assessment on Synthetic Data.} We find that for this synthetic data -- in the absence of noise, lighting changes, and other imaging non-idealities -- we are able to recover nearly ground truth reconstructions of both the objects and background planes. This supports our plane plus offset depth model, which fits the simple plane to the out-of-focus background content instead of generating spurious depth estimates for regions without reliable parallax information. Though the colorful object textures make single-view depth estimation visually difficult, as illustrated by artifacts in the MiDaS reconstructions, these high-contrast cues allow our method to reconstruct even tiny features such as the tusks of the \emph{Synth-Ganesha}. This validates that even with small camera motion, given sufficient image texture we converge to geometrically correct solutions. In Fig.~\ref{figsoap:synthetic_results} we also see how the camera motion estimates converge close to ground truth as our method jointly refines depth and camera trajectory estimates during training. 

\section{Depth and Image Matting}
\label{secsoap:experiments}
\begin{figure*}[t!]
    \centering
    \includegraphics[width=\linewidth]{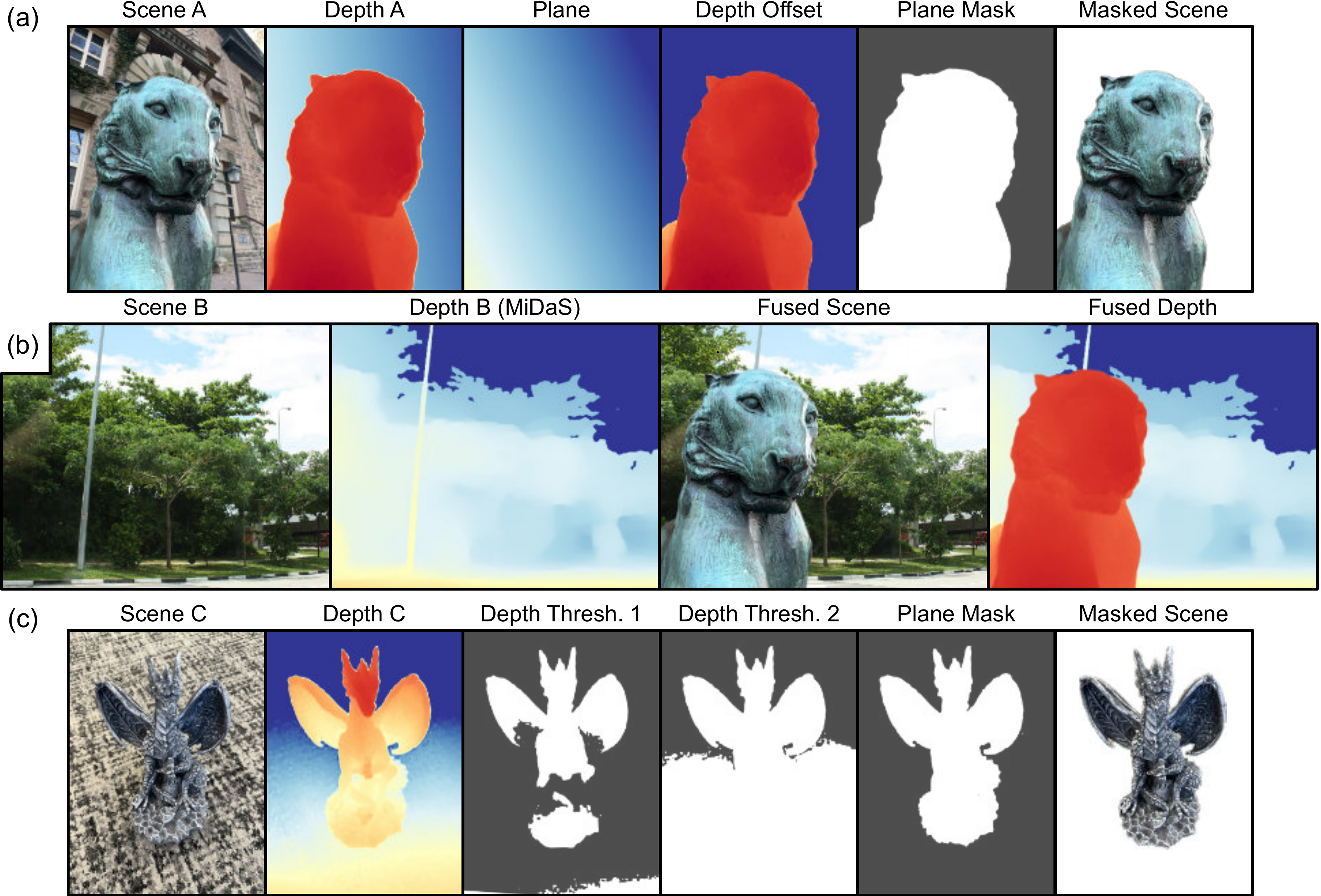}
    \caption{Image and Depth Matting. Example of scene editing enabled by our plane plus offset forward model. We can (a) threshold the depth offset component $d(u,v) - d_\textsc{p}$ to recover a mask of the object in focus and then (b) superimpose it over a new scene. (c) This works even for visually ambiguous scenes where simple depth thresholding fails.}
    \label{figsoap:editting}
\end{figure*}
	
\noindent\textbf{Forward Model Decomposition.} In the proposed plane plus offset depth model, regions that do not generate sufficient parallax information are pulled towards the plane by the regularization term $R$. While we cannot recover meaningful depth from multiview in these regions, they prove \emph{useful for scene segmentation and editing}. Illustrated in Fig.~\ref{figsoap:editting} (a), by masking what parts of the image produce negligible depth offset, we are able to cleanly segment the tiger statue in \emph{Scene A} from its background. In Fig.~\ref{figsoap:editting} (b) we then superimpose this masked image over \emph{Scene B}, a separately captured tree-covered street. We run \emph{Scene B} through MiDaS to hallucinate the depth of the background trees, and overlay this with our geometrically-estimated depth of the tiger to produce a fused depth representation. In this way we leverage multiview information where we have it, and learned image priors where we do not. In Fig.~\ref{figsoap:editting} (c) we see an advantage of using this plane separation technique for segmentation over depth thresholding. As the floor under the dragon figure extends both in front of and behind the figure itself, setting a depth cutoff will always either miss a part of the figure, or include the area around it. Whereas as our plane here represents the depth of the floor, we can threshold the depth offset just like in Fig.~\ref{figsoap:editting} (a) to recover a \emph{high-quality mask of the object}. Thanks to being based on depth rather than image features, this approach has no problems with the visual ambiguity of the dragon and its background, which both contain high-frequency black and white textures.


\graphicspath{{chapters/1/ImplicitSupplemental/}}
\section{Hand Shake Analysis}
\label{sechndr:hand_vis}
\begin{figure*}[h]
    \centering
    \includegraphics[width=\linewidth]{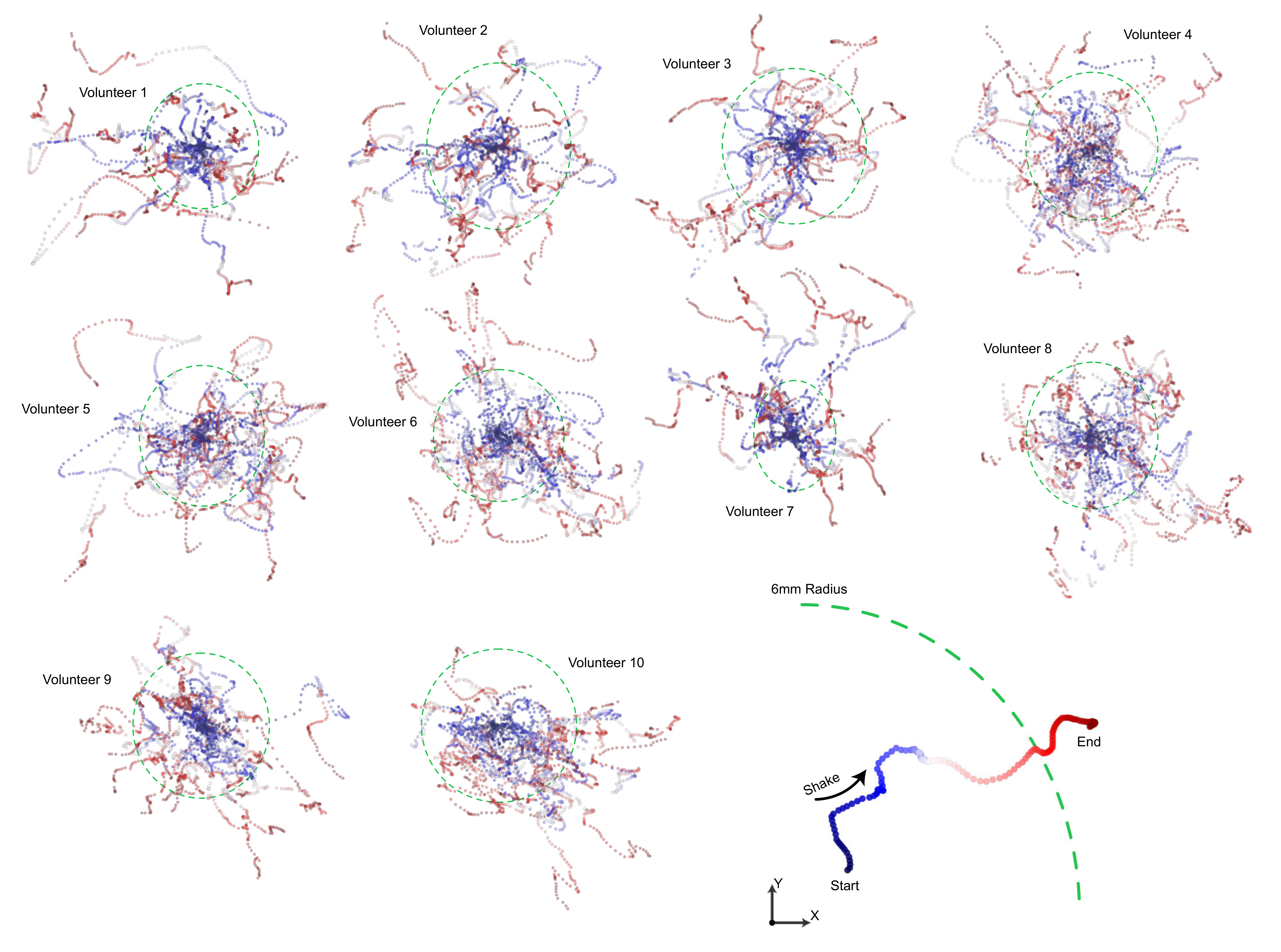}
    \caption{Hand-shake point clouds for individual volunteers. Each hand shake path is recentered to start at the same location, and follows the progression from blue to red over the recorded $N=120$ frames. Each green dashed oval illustrates a radius of 6mm from the center of the starting point of these hand shake paths, and serves to denote the scale of the point cloud. We find that both the scale and symmetry of hand shake paths is user-dependent, with volunteer 7 exhibiting large asymmetric hand tremors, and volunteer 4 small relatively symmetric hand shakes. }
    \label{fighndr:hand_shake_clouds}
\end{figure*}

In Figure~\ref{fighndr:hand_shake_clouds} we present per-individual point clouds which we create by aggregating all the hand shake paths recorded by each of our 10 volunteers, as outlined in Section 4 of \cite{chugunov2022implicit}.

\newpage

\begin{table}[h]
    \centering
    \resizebox{\linewidth}{!}{%
    \begin{tabular}{ c c c c  c  c  c c}
        \toprule
        Bundle Length $N$: & 120 Frames & 100 Frames & 80 Frames & 60 Frames & 30 Frames & 15 Frames \\
        \midrule
        \midrule
        \makecell{Median Effective \\ Baseline [mm]} & 5.75 & 5.34 & 4.60 & 4.021 & 2.62 & 1.5\\
        \midrule
        \makecell{Fraction of Effective \\ Baselines $>$ 5mm} & 0.589 & 0.517 & 0.461 & 0.369 & 0.183 & 0.06\\ %
        \midrule
        \makecell{Fraction of Effective \\ Baselines $>$ 3mm} & 0.844 & 0.794 & 0.756 & 0.661 & 0.433 & 0.19\\ %
        \bottomrule
    \end{tabular}%
    }
    \vspace{0.1em}
    \caption{\label{tabhndr:hand_shake_stats}%
    Quantitative analysis of the maximum displacement, or equivalently the effective baseline, of recorded $N$ frame bundles. We investigate the median effective baseline of these bundles, as well as the fraction of bundles which achieve a greater than 5mm and 3mm effective baseline.
    }
\end{table}

\begin{figure*}[h]
    \centering
    \includegraphics[width=0.9\linewidth]{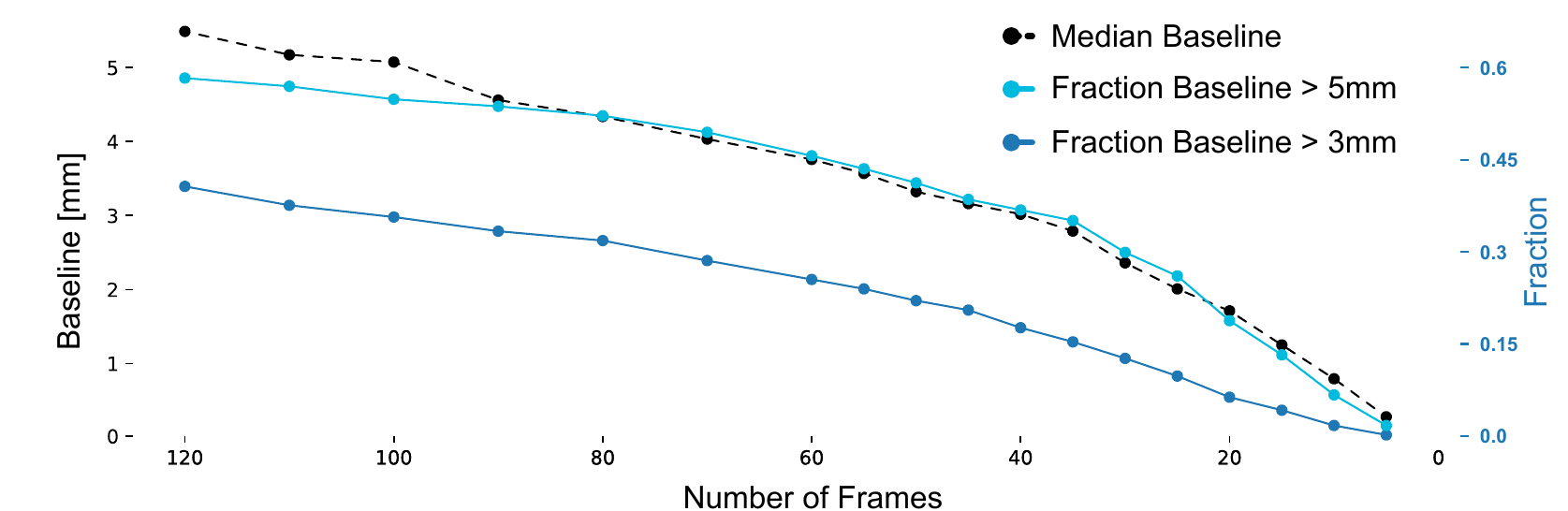}
    \caption{Plot to illustrate the hand shake statistics presented in Table~\ref{tabhndr:hand_shake_stats}. We see that for frame count $120 \geq N \geq 60$ there is not a sharp drop in the fraction of bundles which record a  $>$5mm maximum displacement.}
    \label{fighndr:displacement}
    \vspace{-1em}
\end{figure*}

\vspace{0.5em}\noindent\textbf{Probabilistic View of Hand Shake Bundles.}\hspace{0.1em} 
Given the randomness of natural hand tremor present during a snapshot recording, there is a corresponding randomness in the effective baseline of a recorded length $N$ bundle. While we can consider the direct correlation between $N$ and phone displacement during recording, where we expect the phone to move farther as we record longer, we offer a more probabilistic view of this process. Given a target for micro-baseline depth reconstruction, for example an effective 5mm or 3mm baseline, we can ask with what probability we expect a length $N$ recorded bundle to provide meet this target. Looking to Table~\ref{tabhndr:hand_shake_stats}, we find that for our collected hand shake data $N=100$ frames is sufficient to record a 5mm maximum displacement approximately half the time. As seen in Figure~\ref{fighndr:displacement}, bundles between $N=60$ and $N=120$ frames in length offer a reasonable probability ($>\frac{1}{3}$) of capturing $>$5mm baseline data. Thus, much like in multi-frame superresolution~\cite{tsai1984multiframe,wronski2019handheld}, we can limit the amount of data we acquire during one snapshot bundle if we expect the phone photographer to take multiple snapshots of an object of interest -- looking for the perfect angle, a sharp photo -- one of which we expect to achieve our target baseline.

\chapter{Layer Separation\label{ch:layer}}

This chapter explores the problem of image layer separation: recovering both a scene hidden behind obstructions such as occlusions and reflections and reconstructing the obstruction itself. We operate under the same capture setting as described in Chapter \ref{ch:depth}, where images are acquired with small natural hand motion between frames during view-finding. To separate layers, we leverage the concept of \textit{common fate}~\cite{ullman1979interpretation}: that image features that ``move together belong together". However, this requires us again to densely track the motion of millions of pixels, motion that is no longer constrained to rigid 3D projection. To address this, we develop a novel flow representation which we call a \textit{Neural Spline Field} (NSF). This is a neural field model which maps input spatial coordinates to vectors of spline control points, that can be later indexed into and interpolated to calculate flow values. This NSF model is designed to be highly controllable, with spatial smoothness determined by the network's coordinate encoding parameters, and temporal behavior determined by the parametrization of the interpolating spline function. We demonstrate how these NSFs can fit smooth image motion without any regularization function, accurately separating reflections, occlusions, shadows from background scene content.

\vspace{1em}
\hrule
\vspace{1em}
\noindent  \textit{This chapter is based on the work ``Neural Spline Fields for Burst Image Fusion and Layer Separation"~\cite{chugunov2024neural} by Ilya Chugunov, David Shustin, Ruyu Yan, and Felix Heide presented at CVPR 2024.}

\graphicspath{{chapters/2/NSF/}}
 \vspace{-1em}
\section{Introduction}
%
Over the last decade, as digital photos have increasingly been produced by smartphones, smartphone photos have increasingly been produced by burst fusion. To compensate for less-than-ideal camera hardware -- typically restricted to a footprint of less than 1cm$^3$~\cite{blahnik2021smartphone} -- smartphones rely on their advanced compute hardware to process and fuse multiple lower-quality images into a high-fidelity photo~\cite{delbracio2021mobile}. This proves particularly important in low-light and high-dynamic-range settings~\cite{liba2019handheld, hasinoff2016burst}, where a single image must compromise between noise and motion blur, but multiple images afford the opportunity to minimize both~\cite{kalantari2017deep}. But even as mobile night- and astro-photography applications~\cite{google2018night,google2019astrophotography} use increasingly long sequences of photos as input, their output remains a static single-plane image. Given the typically non-static and non-planar nature of the real world, a core problem in burst image pipelines is thus the alignment~\cite{lecouat2022high, mildenhall2018burst} and aggregation~\cite{wronski2019handheld, bhat2021deep} of pixels into an image array -- referred to as the \textit{align-and-merge} process.

\begin{figure}[t]
    \centering
    \includegraphics[width=0.7\linewidth]{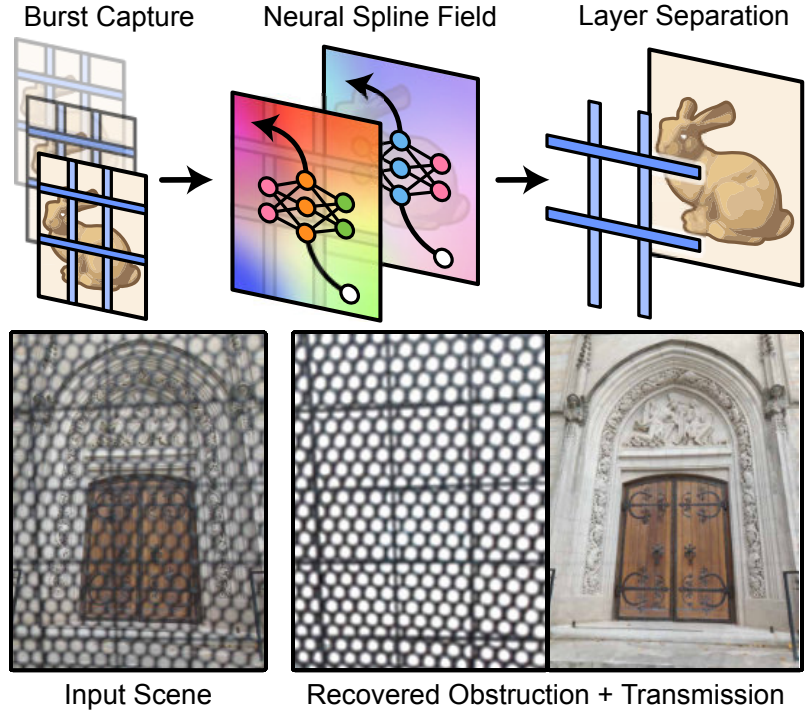}
    \caption{Fitting our two-layer neural spline field model to a stack of images we're able to directly estimate and separate even severe, out-of-focus obstructions to recover hidden scene content.}
    \label{fignsf:teaser}
\end{figure}

While existing approaches treat pixel motion as a source of noise and artifacts, a parallel direction of work~\cite{chugunov2022implicit,yu20143d,ha2016high} attempts to extract useful parallax cues from this pixel motion to estimate the geometry of the scene. Recent work by Chugunov et al.~\cite{chugunov2023shakes} finds that maximizing the photometric consistency of an RGB plus depth neural field model of an image sequence is enough to distill dense depth estimates of the scene. While this method is able to jointly estimate high-quality camera motion parameters, it does not perform high-quality image reconstruction, and rather treats its image model as ``a vehicle for depth optimization"~\cite{chugunov2023shakes}. In contrast, work by Nam et al.~\cite{nam2022neural} proposes a neural field fitting approach for multi-image fusion and layer separation which focuses on the quality of the reconstructed ``canonical view". By swapping in different motion models, they can separate and remove layers such as occlusions, reflections, and moir\'e patterns during image reconstruction -- as opposed to in a separate post-processing step~\cite{shih2015ghosting,gupta2019fully}. This approach, however, does not make use of a realistic camera projection model, and relies on regularization penalties to discourage its motion models from representing non-physical effects -- e.g., pixel tearing or teleportation.

In this work, we propose a versatile layered neural image representation~\cite{nam2022neural} with a projective camera model~\cite{chugunov2023shakes} and novel neural spatio-temporal spline~\cite{ye2022deformable} parametrization. Our model takes as input an unstabilized 12-megapixel RAW image sequence, camera metadata, and gyroscope measurements -- available on all modern smartphones. During test-time optimization, it fits to produce a high-resolution reconstruction of the scene, separated into \textit{transmission} and \textit{obstruction} image planes. The latter of which can be extracted to perform occlusion removal, reflection suppression, and other layer separation applications. To this end, we decompose pixel motion between burst frames into planar motion, from the camera's pose change in 3D space relative to the image planes, and a generic flow component which accounts for depth parallax, scene motion, and other image distortions. We model these flows with neural spline fields (NSFs): networks trained to map input coordinates to spline control points, which are then interpolated at sample timestamps to produce flow field values. As their output dynamics are strictly bound by their spline parametrization, these NSFs produce \textit{temporally} consistent flow with no regularization, and can be controlled \textit{spatially} through the manipulation of their positional encodings. 

\noindent In summary, we make the following contributions: \vspace{-0.5em}
\begin{itemize}
    \item  An end-to-end neural scene fitting approach which fits to a burst image sequence to distill high-fidelity camera poses, and high-resolution two layer transmission plus occlusion image decomposition. \vspace{-0.5em}
    \item A compact, controllable neural spline field model to estimate and aggregate pixel motion between frames. \vspace{-0.5em}
    \item Qualitative and quantitative evaluations which demonstrate that our model outperforms existing single image and multi-frame obstruction removal approaches. \vspace{-0.5em}
\end{itemize}
Code, data, videos, and additional materials are available on our project website {\color{URLBlue}\href{https://light.princeton.edu/publication/nsf/}{light.princeton.edu/nsf}}.

\section{Related Work} 

\noindent\textbf{Burst Photography.}\hspace{0.1em}
A large body of work has explored methods for burst image processing~\cite{delbracio2021mobile} to achieve high image quality in mobile photography settings. During burst imaging, the device records a sequence of frames in rapid succession -- potentially a \textit{bracketed sequence} with varying exposure parameters~\cite{mertens2009exposure} -- and fuses them post-capture to produce a demosaiced~\cite{tan2017joint}, denoised~\cite{mildenhall2018burst, godard2018deep}, superresolved~\cite{wronski2019handheld,lecouat2022high}, or otherwise enhanced reconstruction. Almost all modern smartphone devices rely on burst photography for low-light~\cite{liba2019handheld,hasinoff2016burst} and high dynamic range reconstruction from low dynamic range sensors~\cite{hasinoff2016burst,gallo2015locally}. While existing methods typically use sequences of only 2-8 frames as input, a parallel field of micro-video~\cite{yu20143d,im2015high} or ``long-burst photography"~\cite{chugunov2023shakes} research -- which also encompasses widely deployed Apple Live Photos, Android Motion Photos, and night photography~\cite{google2018night,google2019astrophotography} -- consumes sequences of images up to several seconds in length, acquired naturally during camera viewfinding. Though not limited to long-burst photography, we adopt this setting to leverage the parallax~\cite{xue2015computational} and pixel motion cues in these extended captures for separation of obstructed and transmitted scene content.

\noindent\textbf{Obstruction Removal and Layer Separation.}\hspace{0.1em} 
While their use of visual cues is diverse -- e.g., identifying reflections from ``ghosting" cues on thick glass~\cite{shih2015ghosting} or detecting lattices for fence deletion~\cite{park2011image} -- single-image obstruction removal is fundamentally a segmentation~\cite{liu2022semantic, kume2023singlefft} and image recovery~\cite{hu2023single, gandelsman2019double} problem. In the most severe cases, with fully opaque occluders, this image recovery problem becomes an in-painting task~\cite{xiong2019foreground, farid2016image} to synthesize missing content. This is in contrast to approaches which rely on multiple measurements such as multi-focal stacks~\cite{shen2023light,adeel2022defencing}, multi-view images~\cite{niklaus2020learned,liu2020learning}, flash no-flash pairs~\cite{lei2021robust,lei2022robustwild}, or polarization data~\cite{lei2020polarized}. These methods typically treat obstruction removal as an inverse problem~\cite{bertero2021introduction}, estimating a model of transmitted and occluded content consistent with observed data~\cite{li2013exploiting}. This can also be generalized to an image layer separation problem, an example of which is intrinsic decomposition~\cite{chen2013simple}, where the separated layer is the obstruction. These methods typically rely on learned priors~\cite{gandelsman2019double} and pixel motion~\cite{nam2022neural} to decompose images into multiple components. Our work explores the layer separation problem in the burst photography setting, where pixel motion is on a much smaller scale than in video sequences~\cite{lu2021omnimatte}, and a high-resolution unobstructed view is desired as an output. Rather than tailor to a single application, however, we propose a unified model with applications to reflection, occlusion, and shadow separation.

\noindent\textbf{Neural Scene Representations.}\hspace{0.1em} A growing body of work investigating novel view synthesis has demonstrated that coordinate-based neural representations are capable of reconstructing complex scenes~\cite{barron2023zip,barron2021mip} without an explicit structural backbone such as a pixel array or voxel grid. These networks are typically trained from scratch, through \textit{test-time optimization}, on a single scene to map input coordinate encodings~\cite{tancik2020fourier} to outputs such as RGB~\cite{sitzmann2020implicit}, depth~\cite{chugunov2022implicit}, or x-ray data~\cite{sun2021coil}. While neural scene representations require many network evaluations to generate outputs, as opposed to explicit representations which can be considered ``pre-evaluated", recent works have shown great success in accelerating training~\cite{muller2022instant} and inference~\cite{yu2021plenoctrees} of these networks. Furthermore, this per-output network evaluation is what lends to their versatility, as they can be optimized through auto-differentiation with no computational penalties for sparse or non-uniform sampling of the scene~\cite{kjolstad2017tensor}. Several recent approaches make use of neural scene representations in tandem with continuous motion estimation models to fit multi-image~\cite{chugunov2023shakes} and video~\cite{li2023dynibar} data, potentially decomposing it into multiple layers in the process~\cite{nam2022neural, kasten2021layered}. Our work proposes a novel neural spline field continuous flow representation with a projective camera model to separate effects such as occlusions, reflections, and shadows. In contrast to existing approaches, our flow model does not require regularization to prevent overfitting, as its representation power is controlled directly through encoding and spline hyperparameters.

\section{Neural Spline Fields for Burst Photography}

\noindent We begin with a discussion of the proposed neural spline field model of optical flow. We then continue with our full two-layer projective model of burst photography, its loss functions, training procedure, and data collection pipeline.
\subsection{Neural Spline Fields.}
\begin{figure}[t]
    \centering
    \includegraphics[width=0.8\linewidth]{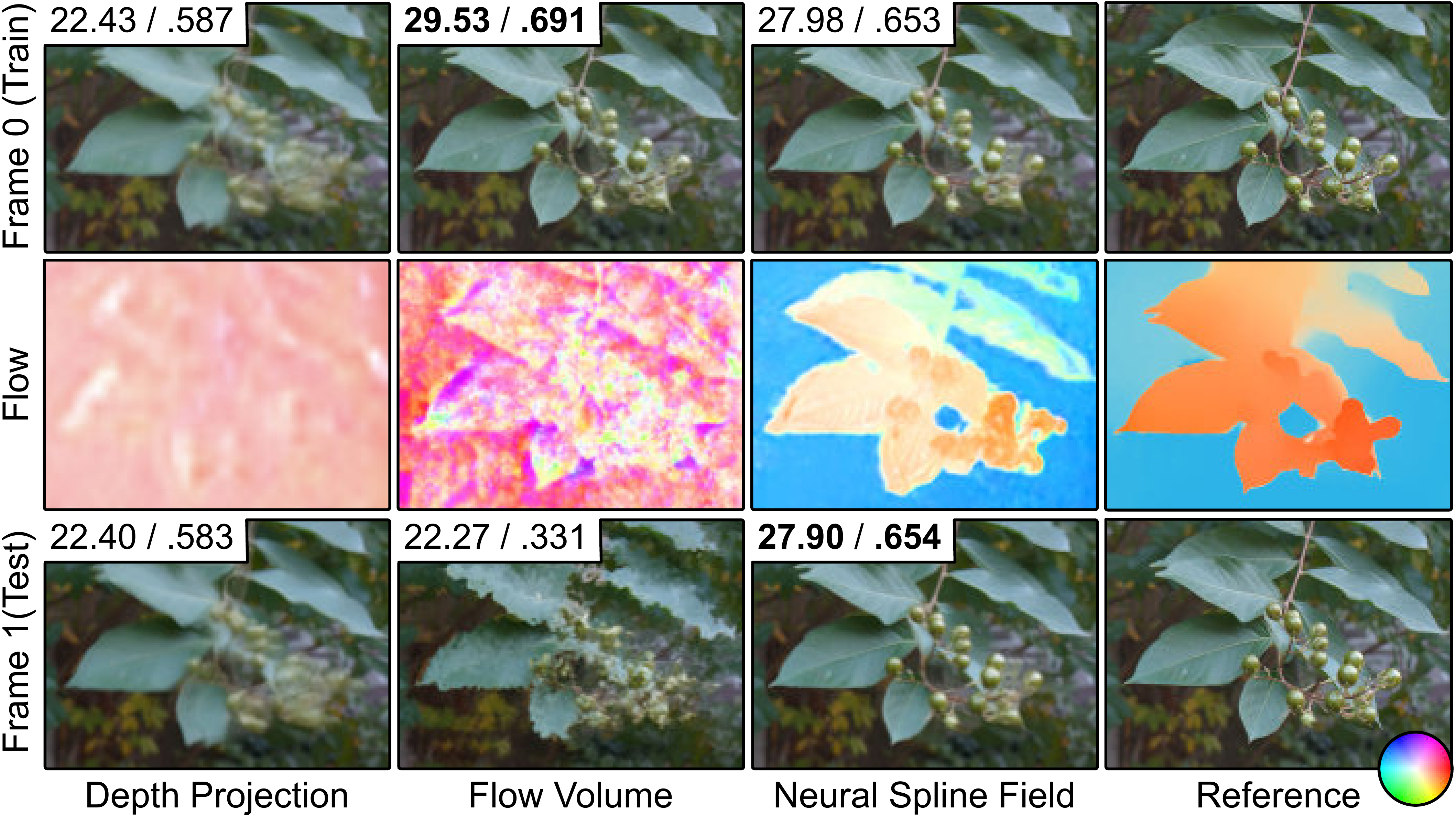}
    \caption{Image and flow estimates for different representations of a short video sequence of a swinging branch; PSNR/SSIM values inset top-left. Depth projection alone is unable to represent both parallax and scene motion, mixing reconstructed content, and an un-regularized 3D flow volume $g(u,v,t)$ trivially overfits to the sequence. With an identical network, spatial encoding, loss function, and training procedure as $g(u,v,t)$, our neural spline field $S(t; \mathbf{P} = h(u,v))$ produces temporally consistent flow estimates well-correlated with a conventional optical flow reference~\cite{lipson2021raft}.}
    \label{fignsf:flow_representations}
    \vspace{-1em}
\end{figure}

\noindent\textbf{Motivation.}\hspace{0.1em} To recover a latent image, existing burst photography methods \textit{align and merge}~\cite{delbracio2021mobile} pixels in the captured image sequence. Disregarding regions of the scene that spontaneously change -- e.g., blinking lights or digital screens -- pixel differences between images can be decomposed into the products of scene motion, illuminant motion, camera rotation, and depth parallax. Separating these sources of motion has been a long-standing challenge in vision~\cite{vogel2013piecewise,teed2021raft} as this is a fundamentally ill-conditioned problem; in typical settings, scene and camera motion are geometrically equivalent~\cite{hartley2003multiple}. One response to this problem is to disregard effects other than camera motion, which can yield high-quality motion estimates for static, mostly-lambertian scenes~\cite{yu20143d,im2015high,chugunov2023shakes}. This can be represented as
\begin{equation}\label{eqnsf:sample_depth}
I(u,v,t) = [R,G,B] = f(\bm{\pi}\bm{\pi}_t^{-1}(u,v)),
\end{equation}
where $I(u,v,t)$ is a frame from the burst stack captured at time $t$ and sampled at image coordinates $u,v\in[0,1]$. Operators $\bm{\pi}$ and $\bm{\pi}_t$ perform 3D reprojection on these coordinates to transform them from time $t$ to the coordinates of a reference image model $f(u,v) \rightarrow [R,G,B]$. To account for other sources of motion, layer separation approaches such as \cite{kasten2021layered, nam2022neural} estimate a generic flow model $\Delta u, \Delta v = g(u,v,t)$ to re-sample the image model
\begin{equation}\label{eqnsf:sample_flow}
I(u,v,t) = f(u + \scalebox{0.7}{$\Delta$} u, v + \scalebox{0.7}{$\Delta$} v).
\end{equation}
However, this parametrization introduces an overfitting risk, the consequences of which are illustrated in Fig.~\ref{fignsf:flow_representations}, as $g(u,v,t)$ and $f(u,v)$ can now act as a generic video encoder~\cite{li2023dynibar}. To combat this, methods often employ a form of gradient penalty such as total variation loss~\cite{nam2022neural}. That is
\begin{align}
    \mathcal{L}_{\text {TVFlow }}=\sum\left\|J_{g}(u, v, t)\right\|_1,\nonumber
\end{align}
where $J_g(u, v, t)$ is the Jacobian of the flow model. During training, this can prove computationally expensive, however, as now each sample requires its local neighborhood to be evaluated to numerically estimate the Jacobian, or a second gradient pass over the model. In both cases, a large number of operations are spent to limit the reconstruction of high frequency spatial and temporal content.

\noindent\textbf{Formulation.}\hspace{0.1em} We propose a neural spline field (NSF) model of flow, a learned spatio-temporal spline~\cite{ye2022deformable} representation which provides strong controls on reconstruction directly through its parametrization. This model splits flow evaluation into two components
\begin{equation}\label{eqnsf:spline_field}
    \Delta u, \Delta v = g(u,v,t) = S(t; \mathbf{P} = h(u,v)).
\end{equation}
Here $h(u,v)$ is the NSF, a network which maps image coordinates to a set of spline control points $\mathbf{P}$. Then, to estimate flow for a frame at time $t$ in the burst stack, we evaluate the spline at $S(t;\mathbf{P})$. We select a cubic Hermite spline
\begin{align}\label{eqnsf:cubic_spline}
    S(t,\mathbf{P}) &= (2t_r^3 - 3t_r^2 + 1) \mathbf{P}_{\lfloor t_s \rfloor} + (-2t_r^3 + 3t_r^2) \mathbf{P}_{\lfloor t_s \rfloor + 1} \nonumber\\
    &+ (t_r^3 - 2t_r^2 + t_r)(\mathbf{P}_{\lfloor t_s \rfloor} - \mathbf{P}_{\lfloor t_s \rfloor - 1})/2 \nonumber \\ 
      &+ (t_r^3 - t_r^2)(\mathbf{P}_{\lfloor t_s \rfloor + 1} - \mathbf{P}_{\lfloor t_s \rfloor})/2\nonumber \\
    t_r &= t_s - \lfloor t_s \rfloor, \quad t_s = t \cdot |\mathbf{P}|,
\end{align}
as it guarantees continuity in time with respect to its zeroth, first, and second derivatives and allows for fast local evaluation -- in contrast to B\'ezier curves~\cite{chugunov2023shakes} which require recursive calculations. We emphasize that the use of splines in graphics problems is {extensive}~\cite{farin2002curves}, and that there are many alternate candidate functions for $S(t,\mathbf{P})$. E.g., if the motion is expected to be a straight line, a piece-wise linear spline with $|\mathbf{P}|=2$ control points would insure this constraint is satisfied irrespective of the outputs of $h(u,v)$.

\begin{figure}[t]
    \centering
    \includegraphics[width=0.8\linewidth]{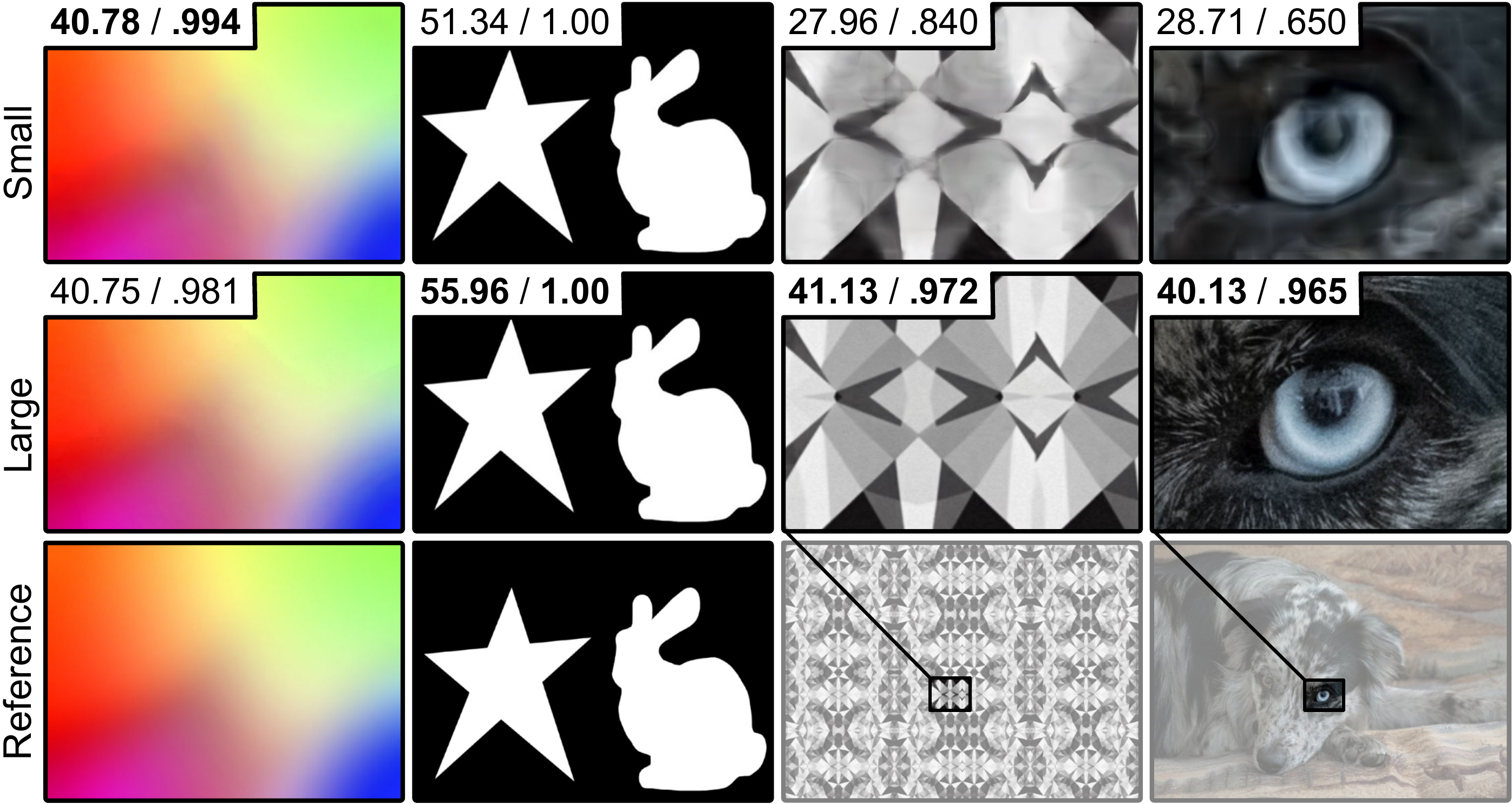}
    \caption{Image fitting results for coordinate networks with \textit{Small} ($\mathrm{L}^\gamma{=}8$) and \textit{Large} ($\mathrm{L}^\gamma{=}16$) multi-resolution hash encodings and identical other parameters; PSNR/SSIM values inset top-left. Unlike a traditional band-limited representation~\cite{yang2022polynomial}, the \textit{Small} resolution network is able to fit both low-frequency smooth gradients and sharp edge mask images, but fails to fit a high density of either. This makes it a promising candidate representation for scene flow and alpha mattes which are comprised of smooth gradients and a limited number of object edges. }
    \label{fignsf:neural_fitting}
    \vspace{-1em}
\end{figure}

\noindent Where the choice of $S(t,\mathbf{P})$ and $|\mathbf{P}|$ determines the temporal behavior of flow, $h(u,v)$ controls its spatial properties. While our method, in principle, is not restricted to a specific spatial encoding function, we adopt the multi-resolution hash encoding $\gamma(u,v)$ presented in M\"uller et al.~\cite{muller2022instant}
\begin{align}\label{eqnsf:multires-hash}
    h(u,v) &= \mathbf{h}(\gamma(u,v;\, \mathrm{params}_\gamma);\, \theta)\nonumber \\
    \mathrm{params}_\gamma &= \left\{\mathrm{B}^\gamma, \mathrm{S}^\gamma, \mathrm{L}^\gamma, \mathrm{F}^\gamma, \mathrm{T}^\gamma\right\},    
\end{align}
as it allows for fast training and strong  spatial controls given by its encoding parameters $\mathrm{params}_\gamma$: base grid resolution $\mathrm{B}^\gamma$, per level scale factor $\mathrm{S}^\gamma$, number of grid levels $\mathrm{L}^\gamma$, feature dimension $\mathrm{F}^\gamma$, and backing hash table size $\mathrm{T}^\gamma$. Here, $\mathbf{h}(\gamma(u,v);\theta)$ is a multi-layer perceptron (MLP)~\cite{hornik1989multilayer} with learned weights $\theta$. Illustrated in Fig.~\ref{fignsf:neural_fitting} with an image fitting example, the number of grid levels $\mathrm{L}^\gamma$ -- which, with a fixed $\mathrm{S}^\gamma$, sets the maximum grid resolution -- provides controls on the maximum ``spatial complexity" of the output while still permitting accurate reconstruction of image edges.


\subsection{Projective Model of Burst Photography} 
\begin{figure*}[t!]
    \centering
    \includegraphics[width=\linewidth]{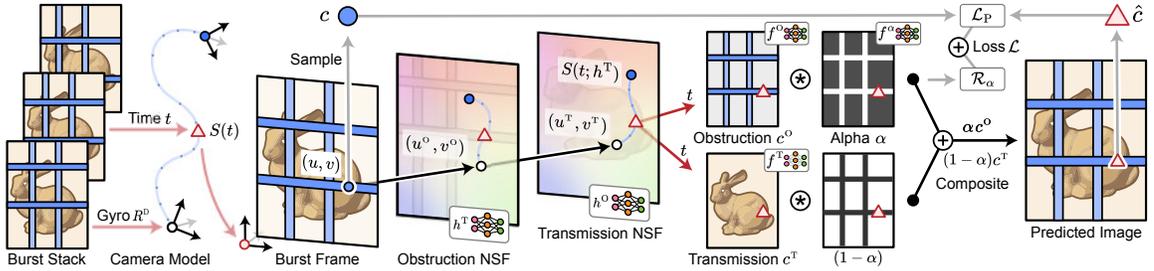}
    \caption{We model an input image sequence as the alpha composition of a \textit{transmission} and \textit{obstruction} plane. Motion in the scene is expressed as the product of a rigid camera model, which produces global rotation and translation, and two neural spline field models, which produce local flow estimates for the two layers. Trained to minimize photometric loss, this model separates content to its respective layers.}
    \label{fignsf:method}
    \vspace{-1.5em}
\end{figure*}

\noindent\textbf{Motivation.}\hspace{0.1em} With a flow model $g(u,v,t)$, and a canonical image representation $f(u,v)$ in hand, we theoretically have all the components needed to model an arbitrary image sequence~\cite{kasten2021layered,nam2022neural}. However, handheld burst photography does \textit{not} produce arbitrary image sequences; it has well-studied photometric and geometric properties~\cite{chugunov2023shakes,chugunov2022implicit,ha2016high,wronski2019handheld}. This, in combination with the abundance of physical metadata such as gyroscope values and calibrated intrinsics available on modern smartphone devices~\cite{chugunov2023shakes}, provides strong support for a physical model of image formation.  \\
\noindent\textbf{Formulation.}\hspace{0.1em} We adopt a forward model similar to traditional multi-planar imaging~\cite{hartley2003multiple}. We note that this departs from existing work~\cite{chugunov2022implicit,chugunov2023shakes}, which employs a backward projection camera model -- ``splatting" points from a canonical representation to locations in the burst stack. A multi-plane imaging model allows for both simple composition of multiple layers along a ray -- a task for which backward projection is not well suited -- and fast calculation of ray intersections without the ray-marching needed by volumetric representations like NeRF~\cite{mildenhall2020nerf}. For simplicity of notation, we outline this model for a single projected ray below. We also illustrate this process in Fig.~\ref{fignsf:method}. Let
\begin{align}
    c = [\mathrm{R},\mathrm{G},\mathrm{B}]^\top = I(u,v,t)
\end{align}
be a colored point sampled at time $t$ in the burst stack at image coordinates  $u,v\in [0,1]$. Note that these coordinates are relative to the camera pose at time $t$; for example $(u,v)=(0,0)$ is always the bottom-left corner of the image. To project these points into world space we introduce camera translation $T(t)$ and rotation $R(t)$ models
\begin{align}\label{eqnsf:translation_rotation}
    T(t) &= S(t,\mathbf{P}^\textsc{t}), \quad R(t) = R^\textsc{d}(t) + \eta_\textsc{r}S(t,\mathbf{P}^\textsc{r})\nonumber \\
    \mathbf{P}^\textsc{t}_i &= 
    \left[\arraycolsep=2.0pt
    \begin{array}{c}
    x \\
    y \\
    z \\
    \end{array}\right], \quad \mathbf{P}^\textsc{r}_i = \left[\begin{array}{ccc}
0 & -r^z & r^y \\
r^z & 0 & -r^x \\
-r^y & r^x & 0
\end{array}\right].
\end{align}
Here $S(t,\mathbf{P})$ is the same cubic spline model from Eq.~\eqref{eqnsf:cubic_spline}, evaluated element-wise over the channels of $\mathbf{P}$. We note there are \textit{no coordinate networks} employed in these models. Translation $T(t)$ is learned from scratch, $\mathbf{P}^\textsc{t}$ initialized to all-zeroes. Rotation $R(t)$ is learned as a small-angle approximation offset~\cite{im2015high} to device rotations $R^\textsc{d}(t)$ recorded by the phone's gyroscope -- or alternatively, the identity matrix if such data is not available. With these two models, and calibrated intrinsic matrix $K$ from the camera metadata, we now generate a ray with origin $O$ and direction $D$ as
\begin{align}\label{eqnsf:ray_generation}
    O \,{=}\, \left[\arraycolsep=2.0pt
    \begin{array}{c}
    O_x \\
    O_y \\
    O_z \\
    \end{array}\right] \,{=}\, T(t), \, \, D \,{=}\, \left[\arraycolsep=2.0pt
    \begin{array}{c}
    D_x \\
    D_y \\
    1 \\
    \end{array}\right] \,{=}\, \frac{R(t)K^{-1}}{D_z} \left[\arraycolsep=2.0pt
    \begin{array}{c}
    u \\
    v \\
    1 \\
    \end{array}\right],
\end{align}
where $D$ is normalized by its z component. We define our transmission and obstruction image planes as $\Pi^\textsc{t}$ and $\Pi^\textsc{o}$, respectively. As XY translation of these planes conflicts with changes in the camera pose, we lock them to the z-axis at depth $\Pi_z$ with canonical axes $\Pi_u$ and $\Pi_v$. Thus, given ray direction $D$ has a z-component of 1, we can calculate the ray-plane intersection as $Q = O + (\Pi_z - O_z)D$ and project to plane coordinates
\begin{align}\label{eqnsf:plane_projection}
    u^{\scalebox{0.5}{$\Pi$}},v^{\scalebox{0.5}{$\Pi$}} = \langle Q, \, \Pi_u \rangle / (\Pi_z - O_z),\, \langle Q, \, \Pi_v \rangle / (\Pi_z - O_z),
\end{align}
scaled by ray length to preserve uniform spatial resolution. Let $u^\textsc{t},v^\textsc{t}$ and $u^\textsc{o},v^\textsc{o}$ be the intersection coordinates for the transmission and obstruction plane, respectively. We alpha composite these layers along the ray as

\begin{align}\label{eqnsf:compositing}
\hat{c} &= (1-\alpha)c^\textsc{t} + \alpha c^\textsc{o}\nonumber \\
c^\textsc{t} &= f^\textsc{t}(u^\textsc{t} \,{+}\, \scalebox{0.7}{$\Delta$} u^\textsc{t}, v^\textsc{t} \,{+}\, \scalebox{0.7}{$\Delta$} v^\textsc{t}),\, \scalebox{0.7}{$\Delta$} u^\textsc{t}, \scalebox{0.7}{$\Delta$} v^\textsc{t} = S(t; h^\textsc{t}(u^\textsc{t}, v^\textsc{t})) \nonumber \\
c^\textsc{o} &\,{=}\, f^\textsc{o}(u^\textsc{o} \,{+}\, \scalebox{0.7}{$\Delta$} u^\textsc{o}, v^\textsc{o} \,{+}\, \scalebox{0.7}{$\Delta$} v^\textsc{o}),\, \scalebox{0.7}{$\Delta$} u^\textsc{o} \ucomma \scalebox{0.7}{$\Delta$} v^\textsc{o} =  S(t; h^\textsc{o}(u^\textsc{o}\ucomma v^\textsc{o}))\nonumber \\
\alpha&=\sigma(\tau_{\sigma} f^\alpha(u^\textsc{o} \,{+}\, \scalebox{0.7}{$\Delta$} u^\textsc{o},v^\textsc{o} \,{+}\, \scalebox{0.7}{$\Delta$} v^\textsc{o})),
\end{align}
where $\hat{c}$ is the composite color point, the weighted sum by $\alpha$ of the transmission color $c^\textsc{t}$ and obstruction color $c^\textsc{o}$. Each is the output of an image coordinate network $f(u,v)$ sampled at points offset by flow from an NSF $h(u,v)$. The sigmoid function $\sigma\,{=}\,1/(1+e^{-x})$ with temperature $\tau_\sigma$ controls the transition between opaque $\alpha\,{=}\,1$ and partially translucent $\alpha\,{=}\,0.5$ obstructions. This proves particularly helpful for learning hard occluders -- e.g., a fence -- where large $\tau_\sigma$ creates a steep transition between $\alpha\,{=}\,0$ and $\alpha\,{=}\,1$, which discourages $f^\alpha(u,v)$ from mixing content between layers.
\subsection{Training Procedure}
\begin{figure}[h!]
    \centering
    \includegraphics[width=0.8\linewidth]{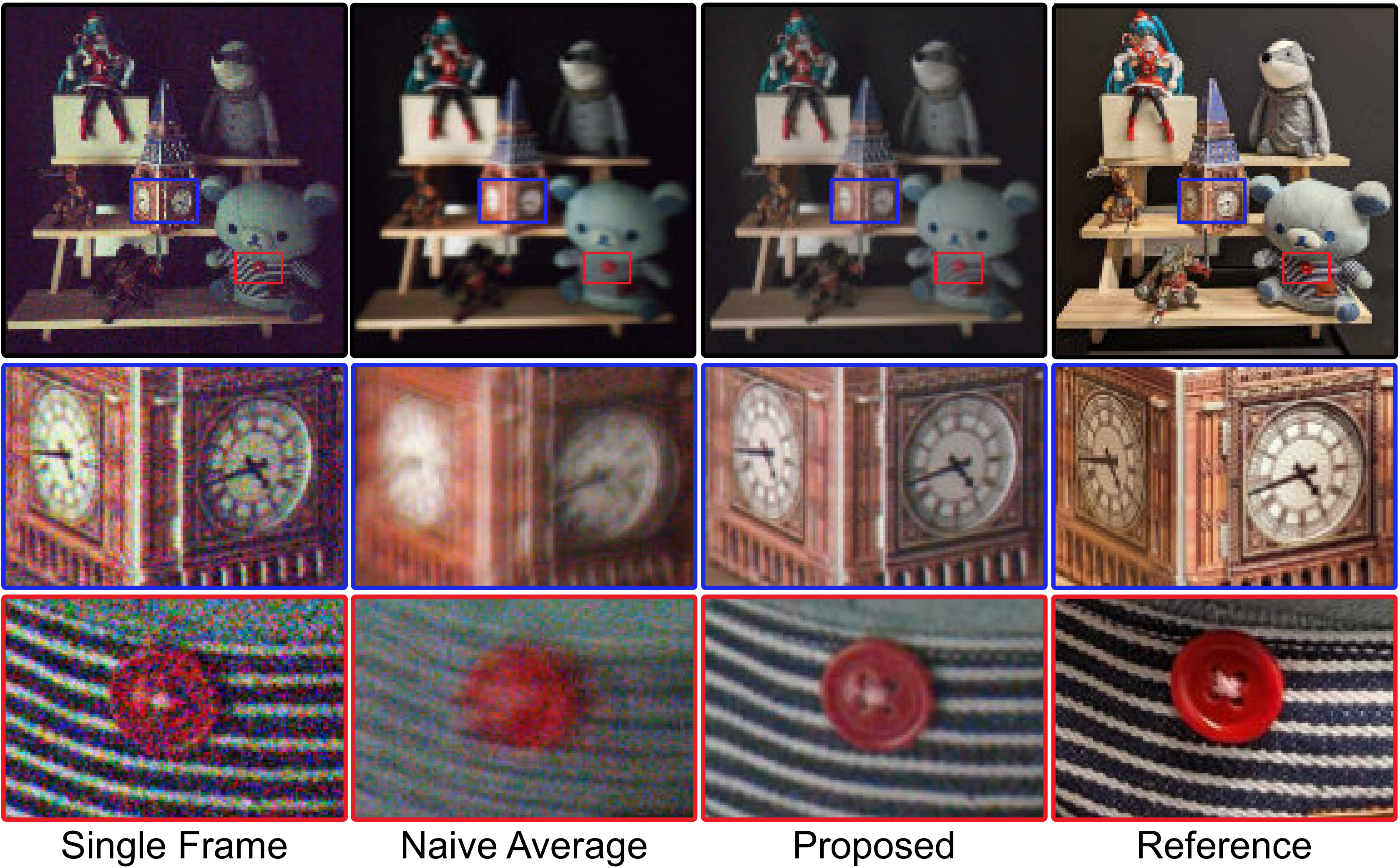}
    \caption{Reconstruction results for noisy, low-light conditions; exposure time 1/30, ISO 5000. The proposed model is able to robustly merge frames into a denoised image representation.}
    \label{fignsf:lowlight}
\end{figure}

\noindent\textbf{Losses.}\hspace{0.1em} Given all the components of our model are fully differentiable, we train them end-to-end via stochastic gradient descent. We define our loss  function $\mathcal{L}$ as
\begin{align}
    \mathcal{L} &= \mathcal{L}_\textsc{p} + \eta_\alpha\mathcal{R}_\alpha \label{eqnsf:loss1}\\
    \mathcal{L}_\textsc{p} &= |(c - \hat c)/(\mathrm{sg}(c) + \epsilon)|, \quad  \mathcal{R}_\alpha = |\alpha|, \label{eqnsf:loss2}\nonumber
\end{align}
where $\mathcal{L}_\textsc{p}$ is a relative photometric reconstruction loss~\cite{mildenhall2022nerf,chugunov2023shakes}, and $\mathrm{sg}$ is the stop-gradient operator. Shown in Fig.~\ref{fignsf:lowlight}, when combined with linear RAW input data this loss proves robust in noisy imaging settings~\cite{mildenhall2022nerf}, appropriate for in-the-wild scene reconstruction with unknown lighting conditions. Regularization term $\mathcal{R}_\alpha$ with weight $\eta_\alpha$ penalizes content in the obstruction layer, discouraging it from duplicating features from the transmission layer. 
\begin{figure*}[t!]
    \centering
    \includegraphics[width=\linewidth]{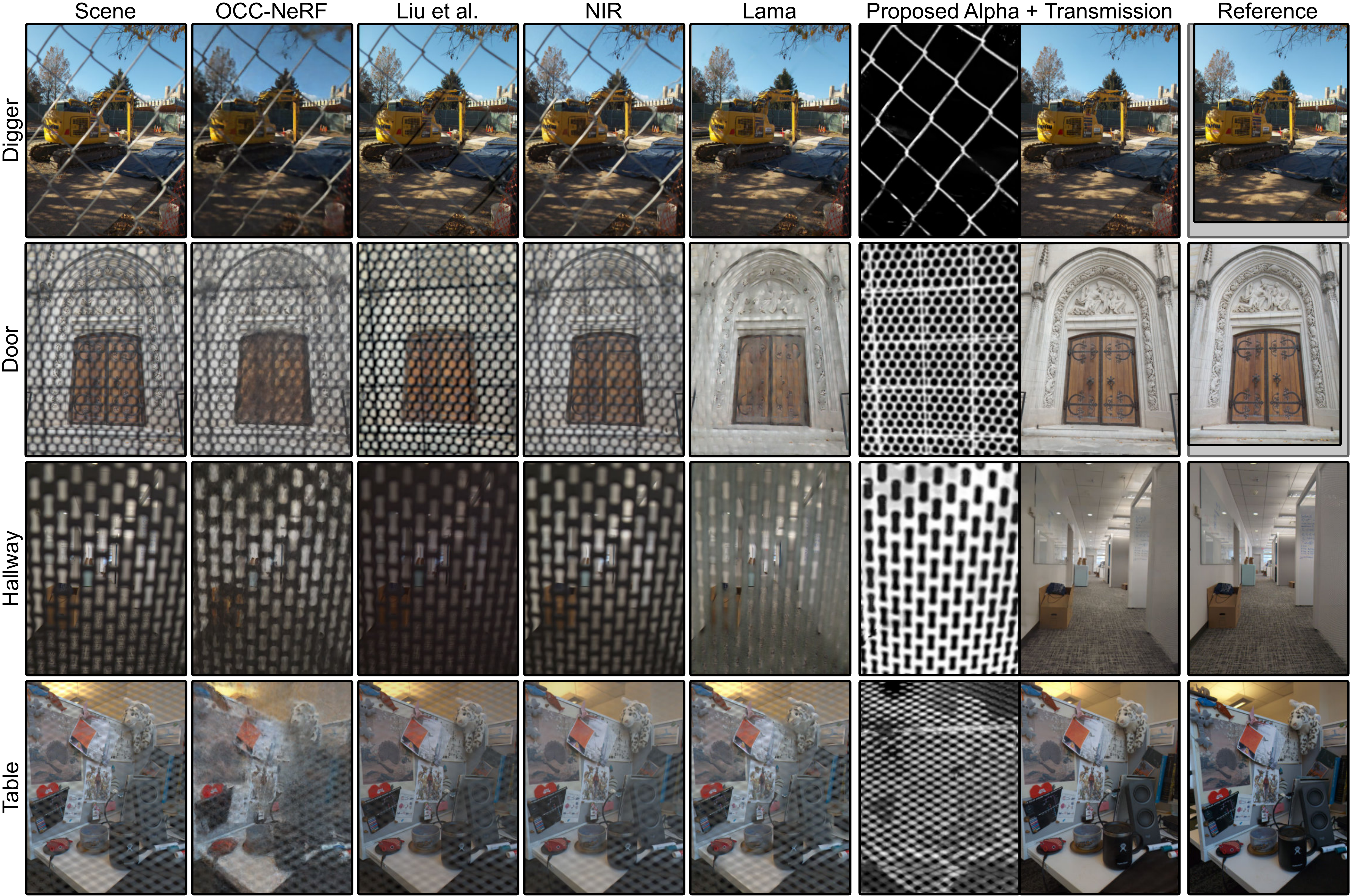}
    \caption{Occlusion removal results and estimated alpha maps for a set of captures with reference views; comparisons to single image, multi-view, and NeRF fitting approaches. See \href{https://light.princeton.edu/publication/nsf}{video materials} for visualization of input data and scene fitting.}
    \label{fignsf:results_occlusion}
\end{figure*}
\begin{figure}[h!]
    \centering
    \includegraphics[width=0.8\linewidth]{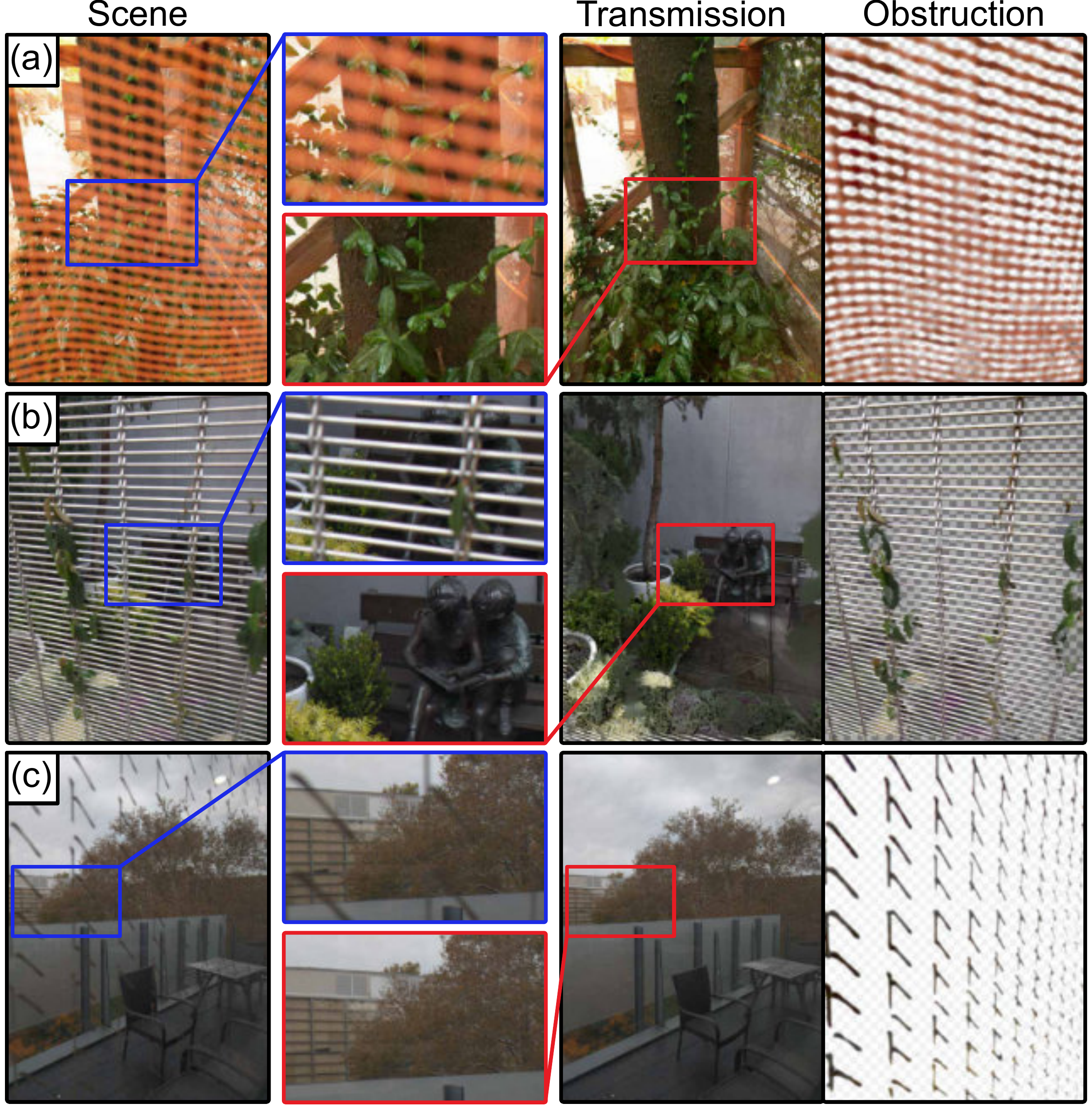}
    \caption{Layer separation results in unique real-world cases enabled by  our generalizable two-layer image model: (a) orange planter, (b) fenced garden, (c) stickers on balcony glass.}
    \label{fignsf:misc_scenes}
\end{figure}
\begin{figure}[h!]
    \centering
    \includegraphics[width=0.8\linewidth]{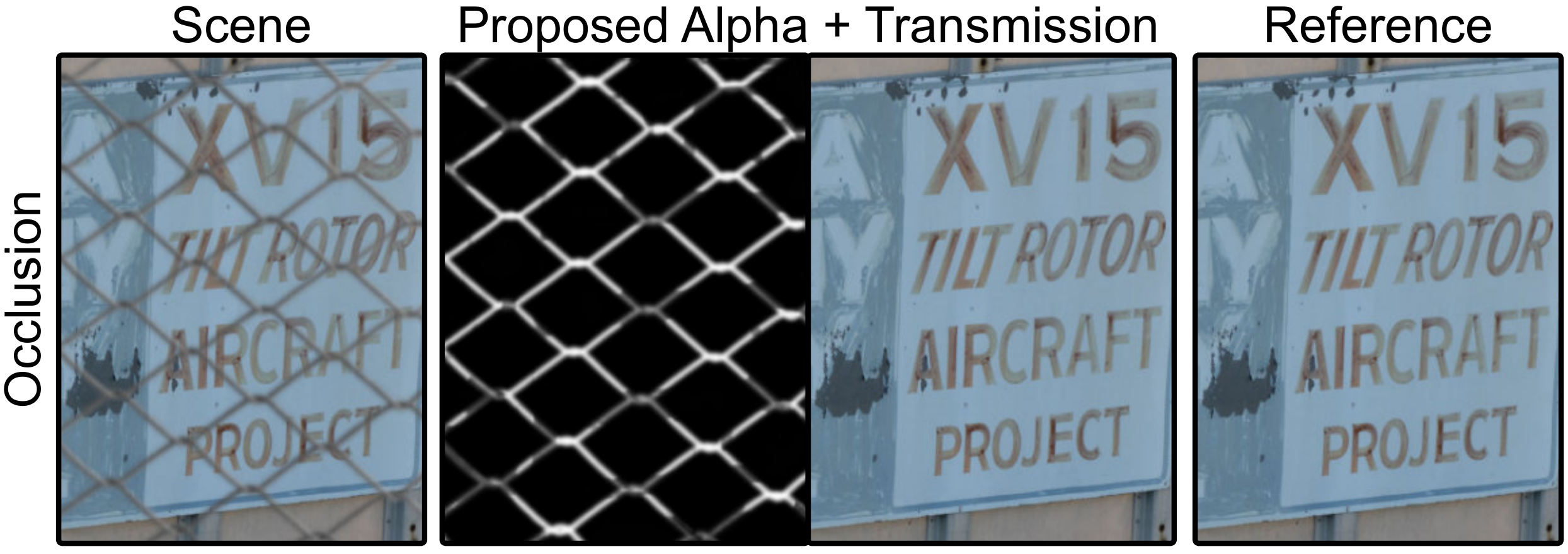}
        \resizebox{0.8\linewidth}{!}{
     \begin{tabular}[b]{ccccc}
     \midrule
          OCC-NeRF & Liu et al. & NIR & Lama & Proposed \\
	\midrule
     19.58$/$0.625 & 19.52$/$0.823 & 19.55$/$0.685 & 20.82$/$0.694 & \textbf{41.20}$/$\textbf{0.982} \\
    \bottomrule
    \vspace{-0.5em}
  \end{tabular}
  }
\includegraphics[width=0.8\linewidth]{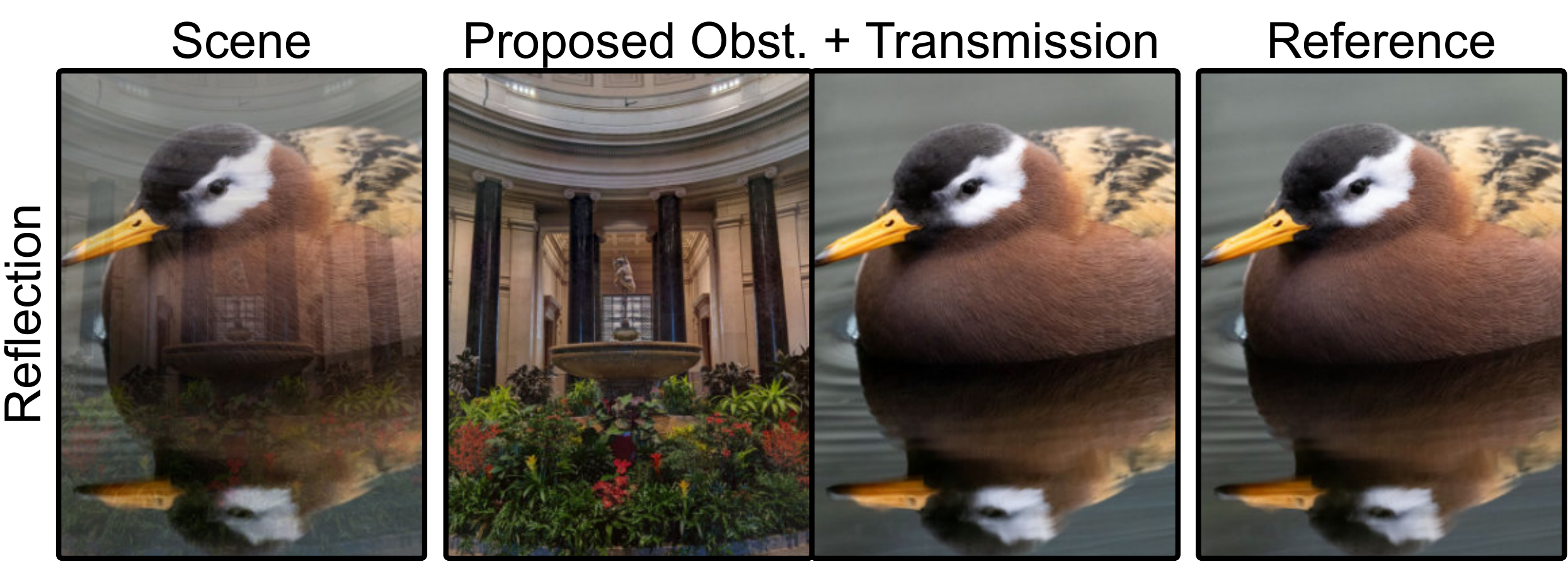}
\resizebox{0.8\linewidth}{!}{
     \begin{tabular}[b]{ccccc}
     \midrule
          NeRFReN & Liu et al. & NIR & DSR-Net & Proposed \\
	\midrule
     19.26$/$0.708 & 19.82$/$0.815 & 21.86$/$0.829 & 18.55$/$0.760 & \textbf{33.84}$/$\textbf{0.985} \\
    \bottomrule
  \end{tabular}
  }
    \caption{Qualitative and quantitative obstruction removal results for a set of synthetic scenes with paired ground truth, camera motion simulated from real measured hand shake data~\cite{chugunov2022implicit}. Evaluation metrics formatted as PSNR/SSIM.}
    \label{fignsf:results_synthetic}
\end{figure}

\begin{figure*}[t!]
    \centering
    \includegraphics[width=\linewidth]{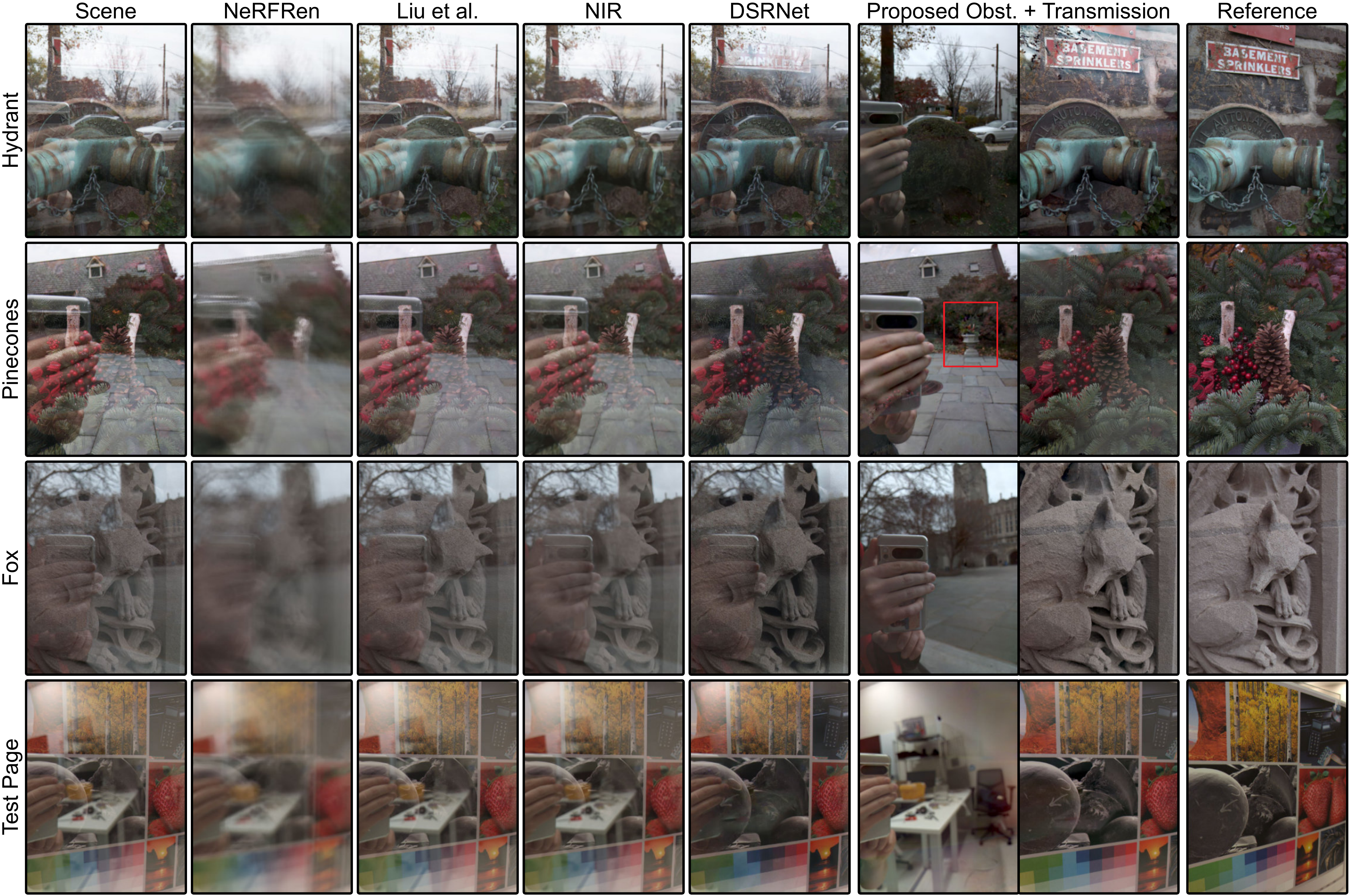}
    \caption{Reflection removal results and estimated alpha maps for a set of captures with reference views; comparisons to single image, multi-view, and NeRF fitting approaches. See \href{https://light.princeton.edu/publication/nsf}{video materials} for visualization of input data and scene fitting.}
    \label{fignsf:results_reflection}
\end{figure*}
\begin{figure}[h!]
    \centering
    \includegraphics[width=0.8\linewidth]{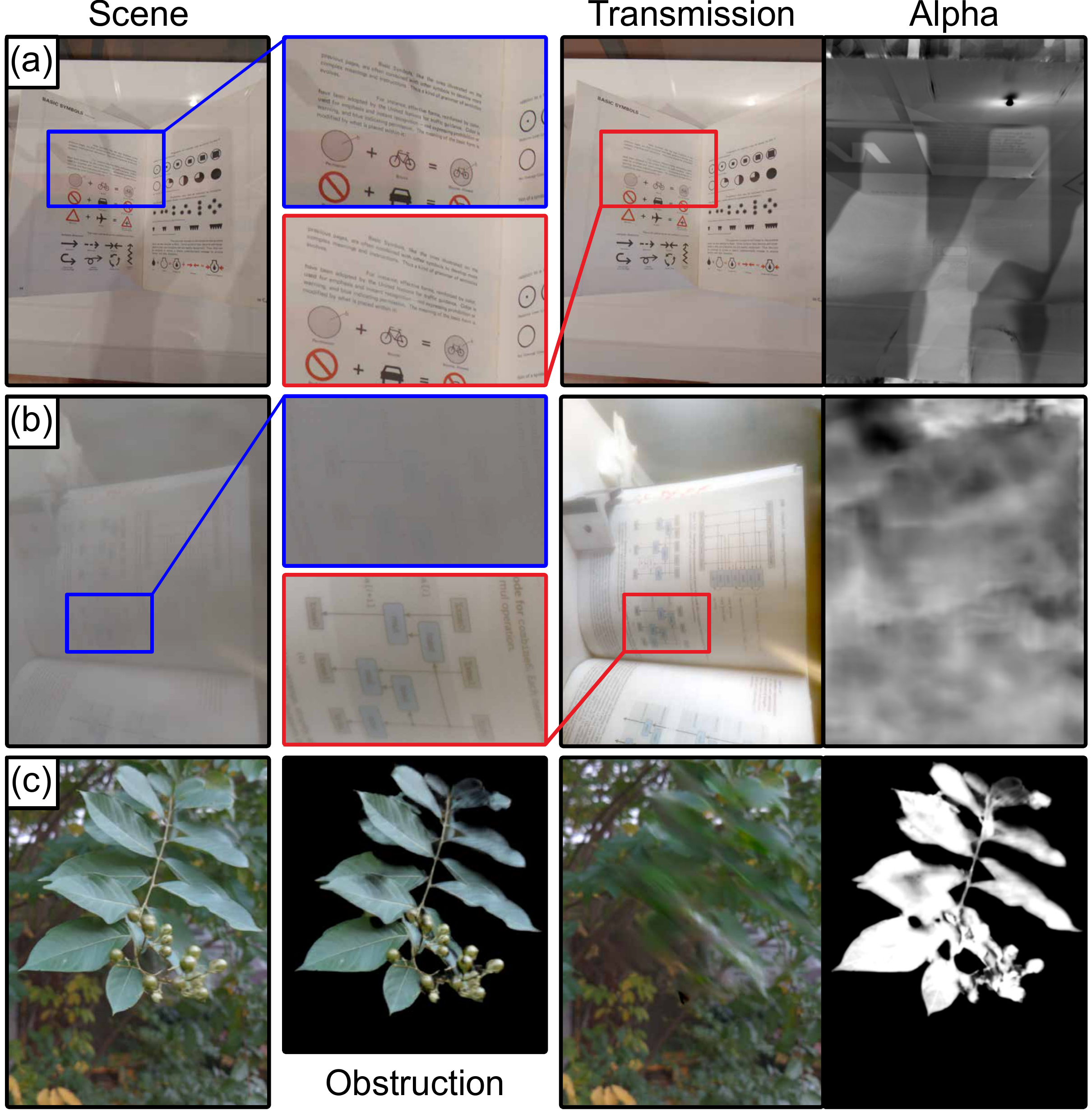}
    \caption{Layer separation results for additional example applications: (a) shadow removal, (b) image dehazing, and (c)  video motion segmentation (see \href{https://light.princeton.edu/publication/nsf}{video materials} for visualization).}
    \label{fignsf:in_the_wild}
\end{figure}

\noindent\textbf{Training.} Given the high-dimensional problem of jointly solving for camera poses, image layers, and neural spline field flows, we turn to coarse-to-fine optimization to avoid low-quality local minima solutions. We mask the multi-resolution hash encodings $\gamma(u,v)$ input into our image, alpha, and flow networks, activating higher resolution grids during later epochs of training:
\begin{align}
    \gamma_i(u,v) &= 
\begin{cases} 
\gamma_i(u,v) & \text{if } \,\, i/|\gamma| < 0.4 + 0.6(\text{sin\_epoch}) \\
0 & \text{if } \,\, i/|\gamma| > 0.4 + 0.6(\text{sin\_epoch})
\end{cases}\nonumber\\
\text{sin\_epoch}&=\mathrm{sin}(\text{epoch}/\text{max\_epoch}),
\end{align}
 This strategy results in less noise accumulated during early training as spurious high-resolution features do not need to be ``unlearned"~\cite{chugunov2023shakes,li2023neuralangelo} during later stages of refinement. 

\vspace{-0.5em}
\section{Applications}\label{secnsf:results}
\noindent\textbf{Data Collection.} To collect burst data we modify the open-source Android camera capture tool \href{https://github.com/Ilya-Muromets/Pani}{Pani} to record continuous streams of RAW frames and sensor metadata. During capture, we lock exposure and focus settings to record a 42 frame, two-second ``long-burst" of 12-megapixel images, gyroscope measurements, and camera metadata. We refer the reader to Chugunov et al.~\cite{chugunov2023shakes} for an overview of the long-burst imaging setting and its geometric properties. We capture data from a set of Pixel 7, 7-Pro, and 8-Pro devices, with no notable differences in overall reconstruction quality or changes in the training procedure required. We train our networks directly on Bayer RAW data, and apply device color-correction and tone-mapping for visualization.


\noindent\textbf{Implementation Details.} During training, we perform stochastic gradient descent on $\mathcal L$ for batches of $2^{18}$ rays per step for $6000$ steps with the Adam optimizer~\cite{kingma2014adam}. All networks use the multi-resolution hash encoding described in Eq.~\eqref{eqnsf:multires-hash}, implemented in tiny-cuda-nn~\cite{muller2021real}. Trained on a single Nvidia RTX 4090 GPU, our method takes \textit{approximately 3 minutes} to fit a full 42-frame image sequence. All networks have a base resolution $B^\gamma{=}4$, and scale factor $S^\gamma{=}1.61$, but while flow networks $h^\textsc{t}$ and $\textsc{o}$ are parameterized with a low number of grid levels $L^\gamma{=}8$, networks which represent high frequency content have $L^\gamma{=}12$ or $L^\gamma{=}16$ levels. These settings are task-specific, and full implementation details and results for short (4-8 frame) image bursts are included in the Supplementary Material.

\begin{figure}[t!]
    \centering
    \includegraphics[width=0.9\linewidth]{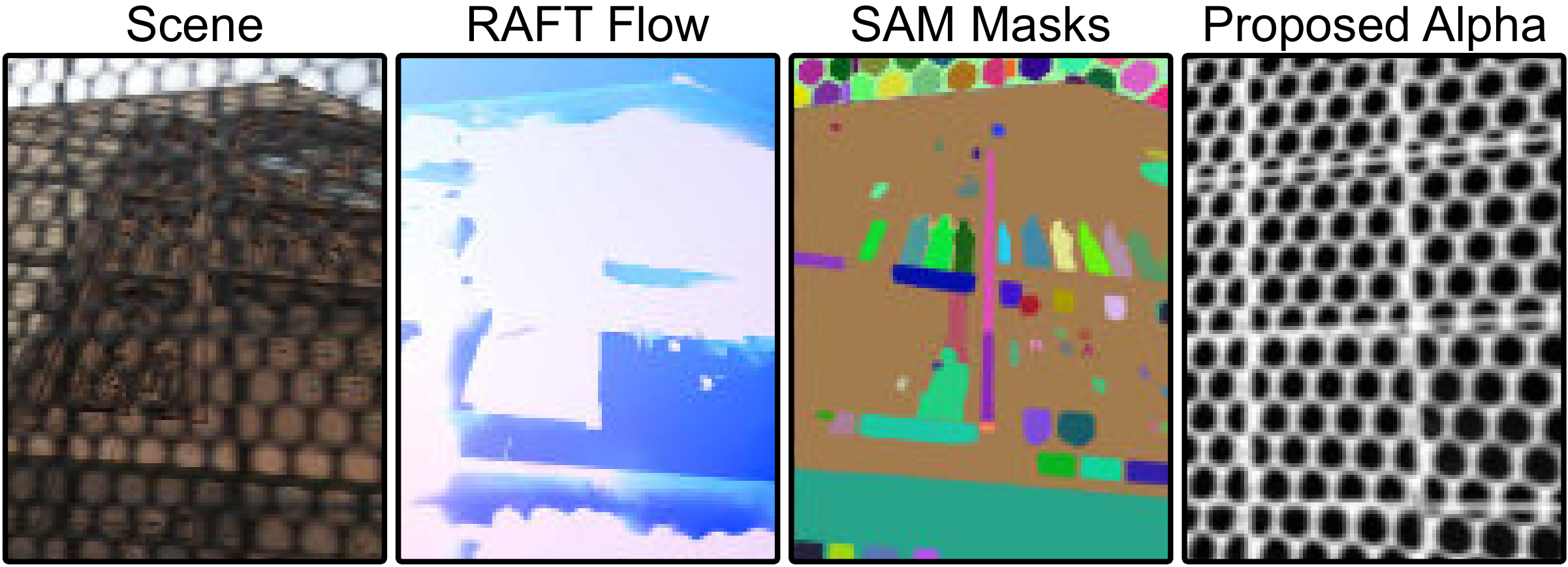}
    \caption{Learned flow estimator RAFT~\cite{teed2020raft} and segmentation model SAM~\cite{kirillov2023segment} struggle to produce meaningful outputs for a small-motion scene with an out-of-focus occluder. SAM successfully segments some objects behind the occluder (e.g., the statues on the building) but does not correctly segment the occluder itself. }
    \label{fignsf:masking}
\end{figure}

\noindent\textbf{Occlusion Removal.}\hspace{0.1em} Initializing the obstruction plane closer to the camera than the transmission plane, that is $\Pi^\textsc{o}_z < \Pi^\textsc{t}_z$, we find that the $f^\textsc{o}(u,v)$ naturally reconstructs foreground content in the scene. Given a scene with content hidden behind a foreground occluder -- e.g., imaging through a fence --  we can then perform occlusion removal with the proposed method by setting $\alpha=0$. We report results in Fig.~\ref{fignsf:results_occlusion} for a set of captures collected with reference views using a tripod-mounted occluder. We compare here to the multiview plus learning method presented in \textit{Liu et al.}~\cite{liu2020learning}, the neural radiance field approach \textit{OCC-NeRF}~\cite{zhu2023occlusion}, the flow + homography neural image model \textit{NIR}~\cite{nam2022neural}, and the single image inpainting method \textit{Lama}~\cite{suvorov2021resolution} as these methods demonstrate a broad range of techniques for occlusion detection and removal with varying assumptions on camera motion. We find that in this small baseline burst photography setting, existing multi-view methods fail to achieve meaningful occlusion removal; as the occluder maintains a high level of self-overlap for the whole image sequence. While the single-image method, \textit{Lama} is able to in-paint occluded regions based on un-occluded content, it cannot faithfully recover lost details such as the carvings in the \textit{Door} scene. Furthermore, \textit{Lama} does not produce an alpha matte, and rather \textit{requires a hand-annotated mask as input}. Illustrated in Fig.~\ref{fignsf:masking}, even otherwise robust mask segmentation networks such as the Segment Anything Model (SAM)~\cite{kirillov2023segment} fail to correctly detect complex occluders. In contrast, our approach distills information from all input frames to accurately recover temporarily occluded content, and jointly produces a high-quality alpha matte. In Fig.~\ref{fignsf:in_the_wild} we present additional layer separation results for real in-the-wild scenes with complex occluders, which demonstrate the versatility of the obstruction image model $f^\textsc{o}(u,v)$. \\
\noindent\textbf{Reflection Removal.}\hspace{0.1em} We show in Fig.~\ref{fignsf:results_reflection} how by flipping the plane depths $\Pi^\textsc{o}_z > \Pi^\textsc{t}_z$,  our model is also able to separate reflected from transmitted content. Here, we compare again to \textit{Liu et al.}~\cite{liu2020learning} and \textit{NIR}~\cite{nam2022neural}, as well as the reflection-specific neural radiance approach \textit{NeRFReN}~\cite{guo2022nerfren} and single-image reflection removal network \textit{DSR-Net}~\cite{hu2023single}. Similarly to occlusion removal, we observe that given small-baseline inputs the multi-view methods fail to achieve meaningful layer separation, and \textit{NeRFRen} struggles to converge on a sharp reconstruction. Only \textit{DSR-Net} is able to suppress even small parts of the reflection such as the car in the \textit{Hydrant} scene. In contrast, the proposed method not only estimates nearly reflection-free transmission layers, but is also able to recover hidden content -- such as the flowerpot highlighted in \textit{Pinecones} -- in the reflection layer. \\
\noindent\textbf{Synthetic Validation.}\hspace{0.1em} Given in-the-wild captures do not have perfectly aligned reference images, to further validate our method we construct a set of rendered scenes with paired ground truth data. Quantitative and qualitative results in Fig.~\ref{fignsf:results_synthetic} and the Supplementary Material align with our findings from real-world captures, with significant PSNR and SSIM improvements across all scenes.\\
\noindent\textbf{Image Enhancement through Layer Separation.} In addition to occlusion and reflection removal, a wide range of other computational photography applications can be viewed through the lens of layer separation. We showcase several example tasks in Fig.~\ref{fignsf:misc_scenes}, including shadow removal, image dehazing, and video motion segmentation. The key relationship between all these tasks is that the two effects undergo different motion models -- e.g., photographer-cast shadows move with the cellphone, while the paper target stays static. By grouping color content with its respective motion model, $f^\textsc{t}(u,v)$ with $h^\textsc{t}(u,v)$ and $f^\textsc{o}(u,v)$ with $h^\textsc{o}(u,v)$, just as in the occlusion case, we can remove the effect by removing its image plane. Fig.~\ref{fignsf:misc_scenes} (c), which fits our two-layer model for an image sequence of a moving tree branch, also highlights that our method does not rely solely on camera motion. Scene motion itself can also be used as a mechanism for layer separation in image bursts, similar to approaches in video masking~\cite{lu2021omnimatte,kasten2021layered}.

\vspace{-1em}\section{Discussion}\vspace{-0.5em}
\noindent In this work, we present a versatile representation of burst photography built on a novel neural spline field model of flow, and demonstrate image fusion and obstruction removal results under a wide array of conditions. In future work, we hope this generalizable model can be tailored to specific layer separation and image fusion applications: \\
\noindent\textbf{Learned Features.} Video layer separation works~\cite{lu2021omnimatte,ye2022deformable,kasten2021layered} make use of pre-trained segmentation networks and optical flow estimators to help guide reconstruction. However, shown in Fig.~\ref{fignsf:masking}, we found these could not be directly applied to small-motion data with large obstructions, as this is far outside the domain of their training data. Adapting these models to complex burst photography settings could potentially help disambiguate image layers in areas without reliable parallax or motion information.

\noindent\textbf{Physical Priors.} Our generic image plus flow representation can accommodate task-specific modules for applications where there are known physical models, such as chromatic aberration removal or refractive index estimation. \\
\noindent\textbf{Beyond Burst Data.} There exist many other sources of multi-image data to which the method can potentially be adapted -- e.g., microscopes, telescopes, and light field, time-of-flight, or hyperspectral cameras.

\section{Implementation Details}
\label{secnsf:supp_implementation}
\begin{figure}[t]
    \centering
    \includegraphics[width=0.6\linewidth]{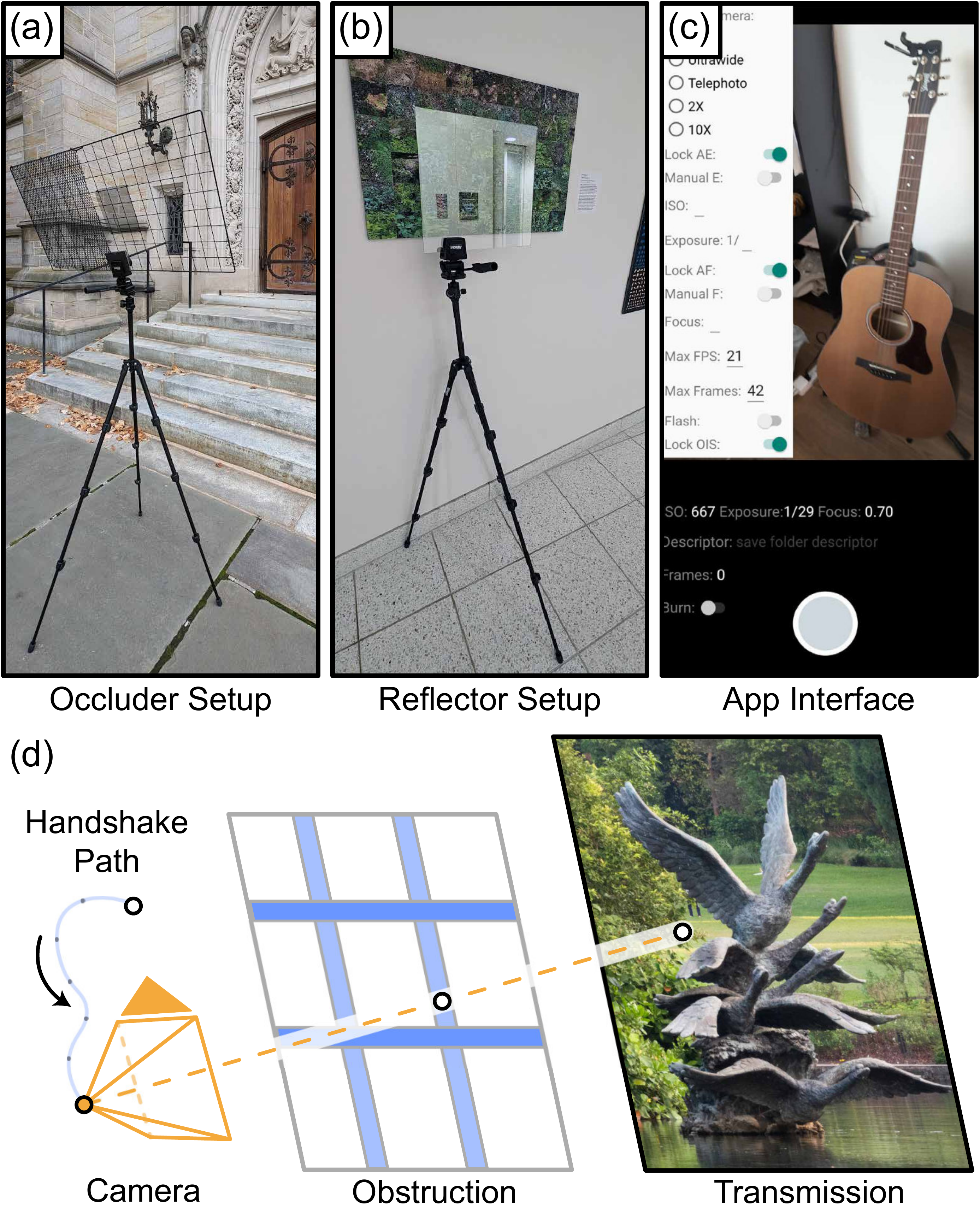}
    \caption{(a) Tripod-mounted occluder setup for capturing paired occlusion removal data. (b) Tripod-mounted reflector setup for capturing paired reflection removal data. (c) Capture app interface with the extended settings menu. (d-e) Example 3D scene with simulated occluder, camera frustum highlighted in orange.}
    \label{fignsf:sup_data_collection}
    \vspace{-1em}
\end{figure}

\noindent\textbf{Data Acquisition}\hspace{0.1em} To acquire paired obstructed and unobstructed captures, we construct two tripod-mounted rigs as illustrated in Fig.~\ref{fignsf:sup_data_collection} (a-b). We begin by capturing a still of the scene without the obstruction, before rotating the tripod into position to capture a 42-frame obstructed long-burst~\cite{chugunov2022implicit} of 12-megapixel RAW frames. As the rig is only used to hold the obstruction -- i.e., the smartphone is not attached to it -- it does not affect natural hand motion during capture.  For accessible natural occluders, such as the fences in Fig.~\ref{fignsf:supp_results_occlusion}, we acquire reference views by positioning the phone at a gap in the occluder -- though this sometimes cannot perfectly remove the occluder as in the case of Fig.~\ref{fignsf:supp_results_occlusion} \textit{Pipes}. We collect data with our modified \href{https://github.com/Ilya-Muromets/Pani}{Pani} capture app, illustrated in Fig.~\ref{fignsf:sup_data_collection} (c), built on the Android camera2 API. During capture, we also record metadata such as camera intrinsics, exposure settings, channel color correction gains, tonemap curves, and other image processing and camera information during capture. We stream gyroscope and accelerometer measurements from on-board sensors as $\approx$100Hz, though we find accelerometer values to be highly unreliable for motion on the scale of natural hand tremor, and so disregard these measurements for this work. We apply minimal processing to the recorded 10-bit Bayer RAW frames -- only correcting for lens shading and BGGR color channel gains -- before splitting them into a 3-plane RGB color volume. We do not perform any further demosaicing on this volume, as these processes correlate local signal values, and instead input it directly into our model for scene fitting. For visualization, we apply the default color correction matrix and tone-curve supplied in the capture metadata.

\begin{figure}[t!]
    \centering
    \includegraphics[width=0.6\linewidth]{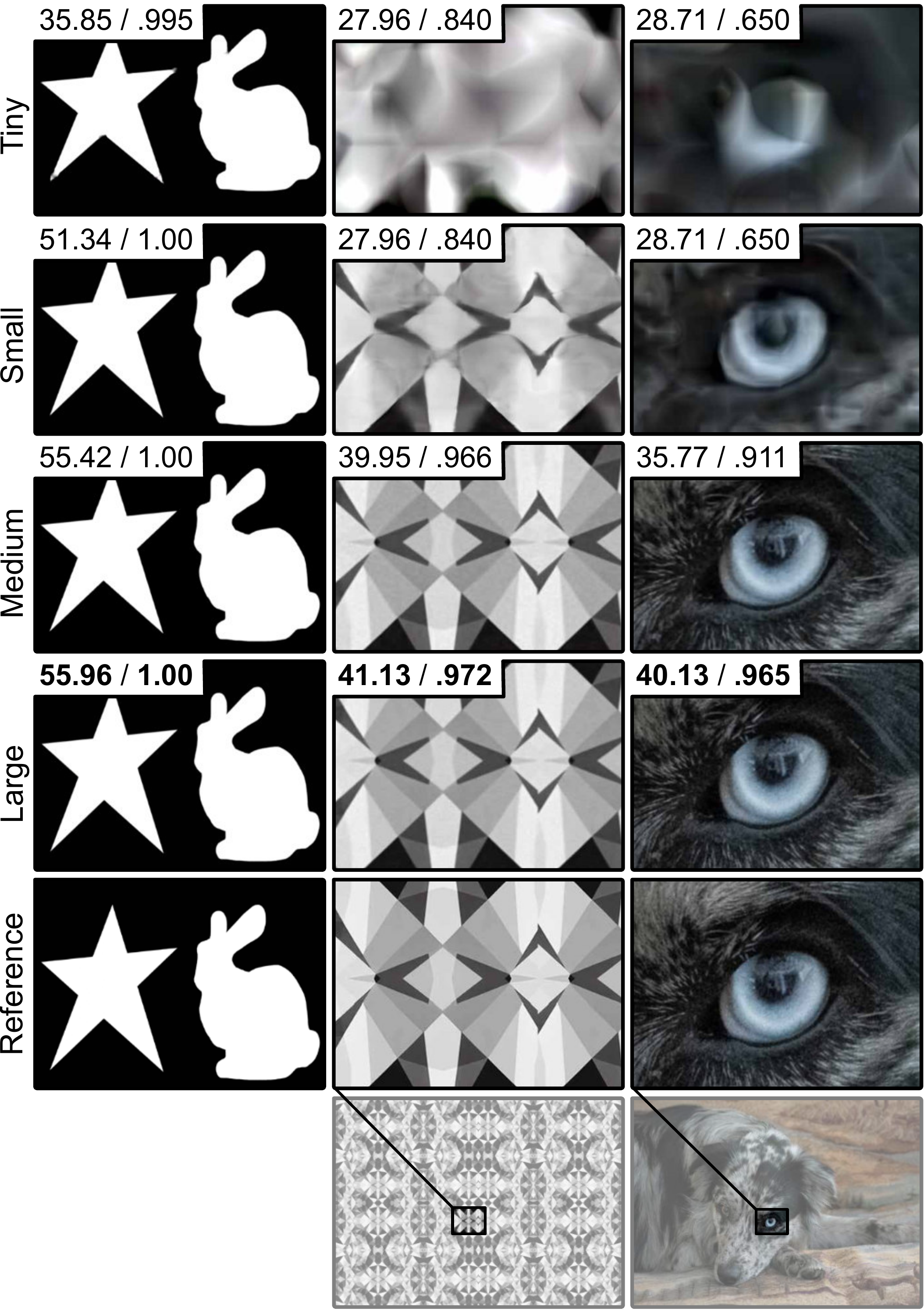}
    \caption{Image fitting results for network encoding configurations as described in Tab.~\ref{tabnsf:network-sizes}, other training and network parameters held constant: 5-layer MLP coordinate networks, hidden dimension 64, ReLU activations. PSNR/SSIM values inset top-left.}
    \label{fignsf:supp_image_fitting}
\end{figure}

\noindent\textbf{Synthetic Data Generation}\hspace{0.1em} Capturing aligned ground-truth data for obstruction removal is a long-standing problem in the field~\cite{wei2019single}, greatly exacerbated by the requirement in our setting of \textit{a sequence} of unstabilized frames with its base frame aligned to an unobstructed image. Thus, to help validate our method, we turn to synthetic captures created through image reprojection. We use 61-megapixel digital camera (Sony A7RIV) captures to simulate the transmission layer, and either hand-segmented occluders or a second 61-megapixel ``reflection" image to simulate the obstruction. These are simulated as 3D planes in space at depths $\Pi^\textsc{o}_z$ and $\Pi^\textsc{t}_z$ respectively -- $\Pi^\textsc{o}_z < \Pi^\textsc{t}_z$ for occluders and $\Pi^\textsc{o}_z > \Pi^\textsc{t}_z$ for reflectors -- and apply a random tilt to the planes with angle $\theta\in[-20^\circ,20^\circ]$. To generate realistic camera motion, we record samples of natural hand tremor with a pose-capture application built on the Apple ARKit library~\cite{chugunov2022implicit}. We then apply this motion path to a projective camera model, re-sample the image planes, and alpha-composite the outputs to produce the simulated burst stack. We emphasize that this data does not capture all the imaging effects present in real burst photography -- e.g., lens distortion, scene deformation, motion blur, chromatic aberrations, or sensor and microlens defects -- and use it as a tool for validating correct layer separation rather than a benchmark for overall performance. Reconstruction results for these simulated bursts are shown in Fig.~\ref{fignsf:supp_occlusion_synthetic} and Fig.~\ref{fignsf:supp_reflection_synthetic}.\\
\noindent\textbf{Implementation Details}\hspace{0.1em} While the overarching model structure is held constant between all applications -- identical projection, image generation, and flow models for all tasks -- elements such as the neural spline field $h(u,v)$ encoding parameters $\mathrm{params}_\gamma$ can be tuned for specific tasks:
\begin{align}\label{eqnsf:supp_multires-hash}
    h(u,v) &= \mathbf{h}(\gamma(u,v;\, \mathrm{params}_\gamma);\, \theta)\nonumber \\
    \mathrm{params}_\gamma &= \left\{\mathrm{B}^\gamma, \mathrm{S}^\gamma, \mathrm{L}^\gamma, \mathrm{F}^\gamma, \mathrm{T}^\gamma\right\}.   
\end{align}
\begin{table}[t!]
\centering
\begin{tabular}{cccccc}
\hline
 & base & scale & levels & feat.  & table \\
Size & $\mathrm{B}^\gamma$ & $\mathrm{S}^\gamma$ & $\mathrm{L}^\gamma$ & $\mathrm{F}^\gamma$ & $\mathrm{T}^\gamma$ \\\hline \hline
\textit{Tiny} (T) & 4 & 1.61 & \textbf{6} & 4 & \textbf{12} \\
\textit{Small} (S) & 4 & 1.61 & \textbf{8} & 4 & \textbf{14} \\
\textit{Medium} (M) & 4 & 1.61 & \textbf{12} & 4 & \textbf{16} \\
\textit{Large} (L) & 4 & 1.61 & \textbf{16} & 4 & \textbf{18} \\ \hline
\end{tabular}
\caption{Multi-resolution hash-table encoding parameters for different ``sizes'' of network, with larger encodings intended to fit higher-resolution data. Note that we only vary the number of grid levels $\mathrm{L}^\gamma$, and match the backing table size $\mathrm{T}^\gamma$ accordingly to avoid hash collisions. The base grid resolution $\mathrm{B}^\gamma$, grid per-level scale $\mathrm{S}^\gamma$, and feature encoding size $\mathrm{F}^\gamma$ are kept constant. }
\label{tabnsf:network-sizes}
\vspace{1em}
\centering
\begin{tabular}{ccccccc}
\multicolumn{7}{l}{\textit{\textbf{occlusion removal}}:} \\
\hline
& flow $h$ & $|h|$ & rgb $f$ & $f^\alpha$  & depth $\Pi_z$ & $\eta_\alpha\mathcal{R}$ \\
\hline \hline
$Tr$: & T & 11 & L &   & 1.0 & \multirow{2}{*}{0.02} \\
$Ob$: & T & 11 & M & M  & 0.5 & \vspace{0.5em}\\ 
\multicolumn{7}{l}{\textit{\textbf{reflection removal}}:} \\
\hline
& flow $h$ & $|h|$ & rgb $f$ & $f^\alpha$  & depth $\Pi_z$ & $\eta_\alpha\mathcal{R}$ \\
\hline \hline
$Tr$: & T & 11 & L &   & 1.0 & \multirow{2}{*}{0.0} \\
$Ob$: & T & 11 & T & L  & 2.5 & \vspace{0.5em}\\ 
\multicolumn{7}{l}{\textit{\textbf{video segmentation}}:} \\
\hline
& flow $h$ & $|h|$ & rgb $f$ & $f^\alpha$  & depth $\Pi_z$ & $\eta_\alpha\mathcal{R}$ \\
\hline \hline
$Tr$: & S & 15 & L &   & 1.0 & \multirow{2}{*}{0.002} \\
$Ob$: & S & 15 & L & M & 2.0 &  \vspace{0.5em}\\
\multicolumn{7}{l}{\textit{\textbf{shadow removal}}:} \\
\hline
& flow $h$ & $|h|$ & rgb $f$ & $f^\alpha$  & depth $\Pi_z$ & $\eta_\alpha\mathcal{R}$ \\
\hline \hline
$Tr$: & T & 11 & L &   & 1.0 & \multirow{2}{*}{0.0} \\
$Ob$: & T & 11 & T & M  & 2.0 &  \vspace{0.5em}\\
\multicolumn{7}{l}{\textbf{\textit{dehazing}:}} \\
\hline
& flow $h$ & $|h|$ & rgb $f$ & $f^\alpha$  & depth $\Pi_z$ & $\eta_\alpha\mathcal{R}$ \\
\hline \hline
$Tr$: & T & 11 & L &   & 1.0 & \multirow{2}{*}{0.01} \\
$Ob$: & T & 11 & T & S  & 0.5 & \vspace{0.5em}\\ 
\multicolumn{7}{l}{\textit{\textbf{image fusion}}:} \\
\hline
& flow $h$ & $|h|$ & rgb $f$ & $f^\alpha$  & depth $\Pi_z$ & $\eta_\alpha\mathcal{R}$ \\
\hline \hline
$Tr$: & S & 31 & L &  & 1.0 & 0.0 \\
\end{tabular}
\caption{Network encoding, flow, and loss configurations used for several layer-separation applications, separated into rows individually defining transmission $Tr$ and obstruction $Ob$ layers. Encoding parameters are defined by the corresponding (T,S,M,L) row of Tab.~\ref{tabnsf:network-sizes}. Flow size $|h|$ indicates the number of spline control points used for interpolation of the corresponding neural spline field $S(t,h(u,v))$. }
\label{tabnsf:application-configs}
\vspace{-1em}
\end{table}
\hspace{-0.5em}By manipulating the parameters of Eq.~\ref{eqnsf:supp_multires-hash} as defined in Tab.~\ref{tabnsf:network-sizes} we construct four different ``sizes'' of network encodings: \textit{Tiny}, \textit{Small}, \textit{Medium}, and \textit{Large}. Image fitting results in Fig.~\ref{fignsf:supp_image_fitting} illustrate what scale of features each of these configurations is able to reconstruct, with larger encoding reconstructing denser and higher-frequency content. Then, assembling together multiple image and flow networks with varying encoding sizes as defined in Tab.~\ref{tabnsf:network-sizes}, we are able to leverage this feature scale control for layer separation tasks such as occlusion, reflection, or shadow removal. 

For tasks such as video segmentation, it is important that both the transmission layer and obstruction layer are able to represent high-resolution images, as the purpose here is to divide and compress video content into two canonical views, alpha matte, and optical flow. Hence for the video segmentation task in Tab.~\ref{tabnsf:network-sizes} both layers have \textit{Large} network encodings. Conversely, for a task such as shadow removal we want to minimize the amount of color and alpha information the shadow obstruction layer is able to represent -- as shadows, like the mask example in Fig.~\ref{fignsf:supp_image_fitting}, are comprised of mostly low-resolution image features. Correspondingly, the shadow removal task in Tab.~\ref{tabnsf:network-sizes} has a \textit{Tiny} image color encoding and only a \textit{Medium} size alpha encoding. We keep these \emph{parameters constant between all tested scenes} for clarity of presentation, however we emphasize that these model configurations are not prescriptive; all neural scene fitting approaches~\cite{mildenhall2020nerf} have per-scene optimal parameters. Given the relatively fast training speed of our approach, approximately 3mins on a single Nvidia RTX 4090 GPU, in settings where data acquisition is costly  -- e.g., scientific imaging settings such as microscopy -- it may even be tractable to sweep model parameters to optimally reconstruct each individual capture.

\section{Additional Reconstruction Results}
\label{secnsf:supp_results}
\begin{figure*}[t!]
    \centering
    \includegraphics[width=\linewidth]{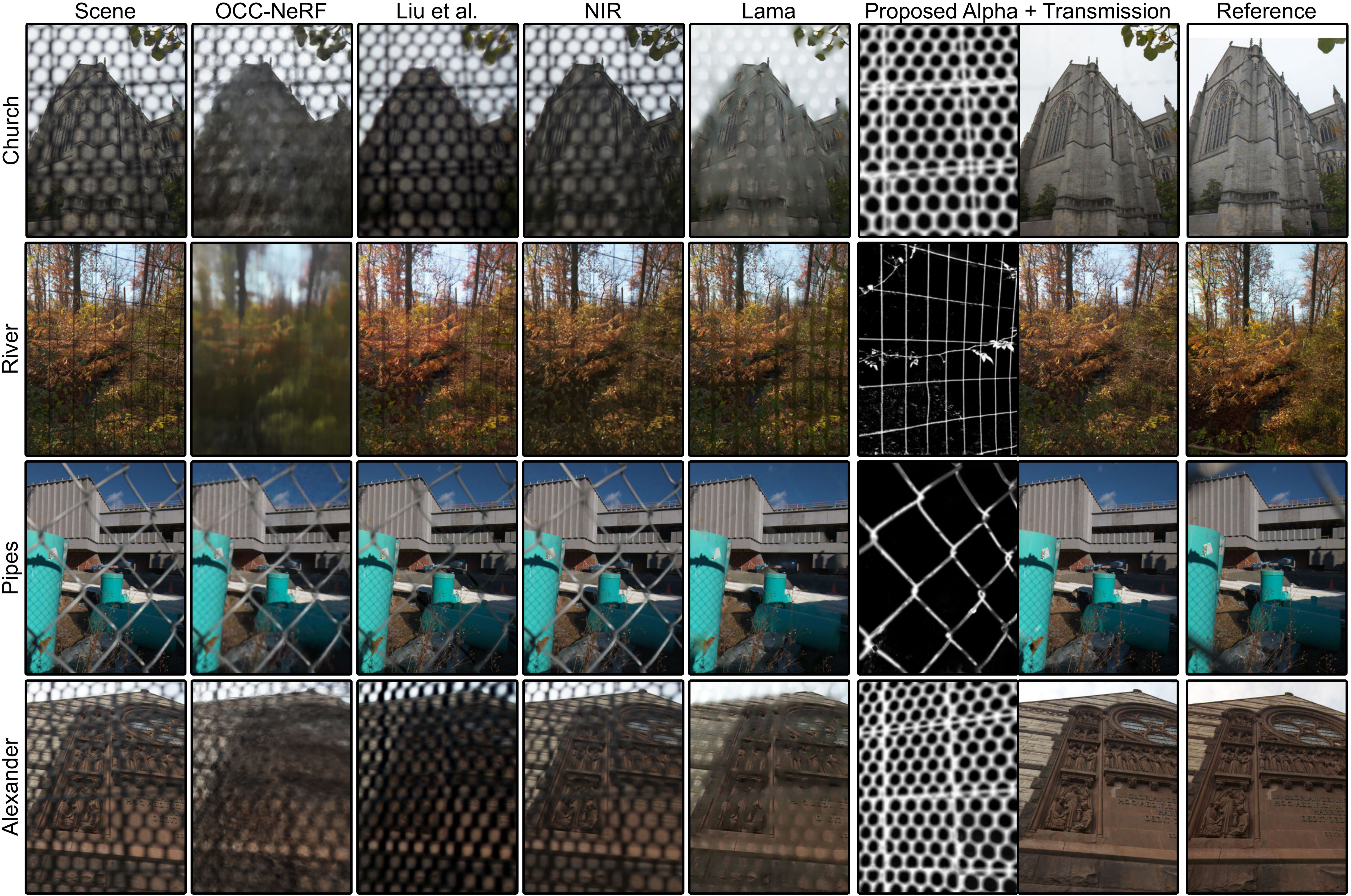}
    \caption{Occlusion removal results and estimated alpha maps for a set of captures with reference views, with comparisons to single image,
multi-view, and NeRF fitting approaches. See video materials for visualization of input data and scene fitting.}
    \label{fignsf:supp_results_occlusion}
\end{figure*}
\noindent In this section, we provide additional quantitative and qualitative obstruction removal results, comparing our proposed model against a range of multi-view and single-image methods. We include discussion of challenging imaging settings and potential directions of future work to address them.\\
\noindent\textbf{Occlusion Removal}\hspace{0.1em} We include a set of additional occlusion removal results in Fig.~\ref{fignsf:supp_results_occlusion} with natural environmental occluders such as fences and grates. We evaluate our results against the multi-image learning-based obstruction removal method Liu et al.~\cite{liu2020learning}, the NeRF-based method OCC-NeRF~\cite{zhu2023occlusion}, the flow plus homography neural image representation NIR~\cite{nam2022neural}, and the single image inpainting approach Lama~\cite{suvorov2021resolution} -- to which we provide hand-drawn masks of the occlusion. We find that, as observed in the main text, the multi-image methods struggle to remove significant parts of the obstruction. Though in some scenes, the multi-image baselines are able to decrease the opacity of the occluder to reveal details behind it. Nevertheless, in all cases the obstruction is still clearly visible after applying each baseline. Given the small camera baseline setting of our input data, the volumetric OCC-NeRF approach struggles to converge on a cohesive 3D scene representation, producing blurred or otherwise inconsistent image reconstructions -- as is the case for the \textit{Church} scene. We find that the the homography-based NIR method also struggles in this small baseline setting, often identifying the entire scene as the canonical view rather than partly obstructed.
Given hand annotated masks, single image methods such as DALL·E and Lama~\cite{suvorov2021resolution} can successfully inpaint sparse occluders such as the fence in the \textit{Pipes}  scene, but struggle to recover content behind dense occluders such as in \textit{Alexander} and \textit{Church} in Fig.~\ref{fignsf:supp_results_occlusion}. As they have no way to aggregate content between frames, they ``recover" hidden content from visual priors on the scene, which may not be reliable when the scene is severely occluded.

\begin{figure*}[t!]
    \centering
    \includegraphics[width=\linewidth]{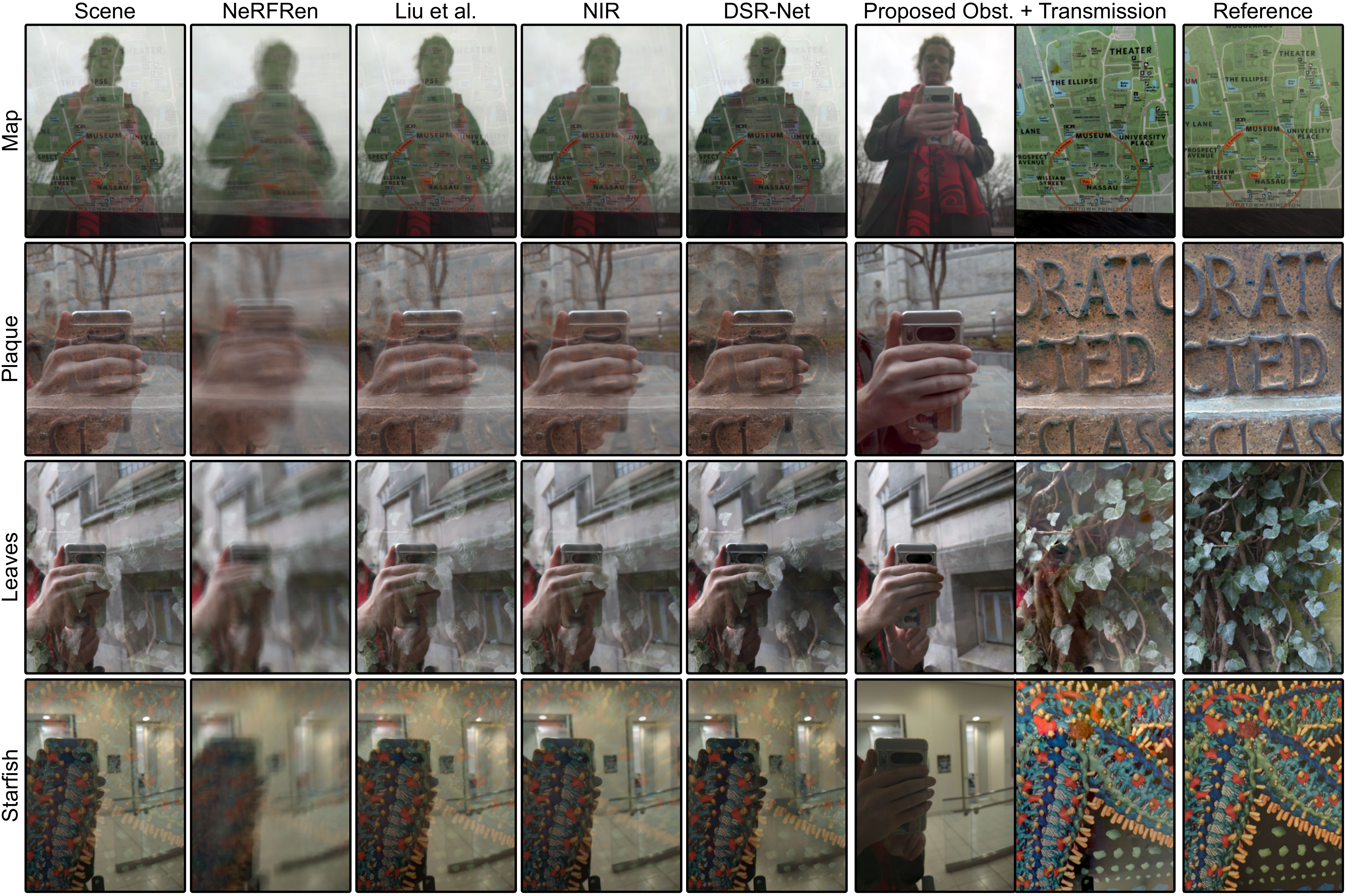}
    \vspace{-1.5em}
    \caption{Reflection removal results and estimated alpha maps for a set of captures with reference views, with comparisons to single image,
multi-view, and NeRF fitting approaches. See video materials for visualization of input data and scene fitting.}
    \label{fignsf:supp_results_reflection}
\end{figure*}
\begin{figure}[h!]
    \centering
    \includegraphics[width=0.8\linewidth]{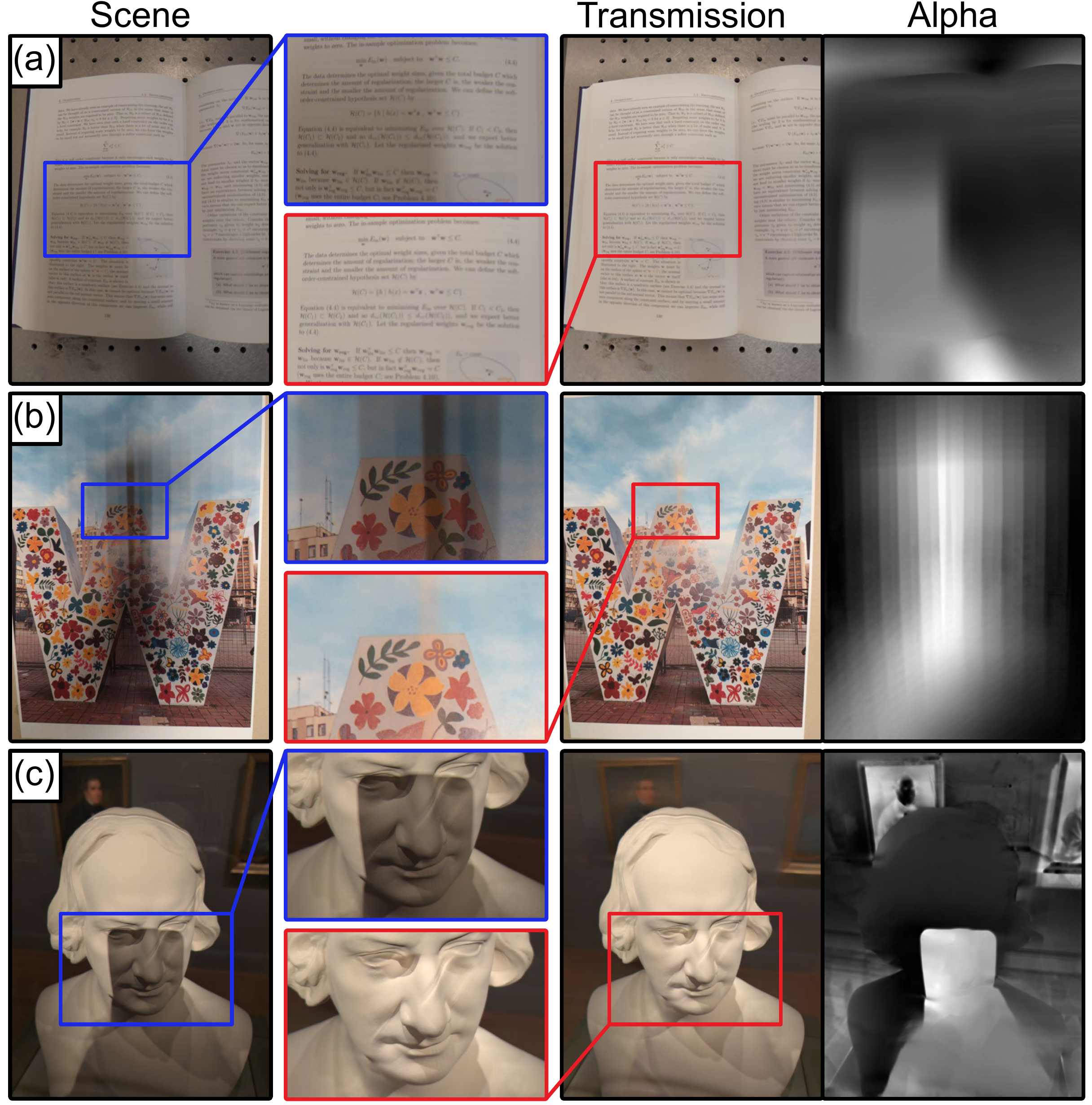}
    \caption{Shadow removal results under different lighting conditions: (a) partially diffuse, (b) multiple point, (c) single point.}
    \label{fignsf:supp_shadow}
\end{figure}
\begin{figure}[h!]
    \centering
    \includegraphics[width=0.8\linewidth]{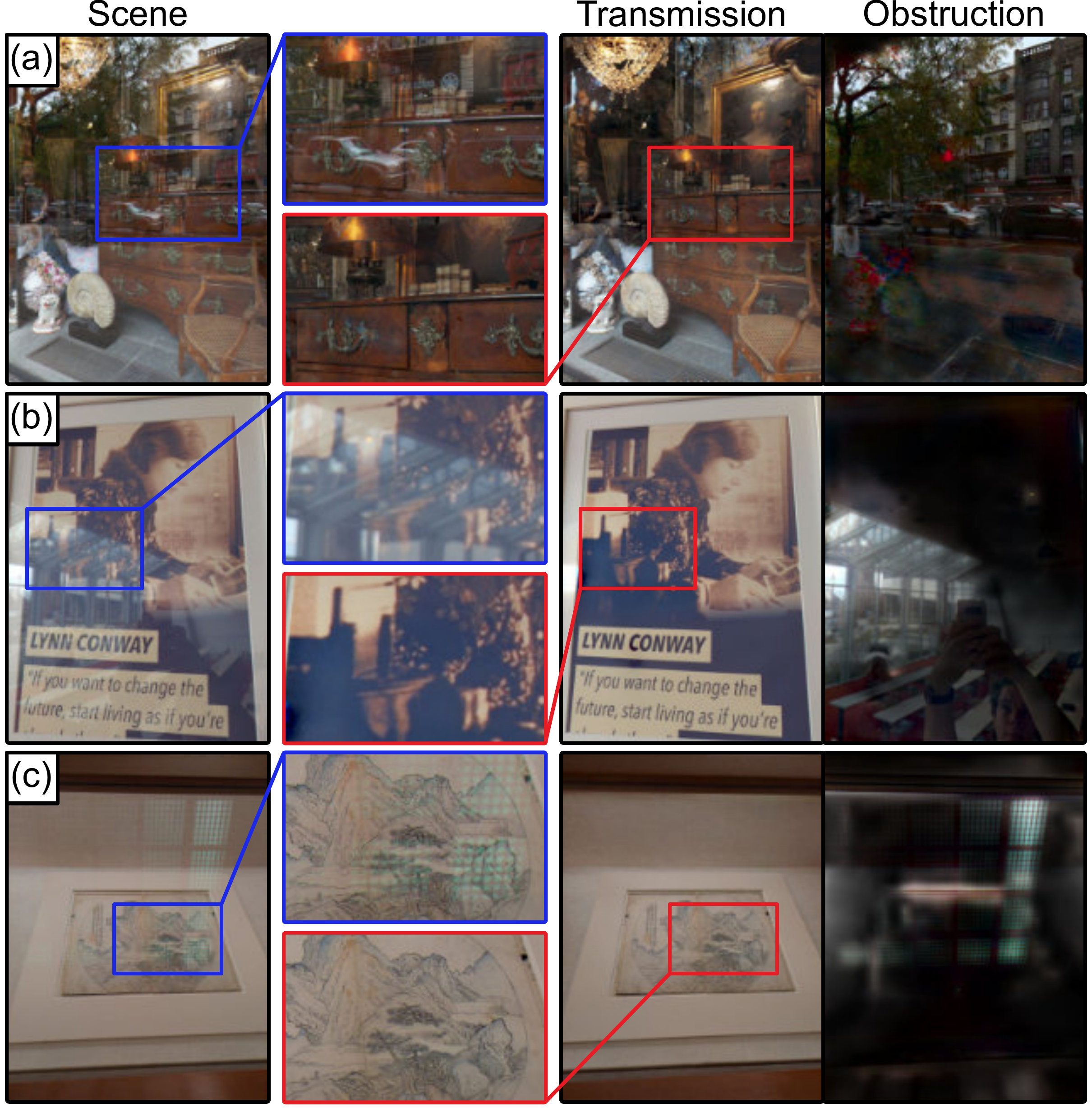}
    \caption{Reflection removal results for challenging in-the-wild scenes: (a) storefront window, (b) poster, (c) museum painting.}
    \label{fignsf:wild_reflection}
\end{figure}

\begin{figure*}[h!]
    \vspace{2em}
    \centering
    \includegraphics[width=\linewidth]{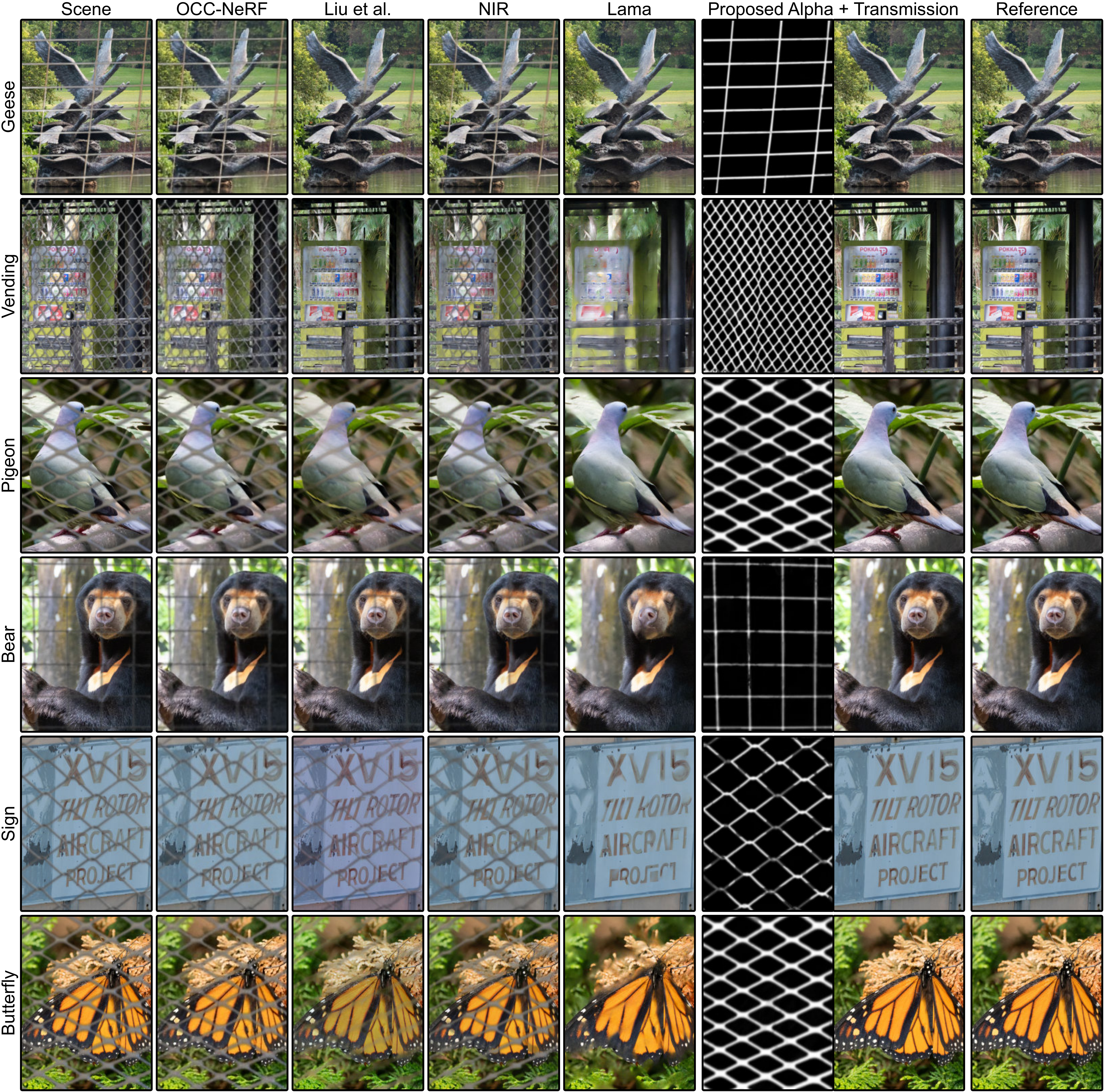}\vspace{1em}
        \resizebox{\linewidth}{!}{
     \begin{tabular}[b]{lccccclccccc}
     \midrule
          \textbf{Occlusion} & OCC-NeRF & Liu et al. & NIR & Lama & Proposed & \textbf{Occlusion} & OCC-NeRF & Liu et al. & NIR & Lama & Proposed\\
	\midrule
	\midrule
    \textit{Geese} & 19.49$/$0.578  & 32.24$/$0.970 & 20.89$/$0.696  & 21.96$/$0.760 & \textbf{41.80}$/$\textbf{0.986} &
    \textit{Vending}  & 18.05$/$0.550 & 15.10$/$0.754 & 17.96$/$0.625 & 17.42$/$0.591 & \textbf{39.62}$/$\textbf{0.981} \\
    \textit{Pigeon} & 18.60$/$0.691 & 15.17$/$0.725 & 18.74$/$0.691 & 21.55$/$0.753 & \textbf{40.33}$/$\textbf{0.965} &
    \textit{Bear}   & 23.72$/$0.696 & 26.32$/$0.930 & 23.28$/$0.746 & 23.84$/$0.815 & \textbf{40.88}$/$\textbf{0.980} \\
    \textit{Sign} & 24.34$/$0.870 & 24.11$/$0.952 & 22.84$/$0.905 & 28.57$/$0.932 & \textbf{48.63}$/$\textbf{0.994} & 
    \textit{Butterfly} & 17.67$/$0.674 & 15.43$/$0.828 & 18.25$/$0.750 & 17.89$/$0.722 & \textbf{39.53}$/$\textbf{0.980} \\
    \bottomrule
  \end{tabular}
  }
    \caption{Qualitative and quantitative occlusion removal results for a set of 3D rendered scenes with paired ground truth. Evaluation metrics formatted as PSNR/SSIM.}
    \label{fignsf:supp_occlusion_synthetic}
    \vspace{2em}
\end{figure*}

\begin{figure*}[h!]
    \vspace{2em}
    \centering
    \includegraphics[width=\linewidth]{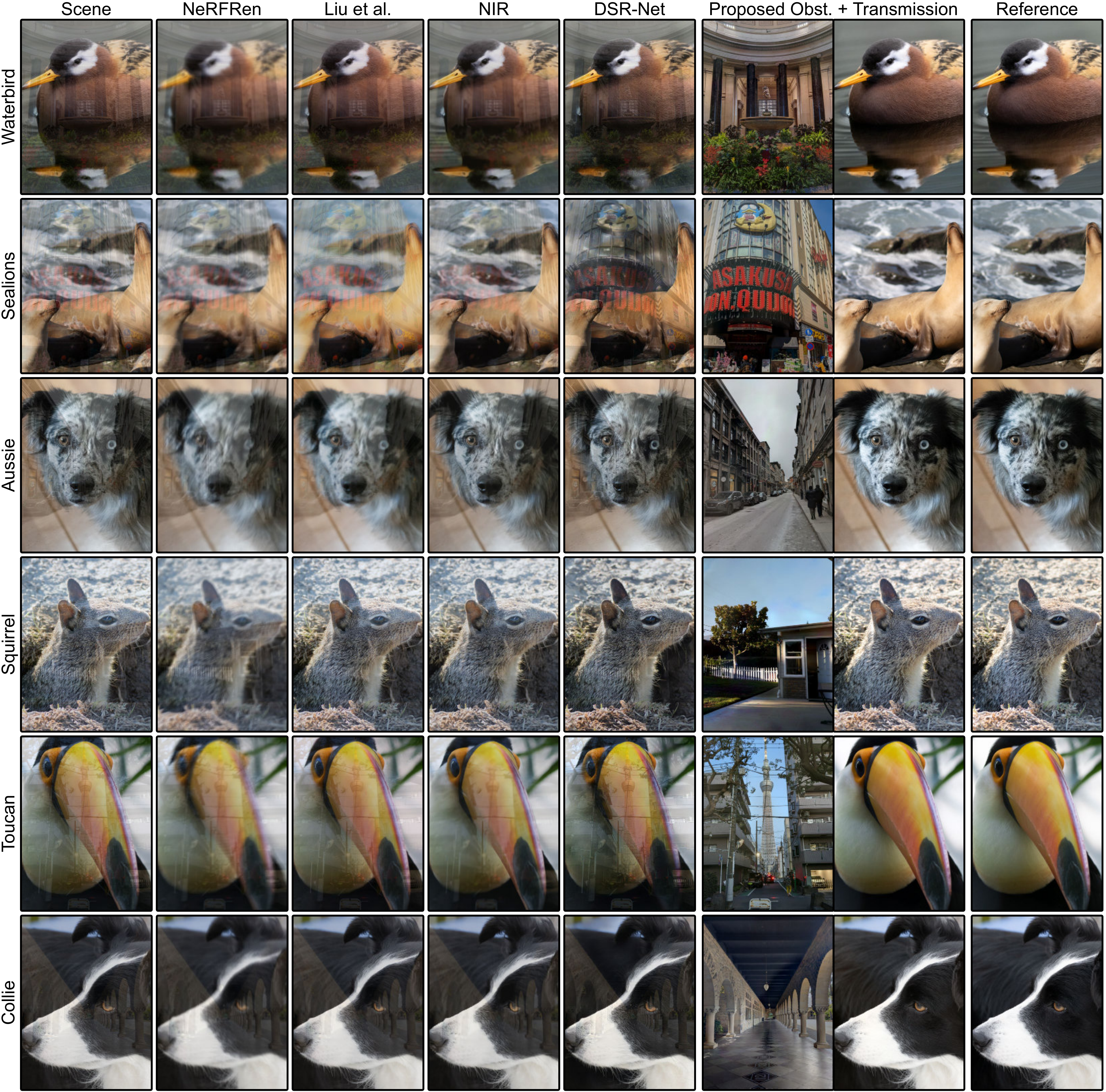}\vspace{1em}
 \resizebox{\linewidth}{!}{
     \begin{tabular}[b]{lccccclccccc}
     \midrule
          \textbf{Reflection} & NeRFReN & Liu et al. & NIR & DSR-Net & Proposed & \textbf{Reflection} & NeRFReN & Liu et al. & NIR & DSR-Net & Proposed\\
	\midrule
	\midrule
    \textit{Waterbird} & 21.94$/$0.695 & 23.68$/$0.811 & 24.08$/$0.751 & 19.95$/$0.753 & \textbf{39.16}$/$\textbf{0.982} & 
    \textit{Sealion}  & 20.28$/$0.811 & 11.45$/$0.726 & 22.36$/$0.899 & 13.27$/$0.657 & \textbf{32.31}$/$\textbf{0.993}
    \\
    \textit{Aussie} & 18.88$/$0.561 & 18.09$/$0.634 & 20.54$/$0.665 & 19.56$/$0.738 & \textbf{30.90}$/$\textbf{0.971} & 
    \textit{Squirrel}   & 17.15$/$0.431 & 23.55$/$0.950 & 22.04$/$0.789 & 19.05$/$0.860 & \textbf{33.34}$/$\textbf{0.988} \\
    \textit{Toucan} & 19.98$/$0.817 & 21.14$/$0.837 & 21.67$/$0.873 & 17.63$/$0.717 & \textbf{36.00}$/$\textbf{0.985} &
    \textit{Collie} & 18.60$/$0.706 & 22.34$/$0.862 & 22.08$/$0.801 & 21.96$/$0.847  & \textbf{32.98}$/$\textbf{0.978}
     \\
    \bottomrule
  \end{tabular}
  }
    \caption{Qualitative and quantitative reflection removal results for a set of 3D rendered scenes with paired ground truth. Evaluation metrics formatted as PSNR/SSIM.}
    \label{fignsf:supp_reflection_synthetic}
    \vspace{2em}
\end{figure*}

In contrast, our method automatically distills a high-quality alpha matte for the obstruction and reconstructs the underlying transmission layer using information from multiple views. This mask is of similar quality regardless of whether the scene is obstructed by a dense occluder or a sparse occluder, so long as there is sufficient parallax between the two layers. The depth-separation properties of our alpha estimation are showcased in the \textit{River} example, where the obstruction layer isolated not only the grid of the fence, but also the branches and leaves weaved through the fence. Our method reconstructs the transmitted layer behind the occlusion with favorable results compared to all baseline methods.

\noindent\textbf{Reflection Removal}\hspace{0.1em} 
For reflection removal, we compare with the reflection-aware NeRF-based method NeRFReN~\cite{guo2022nerfren} in addition to NIR~\cite{nam2022neural}, Liu et al.~\cite{liu2020learning}, and the single-image reflection removal method DSRNet~\cite{hu2023single}. We show reflection removal results in Fig.~\ref{fignsf:supp_results_reflection}. We observe results with a similar trend to those in the obstruction removal task. The volumetric method NeRFReN struggles to reconstruct a high-fidelity scene representation, as Liu et al. and NIR also struggle with the small baseline of the camera motion. The single-image method DSRNet performs best among the baselines, as it has no priors on image motion. However, without the ability to draw information from multiple views, DSRNet uses learned priors to disambiguate reflected and transmitted content. This appears not to be very effective for high opacity reflections, such as the \textit{Leaves} example and the phone in the \textit{Plaque} scene. Our method achieves the highest-quality reconstruction and layer separation among all methods tested, across all scenes, with our estimated obstruction revealing the detailed structure of the scene being reflected. In Fig.~\ref{fignsf:wild_reflection} we also showcase our model's performance on challenging, in-the-wild scenes where we do not have the ability to acquire reference views. We observe robust reflection removal, matching the reconstruction quality observed for scenes acquired with our tripod setup.

\noindent\textbf{Validation on Synthetic Scenes}\hspace{0.1em}
We generate synthetic scenes as described in Sec.~\ref{secnsf:supp_implementation}, and compare our obstruction removal results to the same baselines outlined in the previous sections, including: OCC-NeRF~\cite{zhu2023occlusion}, NeRFReN~\cite{guo2022nerfren}, Liu et al.~\cite{liu2020learning}, NIR~\cite{nam2022neural}, Lama~\cite{suvorov2021resolution} and DSRNet~\cite{hu2023single}. We show quantitative and qualitative results for occlusion removal and reflection removal in Fig.~\ref{fignsf:supp_occlusion_synthetic} and Fig.~\ref{fignsf:supp_reflection_synthetic} respectively. We also provide NeRF-based methods with ground truth camera poses, which results in higher fidelity NeRF-based reconstruction than on real-world data. Overall, we observe similar trends to the real-world examples, with most multi-image based methods failing to remove the majority of the obstructions for the majority of scenes. This is with the exception of Liu et al.~\cite{liu2020learning} for the \textit{Geese}, \textit{Vending} and \textit{Butterfly} scenes in Fig.~\ref{fignsf:supp_occlusion_synthetic}, where it succeeds at removing a large portion of the fence occluders. We believe this is a strong indication that this method relies heavily on visual cues to identify the occluder (e.g., gray mostly-in-focus fences), and helps to explain its failure to identify and remove other categories of obstructions such as the black hexagonal grids in Fig.~\ref{fignsf:supp_results_occlusion}. Lama~\cite{suvorov2021resolution}, when provided with a ground-truth occlusion mask, is able to reconstruct a relatively coherent transmission layer. However, upon closer inspection the results are missing details in the ground-truth transmission layer, such as the distorted text in \textit{Sign}  and missing beak of \textit{Pigeon} in Fig.~\ref{fignsf:supp_occlusion_synthetic}. We observe that both multi-image methods and DSRNet~\cite{hu2023single} fail to effectively remove reflections in Fig.~\ref{fignsf:supp_reflection_synthetic}, with DSRNet~\cite{hu2023single} accidentally enhancing the reflected content in the \textit{Sealions} scene. These observations are supported by quantitative results, with our method achieving the highest PSNR and SSIM  across all scenes tested. We observe an average PSNR increase of more than 10db, with near-perfect reconstruction of both obstructions and obstructed content; though emphasize that these results represent a validation of the models in a simplified imaging setting, and are not fully representative of performance across diverse in-the-wild scenarios.

\begin{figure}[h!]
    \centering
    \includegraphics[width=0.6\linewidth]{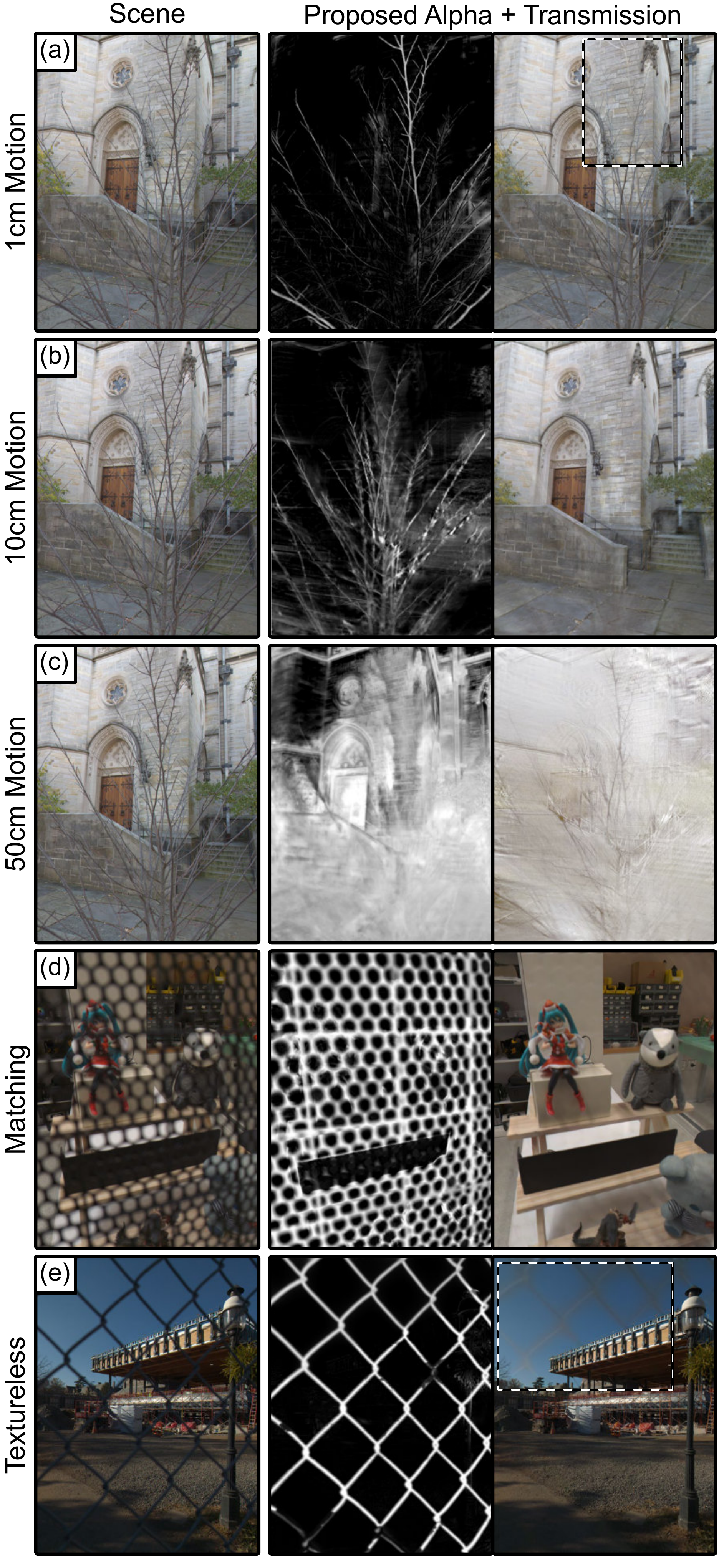}
    \caption{Challenging image reconstruction cases including varying scales of camera motion, overlap between occluder and transmission colors, and residual signal left on scene content in low-texture regions. Areas of interest highlighted with dashed border.}
    \label{fignsf:supp_challenging}
    \vspace{-1.5em}
\end{figure}
\begin{figure}[t!]
    \centering
    \includegraphics[width=0.7\linewidth]{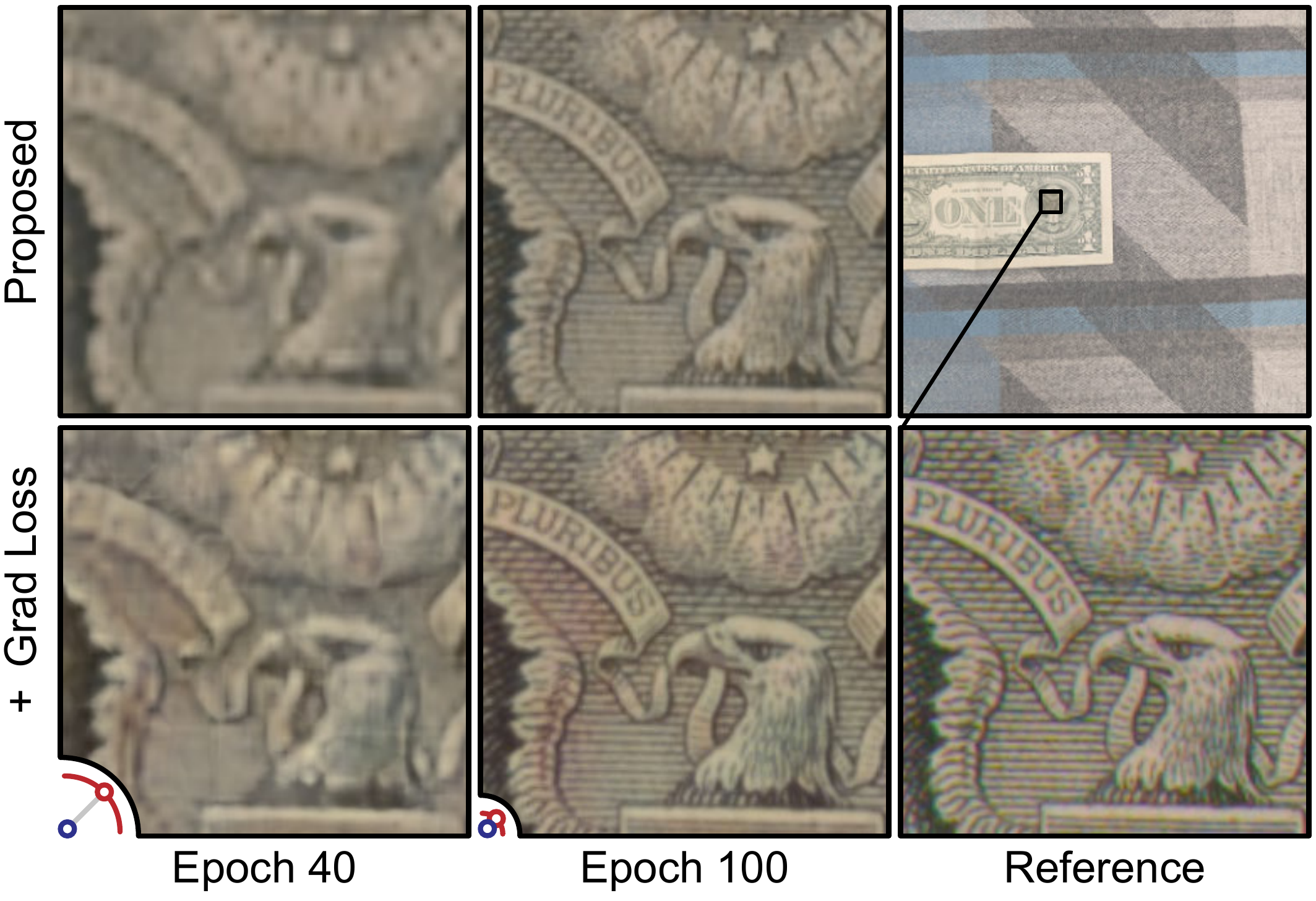}
    \caption{Visualization of the effects of gradient loss $\mathcal{L}_\textsc{g}$ on image reconstruction at 25x zoom. Inset bottom left is the radius of perturbation at epoch 40 and epoch 100, the end of training. }
    \label{fignsf:supp_gradient_loss}
\end{figure}
\noindent\textbf{Shadow Removal}\hspace{0.1em} In Fig.~\ref{fignsf:supp_shadow} we demonstrate shadow removal results for scenes with disparate lighting conditions: (a) a book illuminated by a diffuse overhead lamp, (b) a poster illuminated by an array of LEDs, and (c) a bust illuminated by a strong point light source. We note that the grid of LEDs act as a set of point light sources, producing multiple copies of the shadow to be overlayed on the scene. In all settings we are able to extract the shadow with the same obstruction network defined in the \textit{shadow removal} application in Tab.~\ref{tabnsf:application-configs}, further reinforcing the our image fitting findings from Fig.~\ref{fignsf:supp_image_fitting}. Namely that coordinate networks with low-resolution multi-resolution hash encodings are able to effectively fit both scenes comprised of smooth gradients, as in the diffuse shadow case, and limited numbers of image discontinuities, as in the multiple point source case. In (c) we furthermore see that while the photographer-cast shadow is successfully removed from the bust, the shadows cast by other light sources are left intact. This reinforces that our proposed model is separating shadows based not only on their color, but on the motion they exhibit in the scene; as the other shadows cast on the bust undergo the same parallax motion as the bust itself.

\noindent\textbf{Challenging Settings}\hspace{0.1em} We compile a set of challenging imaging settings in Fig.~\ref{fignsf:supp_challenging} which highlight areas where our proposed approach could be improved. One limitation of our work is that it cannot generate unseen content. While this means it cannot hallucinate features from unreliable image priors, it also means that it is highly parallax-dependent for generating accurate reconstructions. This is highlighted in Fig.~\ref{fignsf:supp_challenging} (a-c), where with hand motion on the scale of 1cm is only enough to separate and remove the topmost branch of the occluding plant. Motion on the scale of 10cm is enough to remove most of the branches, but larger motion on the scale of half a meter in diameter causes the reconstruction to break down. This is likely due to the small motion and angle assumptions in our camera model, as it is not able to successfully jointly align the input image data and learn its multi-layer representation. Thus work on large motion or wide-angle data for large obstruction removal --  e.g., removing telephone poles blocking the view of a building  -- remains an open problem. Fig.~\ref{fignsf:supp_challenging} (d) demonstrates the challenge of estimating an accurate alpha matte when the transmitted and obstructed content are matching colors. In this case, although the obstruction is ``removed", we see that the alpha matte is missing a gap around the black object in the scene behind the occluder. In this region the model does not need to use the obstruction layer to represent pixels that are already black in the transmission layer -- in fact, the alpha regularization term $\mathcal{R}_\alpha$ would penalize this. Thus the alpha matte is actually a produce of both the actual alpha of the obstruction and its relative color difference with what it is occluding. Fig.~\ref{fignsf:supp_challenging} (e) highlights a related problem. In regions where the transmission layer is low-texture, and lacks parallax cues, it is ambiguous what is being obstructed and where the border of the obstruction lies. Thus ghosting artifacts are left behind in areas such as the sky of the \textit{Textureless} scene. What is noteworthy, however, is that these are also exactly the regions in which in-painting methods such as Lama~\cite{suvorov2021resolution} are most successful, as there are no complex textures that need to be recovered from incomplete data, leaving a hybrid model as an interesting direction for future work.


\noindent
fignsf:\section{Additional Experiments and Analysis}
\label{secnsf:supp_experiments}

\noindent\textbf{Gradient Loss}\hspace{0.1em} A significant challenge posed by the task of aggregating long-burst data is the so-called problem of ``regression to the mean''. When minimizing a metric such as relative mean-square error, which penalizes small color differences significantly less than large discrepancies, the final reconstruction is encouraged to be smoother than the original input data~\cite{bahat2020explorable}. Thus, in developing our approach we explored -- but ultimately did not use -- a form of gradient penalty loss:
\begin{align}
    \mathcal{L}_\textsc{g} &= |(\scalebox{0.7}{$\Delta$}c - \scalebox{0.7}{$\Delta$}\hat c)/(\mathrm{sg}(\scalebox{0.7}{$\Delta$}c) + \epsilon)|^2 \nonumber.
\end{align}
Rather than sample a grid of points around $u^\textsc{o},v^\textsc{o}$ and $u^\textsc{t},v^\textsc{t}$ or perform a second pass over the image networks~\cite{nam2022neural} to compute Jacobians, we compute color gradients $\scalebox{0.7}{$\Delta$} c$ by pairing each ray with an input perturbed in a random direction
\begin{align}\label{eqnsf:perturb}
    \scalebox{0.7}{$\Delta$} c &= I(u,v,t) - I(\tilde u,\tilde v,t) \\
    \tilde u,\tilde v &= u + r\mathrm{cos}(\phi), \,\, v + r\mathrm{sin}(\phi), \quad \phi \sim \mathcal{U}(0,2\pi)\nonumber,
\end{align}
where $r$ determines the magnitude of the perturbation. The estimated color gradient $\scalebox{0.7}{$\Delta$} \hat c$ is similarly calculated for the output colors of our model. Illustrated in Fig.~\ref{fignsf:supp_gradient_loss}, by reducing radius $r$ from multi-pixel to sub-pixel perturbations during training, we are able to improve fine feature recovery in the final reconstruction via gradient loss $\mathcal{L}_\textsc{g}$ without significantly impacting training time -- as perturbed samples are also re-used for regular photometric loss calculation $\mathcal{L}_p$. However, as we do not apply any demosaicing or post-processing to our input Bayer array data, we find this loss can also lead to increased color-fringing artifacts -- the red tint in the bottom row of Fig.~\ref{fignsf:supp_gradient_loss}. For these reasons, and poor convergence in noisy scenes, we did not include this loss in the final model. However, there may be potentially interesting avenue of future research into a jointly trained demosaicing module to robustly estimate real color gradient directly from quantized and discretized Bayer array values. 

\begin{figure*}[t]
    \centering
    \includegraphics[width=\linewidth]{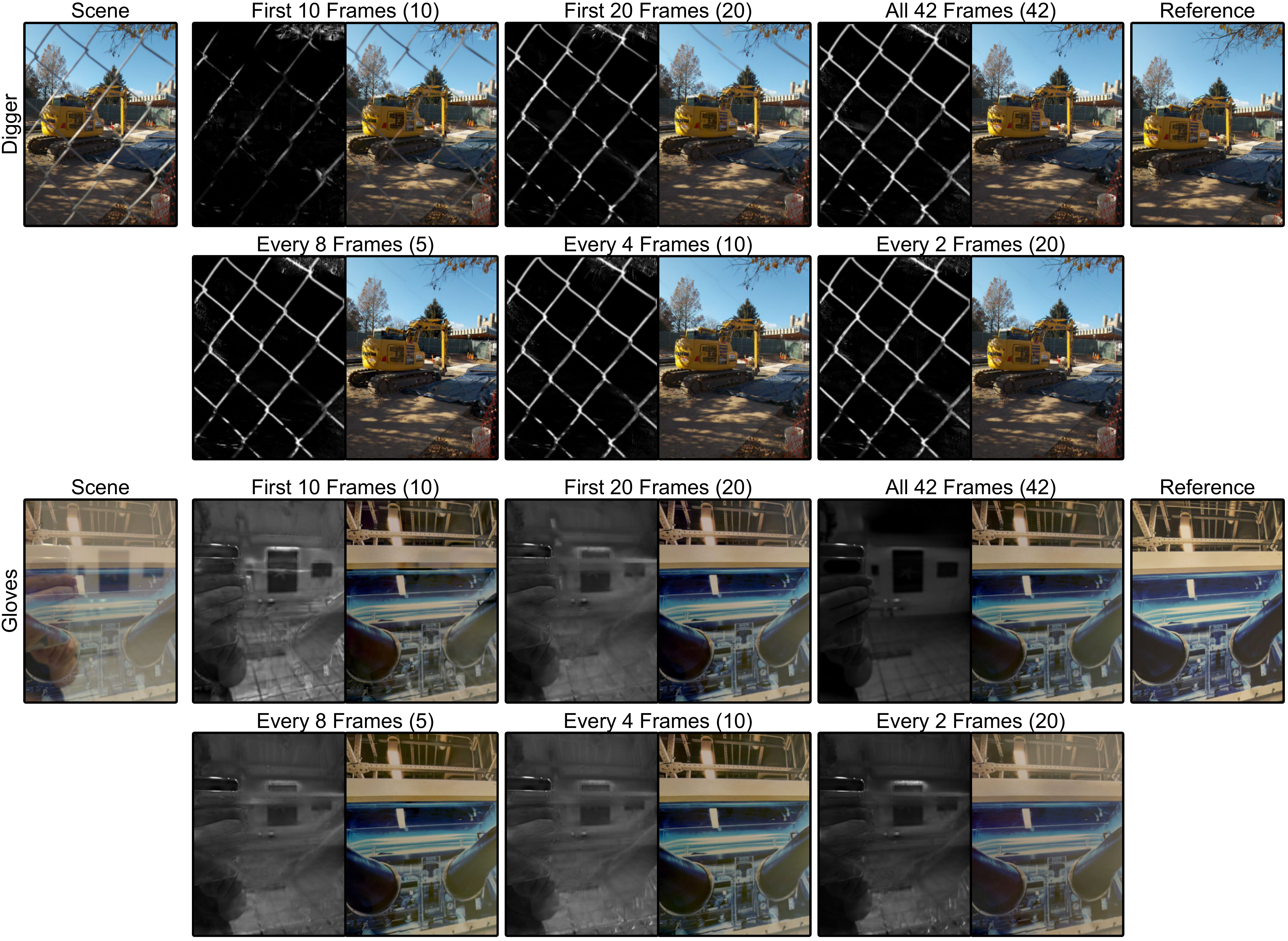}
    \caption{Ablation study on the effects of the number of input frames or duration of capture on transmission layer reconstruction and estimated alpha matte. Total number of frames input into the model denoted by the number in parentheses-- e.g., (10) = ten frames.}
    \label{fignsf:supp_ablation_frames}
\vspace{1em}
\end{figure*}
\begin{figure*}[h!]
    \centering
    \includegraphics[width=0.9\linewidth]{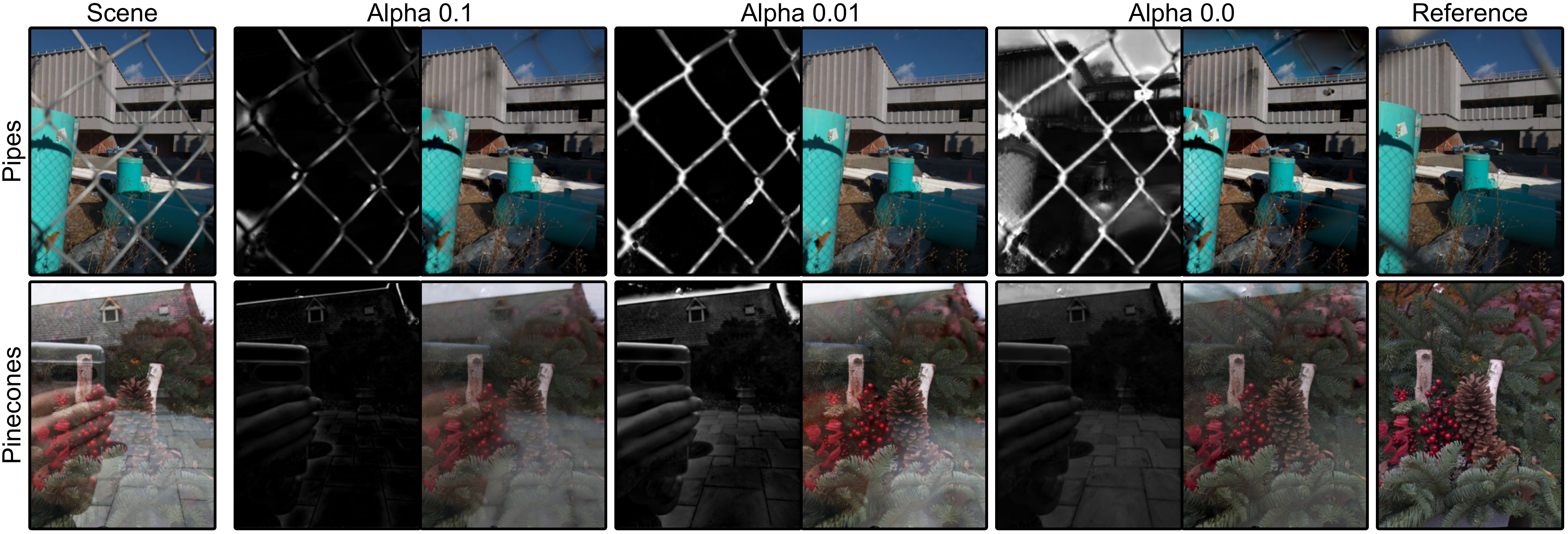}
    \caption{Ablation study on the effects of alpha regularization weight $\eta_\alpha$ on transmission layer reconstruction and estimated alpha matte.}
    \label{fignsf:supp_ablation_alpha}
    \vspace{1em}
\end{figure*}
\begin{figure*}[t!]
    \centering
    \includegraphics[width=\linewidth]{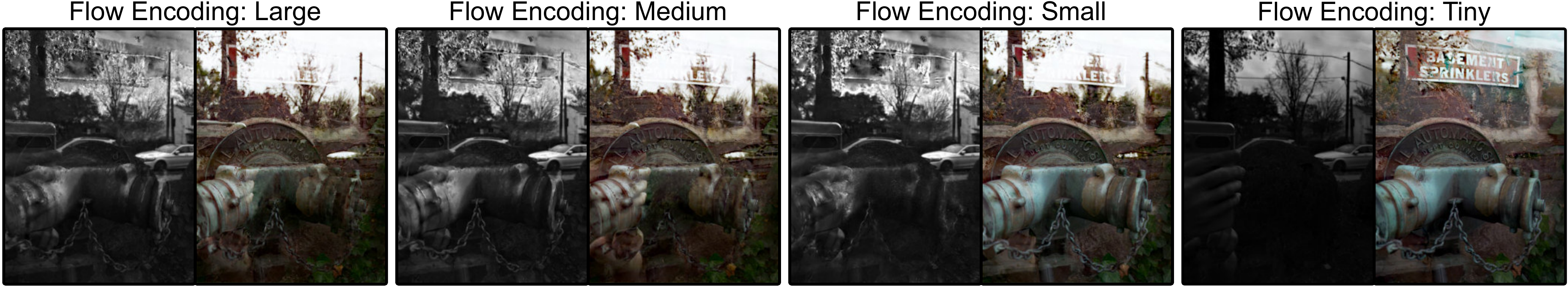}
    \vspace{-1.5em}
    \caption{Ablation study on the effects of flow encoding size (Tab.~\ref{tabnsf:network-sizes}) on transmission layer reconstruction and estimated alpha matte.}
    \label{fignsf:supp_ablation_network}
\end{figure*}
\begin{figure}[h]
    \centering
    \includegraphics[width=0.75\linewidth]{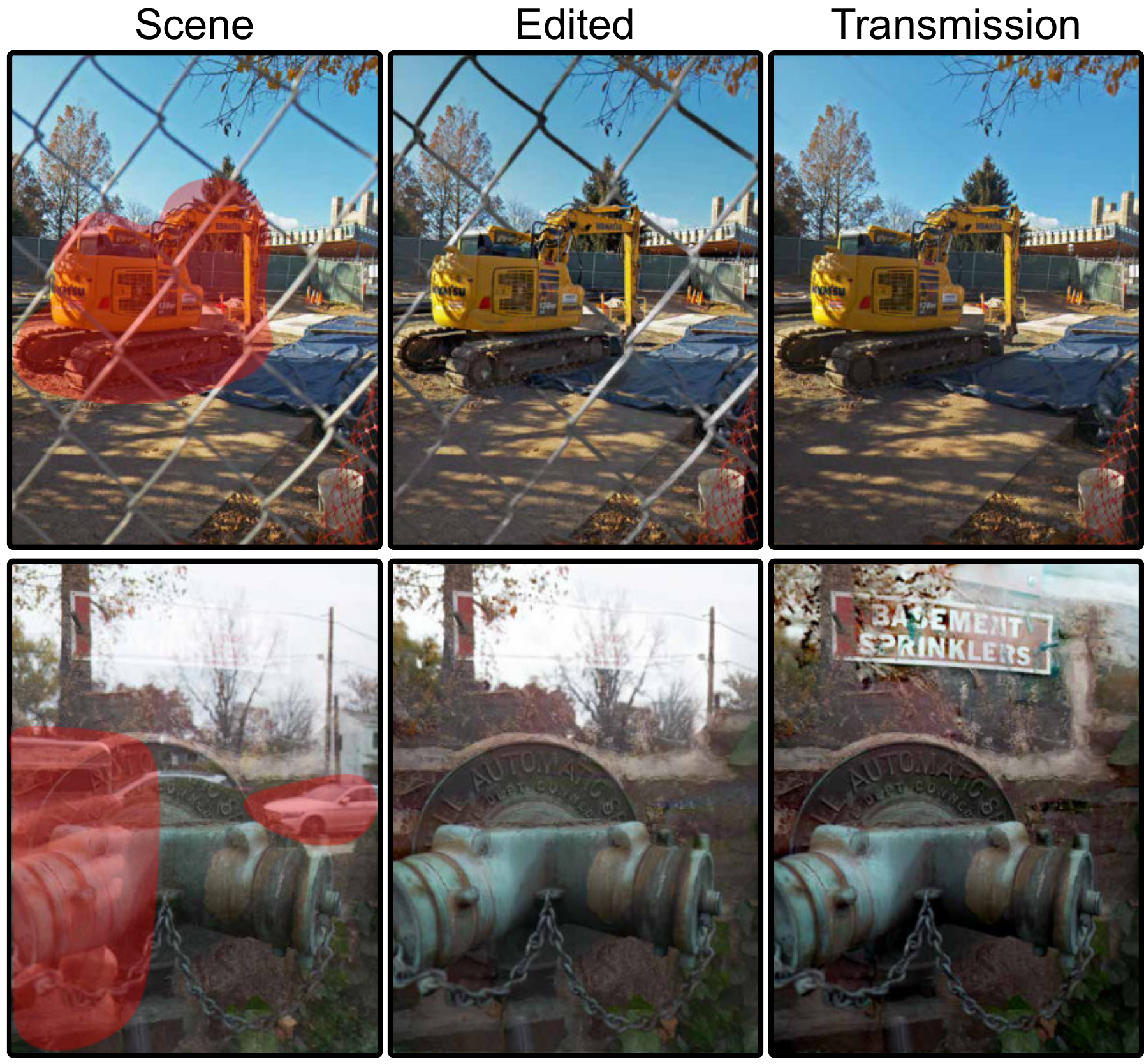}
    \caption{Demonstration of user-interactive scene editing facilitated by layer separation. Only the user-selected region of the obstruction, highlighted in red, is removed without affecting surrounding scene content, see text.}
    \label{fignsf:supp_editting}
\end{figure}
\noindent\textbf{Alpha Regularization Ablation}\hspace{0.1em} In Fig.~\ref{fignsf:supp_ablation_alpha}, we visualize the effects of alpha regularization weight $\eta_\alpha$ on reconstruction. The primary function of this regularization is remove low-parallax content from the obstruction layer, as there is no alpha penalty for reconstructing the same content via the transmission layer. As seen in the \textit{Pipes} example, without alpha regularization the obstruction layer is able to freely reconstruct part of the transmitted scene content such as the sky, the pipes, and the walls of the occluded buildings. A small penalty of $\eta_\alpha=0.01$ is enough to remove this unwanted content from the obstruction layer, while $\eta_\alpha=0.1$ is enough to also start removing parts of the actual obstruction. Contrastingly, in the case of reflection scenes such as \textit{Pinecones}, even a relatively small alpha regularization weight of $\eta_\alpha=0.01$ removes part of the actual reflection -- leaving behind a grey smudge in the bottom right corner of the reconstruction. As reflections are typically partially transparent obstructions, and can occupy a large area of the scene, removing them purely photometrically is ill-conditioned. There is no visual difference between a gray reflector covering the entire view of the camera and the scene actually being gray. Thus $\eta_\alpha$ can also be a user-dependent parameter tuned to the desired ``amount'' of reflection removal.\\
\noindent\textbf{Frame Count Ablation}\hspace{0.1em} Thusfar we have used all 42 frames in each long-burst capture as input to our method, but we highlight that this is not a requirement of the approach. The training process can be applied to any number of frames -- within computational limits. In Fig.~\ref{fignsf:supp_ablation_frames} we showcase reconstruction results for both subsampled captures, where only every $k$-th frame of the image sequence is kept for training, and shortened captures, where only the first $n$ frames are retained. Similar to the problem of depth reconstruction~\cite{chugunov2023shakes}, we find that obstruction removal performance directly depends on the total amount of parallax in the input. Sampling the \textit{first} 10 frames -- approximately 0.5 seconds of recording -- results in diminished obstruction removal for both the \textit{Digger} and \textit{Gloves} scenes as the obstruction exhibits significantly less motion during the capture. In contrast, given \textit{a five frame input sampled evenly across the full two-second capture}, our proposed approach is able to successfully reconstruct and remove the obstruction. This subsampled scene also \textit{trains considerably faster}, converging in only 3 minutes as less frames need to be sampled per batch -- or equivalently more rays can be sampled from each frame for each iteration. This further validates the benefit of a long burst capture.\\
\noindent\textbf{Flow Encoding Size Ablation}\hspace{0.1em} A key model parameter which controls layer separation, as discussed in Section ~\ref{secnsf:supp_implementation}, is the size of the encoding for our neural spline flow fields. In Fig.~\ref{fignsf:supp_ablation_network} we illustrate the effects on obstruction removal of over-parameterizing this flow representation. When the two layers are undergoing simple motion caused by parallax from natural hand tremor, a \textit{Tiny} flow encoding is able to represent and pull apart the motion of the reflected and transmitted content. However, high-resolution neural spline fields, just like a traditional flow volume $h(u,v,t)$, can quickly overfit the scene and mix content between layers. We can see this clearly in the \textit{Large} flow encoding example where the reflected phone, trees, and parked car appear in both the obstruction alpha matte and transmission image. Thus it is critical to the success of our method to construct a task-specific neural spline field representation appropriate for the expected amount and density of scene motion.

\noindent\textbf{Applications to Scene Editing}\hspace{0.1em} In Fig.~\ref{fignsf:supp_editting} we showcase the scene editing functionality facilitated by our proposed methods layer separation. As we estimate an image model for both the transmission and obstruction, we are not limited to only removing a layer but can independently manipulate them. In this example we rasterize both layers to RGBA images and input them into an image editor. The user is then able to highlight and delete a portion of the occlusion while retaining its other content. Thus we can create physically unrealizable photographs such as only the fence appearing to be behind the \textit{Digger}, or selectively remove the photographer's hand and parked car from the \textit{Hydrant} scene.

\chapter{Image Stitching\label{ch:pan}}
In this chapter, we revisit the topic of panoramic image stitching. Unlike the image data in Chapters \ref{ch:depth} and \ref{ch:layer}, where the focus was on extracting information from small motion, the primary challenge here lies in aggregating information over large rotational motion, as the camera rotates 90, 180, or even 360 degrees during capture. Additionally, we aim to create an interactive scene model which is able to reconstruct effects such as scene motion, lighting changes, and depth parallax, while remaining compact and able to render views in real time. To achieve this, we propose a novel Neural Light Sphere (NeuLS) model, a combination neural field model for camera ray deformation and view-dependent color. The ray deformation model handles effects such as parallax and local scene flow, bending rays to follow the apparent non-rigid motion of pixels between images, while the view-dependent color model captures effects such as reflections and lighting changes. This single layer model avoids expensive volume sampling operations, as we do not have to march camera rays through space to aggregate color information, and results in a total model size of 80 MB per scene, and real-time (50 FPS) rendering at 1080p resolution
\vspace{1em}
\hrule
\vspace{1em}
\noindent  \textit{This chapter is based on the work ``Neural Light Spheres for Implicit Image Stitching and View Synthesis"~\cite{chugunov2024neuls} by Ilya Chugunov, Amogh Joshi, Kiran Murthy, Francois Bleibel, and Felix Heide presented at SIGGRAPH Asia 2024.}

\graphicspath{{chapters/3/PanPan/}}
  
 \vspace{-1em}
\section{Introduction}
The \textit{panorama} of the 19th century was typically a commissioned collection of paintings in a cylindrical arrangement with a dedicated viewing platform to maximize observers' immersion in the work \cite{trumpener2020viewing}. The digital panorama of the 21st century is typically a long rectangle left un-shared -- or un-\textit{viewed} -- in the storage space of the cellphone used to capture it. Yet, arguably the most common form of digital panorama might be the one that is un-\textit{taken}, where the user decides that the hassle of acquisition -- e.g., slowly and carefully sweeping the camera in a level arc across a scene -- is not worth the final product.

To address this imbalance, we can simplify acquisition, increase the appeal of the final product, or (preferably) do both. Moving from cylindrical warping~\cite{szeliski1997creating} and seam matching~\cite{zomet2006seamless} approaches to more parallax-tolerant image stitching processes~\cite{zaragoza2013projective, zhang2014parallax} allows the photographer to take a less restricted camera path and still produce a high-quality panorama. However, the end result remains a single static image. Work on multi-layer depth panoramas~\cite{zheng2007layered, lin2020deep} and panoramic mesh reconstruction~\cite{hedman2017casual} offer a more interactive experience than a traditional panorama, able to use parallax information to render novel views of the scene.  The recent explosion in radiance field methods~\cite{mildenhall2021nerf, kerbl20233d} can be seen as an evolution of ``interactive panoramas'', with a line of connected works from image-based rendering~\cite{chen1993view} to direct view synthesis~\cite{flynn2016deepstereo} and hybrid 3D and image feature approaches~\cite{sitzmann2019deepvoxels}. Neural radiance field (NeRF) methods can produce fast~\cite{muller2022instant} scene reconstructions which model for both parallax and view-dependent lighting effects with high visual quality~\cite{barron2023zip} and from unstructured and unknown poses~\cite{lin2021barf}. However, outward or front-facing panoramas present a major challenge for these volumetric representations, as large parts of the scene are only observed for a few frames before falling out of view, turning scene reconstruction into a collection of sparse view problems~\cite{niemeyer2022regnerf}.

In this work we explore a compact neural light field~\cite{attal2022learning} model for panoramic image stitching and view synthesis; capable of encoding depth parallax, view-dependent lighting, and local scene motion and color changes. We represent the scene as a color-on-a-sphere model decomposed into two components: a view-dependent ray offset model for parallax, lens distortion, and smooth motion; and a view-dependent color model for occluded content, reflections, refraction, and color changes. Taking as input an arbitrary path panorama -- vertical, horizontal, random-walk -- we fit our model at \textit{test-time} to jointly estimate the camera path, and produce a high-resolution stitched representation of the scene. We demonstrate how this model enables geometrically consistent field-of-view expansion, transforming portrait-mode panoramas into immersive, explorable wide-view renders.

Specifically, we make the following contributions:
\begin{itemize}
  \item A compact and efficient (80 MB model size per scene, 50 FPS rendering at 1080p resolution) two-stage neural light sphere model of panoramic photography.
  \item Validation of panoramic image stitching and view synthesis performance under varying imaging settings, including low-light conditions, with comparisons to traditional image stitching and radiance field approaches. 
  \item An Android-based data collection tool for streaming and recording full-resolution RAW image arrays, camera and system metadata, and on-board device measurements such as gyroscope and accelerometer values.
  \item A diverse collection of 50 indoor and outdoor handheld panoramic scenes recorded from all three on-device cameras with full 10-bit color depth, 12-megapixel resolution.
\end{itemize}

\noindent We make our code, data, and data collection app available open-source on our project website: {\color{URLBlue}\href{https://light.princeton.edu/NeuLS}{light.princeton.edu/NeuLS}}


\section{Related Work}
\paragraph{Image Stitching} There is a rich history of methods for stitching or \textit{mosaicing}~\cite{burt1983multiresolution} multiple images into one, with demand for the task long pre-dating the invention of digital photography~\cite{shepherd1925interpretation}. A common approach is to first extract image features, either directly calculated~\cite{brown2007automatic, lowe2004distinctive} or learned~\cite{sarlin2020superglue}, which are matched to position and warp images together~\cite{gao2011constructing}. Allowing for image transforms beyond simple homographies~\cite{hartley2003multiple} can allow for parallax-tolerant image warping and stitching~\cite{shum2002construction, zhang2014parallax}, reducing blur from pixel disparity between views. Seam-carving approaches dynamically adjust the stitching boundaries to better match visual features~\cite{agarwala2004interactive, gao2013seam}, helping to avoid artifacts from mismatched content on image boundaries. Inspired by local deformation image stitching~\cite{zaragoza2013projective} and panoramic video texture~\cite{agarwala2005panoramic} work, we develop a neural field model which can accommodate for both parallax and scene motion during reconstruction. However, rather than use sparse pre-computed features and break the reconstruction pipeline into multiple discrete steps, we leverage a neural scene representation and fast ray sampling to optimize our model end-to-end over dense pixel-wise photometric loss.

\paragraph{Layered and Depth Panoramas} Concentric mosaic~\cite{shum1999rendering} and layered depth map~\cite{shade1998layered} representations offer a compact way to model the effects of parallax and occlusion in a scene. Layered depth panoramas~\cite{zheng2007layered} make use of a layered representation to produce an interactive image stitching reconstruction, able to render novel views through trigonometric reprojection. Follow-on work extends this reconstruction to mesh representations~\cite{hedman2017casual,hedman2018instant} and learned features~\cite{lin2020deep}, offering improved reconstruction of object surfaces which are otherwise occluded between depth layers. Work in this space often targets VR applications~\cite{bertel2020omniphotos, lai2019real, attal2020matryodshka}, as they drive demand for high-quality immersive and interactive user experiences in 3D environments. Also related are video mosaic approaches~\cite{rav2008unwrap, kasten2021layered}, which forgo re-rendering to decompose a video into a direct 2D-to-2D pixel mapping onto a set of editable atlases. In this work, we target reconstructions that can provide an interactive user experience with minimal hardware or camera motion requirements~\cite{bertel2020omniphotos}, and which are able to tolerate moderate scene motion and color changes.

\paragraph{Light Field Methods} Modeling ray color as a product of three dimensional spatial and angular components, a light field can fully represent effects of depth parallax, reflections, and refraction in a scene~\cite{lfr,ng2005light} at the cost of high data, storage, and computational requirements~\cite{wilburn2005high}. Lumigraphs~\cite{gortler1996lumigraph} make use of a simpler geometric proxy -- e.g., the crossing points of a ray intersecting with two planes -- to represent the spatial and angular components of a light field, greatly lowering data and computational requirements for reconstruction and rendering~\cite{chai2000plenoptic}. Motivated by recent work in neural light field representations~\cite{attal2022learning,suhail2022light}, we develop a compact spherical representation which decomposes the scene into view-dependent ray offset -- for effects such as parallax and local motion -- and view-dependent color for occlusions and time-dependent content.
\paragraph{Neural Scene Representations} Recent work in neural scene representations, particularly in the area of neural radiance fields (NeRFs) \cite{barron2023zip,mildenhall2021nerf}, has demonstrated that high quality scene reconstruction can be achieved without pixel arrays, voxel grids, or other explicit backing representations. These approaches train a neural network at \textit{test time} -- starting with an untrained network, overfit to a single scene -- to map from encoded~\cite{tancik2020fourier} coordinates to output parameters such as color~\cite{nam2022neural}, opacity~\cite{martin2021nerf}, density~\cite{corona2022mednerf}, depth~\cite{chugunov2023shakes}, camera lens parameters~\cite{xian2023neural}, and surface maps~\cite{morreale2021neural}. While they are not neural scene representations, forward projection ``Gaussian Splatting''~\cite{kerbl20233d} models have recently exploded in popularity as an alternative to NeRF scene representations, offering increased rendering speed by avoiding costly volume sampling operations. However, outward panoramic captures with largely rotational motion present a challenge for these methods, which rely on large view disparity to localize content in 3D space. We instead propose a view-dependent ray offset and color model to reconstruct local parallax and view-dependent effects from minimal view disparity. By embedding this representation on a spherical surface, we also substitute costly NeRF volume sampling with efficient ray-sphere crossings, resulting in a compact 80 MB model capable of real-time 1920x1080px rendering at 50 FPS.

\section{Neural Light Sphere Reconstruction}
In this section we describe our proposed neural light sphere model for implicit image stitching and re-rendering. We begin with an overview of our backward projection model for unstructured panoramic captures. We then discuss the neural field representations backing this model, its loss and training procedure, how we collect scene data for reconstruction, and implementation details.

\begin{figure}[t!]
{\includegraphics[width=\textwidth]{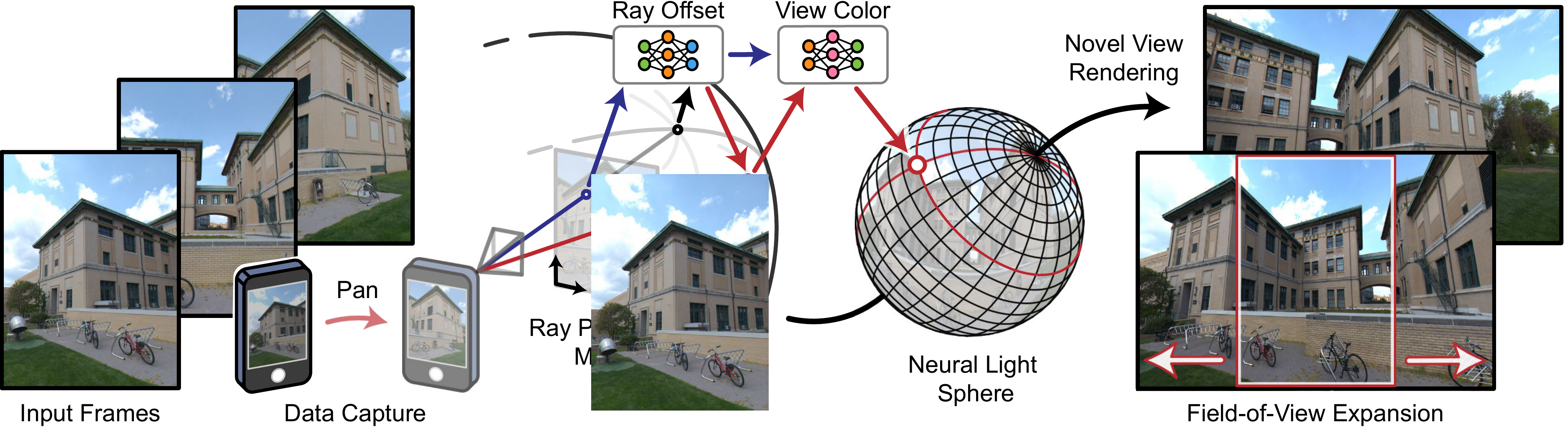}
\caption{Fit during test-time directly to an input panoramic video capture, with no pre-processing steps, our neural light sphere model produces a parallax, lighting, and motion-tolerant reconstruction of the scene. Placing a virtual camera into the sphere, we can generate high-quality wide field-of-view renders of the environment, turning what would otherwise be a static panorama into an interactive viewing experience.}
\label{figpan:teaser}}
\end{figure}

\subsection{Projective Model of Panoramic Imaging}
\label{secpan:projective_model}
\begin{figure*}[t!]
 \centering
\includegraphics[width=\textwidth]{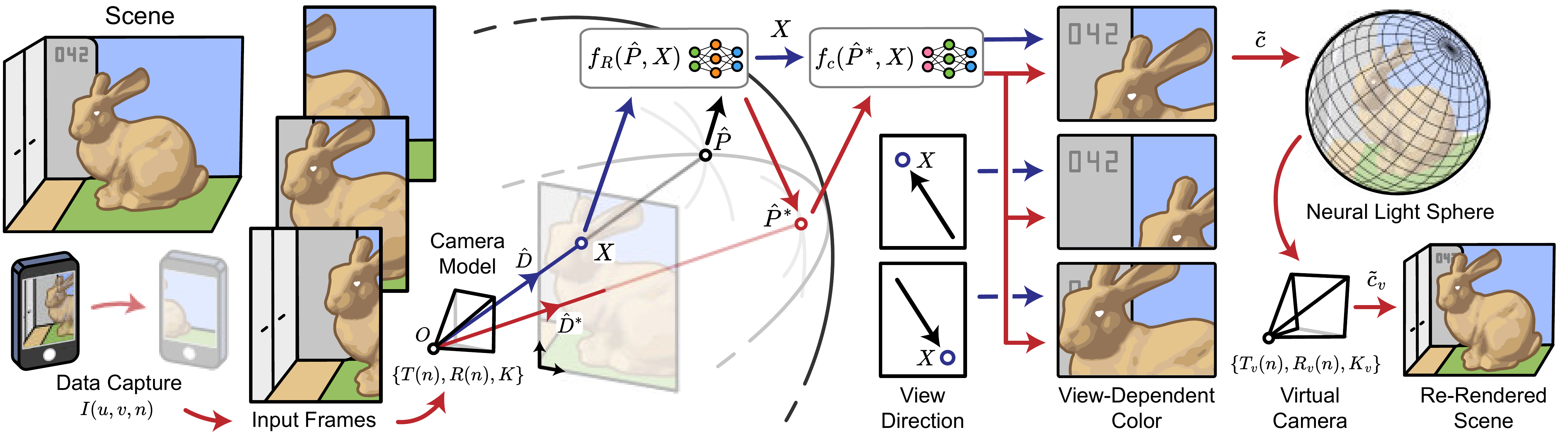} \vspace{-1em}
  \caption{\textbf{Neural Light Sphere Model.} Taking as input panoramic video capture $I(u,v,n)$, we perform backward camera projection from a point $X=(u,v)$ into a spherical hull to estimate an initial intersection point $P$. Ray offset model $f_{\textsc{r}}(\hat{P},X)$ then bends this ray to a corrected point $\hat{P}^*$, which is used to sample the view-dependent color model $f_{\textsc{c}}(\hat{P}^*,X)$. Simulating a new virtual camera with our desired position and FOV, we use this neural light sphere model to re-render the scene to novel views.} \vspace{-1em}
\label{figpan:method}
\end{figure*}

In this work, we adopt a spherical backward projection model~\cite{szeliski2007image} for our scene representation. That is, we model each image in the input video as the product of rays originating at the camera center intersecting with the inner surface of a sphere. To simplify notation, we outline this process for a single ray below, illustrated in Fig.~\ref{figpan:method}, and later generalize to batches of rays. Let
\begin{align}
    c = [\mathrm{R},\mathrm{G},\mathrm{B}]^\top = I(u,v,n)
\end{align}
be a colored point sampled at image coordinates $u,v\in [0,1]$ from a frame $n \in [0,N{-}1]$ in a video $I(u,v,n)$, where $N$ is the total number of captured video frames. To project this point to a camera ray, we introduce camera rotation $R(n)$ and translation $T(n)$ models
\begin{align}\label{eqpan:translation_rotation}
    &T(n) = \mathbf{T}_{n}, \quad R(n) = \mathrm{rot}(\eta_\textsc{r}\mathbf{R}_n) \mathbf{G}_n \nonumber \\
    &\mathbf{T}_{n} =  
    \left[\arraycolsep=2.0pt
    \begin{array}{c}
    t_x \\
    t_y \\
    t_z \\
    \end{array}\right],\,\,
    \mathbf{R}_{n} =  
    \left[\arraycolsep=2.0pt
    \begin{array}{c}
    r_x \\
    r_y \\
    r_z \\
    \end{array}\right],\,\,
    \mathrm{rot}(\mathbf{R}_n)  = \left[\begin{array}{ccc}
1 & -r_z & r_y \\
r_z & 1 & -r_x \\
-r_y & r_x & 1
\end{array}\right].
\end{align}
Here, we model translation for frame $n$ as three dimensional motion, initialized at zero. $R(n)$ is a small-angle approximation~\cite{boas2006mathematical} offset $\mathbf{R}_n$ to device rotation $\mathbf{G}_n$ recorded from the phone onboard gyroscope, weighted by $\eta_\textsc{r}$. With calibrated intrinsics matrix $K$, sourced from device camera metadata, we project the point at $u,v$ sampled from frame $n$ to a ray with origin $O$ and direction $D$ as
\begin{align}\label{eqpan:ray_generation}
    O \,{=}\, \left[\arraycolsep=2.0pt
    \begin{array}{c}
    O_x \\
    O_y \\
    O_z \\
    \end{array}\right] = T(n), \, \, \quad D = \left[\arraycolsep=2.0pt
    \begin{array}{c}
    D_x \\
    D_y \\
    D_z \\
    \end{array}\right] = R(t)K^{-1} \left[\arraycolsep=2.0pt
    \begin{array}{c}
    u \\
    v \\
    1 \\
    \end{array}\right].
\end{align}
We normalize the direction vector $\hat{D} = D/\left\|D\right\|$ to simplify reprojection steps. Next, we define our image model to lie on the surface of a sphere, and calculate its intersection point $P$ with this ray as
\begin{align}\label{eqpan:sphere_intersection}
   \hat{P} &=  P/\left\|P\right\|, \quad P \,{=}\, \left[\arraycolsep=2.0pt
    \begin{array}{c}
    P_x \\
    P_y \\
    P_z \\
    \end{array}\right] = O + t\hat{D}\nonumber\\
    t &= -\left(O \cdot \hat{D}\right) + \sqrt{(O \cdot \hat{D})^2 - (\left\| O\right\|^2 - 1)},
\end{align}
assuming a sphere of radius 1, centered at $[0,0,0]^{\top}$, with the ray originating within its radius ($\left\| O\right\|^2 < 1$). However, as this sphere model, in general, does not match the true scene geometry, we introduce a ray offset model $f_{\textsc{r}}(\hat{P},X)$ to offset the ray direction as

\begin{align}\label{eqpan:ray_distortion}
   \hat{D}^* =  D^*/\left\|D^*\right\|, \quad D^* = \mathrm{rot}\left(\mathbf{R} = f_\textsc{r}(\hat{P}, X)\right)\hat{D},
\end{align}
where $X = [u,v]^\top$ is the ray's originating image coordinates, and $\mathrm{rot(\mathbf{R})}$ is the small-angle rotation model from Eq.~\ref{eqpan:translation_rotation}. We can observe that this model generalizes effects such as parallax (deflecting rays as a function of position via $\hat{P}$) and lens distortion (deflecting rays as a function of their angle relative to the camera center via $X$). With this corrected ray $(O, \hat{D}^*)$, we re-sample our sphere via Eq.~\ref{eqpan:sphere_intersection} to generate a new intersection point $\hat{P}^*$. To map this point to estimated scene color $\tilde{c}$, we introduce a view-dependent color model $f_\textsc{c}$, where
\begin{align}\label{eqpan:ray_distortion}
   \tilde{c} = [\tilde{\mathrm{R}},\tilde{\mathrm{G}},\tilde{\mathrm{B}}]^\top = f_\textsc{c}(\hat{P}^*, X).
\end{align}
This model takes as input the camera coordinate $X$, which allows for modeling of view-dependent effects such as occlusions, reflections, motion, and generated content (e.g., flashing lights), and maps it together with the ray intersection on the sphere $\hat{P}^*$ to an output estimated RGB value $\tilde{c}$. To generate novel views we take as input virtual camera intrinsics $K_v$, translation $T_v(n)$ and rotation $R_v(n)$, and repeat Eq.~\eqref{eqpan:translation_rotation}--\eqref{eqpan:ray_distortion} with these new parameters to generate a colored point $\tilde{c}_v$.
\begin{figure}[t!]
 \centering
\includegraphics[width=0.8\linewidth]{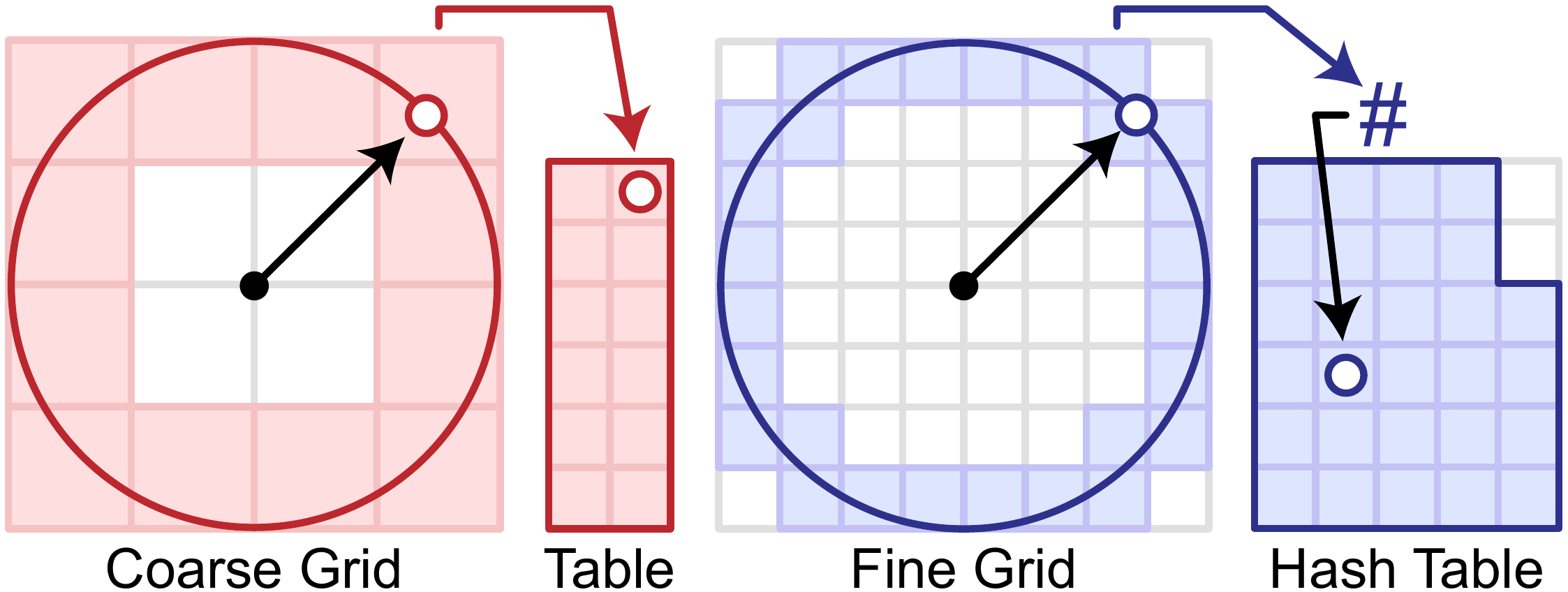}
  \caption{\textbf{Hash Grid Spheres.} In this 2D example we can observe how, for points on a circle, the number of accessed elements in the backing grid roughly doubles for a squaring of grid elements. Given an efficient mapping from grid location to element -- e.g., hash table lookup -- this forms a compact representation even at high resolutions, where storing a dense grid would be computationally intractable.} \vspace{-1em}
\label{figpan:hash_sphere}
\end{figure}

\subsection{Neural Field Representations} \label{secpan:neural_fields}
In the section above, we introduce, but do not \textit{define}, our two core models: $f_\textsc{r}$ for ray offset, and $f_c$ for view-dependent color estimation. Much of the diversity in image stitching and view synthesis approaches can be seen as design choices for these models. For example, $f_\textsc{r}$ could be a layered depth model~\cite{shade1998layered} or cylindrical projection~\cite{mcmillan1995plenoptic}, and $f_\textsc{c}$ could be an explicit color blending~\cite{buehler2001unstructured} or implicit radiance field~\cite{mildenhall2021nerf}, each with tradeoffs in representation power, extrapolation, and input data requirements. With this in mind, we aim to design $f_\textsc{r}$ and $f_\textsc{c}$ to produce a system which is:
\begin{enumerate}
    \item \textbf{Compact}: such that that model is simple to train and has low memory and disk space usage. Thus we minimize the number of components, networks, loss and regularization functions, and avoid pre-processing steps (such as COLMAP~\cite{schonberger2016structure}).
    \item \textbf{Robust}: able to reconstruct a wide range of capture settings (indoor, outdoor, night-time), capture paths, and scene dynamics (e.g., moving clouds, blinking lights). Failing gracefully for hard-to-model effects, with localized reconstruction errors.
\end{enumerate}
\begin{figure}[t!]
 \centering
    \includegraphics[width=0.8\linewidth]{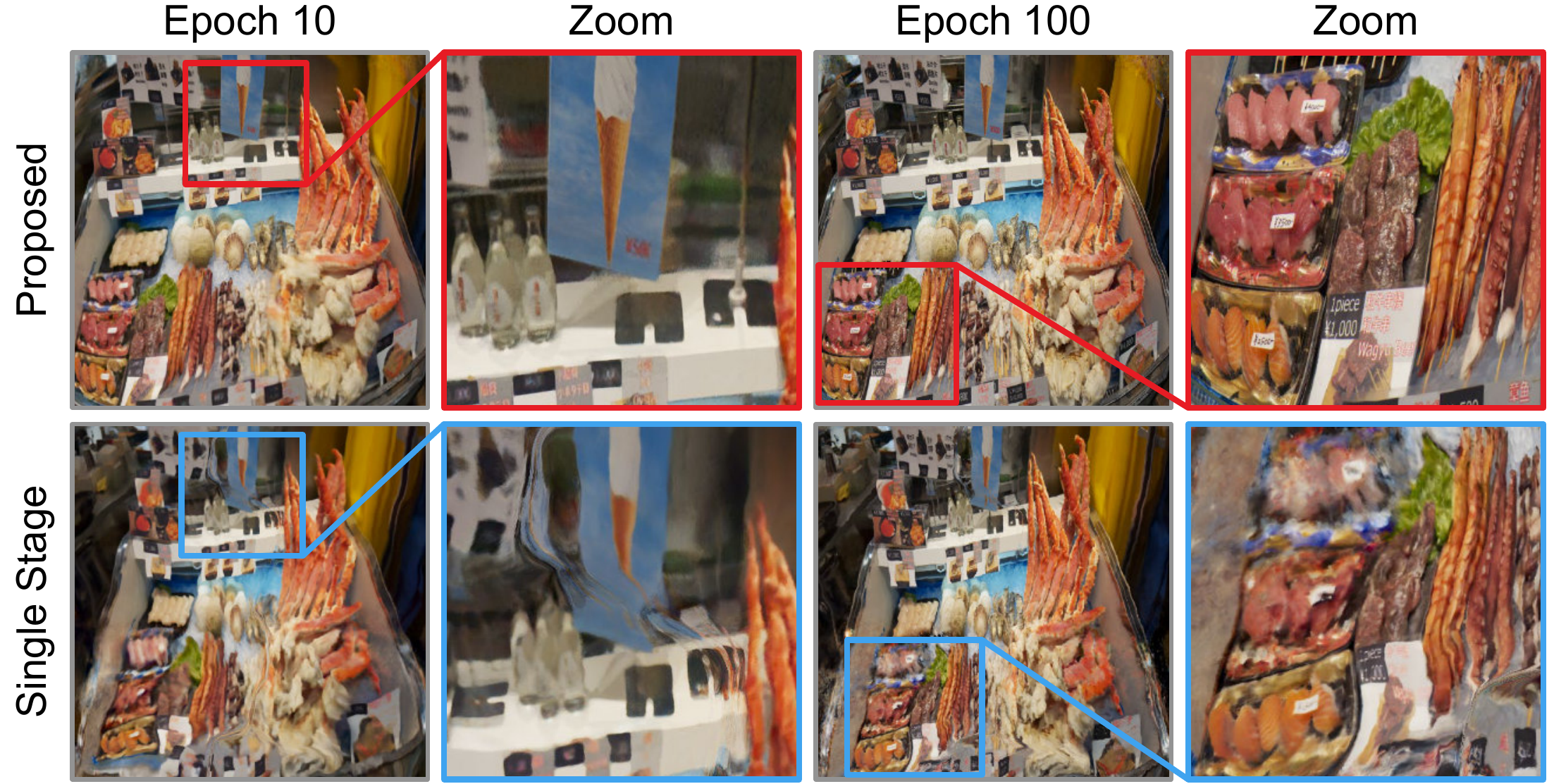}
  \caption{\textbf{Two Stage Training.} Breaking training into two stages allows the camera pose and static image model to first fit an approximation of the scene before view-dependent effects are introduced via $h_\textsc{r}$ and $h_\textsc{d}$. This helps avoid artifacts during early training, like the discontinuities around the sign in the \textit{Single Stage} example, which result in poor final reconstruction quality.} \vspace{-1em}
\label{figpan:two_stage}
\end{figure}
Neural scene representations, particularly with high-level hardware-optimized implementations~\cite{muller2022instant}, offer compelling solutions to this design challenge. By implicitly representing the scene in the weights of a multi-layer perceptron (MLP)~\cite{hornik1989multilayer}, we can effectively turn data storage and retrieval into a component of our inverse imaging model. Correspondingly, we represent ray offset $f_\textsc{r}$ as
\begin{align}
    f_\textsc{r}(\hat{P}, X) = h_\textsc{r}(\gamma_1 (\hat{P}) \oplus \gamma_1 (X);\, \theta_\textsc{r}),
\end{align}
where $\oplus$ denotes concatenation. Here, $h_\textsc{r}$ is an MLP with learned weights $\theta_\textsc{r}$, and $\gamma_1$ is the multi-resolution hash grid encoding from \cite{muller2022instant}, sampled with 3D normalized ray intersection $\hat{P}$ and 2D camera coordinate $X$. During training, $h_\textsc{r}$ learns a mapping between these encoded vectors and the offset applied to $\hat{D}\rightarrow\hat{D}^*$. We similarly construct the view-dependent color model $f_\textsc{c}$ as
\begin{align}
    f_\textsc{c}(\hat{P}^*, X) = h_\textsc{c}\left(h_\textsc{p}(\gamma_2 (\hat{P}^*); \,\theta_\textsc{p}) + h_\textsc{d}(\gamma_1 (X);\, \theta_\textsc{d}); \, \theta_\textsc{c}\right),
\end{align}
The network $h_\textsc{d}$ takes as input camera coordinate $X$ and outputs a vector encoding of view direction; network  $h_\textsc{p}$ similarly encodes the corrected position of the sphere crossing. This combined encoding is then mapped to color via $h_\textsc{c}$. Of note is that $\gamma_2$, the multi-resolution hash encoding applied to $\hat{P}$, and $\gamma_1$, the encoding applied to $\hat{P}^*$, operate in 3D world space on the surface of the unit sphere. That is, \textit{we never convert intersections to spherical coordinates}, and avoid the associated non-linear projection~\cite{zelnik2005squaring} and singularity problems. While it would be exceedingly inefficient to store a sphere in a dense representation of sufficient resolution for high-quality image synthesis (e.g., $4000^3$ voxels for 12-megapixels images, the majority of which would be empty), this is made possible thanks to the hash-grid backing of $\gamma$. Illustrated in Fig.~\ref{figpan:hash_sphere}, as the majority of the grid locations inside in the unit cube are never sampled, since they do not intersect with the unit sphere's surface, the corresponding stored entries in $\gamma$ are never queried. Thus the size of the hash table for $\gamma$ -- which determines its latency, memory usage, and storage requirements -- can be on the order of magnitude of the sphere's surface area rather than its volume.

\begin{figure}[t!]
 \centering
\includegraphics[width=0.8\linewidth]{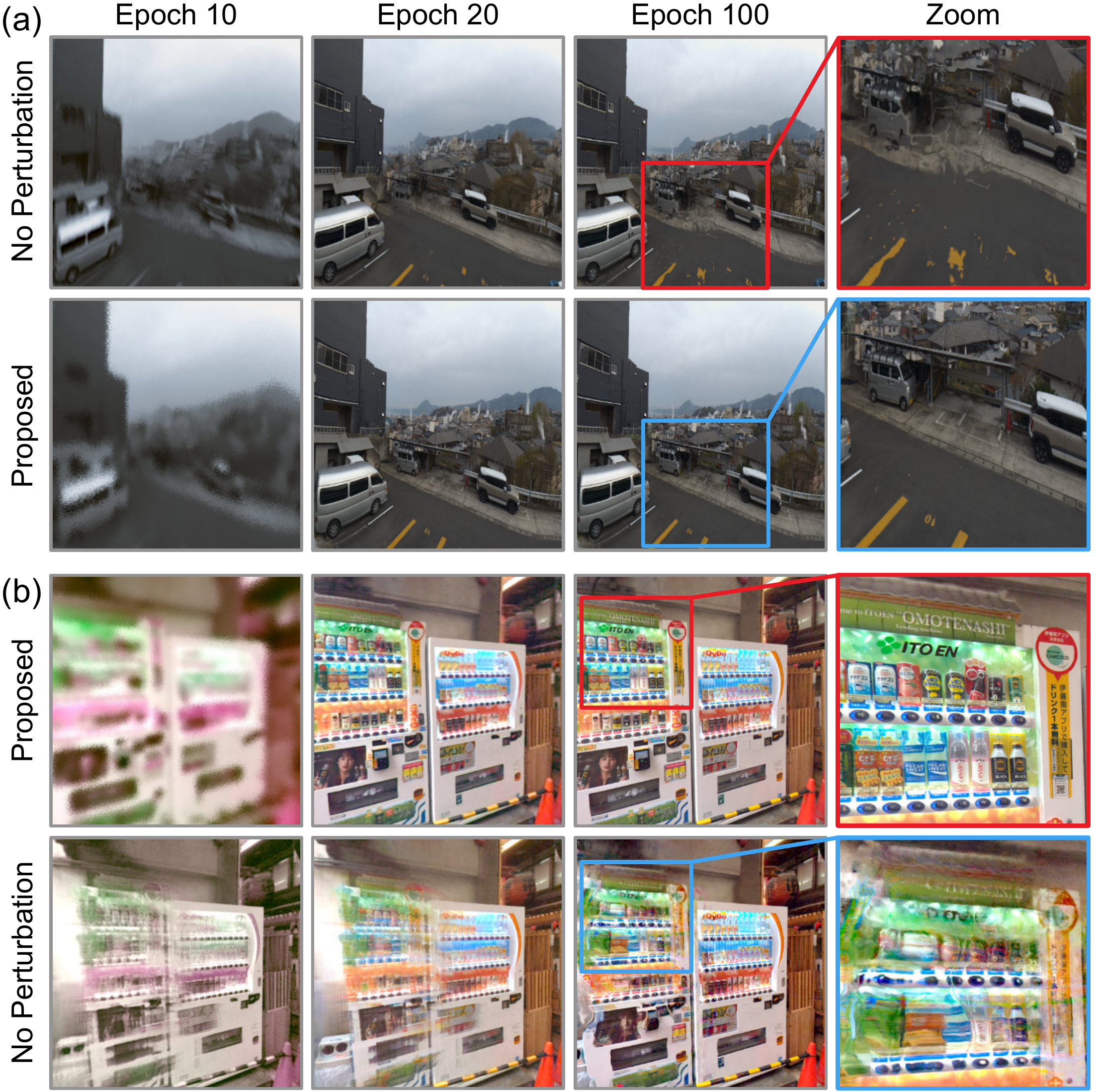}
  \caption{\textbf{Ray Perturbations.} By applying small perturbations to ray origins $O$ we are able to avoid hard-to-escape local minima solutions during early training epochs. In (a) we see how for the road, a region with low image texture, the \textit{No Perturbation} example duplicates content; creating two copies of the \#10 parking spot. In (b) we see how for repeated textures, perturbations can also help avoid ``crunching'' content in early training, where the repeated cans in the vending machine are accidentally aligned on top of each other. }
\label{figpan:perturbations}
\end{figure}

\subsection{Loss and Training Procedure}
With the rotation model $R(n)$ initialized with the device's onboard gyroscope measurements, and the translation model $T(n)$ initialized as all zeroes, we train the networks $\{h_\textsc{r}, h_\textsc{c}, h_\textsc{p}, h_\textsc{d}\}$ from scratch via stochastic gradient descent to fit an input scene. We break training into two stages: in the first, we freeze the ray offset and view-dependent color networks $h_\textsc{r}, h_\textsc{d}$ to allow the model to learn initial camera pose estimates and spherical color map, and in the second stage we unlock all networks to let them jointly continue training. Illustrated in Fig.~\ref{figpan:two_stage}, this helps prevent image artifacts caused by $h_\textsc{r}, h_\textsc{d}$ from accumulating during early training, where it is uncertain if parts of the scene are undergoing view-dependent color changes or simply stereo parallax. A similar problem also occurs for training the sphere color networks $h_\textsc{c}, h_\textsc{p}$, where the multi-resolution hash encoding $\gamma$ allows the network to fit image content \textit{undesirably} fast. This leads to artifacts, as seen in Fig.~\ref{figpan:perturbations}, where the image model learns duplicated or overlapping content faster than the motion model can correct for. We find that an effective and computationally inexpensive way of combating this behavior, shown in Eq.~\eqref{eqpan:perturbation}, is to add small perturbations to rays generated via Eq.~\eqref{eqpan:ray_generation} as
\begin{align}\label{eqpan:perturbation}
    \tilde{O} = O + \eta_p\mathcal{N}(0, 1),
\end{align}
where $\mathcal{N}(0, 1)$ is zero-mean standard Gaussian noise. The weight term $\eta_p$ is gradually decayed to zero over the first stage of training. Similar to prior work~\cite{chugunov2023shakes, li2023neuralangelo} we also mask the highest frequency grids in $\gamma_1$ and $\gamma_2$ to reduce the amount of accumulated noise during early training.

Given linear RAW inputs, we find $L_1$ to be an effective training loss, particularly for high noise reconstruction where zero-mean Gaussian read noise~\cite{brooks2019unprocessing} can be averaged out:
\begin{align}\label{eqpan:loss}
    \mathcal{L} = |c - \tilde{c}|.
\end{align}
We find that, with careful selection of encoding parameters for $\gamma_1$ and $\gamma_2$, \emph{no additional explicit regularization penalties are required} to constrain scene reconstruction~\cite{chugunov2024neural}. 

\begin{figure}[t!]
 \centering
\includegraphics[width=0.8\linewidth]{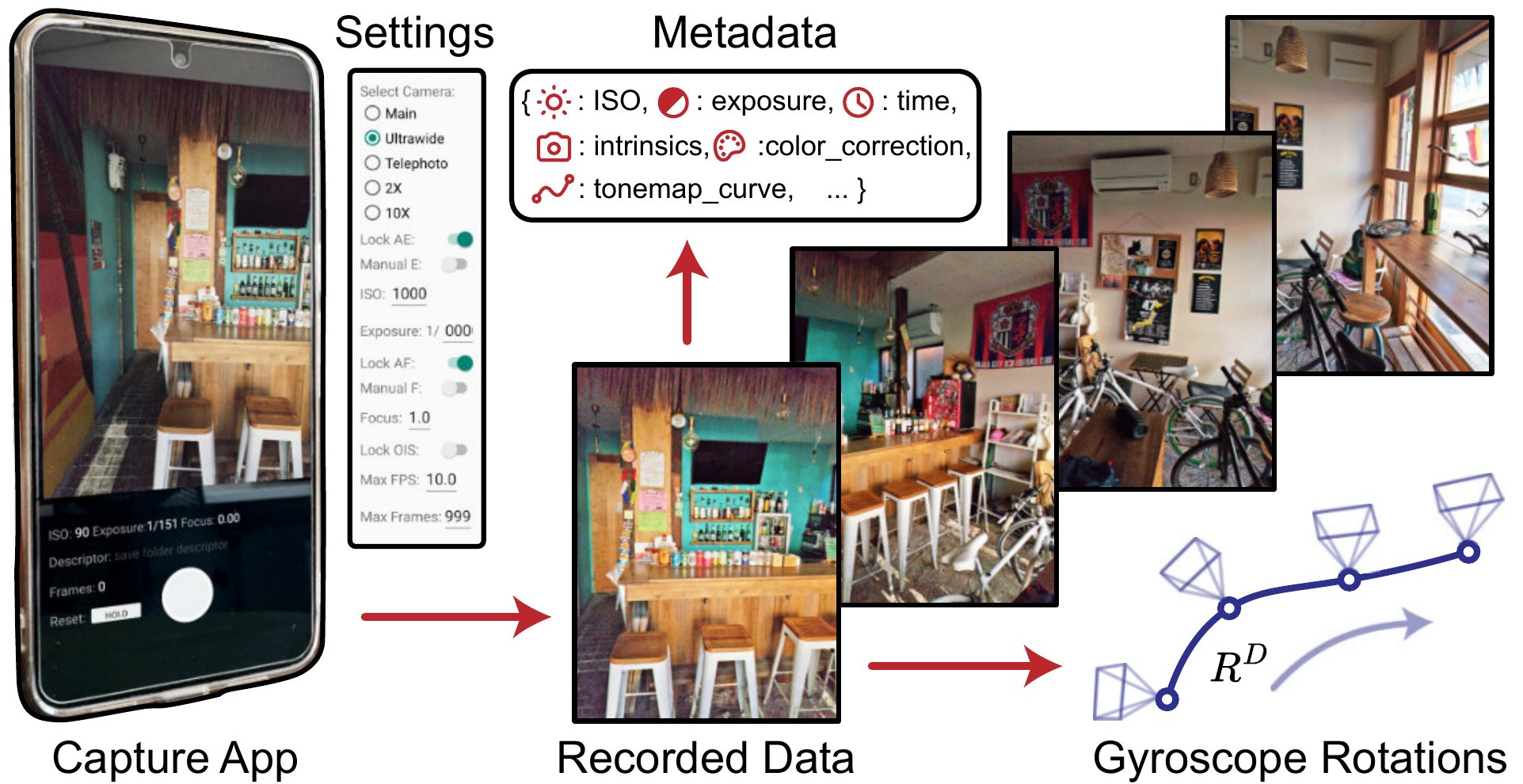}
  \caption{\textbf{Data Capture.} We develop an open-source Android-based mobile application to facilitate in-the-wild capture of scenes. The app's settings allow for camera selection (main, ultrawide, or telephoto) and to either use the device's auto-focus and auto-exposure features for capture, or set their respective values. During capture, we record full resolution Bayer RAW images, device accelerometer and gyroscope measurements, and all exposed camera and frame metadata including: ISO, exposure, timestamps, camera intrinsics, and color and tone correction values. }
\label{figpan:app_layout}
\end{figure}

\subsection{Data Collection} To record in-the-wild panorama video captures in unknown imaging conditions -- ranging from broad daylight to night-time photography -- we developed an Android-based data capture application, illustrated in Fig.~\ref{figpan:app_layout}. The app records a stream of RAW images along with metadata, enabling us to leverage linear sensor data for noise-robust reconstruction. While there exist other RAW video and image recording apps, we found they were paid and closed-source, missing desired functionality (e.g., specifying ISO, exposure, and recording frame-rate), and/or failed to record desired data (e.g., gyroscope measurements). In contrast, our app records full-resolution full bit-depth RAW images at the hardware maximum of 30 frames per second, accelerometer and gyroscope measurements, and nearly all camera and image metadata exposed by the Android APIs -- a list of which is included in the supplementary material. We make this app available open-source at: {\color{URLBlue}\href{https://github.com/Ilya-Muromets/Pani}{github.com/Ilya-Muromets/Pani}}
\begin{figure*}[t!]
 \centering
\includegraphics[width=\textwidth]{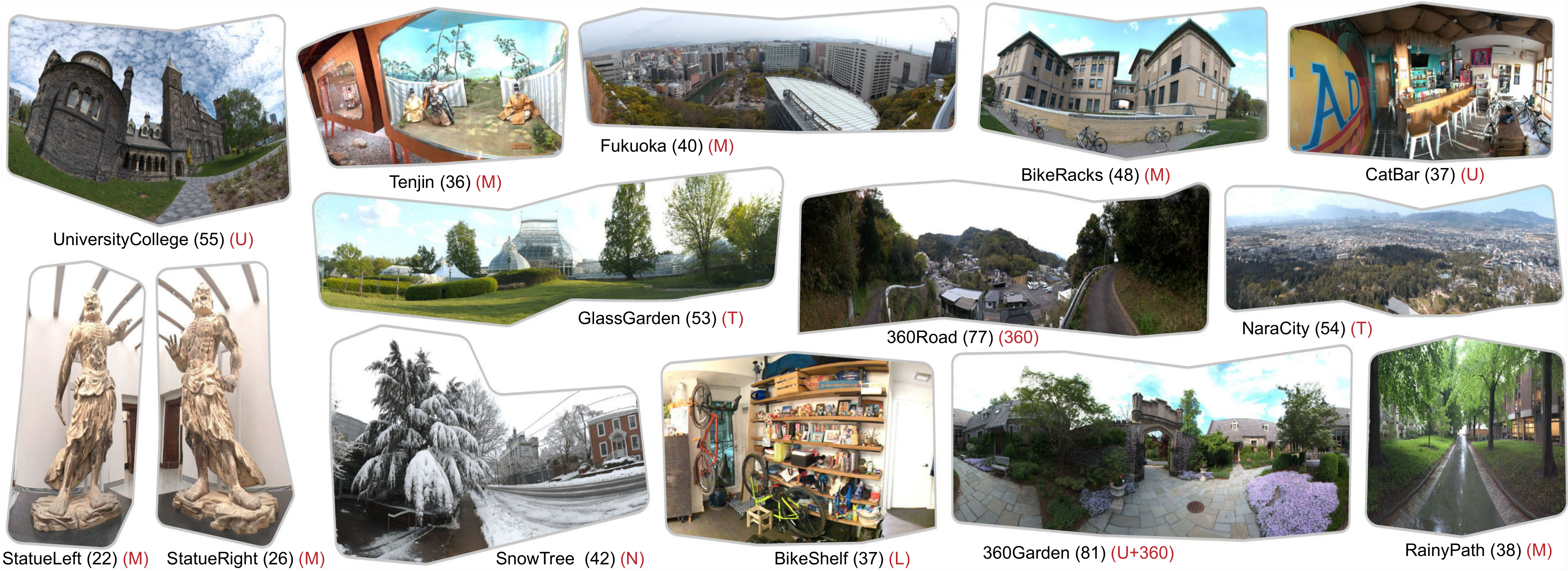}
  \caption{\textbf{Scene Diversity.} Shown above are spherical re-projections of reconstructions for a representative subset of scenes from our collected dataset. These include: (M) 1x main lens, (U) 0.5x ultrawide, (T) 5x telephoto, (L) low-light, (N) non-linear, and (360) full 360 degree captures. Scene titles are formatted as: \textit{Scene Name (Number of Captured Frames in Input)}.   }
\label{figpan:scenes}
\end{figure*}
We used a handheld Google Pixel 8 Pro cellphone to record a set of 50 scenes, a selection of which are presented in Fig.~\ref{figpan:scenes}, which cover a wide span of both imaging settings and capture paths. These include traditional $360^\circ$ and $180^\circ$ panoramas, as well as linear horizontal and vertical pans, back-and-forth pans, and random-walk paths. We use the device's auto-exposure settings for recording, with sensor sensitivity varying from ISO $\approx20$ in daylight to ISO $\approx10,000$ for night-time scenes. Though we restrict exposure time to $\leq1/100$s to minimize motion blur during the relatively fast capture process (3-10 seconds depending on the length of the capture path). Recorded image sequences range between 30 and 100 frames depending, and include captures with the main (1x), ultrawide (0.5x), and telephoto (5x) cameras available on the device.

\subsection{Implementation Details} We implement our model in PyTorch with the tiny-cuda-nn framework~\cite{muller2021real}. It is trained via stochastic gradient descent with the Adam~\cite{kingma2014adam} optimizer ($\beta\,{=}\,[0.9,0.99]$, $\epsilon\,{=}\,10^{-9}$, weight decay $10^{-5}$, learning rate $10^{-3}$) for 100 epochs, with 200 batches of $2^{18}$ rays per epoch. Rotation weight $\eta_\textsc{r}\,{=}\,10^{-3}$. Networks ${h_\textsc{r}, h_\textsc{p}, h_\textsc{d}}$ are all identical 5 layer 128 hidden unit MLPs;  $h_\textsc{c}$ is single 32$\times$3 linear layer to discourage blending of view-dependent and static color. Encoding $\gamma_2$ is a 15-level hash grid, with grid resolutions spanning $4$ to $3145$ by powers of 1.61 for each encoded dimension, and with a backing table size of $2^{19}$. To constrain the spatial frequency of the view-dependent color and ray models, encoding $\gamma_1$ is a significantly lower-resolution grid, with 8 levels spanning resolutions of $4$ to $112$. Trained on a single Nvidia RTX 4090 GPU, our method takes approximately \textit{12 minutes} to fit a 40 frame 12-megapixel sequence, though we include results in the supplementary material for how this can be further accelerated to under 30 seconds for generating ``preview-quality'' reconstructions from 3-megapixel inputs. The image model takes 80 MB of disk space, and can render 1920$\times$1080px frames at 50 FPS. Critical to our core design goals discussed in Sec.~\ref{secpan:neural_fields}, \textit{all parameters and training procedures are identical for all captures tested in all settings} (daytime, night-time, ultrawide, telephoto, etc.).

\section{Assessment}
In this section we compare our method to traditional image stitching and radiance field approaches. We then analyze the contributions of core model components, and confirm its applicability to the reconstruction of night-time scenes with noisy captures. For each scene we render views at 3x their original captured FOV, and include 3 input frames spanning the same FOV.
\subsection{Comparisons To Traditional Image Stitching}
While a large stitched image canvas is not the primary intended output of our neural light sphere model, as we focus on wide-view video rendering, comparisons to traditional image stitching methods help illustrate the challenges of this setting.

\begin{figure*}[h!]
 \centering
\includegraphics[width=\textwidth]{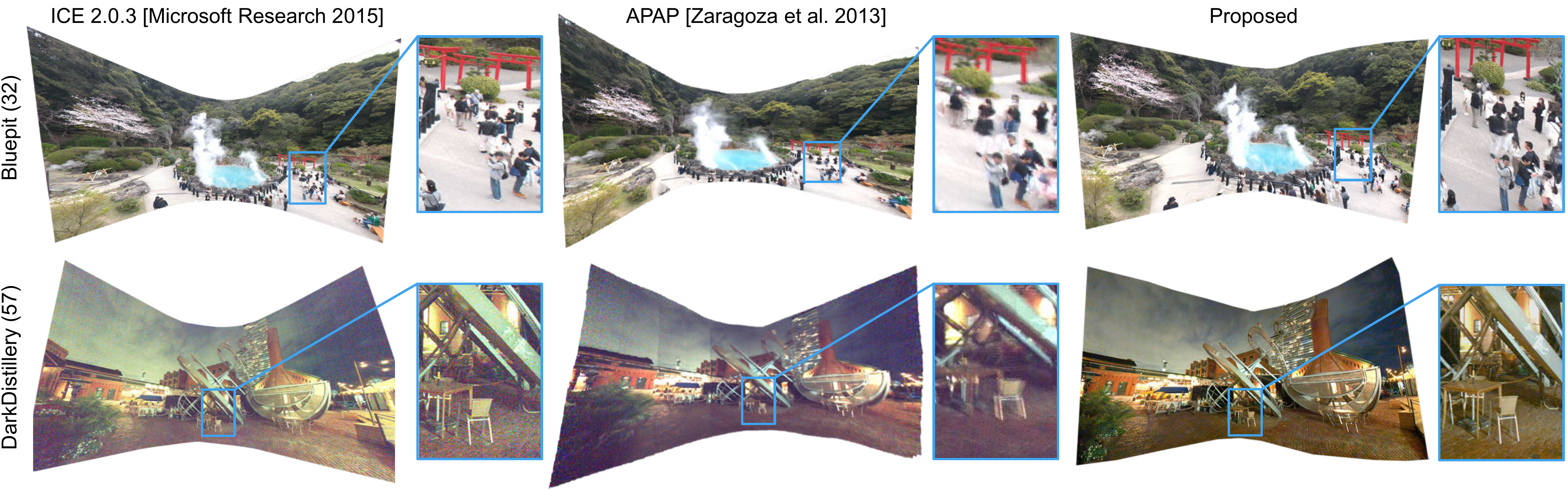}
  \caption{\textbf{Image Stitching Comparisons.} Visualizing rectilinear projections of the stitched panoramas, we see that APAP~\cite{zaragoza2013projective} averages multiple frames in \textit{DarkDistillery} to reduce noise, while ICE~\cite{MicrosoftICE} segments and freezes the motion of pedestrians in \textit{Bluepit}. Our proposed approach aims to do both, averaging multiple rays to reduce noise when possible while also preserving content in areas with local scene motion.
 }
\label{figpan:stitching_comparison}
 \centering
\includegraphics[width=0.98\textwidth]{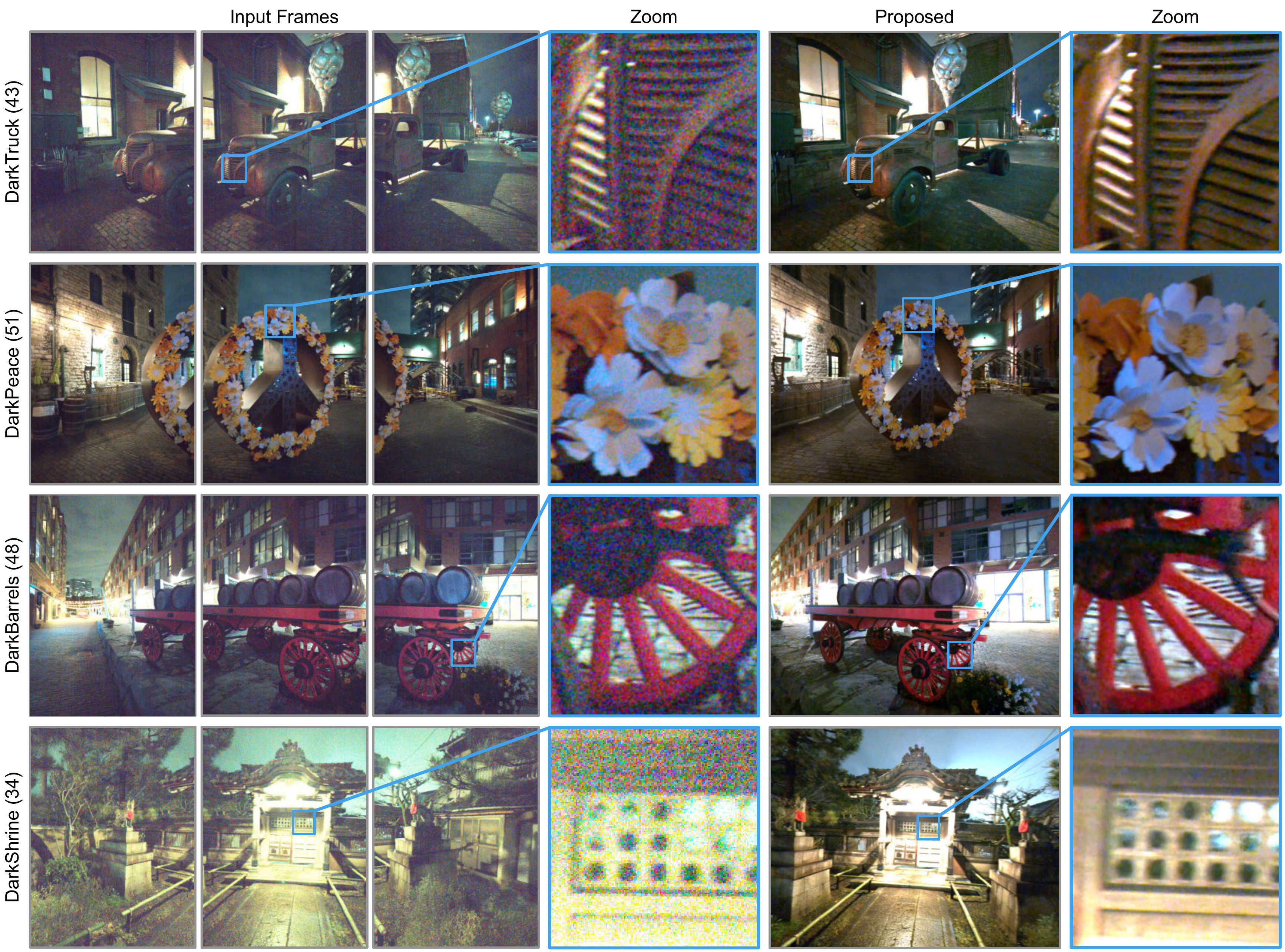}
  \caption{\textbf{Low-light Reconstruction.} Under low-light conditions, with sensor sensitivity at ISO 10,000 and exposure between $1/60\mathrm{s}$ and $1/120\mathrm{s}$, our proposed model is able to not only successfully reconstruct but also considerably denoise the captured scene. We recommend the reader to view the {\color{URLBlue}\href{https://light.princeton.edu/NeuLS}{associated video materials}} to see the effects of this denoising for interactive rendering.}
\label{figpan:lowlight}
\end{figure*}

Presented in Fig.~\ref{figpan:stitching_comparison}, we compare our approach to As-Projective-As-Possible (APAP) image stitching~\cite{zaragoza2013projective}, a robust parallax-tolerant cell-warping approach, and the Microsoft Image Composite Editor (ICE)~\cite{MicrosoftICE}, a polished software suite which performs globally projective warping and seam-blending to hide stitched image borders. APAP is able to warp and average multiple noisy measurements into a cleaner reconstruction, while ICE is restricted to stitching the borders of images together. However, ICE is significantly more resilient to motion-blur, freezing a sharp still frame of moving scene content. Our neural light sphere model offers both of these capabilities, averaging rays for better signal-to-noise ratio in static regions of the scene, while also more faithfully reconstructing dynamic content.

\begin{figure*}[h!]
 \centering
 
\includegraphics[width=\textwidth]{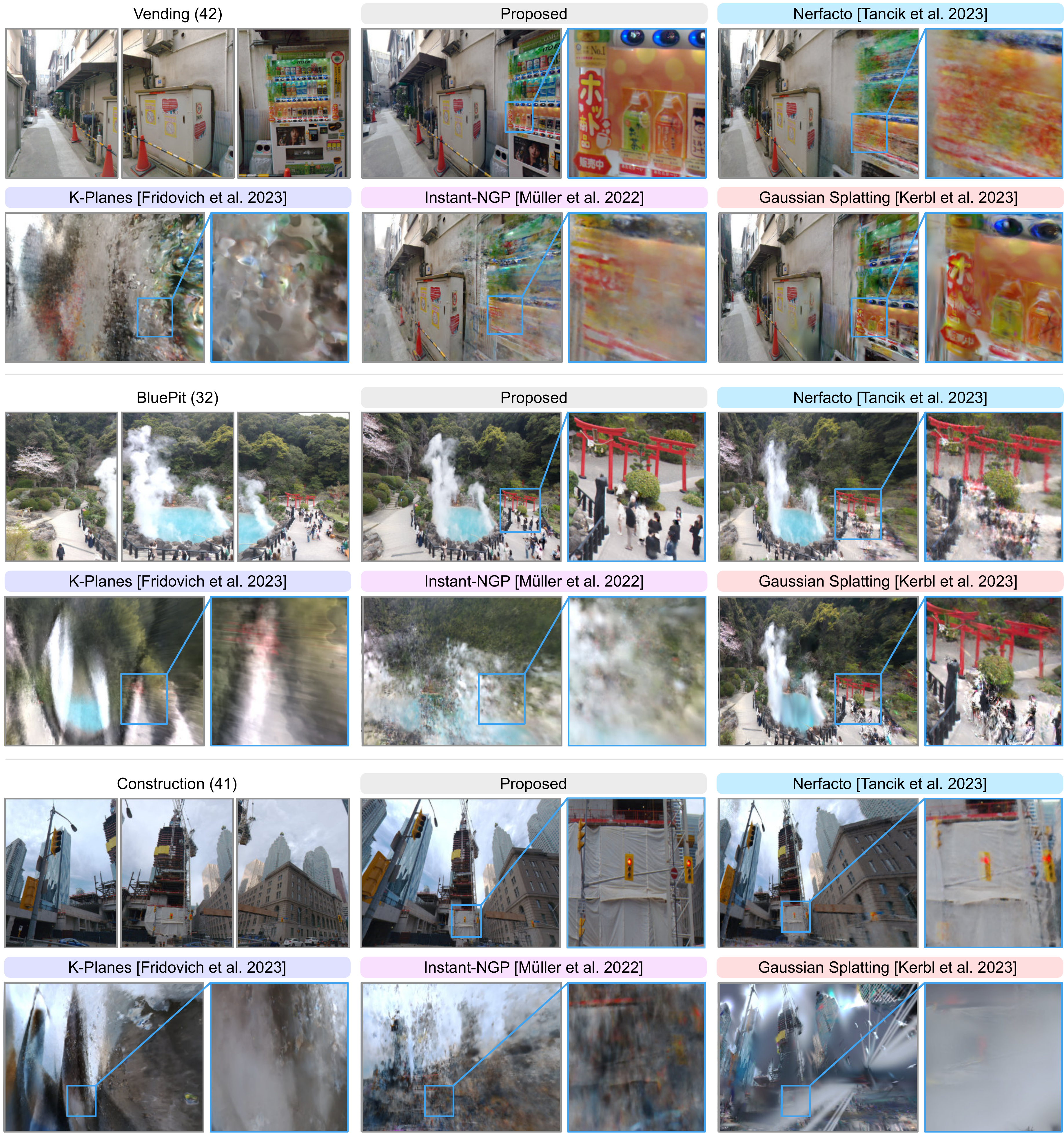}
  \caption{\textbf{Radiance Field Comparisons.} Compared to radiance field approaches, including other multi-resolution hash-based~\cite{tancik2023nerfstudio,muller2022instant} and non-volume-integrating~\cite{kerbl20233d,fridovich2023k} methods, we achieve significantly higher reconstruction quality over a range of settings. While Gaussian Splatting and Nerfacto are able to successfully overfit the center of most scenes (observed content), when the FOV is expanded to sample rays at wide angles they fail to correctly reconstruct fine images textures like the bottle labels in \textit{Vending}. In contrast, our neural light sphere model is able to reconstruct content in motion, like the pedestrians in \textit{BluePit} and fine parallax effects as in the traffic lights in \textit{Construction}. We recommend the reader to view the {\color{URLBlue}\href{https://light.princeton.edu/NeuLS}{associated video materials}} to better visualize these effects.  Scene titles are formatted as: \textit{Scene Name (Number of Captured Frames in Input)}.}
  \vspace{-1em}
\label{figpan:nerf_comparison}
\end{figure*}

figpan:
\subsection{Comparisons To Radiance Field Approaches} In Fig.~\ref{figpan:nerf_comparison} we compare our hash-grid based, non-volume-sampling neural light sphere approach to several related radiance field methods including: \textit{K-Planes} \cite{fridovich2023k}, an explicit representation that also avoids volume sampling by representing the scene as a product of two-dimensional planar features; \textit{Gaussian Splatting} \cite{kerbl20233d}, which also avoids volume sampling through its forward-projection model; \textit{Instant-NGP} \cite{muller2022instant}, which makes use of the same multi-resolution hash-grid backing as our approach; and the \textit{Nerfacto}~\cite{tancik2023nerfstudio} model, a robust combined approach with a hash-grid backing and per-image appearance conditioning. Unfortunately, given the largely rotational motion of panorama captures, even using exhaustive feature matching both COLMAP~\cite{schonberger2016structure} and HLOC~\cite{sarlin2019coarse} failed to reconstruct poses for a significant portion of our tested scenes -- including virtually all telephoto and ultrawide captures. We thus limit the comparison scenes to ones where COLMAP produced valid poses, and enable camera pose optimization in baseline methods which support it. In contrast, we emphasize that, beyond selecting a directory to load from, \textit{there is no human interaction required between capture and reconstruction for our proposed pipeline}. 

Despite tuning feature grid and regularization parameters, we were unable to achieve high-quality reconstructions with \textit{K-Planes}, which appears to produce noisy low-dimensional approximations of the scene. We find that the other baseline methods tend to overfit input captures by placing content a large distance away from the estimated camera position, producing an effect similar to traditional image stitching~\cite{brown2007automatic}. We suspect this is in large part due to inaccurate initial camera pose estimates, which cause content to be incorrectly localized in 3D space, and cause the reconstructions to settle in geometrically inaccurate local-minima solutions. When the FOV of the simulated camera is expanded, and we simulate rays at steeper angles relative to the camera axis as compared to the input data, we see these overfitting artifacts as texture quality on the edges of the baseline renders significantly degrades. \textit{Instant-NGP} in particular struggles to extrapolate from data with low parallax or significant scene motion, such as the billowing steam clouds in \textit{Bluepit}. Conversely, our proposed approach is able to recover fine texture content in these areas, including readable text on the drink labels in \textit{Vending}.

\subsection{Applications to Low-light Photography} Illustrated in Fig.~\ref{figpan:lowlight}, we find that, when trained on 10-bit linear RAW data, our neural light sphere model is robust to sensor noise as experienced in high ISO ($\geq 10,000$) settings during low-light photography. Similar to the findings of \cite{mildenhall2022nerf}, we find that by averaging rays that converge to identical scene points during training, our model learns a mean photometric solution for scene reconstruction, averaging out zero-mean Gaussian read noise. This also proves beneficial for non-light-limited settings, as we can lower exposure time for a single image to reduce motion blur during capture without risking failed reconstruction. Based on these initial findings, we expect a neural neural light sphere-style model could potentially be tailored for applications such as video denoising and astrophotography.

\section{Discussion}
In this work we present a compact and robust neural light sphere model for handheld panoramic scene reconstruction. We demonstrate high-quality texture reconstruction in expanded field-of-view renders, with high tolerance to adverse imaging effects such as noise and localized pixel motion. 

\paragraph{Future Work} We hope that this work, and the accompanying metadata- and measurement-rich dataset, can encourage follow-on research into scene reconstruction under adverse imaging conditions. Many of the scenes, such as those illustrated in Fig.~\ref{figpan:scenes}, purposely contain effects such as lens flare, snow, clouds, smog, reflections, sensor noise, and saturated high-dynamic range content. During in-the-wild data collection we found these effects unavoidable, highlighting the importance of robust reconstruction methods for practical computational photography. 

Beyond conventional photography, we believe this approach can be extended to industrial and scientific imaging settings such as satellite and telescope-based photography, scanning and array microscopes, and infrared or hyperspectral imaging. In particular, with a hardware-optimized hash-grid backing, our model design makes it computationally tractable to fit petapixel-and-larger data produced by these imaging modalities by breaking it into smaller ray batches -- e.g., a hash table size of $2^{22}$ reliably trains with batch size $2^{13}$ on a single Nvidia RTX 4090 GPU.

\begin{figure}[t!]
 \centering
\includegraphics[width=0.8\linewidth]{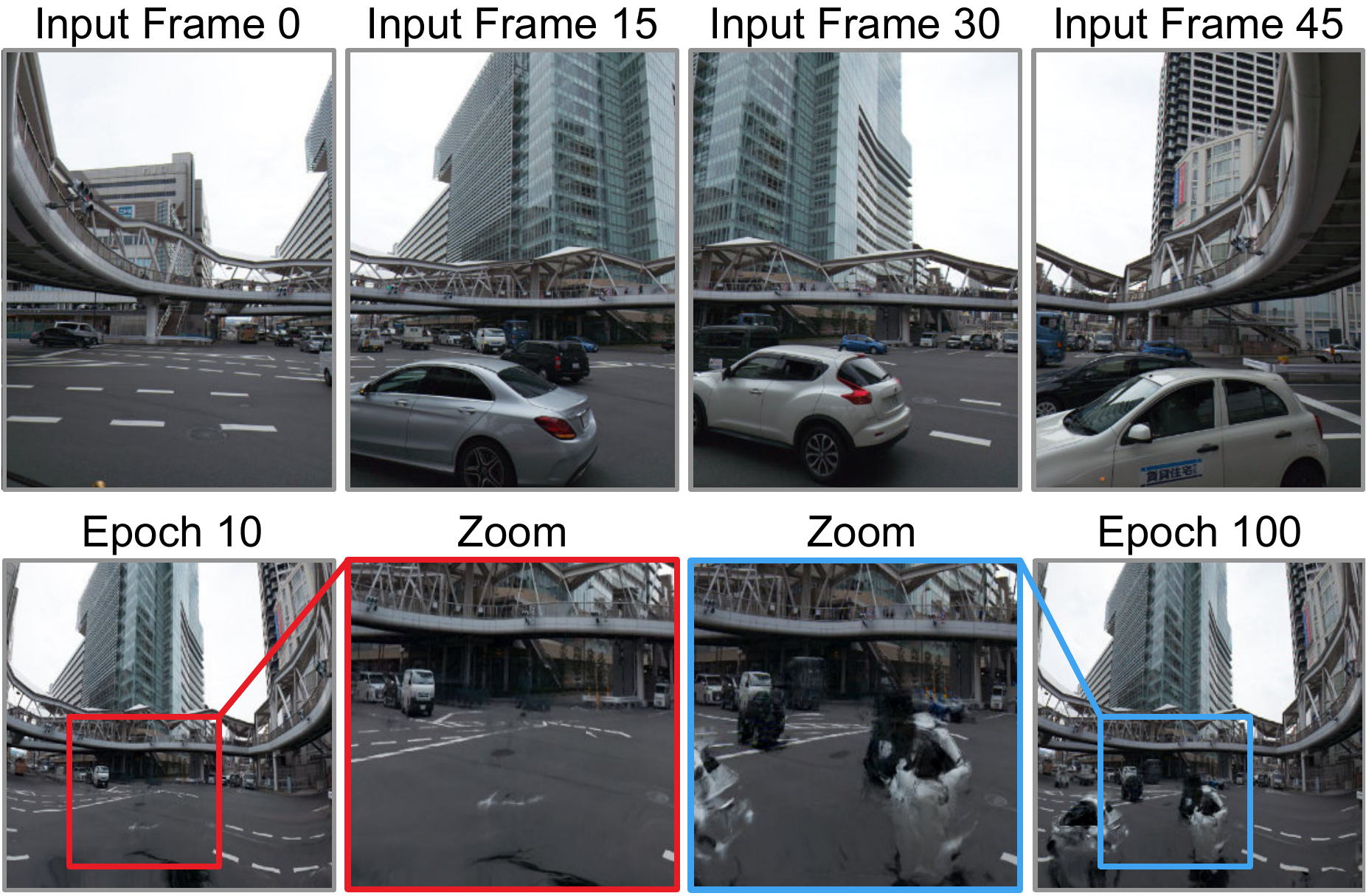}
  \caption{\textbf{Fast Occluders.} Objects such as bikes and cars, which quickly enter and exit the field-of-view of the camera, pose a challenge for scene reconstruction as they cannot be compactly modeled as a view-dependent effect. Shown in the example above, during early training the fast-moving cars are effectively erased from the reconstruction, which fits quickly to the median static pixel color. However, during later training stages, the view-dependent $h_\textsc{r}$ and $h_\textsc{d}$ models attempt (and fail) to reconstruct the content in motion, leading to transient car-shaped artifacts in the reconstruction.}
\label{figpan:fast_cars}
\end{figure}

\paragraph{Limitations} Although the proposed method is robust to localized pixel motion and color changes -- e.g., swaying tree branches, flowing water -- it is not capable of reconstructing large fast-moving obstructions such as vehicles driving through a scene as shown in Fig.~\ref{figpan:fast_cars}. This setting has posed a long-standing challenge for image-stitching and panoramic reconstruction works~\cite{szeliski2007image}, as when there are few observations of these occluders, this becomes a segmentation and tracking problem that is difficult to solve with a purely photometric approach such as ours. Similarly, without the ability to generate novel content, the camera path of the input capture strongly determines view synthesis performance -- e.g., a purely horizontal pan does not provide enough view information to simulate the effects of large vertical camera motion.

figpan:





\section{Implementation Details}
We compile a list of data recorded by our capture app and its uses in Tab.~\ref{table:metadata}. Our image processing pipeline follows the following sequence:
\begin{enumerate}
    \item Rearrange RAW data to BGGR format with \textit{color filter arrangement}
    \item Re-scale color channels as: \textit{(channel - black level)/(white level - black level)}
    \item Multiply by \textit{color correction gain}
    \item Multiply by inverse of \textit{shade map}
    \item Linearly interpolate gaps in mosaic (i.e., three interpolated values per red or blue, two interpolated values per green)
    \item Input into dataloader for training
\end{enumerate}
\vspace{1em}

\noindent To render final output images we then:
\begin{enumerate}
    \item Multiply RGB by the $3\times3$ \textit{color correction matrix}
    \item Re-scale color values with the \textit{tonemap curve}
\end{enumerate}
Or, optionally, skip this color correction to maximize render speed.

During training, we also use \textit{lens distortion} and \textit{rolling shutter skew} values to correct measurements on the ray level. Specifically we apply the lens distortion model as:

\begin{figure*}[t!]
 \centering
\includegraphics[width=\textwidth]{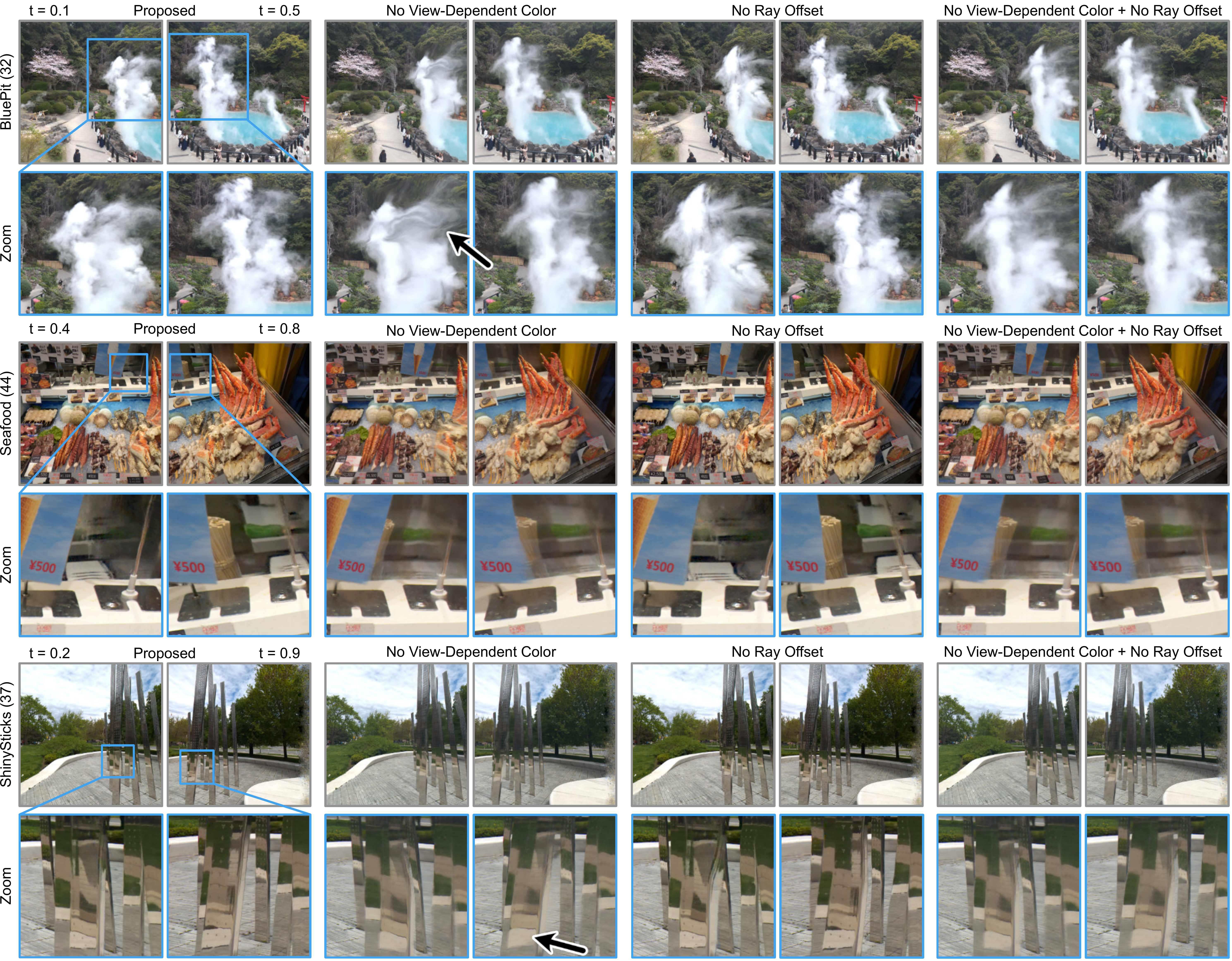}
 \caption{ \textbf{Model Component Analysis.} Shown above are the effects on reconstruction of zeroing out the contribution of the view-dependent color model $h_\textsc{d}\left(\gamma_1 (X);\, \theta_\textsc{d}\right)$, ray offset model $f_\textsc{r}(\hat{P}, X)$, or both models. We can observe that complex dynamic effects such as the steam clouds in \textit{BluePit} are produced by a combination of view-dependent color effects for the cloud texture, and ray offset for bulk motion. This is in contrast to the chopstick canister hidden behind the blue sign in \textit{Seafood}, which is almost entirely reconstructed with view-dependent color alone. In \textit{ShinySticks}, we observe how the sharp content and dots on the surface of the statue disappear when view-dependent color is removed, and large distortions in geometry appear when ray offset is omitted. }
\label{figpan:modelcomponents}
\end{figure*}

\begin{equation}
\begin{aligned}
    x_{\text{dist}} &= x (1 + \kappa_1 r^2 + \kappa_2 r^4 + \kappa_3 r^6 + \kappa_4 r^8 + \kappa_5 r^{10}) \\
    y_{\text{dist}} &= y (1 + \kappa_1 r^2 + \kappa_2 r^4 + \kappa_3 r^6 + \kappa_4 r^8 + \kappa_5 r^{10})
\end{aligned}
\end{equation}
where $( r^2 = x^2 + y^2)$ is the squared radius from the optical center given by the camera \textit{intrinsics}. We also shift the time $n$ at which rays are sampled -- linearly interpolating translation \textit{T(n)} and rotation \textit{R(n)} -- by the row the ray was sampled from multiplied by the row rolling shutter delay given by \textit{rolling shutter skew}/\textit{image height}. We note that this rolling shutter delay had negligible effect on the overall reconstruction, possibly due to view-dependent ray offset model $f_\textsc{r}(\hat{P}, X)$ already able to compensate for it (introducing a row-dependent skew to the rays).

While we do not use data such as \textit{accelerometer values}, which give poor localization performance after double integration for pans, or \textit{ISO} and \textit{exposure time}, we hope that these may be of use in follow-on work. For example, while we keep exposure and ISO locked during our captures, it could be possible to combine bracketing~\cite{delbracio2021mobile} with panoramic capture to reconstruct ultra-HDR scenes.

\begin{table}[t!]
\centering
\begin{tabular}{@{} p{0.4\linewidth} p{0.5\linewidth} @{}}
\toprule
\textbf{Data} & \textbf{Purpose} \\
\midrule
intrinsics & ray projection ($K$) \\
color correction matrix & render output images \\
tonemap curve & render output images \\
shade map & correct RAWs (lens shading) \\
color filter arrangement & correct RAWs (BGGR) \\
lens distortion & correct RAWs (distortion) \\
color filter gains & correct RAW (color) \\ 
whitelevel & scale RAW data (max) \\
blacklevel & scale RAW data (min) \\
gyroscope values & rotation initialization ($G$) \\
timestamps & synchronize measurements \\
rolling shutter skew &  rolling shutter correction \\
accelerometer values & \textbf{unused} \\
ISO & \textbf{unused} \\
exposure time & \textbf{unused} \\
focus distance & \textbf{unused} \\
focal length & \textbf{unused} \\
lens extrinsics & \textbf{unused} \\
lens aperture & \textbf{unused} \\
neutral color point & \textbf{unused} \\
noise profile & \textbf{unused} \\

\bottomrule
\end{tabular}
\caption{\textbf{Recorded Data.} A non-exhaustive list of data, both used and unused in this project, recorded by our capture app.}
\label{table:metadata}
\end{table}

\begin{figure*}[h!]
 \centering
\includegraphics[width=\textwidth]{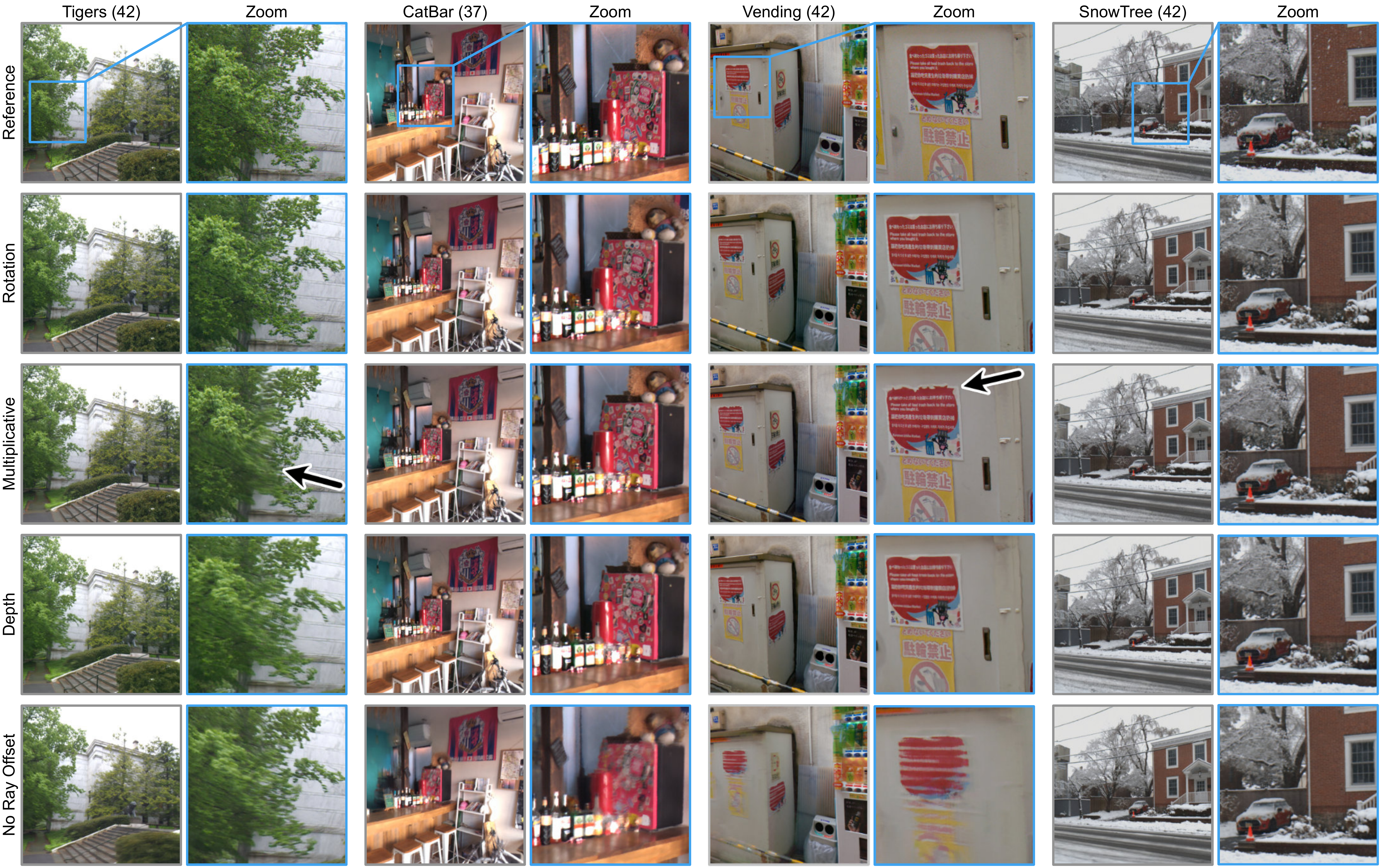}
  \caption{\textbf{Ray Offset Models.} Comparing scene reconstruction results for various ray offset models, it's clear from the \textit{No Ray Offset} results that many scenes such as \textit{CatBar} and \textit{Vending} contain significant parallax effects that a sphere projection model alone cannot compensate for. The \textit{Depth} and \textit{Multiplicative} models significantly improves reconstruction quality, albeit some regions in the \textit{Multiplicative} reconstructions suffer from distortions. The linearized \textit{Rotation} model avoids these artifacts while maintaining high reconstruction quality, recovering legible text in the \textit{Vending} scene.  }
\label{figpan:offsets}
\end{figure*}

\begin{figure*}[t!]
 \centering
\includegraphics[width=\textwidth]{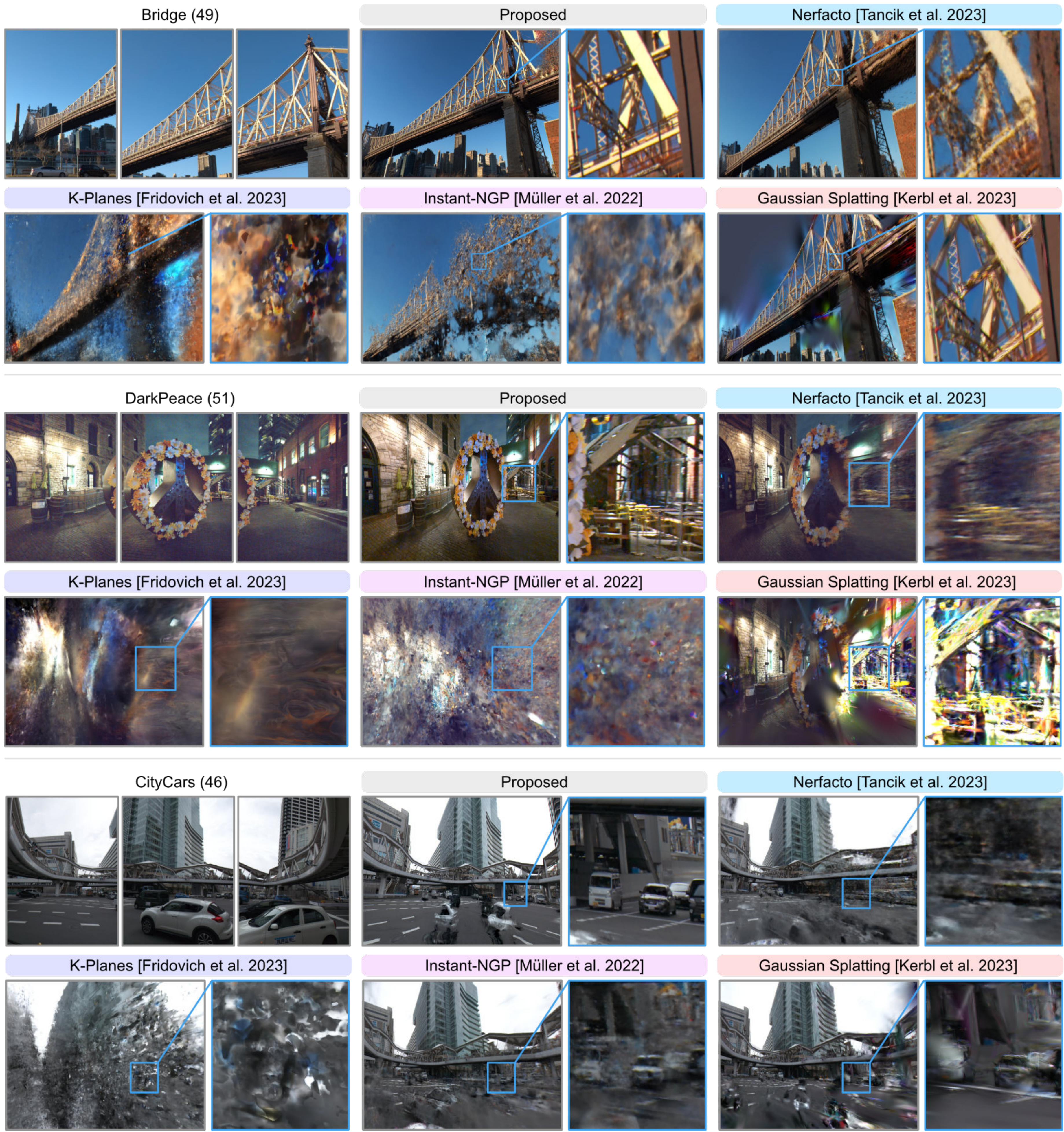}
  \caption{\textbf{Additional Radiance Field Comparisons.} Reconstruction results for a highly detailed back-and-forth \textit{Bridge} capture, night-time \textit{DarkPeace}, and \textit{CityCars} with fast-moving occluders. Scene titles are formatted as: \textit{Scene Name (Number of Captured Frames in Input)}}
\label{figpan:nerf_comparison}
\end{figure*}

\begin{figure*}[h!]
 \centering
\includegraphics[width=\textwidth]{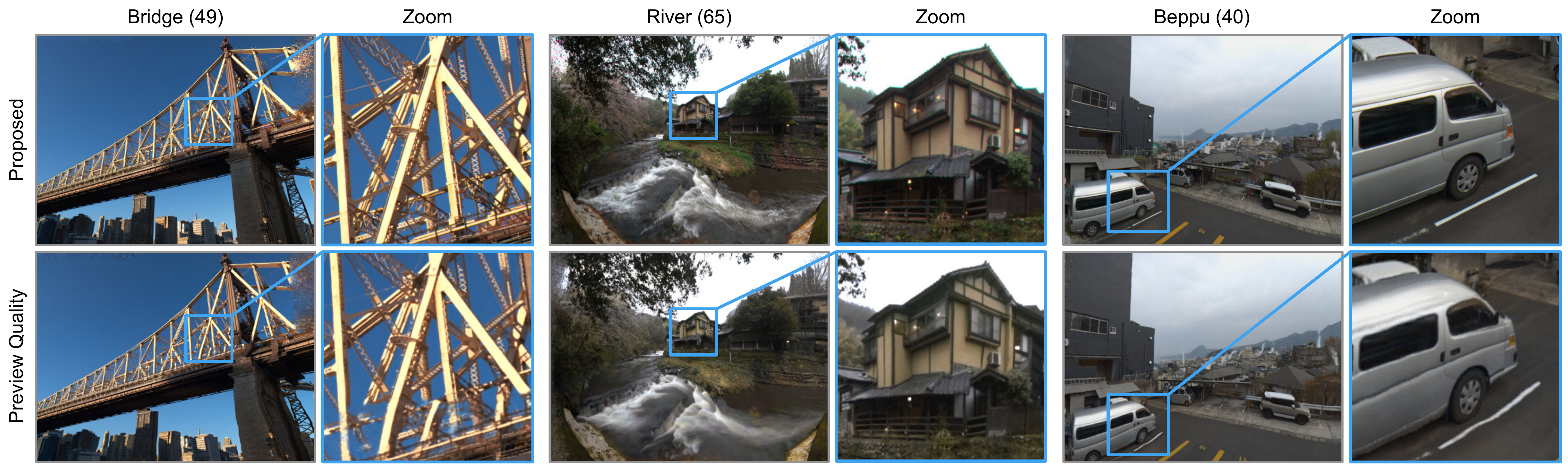}
  \caption{\textbf{Preview Quality Reconstructions.} Trained on 1/4 resolution inputs for 1/10th of the number of epochs, while they don't reach the full reconstruction quality of the proposed method, these ``Preview Quality'' reconstructions take less than 30 seconds of training time per scene.}
\label{preview}
\end{figure*}

\section{Model Component Analysis}
\label{secpan:modelcomponent}
In Fig.~\ref{figpan:modelcomponents} we visualize the independent contributions of the view-dependent color $f_{\textsc{c}}(\hat{P}^*,X)$ and ray offset models $f_\textsc{r}(\hat{P}, X)$ to our neural light sphere reconstructions. We train the model with both of these components active, and \textit{during inference time } we remove the output of the ray offset model
\begin{align}\label{eqpan:ray_distortion}
   \hat{D}^* =  D^*/\left\|D^*\right\|, \quad D^* = \cancelto{1}{\mathrm{rot}\left(\mathbf{R} = f_\textsc{r}(\hat{P}, X)\right)}\hat{D} = \hat{D},
\end{align}
remove the view-dependent color model
\begin{align}
    f_\textsc{c}(\hat{P}^*, X) &= h_\textsc{c}\left(h_\textsc{p}\left(\gamma_2 (\hat{P}^*); \,\theta_\textsc{p}\right) + \cancelto{0}{h_\textsc{d}\left(\gamma_1 (X);\, \theta_\textsc{d}\right)}; \, \theta_\textsc{c}\right)\nonumber \\
    f_\textsc{c}(\hat{P}^*, X) &= h_\textsc{c}\left(h_\textsc{p}\left(\gamma_2 (\hat{P}^*); \,\theta_\textsc{p}\right); \, \theta_\textsc{c}\right),
\end{align}
or remove both. From the resultant reconstructions, we can see how effects in the scene are modeled by one, both, or neither of these models. Static content on the surface of the sphere, such as the background folliage in \textit{BluePit} and \textit{ShinySticks} remains nearly identical in all reconstructions, which is entirely expected as this content exhibits almost no parallax and view-dependent color changes. In contrast, scene elements such as the reflections on the surface of \textit{ShinySticks} and the steam clouds in \textit{BluePit} require both the ray offset and view-dependent color models to work in tandem in order to produce these complex visual effects. This separability of our neural light sphere model also points towards a potentially interesting direction of future work, editing both content and its dynamics after reconstruction similar to a video mosaic~\cite{kasten2021layered} (e.g., turning the motion of the steam clouds into billowing smoke from a fire).
\section{Alternative Ray Offset Models} During the development of this work, we experimented with different ray offset models to model parallax and scene motion. This includes a \textit{Depth} model where we modify Eq. 4 of the main work to individually offset the radius of the sphere by  $f_\textsc{r}(\hat{P}, X)$ for each ray
\begin{align}\label{eqpan:sphere_intersection}
   \hat{P}^* &=  P/\left\|P\right\|, \quad P \,{=}\, \left[\arraycolsep=2.0pt
    \begin{array}{c}
    P_x \\
    P_y \\
    P_z \\
    \end{array}\right] = O + t\hat{D}
\end{align}
\begin{align}
    t &= -\left(O \cdot \hat{D}\right) + \sqrt{(O \cdot \hat{D})^2 - (\left\| O\right\|^2 - (1+f_\textsc{r}(\hat{P}, X)))},
\end{align}
where $\circ$ denotes element-wise multiplication. In Fig.~\ref{figpan:offsets} we can see how this model further sharpens content when compared to the \textit{Depth} model, but leads to blur and distortions in the scene where a large multiplicative offset causes rays to be ``pushed" out of a region in the scene. The final ray offset model we chose was a linearized \textit{Rotation} model
\begin{align}\label{eqpan:ray_distortion}
   \hat{D}^* =  D^*/\left\|D^*\right\|, \quad D^* = \mathrm{rot}\left(\mathbf{R} = f_\textsc{r}(\hat{P}, X)\right)\hat{D},
\end{align}
which we observed to lead to high reconstruction quality without the distortions observed in the \textit{Multiplicative} model. Here a larger $f_\textsc{r}(\hat{P}, X)$ rotates a region of rays together a larger distance, rather than pushing them out of a region on the sphere.

To compare these models, we remove the view-dependent color model $h_\textsc{d}\left(\gamma_1 (X);\, \theta_\textsc{d}\right)$ as outlined in Sec.~\ref{secpan:modelcomponent} \textit{during training}, not just during inference. As otherwise this $h_\textsc{d}\left(\gamma_1 (X);\, \theta_\textsc{d}\right)$ can compensate for content that was not correctly reconstructed by the ray offset model. We compare reconstruction results for these offset models in Fig.~\ref{figpan:offsets}, noting that for scenes such as \textit{Vending} and \textit{CatBar} with large amount of parallax the choice of offset model significantly affects reconstruction quality. Conversely, for \textit{SnowTree}, where content is far from the camera, all models produce similar reconstructions, emphasizing the importance of collecting a diverse set of scenes to holistically evaluate in-the-wild image stitching.

\section{Additional Reconstruction Results}
In Fig.~\ref{figpan:nerf_comparison} we showcase additional reconstruction results and comparisons to radiance field baselines: \textit{K-Planes} \cite{fridovich2023k}, \textit{Gaussian Splatting} \cite{kerbl20233d},  \textit{Instant-NGP} \cite{muller2022instant}, and \textit{Nerfacto}~\cite{tancik2023nerfstudio}. Noteably, we see in \textit{Bridge} the high resolution reconstruction enabled by our method, which is able to correctly resolve the cross-hatch bars in the bridge's support structure. In \textit{DarkPeace} we see that while \textit{Nerfacto} and \textit{Gaussian Splatting} successfully reconstruct the left side of the scene, the area of maximum overlap where the capture started, they produce extremely noisy reconstructions at the end of the capture sequence, with \textit{Instant-NGP} failing to reconstruct any of the scene. In \textit{CityCars} we can observe how, while our neural light sphere model is not able to reconstruct the fast-moving cars, the reconstruction artifacts only disrupt local content. Zooming into the background, we can still resolve the static cars, unlike the baseline methods, which produce reconstructions corrupted by motion artifacts.

\section{Preview-Scale Rendering}
While the reconstructions shown in the main text are relatively fast to train compared to the average neural radiance field approach, during model exploration and development we found it extremely beneficial to be able to quickly test large collections of scenes. By down-sampling the input data from full resolution 12-megapixel images to 1/4 resolution 3-megapixel images, dividing the max epochs by 10, and removing tensorboarding operations we are able to render ``preview quality'' scenes in less than 30 seconds. While there is a notable drop in quality for some scene content, as seen in the deformation of the grey car in the \textit{Beppu} example shown in Fig.~\ref{preview}, other scenes reach high reconstruction quality even in this short training time. Even zooming into the \textit{River} scene it is difficult to see a change in quality between the two reconstructions; suggesting that with some training augmentation, near-instant reconstruction could be possible for some subset of panoramic video captures.

\chapter{\vspace{-1em}Conclusion\label{ch:conclusion}}
In the article \textit{``The Inverse Problems You Carry in Your Pocket"}~\cite{chugunov2025inverse}, I ask if mobile computational photography can be \textit{``an accessible and affordable bridge between modern computer
vision and traditional inverse imaging problems"}, and I believe the answer is a resounding \textit{yes}. 

While the applications I demonstrate in this thesis -- depth reconstruction, layer separation, and image stitching -- are compelling in their own right, with state-of-the-art reconstruction results, I believe the greater value lies in the design principles by which these models were created. In developing the neural field representations presented in the previous chapters I constantly iterated on the goal of finding  \textit{``the simplest, useful model"} of the captured data. What made this more challenging and exciting is that I was also the one in charge of capturing the data. With the help of some Adobe and Google industry collaborators\footnote{Big thanks to Kevin, Francois, Kiran, and Sam.}, I was even able to develop the mobile data capture apps themselves, putting me in charge of the full pipeline from image acquisition to calibration, processing, and reconstruction. This led to a looping series of questions during research:
\begin{enumerate}
    \item Am I recording all the information I need for this task? Do I need additional calibration data, measurements, or device metadata? (e.g., gyroscope readings, color correction values)
    \item Did I correctly process and incorporate this data into my model? (e.g., does the camera model need lens distortion parameters, should I introduce an offset for rolling shutter?) 
    \item Did I capture enough and a good variety of scenes? Do I know when the model fails and why? (e.g., low-light settings, scene motion, adverse weather)
    \item Is the current design of the model optimal for this task? Do the parameters generalize across the captured scene? (e.g., do the encoding parameters and model geometry match the data?)
\end{enumerate}
Not coincidentally, these are the same questions I iterated through during my work with scientific and medical imaging systems. In this way, I believe these kinds of mobile computational photography projects can be an excellent gateway into computational imaging, even without access to an optical or radiological lab. Cell phones are made to be versatile and programmable, the room for creative research and exploration with them is infinite, and the skills are fully transferable. 

Returning to the core scientific questions of this work, while the specific applications vary, I present three distinct models that aggregate information from millions of small adjustments to distill semantically meaningful data -- whether it's for depth estimation, obstruction removal, or constructing a cohesive merged scene. Over the past decade we've seen a rapid evolution in imaging systems, with modern cell phones now equipped with multiple cameras, each with sensors comprised of millions of photodiodes -- which themselves are now being split into more sub-pixels in split pixel sensors~\cite{shi2024split}. Meanwhile, AR/VR devices are experimenting with laser depth sensors, infrared eye-tracking cameras, and an array of other passive and active imaging sensors. What this means is that the sheer amount of sensor data, the measurements pouring out of these devices, is continuing to rise, likely at a faster rate than we can develop power- and sample-efficient ways to process this data. This growing mountain of data underscores the need for compact, efficient methods that can scale to increasingly diverse and high-dimensional sensor streams. I'm particularly excited to see how neural field representations will evolve to tackle challenges in settings with extremely high-dimensional data, such as dynamic MRI, space telescopy, cryo-electron microscopy, and single-photon imaging.
\appendix 

\singlespacing
\bibliographystyle{plain}

\cleardoublepage
\ifdefined\phantomsection
  \phantomsection  
\else
\fi
\addcontentsline{toc}{chapter}{Bibliography}

\bibliography{cvpr}
\end{document}